\documentclass{llncs}
\usepackage{times}
\usepackage{helvet}
\usepackage{courier}

\usepackage{amssymb}
\usepackage{amsmath}
\usepackage{txfonts}
\usepackage{amssymb}
\usepackage{enumerate}
\usepackage{amsfonts}
\usepackage{times}
\usepackage{mathrsfs}
\usepackage{amscd}
\usepackage{stmaryrd}
\usepackage{graphicx}
\usepackage{makeidx}
\usepackage{multirow}

\newtheorem{algorithm}{Algorithm}
\newtheorem{assumption}{Assumption}

\newcommand{\powerset}[1]{{\mathcal P}(#1)}

\newcommand{\activation}{\alpha}

\newcommand{\real}{{\mathbb R}}

\newcommand{\activations}{acts}
\newcommand{\manipulation}{\delta}
\newcommand{\manipulationset}{\Delta}
\newcommand{\legal}{{V}}
\newcommand{\ladder}{ld}
\newcommand{\ladderset}{{\cal L}}

\DeclareMathOperator*{\argmax}{arg\,max}

\usepackage{color}
\usepackage{colortbl}

\begin{document}
\title{Safety Verification of Deep Neural Networks\thanks{This work is supported by the EPSRC Programme Grant on Mobile Autonomy (EP/M019918/1). Part of this work was done while MK was visiting the Simons Institute for the Theory of Computing.}}

\author{Xiaowei Huang, Marta Kwiatkowska, Sen Wang and Min Wu}

\institute{Department of Computer Science, University of Oxford}

\maketitle

\begin{abstract}
Deep neural networks have achieved impressive experimental results in image classification,
but can surprisingly be unstable with respect to adversarial perturbations, that is, minimal changes to the input image that cause the network to misclassify it.
With potential applications including perception modules and end-to-end controllers for self-driving cars, this raises concerns about their safety.
We develop a novel automated verification framework for feed-forward multi-layer neural networks based on Satisfiability Modulo Theory (SMT). 
We focus on safety of image classification decisions with respect to image 
manipulations, such as scratches or changes to camera angle or lighting conditions that would result in the same class being assigned by a human, and define safety for an individual decision in terms of invariance of the classification 
within a small neighbourhood of the original image. 
We enable exhaustive search of the region by
employing discretisation, 
and propagate the analysis layer by layer.
Our method works directly with the network code and, in contrast to existing methods, can guarantee that adversarial examples, if they exist, are found for the given region and family of manipulations. If found, adversarial examples can be shown to human testers and/or used to fine-tune the network. 
We implement the techniques using Z3 and evaluate them on state-of-the-art networks, including regularised and deep learning networks. We also compare against existing techniques to search for adversarial examples and estimate network robustness.
\end{abstract}

\section{Introduction}

\begin{figure}
\parbox{2.9cm}{
\includegraphics[width=1.4cm,height=1.4cm]{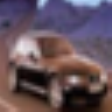}
\includegraphics[width=1.4cm,height=1.4cm]{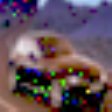}
{automobile to bird}
}
\parbox{2.9cm}{
\includegraphics[width=1.4cm,height=1.4cm]{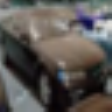}
\includegraphics[width=1.4cm,height=1.4cm]{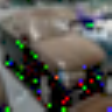}
{automobile to frog }
}
\parbox{2.9cm}{
\includegraphics[width=1.4cm,height=1.4cm]{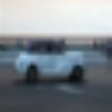}
\includegraphics[width=1.4cm,height=1.4cm]{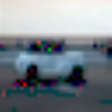}
{automobile to airplane }
}
\parbox{2.9cm}{
\includegraphics[width=1.4cm,height=1.4cm]{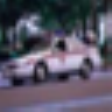}
\includegraphics[width=1.4cm,height=1.4cm]{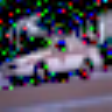}
{automobile to horse }
}
\caption{Automobile images (classified correctly) and their perturbed images (classified wrongly)}
\label{fig:automobile}
\end{figure}

Deep neural networks have achieved impressive experimental results in image classification, matching the cognitive ability of humans~\cite{LBH2015} in complex tasks with thousands of classes. Many applications are envisaged, including their use as perception modules and end-to-end controllers for self-driving cars \cite{NVIDIA2016}. Let $\real^n$ be a vector space of images (points) that we wish to classify and assume that $f: \real^n \to C$, where $C$ is a (finite) set of class labels, models the human perception capability, then
a neural network classifier is a function $\hat{f}(x)$ which approximates $f(x)$ from $M$ training examples 
$\{(x^i,c^i)\}_{i=1,..,M}$.
For example, a perception module of a self-driving car may input an image from a camera and must correctly classify the type of object in its view, irrespective of aspects such as the angle of its vision and image imperfections.
Therefore, though they clearly include imperfections, all four pairs of images in
Figure~\ref{fig:automobile} should arguably be classified as automobiles, since they appear so to a human eye.

Classifiers employed in vision tasks are typically multi-layer networks, which propagate the input image through a series of linear and non-linear operators. They are high-dimensional, often with millions of dimensions, 
non-linear and potentially discontinuous: even a small network, such as that trained to classify hand-written images of digits 0-9, has over 60,000 real-valued parameters and 21,632 neurons (dimensions) in its first layer. 
At the same time, the networks are trained on a finite data set and expected to generalise to previously unseen images. To increase the probability of correctly classifying such an image, regularisation techniques such as dropout are typically used, which improves the smoothness of the classifiers, in the sense that images that are close (within $\epsilon$ distance) to a training point are assigned the same class label.

Unfortunately, it has been observed in~\cite{Biggio2013,SZSBEGF2014} that deep neural networks, including highly trained and smooth networks optimised for vision tasks, are unstable with respect to so called \emph{adversarial perturbations}. Such adversarial perturbations are (minimal) changes to the input image, often imperceptible to the human eye, that cause the network to misclassify the image. Examples include not only artificially generated random perturbations, but also (more worryingly) modifications of camera images~\cite{KGB2016} that correspond to resizing, cropping or change in lighting conditions. They can be devised without access to the training set~\cite{practical-blackbox} and are transferable~\cite{DBLP:journals/corr/GoodfellowSS14}, in the sense that an example misclassified by one network is also misclassified by a network with a different architecture, even if it is trained on different data. Figure~\ref{fig:automobile} gives adversarial perturbations of automobile images that are misclassified as a bird, frog, airplane or horse by a highly trained state-of-the-art network. This obviously raises potential safety concerns for applications such as autonomous driving and calls for automated verification techniques that can verify the correctness of their decisions. 

Safety of AI systems 
is receiving increasing attention, to mention \cite{DBLP:journals/corr/SeshiaS16,DBLP:journals/corr/AmodeiOSCSM16}, in view of their potential to cause harm in safety-critical situations such as autonomous driving. Typically, decision making in such systems is either solely based on machine learning, through end-to-end controllers, or involves some combination of logic-based reasoning and machine learning components, where an image classifier produces a classification, say speed limit or a stop sign, that serves as input to a controller. A recent trend towards ``explainable AI'' has led to approaches that learn not only how to assign the classification labels, but also additional explanations of the model, which can take the form of 
a justification explanation (why this decision has been reached, for example identifying the features that supported the decision)~\cite{Hendricks2016,016:WIT:2939672.2939778Ribeiro:2}. 
In all these cases, the safety of a decision can be reduced to ensuring the correct behaviour of a machine learning component. However, safety assurance and verification methodologies for machine learning are little studied.

The main difficulty with image classification tasks, which play a critical role in perception modules of autonomous driving controllers, is that they do not have a formal specification in the usual sense:
ideally, the performance of a classifier should match the perception ability and class labels assigned by a human. 
Traditionally, the correctness of a neural network classifier is expressed in terms of \emph{risk}~\cite{Vapnik91}, defined as the probability of misclassification of 
a given image, weighted with respect to the input distribution $\mu$ of images. 
Similar (statistical) robustness properties of deep neural network classifiers, which compute the average minimum distance to a misclassification and are independent of the data point, have been studied and can be estimated using tools such as DeepFool~\cite{fross-pract} and cleverhans~\cite{papernot2016cleverhans}. 
However, we are interested in the safety of an \emph{individual decision}, and to this end focus on the key property of the classifier being \emph{invariant} to perturbations \emph{at a given point}. This notion is also known as pointwise robustness~\cite{fross-theory,constraints} or local adversarial robustness~\cite{KBDJK2017}.

{\bf Contributions.}
In this paper we propose a general framework for automated verification of safety of classification decisions made by feed-forward deep neural networks. Although we work concretely with image classifiers, the techniques can be generalised to other settings.
For a given image $x$ (a point in a vector space), we assume that there is a (possibly infinite) region $\eta$ around that point that incontrovertibly supports the decision, in the sense that all points in this region must have the same class. This region is specified by the user and can be given as a small diameter, or the set of all points whose salient features are of the same type. We next assume that there is a family of operations $\manipulationset$, which we call manipulations, that specify modifications to the image under which the classification decision should remain invariant in the region $\eta$. Such manipulations can represent, for example, camera imprecisions, change of camera angle, or replacement of a feature. 
We define a network decision to be \emph{safe} for input $x$ and region $\eta$ with respect to the set of manipulations $\manipulationset$ if applying the manipulations on $x$ will not result in a class change for $\eta$. We employ discretisation to enable a \emph{finite} \emph{exhaustive} search of the high-dimensional region $\eta$ for adversarial misclassifications. 
The discretisation approach is justified in the case of image classifiers since they are typically represented as vectors of discrete pixels (vectors of 8 bit RGB colours).
To achieve scalability, we propagate the analysis \emph{layer by layer}, mapping the region and manipulations to the deeper layers. We show that this propagation is sound, and is complete under the additional assumption of minimality of manipulations, which holds in discretised settings.
In contrast to existing approaches~\cite{SZSBEGF2014,PMJFCS2015}, our framework can guarantee that a misclassification is found if it exists.
Since we reduce verification to a search for adversarial examples, we can achieve safety \emph{verification} (if no misclassifications are found for all layers) or \emph{falsification} (in which case the adversarial examples can be used to fine-tune the network or shown to a human tester).

We implement the techniques using Z3~\cite{z3} in a tool called DLV (Deep Learning Verification)~\cite{DLV} and evaluate them on state-of-the-art networks, including regularised and deep learning networks.
This includes image classification networks trained for classifying hand-written images of digits 0-9 (MNIST), 10 classes of small colour images (CIFAR10), 43 classes of the German Traffic Sign Recognition Benchmark (GTSRB) \cite{Stallkamp2012} and 1000 classes of colour images used for the well-known imageNet large-scale visual recognition challenge (ILSVRC) \cite{ILSVRC}.
We also perform a comparison of the DLV falsification functionality on the MNIST dataset against the methods of~\cite{SZSBEGF2014} and \cite{PMJFCS2015}, focusing on the search strategies and statistical robustness estimation.
The perturbed images in Figure~\ref{fig:automobile} are found automatically using our tool for the network trained on the CIFAR10 dataset.

This invited paper is an extended and improved version of~\cite{HKWW2016}, where an extended version including appendices can also be found.

\section{Background on Neural Networks}

We consider feed-forward multi-layer neural networks~\cite{bishop1995neural}, henceforth abbreviated as neural networks. 
Perceptrons (neurons) in a neural network are arranged in disjoint layers, with each perceptron in one layer connected to the next layer, but no connection between perceptrons in the same layer.
Each layer $L_k$ of a network is associated with an $n_k$-dimensional vector space $D_{L_{k}} \subseteq \real^{n_k}$, in which each dimension corresponds to a perceptron. We write $P_k$ for the set of perceptrons in layer $L_k$ and $n_k=|P_k|$ is the number of perceptrons (dimensions) in layer $L_k$.

Formally, a \emph{(feed-forward and deep) neural} network $N$ is a tuple $(L,T,\Phi)$, where $L=\{L_k~|~k\in \{0,...,n\}\}$ is a set of layers such that layer $L_0$ is the \emph{input} layer and  $L_n$ is the \emph{output} layer, $T\subseteq L\times L$ is a set of sequential connections between layers such that, except for the input and output layers, each layer has an incoming connection and an outgoing connection, and $\Phi=\{\phi_k~|~k\in  \{1,...,n\}\}$ is a set of \emph{activation functions} $\phi_k: D_{L_{k-1}} \to D_{L_{k}}$, one for each non-input layer.
Layers other than input and output layers are called the \emph{hidden} layers.

The network is fed an input $x$ (point in $D_{L_{0}}$) through its input layer, which is then propagated through the layers by successive application of the activation functions. An \emph{activation} for point $x$ in layer $k$ is the value of the corresponding function, denoted $\activation_{x,k}=\phi_k(\phi_{k-1}(...\phi_1(x))) \in D_{L_k}$, where $\activation_{x,0}=x$.
For perceptron $p \in P_k$ we write $\alpha_{x,k}(p)$ for the value of its activation on input $x$.
For every activation $\activation_{x,k}$ and layer $k'<k$, we define $Pre_{k'}(\activation_{x,k})=\{\activation_{y,k'}\in D_{L_{k'}}~|~\activation_{y,k} = \activation_{x,k}\}$ to be the set of activations in layer $k'$ whose corresponding activation in layer $L_k$ is $\activation_{x,k}$. 
The classification decision is made based on the activations 
in the output layer  by, e.g., assigning to $x$ the class $\arg\max_{p\in P_n}\activation_{x,n}(p)$.
For simplicity, we use $\activation_{x,n}$ to denote the class assigned to input $x$, and thus $\activation_{x,n}=\activation_{y,n}$ expresses that two inputs $x$ and $y$ have \emph{the same class}.

The neural network classifier $N$ represents a function $\hat{f}(x)$ which approximates $f(x): D_{L_0} \to C$, a function that models the human perception capability in labelling images with labels from $C$, from $M$ training examples 
$\{(x^i,c^i)\}_{i=1,..,M}$.
Image classification networks, for example convolutional networks,
may contain many layers, which can be non-linear,
and work in high dimensions, which for the image classification problems can be of the order of millions. Digital images are represented as 3D tensors of pixels (width, height and depth, the latter to represent colour), where each pixel is a discrete value in the range 0..255.
The training process 
determines real values for weights used as filters that are convolved with the activation functions.
Since it is difficult to approximate $f$ with few samples in the sparsely populated high-dimensional space, to increase the probability of classifying correctly a previously unseen image, various regularisation techniques such as dropout are employed. They improve the smoothness of the classifier, in the sense that points that are $\epsilon$-close to a training point (potentially infinitely many of them) classify the same. 

In this paper, we work with the code of the network and its trained weights. 

\section{Safety Analysis of Classification Decisions}
In this section we define our notion of safety of classification decisions for a neural network, based on the concept of a manipulation of an image, essentially perturbations that a human observer would classify the same as the original image. Safety is defined for an individual classification decision and is parameterised by the class of manipulations and a neighbouring region around a given image. To ensure finiteness of the search of the region for adversarial misclassifications, we introduce so called ``ladders'', nondeterministically branching and iterated application of successive manipulations, and state the conditions under which the search is exhaustive.

\paragraph{\bf Safety and Robustness}
Our method assumes the existence of a (possibly infinite) region $\eta$ around a data point (image) $x$ 
such that all points in the region 
are indistinguishable by a human, and therefore 
have the same true class. 
This region is understood as supporting the \emph{classification decision} and can usually be inferred from the type of the classification problem. For simplicity, we identify such a region via its diameter $d$ with respect to some user-specified norm, which intuitively measures the closeness to the point $x$. 
As defined in~\cite{fross-theory}, a network $\hat{f}$ approximating human capability $f$ is said to be \emph{not robust at} $x$ if there exists a point $y$ in the region $\eta = \{z \in D_{L_0} \mid || z-x || \leq d\}$ of the input layer such that $\hat{f}(x) \neq \hat{f}(y)$. The point $y$, at a minimal distance from $x$, is known as an \emph{adversarial example}.
Our definition of \emph{safety for a classification decision} (abbreviated \emph{safety at a point}) follows
he same intuition, except that we work layer by layer, and therefore will identify such a region $\eta_k$, a subspace of $D_{L_k}$, at each layer $L_k$, for $k\in \{0,...,n\}$, and successively refine the regions through the deeper layers.
We justify this choice based on the observation~\cite{ALRMTP2016,LBH2015,Mallat2016} that deep neural networks are thought to compute progressively more powerful invariants as the depth increases. In other words, they gradually transform images into a representation in which the classes are separable by a linear classifier. 

\begin{assumption}\label{assump}
For each activation $\activation_{x,k}$ of point $x$ in layer $L_k$, the region $\eta_k(\activation_{x,k})$ contains activations 
that the human observer believes to be so close to $\activation_{x,k}$ that they should be classified the same as $x$. 
\end{assumption}
Intuitively, safety for network $N$ at a point $x$ means that the classification decision is robust at $x$ against perturbations within the region $\eta_k(\activation_{x,k})$. Note that, while the perturbation is applied in layer $L_k$, the classification decision is based on the activation in the output layer $L_n$.

\begin{definition}\label{def:general}
[General Safety]
Let $\eta_k(\activation_{x,k})$ be a region in layer $L_k$ of a neural network $N$ such that $\activation_{x,k}\in \eta_k(\activation_{x,k})$. We say that $N$ is \emph{safe for input} $x$ \emph{and region} $\eta_k(\activation_{x,k})$, written as $N,\eta_k\models x$, if for all activations $\activation_{y,k}$ in $\eta_k(\activation_{x,k})$ we have  $\activation_{y,n}= \activation_{x,n}$. 
\end{definition}

We remark that, unlike the notions of risk~\cite{Vapnik91} and robustness of~\cite{fross-theory,constraints}, we work with safety for a specific point and do not account for the input distribution, but such expectation measures can be considered, see Section~\ref{sec:comparison} for comparison.

\paragraph{\bf Manipulations}\label{sec:manipulations}

A key concept of our framework is the notion of a {\it manipulation}, an operator that intuitively models image perturbations, for example
bad angles, scratches or weather conditions, the idea being that the classification decisions in a region of images close to it should be invariant under such manipulations.
The choice of the type of manipulation is dependent on the application and user-defined, reflecting knowledge of the classification problem to model perturbations that should or should not be allowed. 
Judicious choice of families of such manipulations and appropriate distance metrics is particularly important.
For simplicity, we work with 
operators $\manipulation_k:D_{L_k}\rightarrow D_{L_k}$ over the activations in the vector space of layer $k$,  
and consider
the Euclidean ($L^2$) and Manhattan ($L^1$) norms to measure the distance between an image and its perturbation through $\manipulation_k$, but the techniques generalise to other norms discussed in~\cite{fross-theory,DBLP:journals/corr/GoodfellowSS14,constraints}. 
More specifically, applying a manipulation $\manipulation_k(\activation_{x,k})$ to an activation $\activation_{x,k}$ will result in another activation such that the values of \emph{some or all} dimensions are changed.
We therefore represent a manipulation as
a hyper-rectangle, defined for two activations $\activation_{x,k}$ and $\activation_{y,k}$ of layer $L_k$ by
$
rec(\activation_{x,k},\activation_{y,k}) = \times_{p\in P_k} [min(\activation_{x,k}(p),\activation_{y,k}(p)),~max(\activation_{x,k}(p),\activation_{y,k}(p))].
$
The main challenge for verification is the fact that the region $\eta_k$ contains potentially an uncountable number of activations. Our approach relies on discretisation in order to enable a finite exploration of the region to discover and/or rule out adversarial perturbations. 

For an activation $\activation_{x,k}$ and a set $\manipulationset$ of manipulations, we denote by $rec(\manipulationset,\activation_{x,k})$ the polyhedron which includes all hyper-rectangles that result from applying some manipulation in $\manipulationset$ on $\activation_{x,k}$, i.e., $rec(\manipulationset,\activation_{x,k}) = \bigcup_{\manipulation\in \manipulationset}rec(\activation_{x,k},\manipulation(\activation_{x,k}))$. Let $\manipulationset_k$ be the set of all possible manipulations for layer $L_k$.
To ensure region coverage, we define \emph{valid} manipulation as follows.

\begin{definition}\label{def:legalmanipulations}
Given an activation $\activation_{x,k}$,  a set of manipulations $\legal(\activation_{x,k})\subseteq \manipulationset_k$ is \emph{valid} if $\activation_{x,k}$ is an interior point of $ rec(\legal(\activation_{x,k}),\activation_{x,k})$, i.e., $\activation_{x,k}$ is in $rec(\legal(\activation_{x,k}),\activation_{x,k})$ and  does not belong to the boundary of $rec(\legal(\activation_{x,k}),\activation_{x,k})$.
\end{definition}

\begin{figure}
\center
\includegraphics[width=5cm,height=3cm]{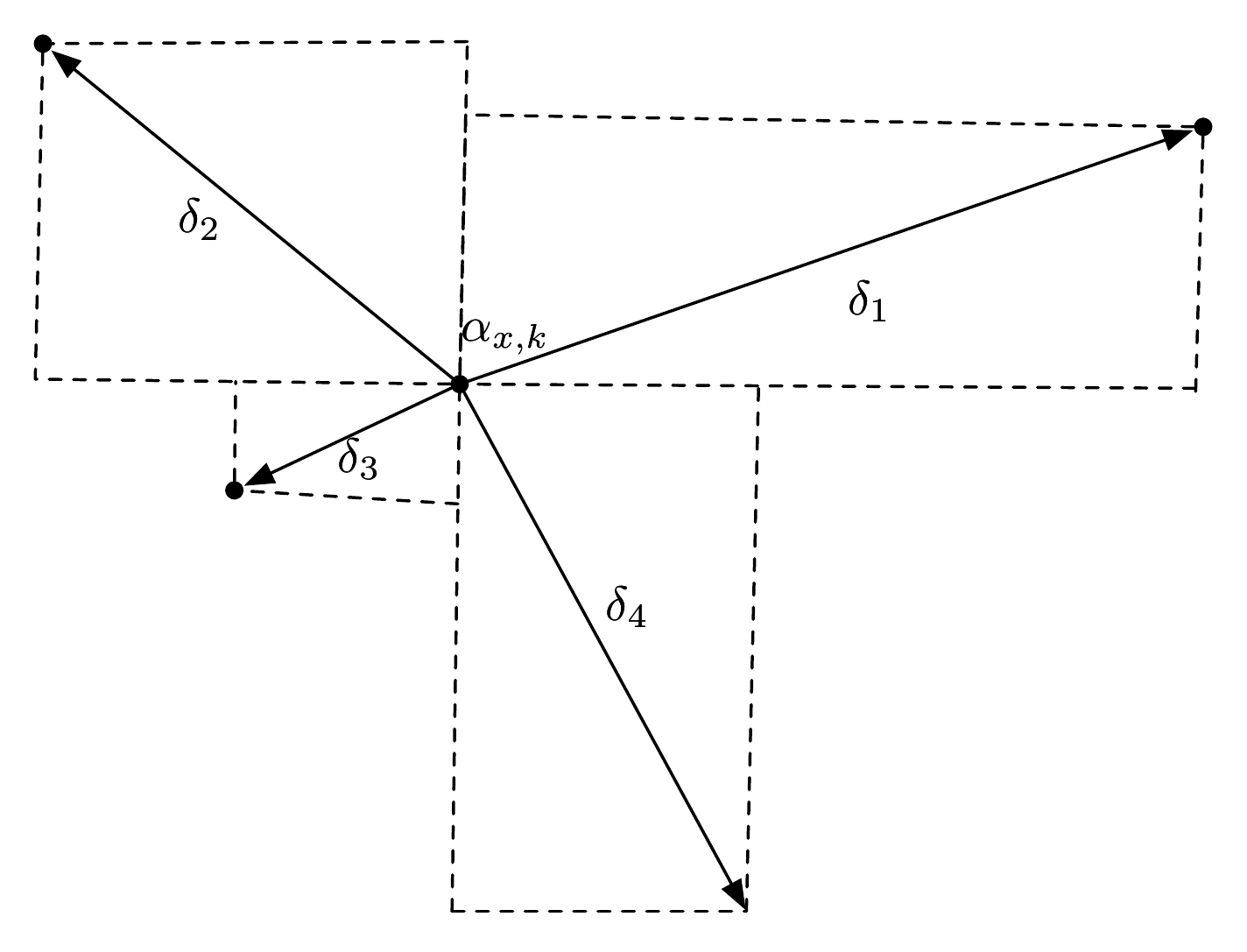}
\caption{Example of a set $\{\manipulation_1,\manipulation_2,\manipulation_3,\manipulation_4\}$ of valid manipulations  in a 2-dimensional space}
\label{fig:lm}
\end{figure}

Figure~\ref{fig:lm} presents an example of valid manipulations in two-dimensional space: each arrow represents a manipulation, each dashed box represents a (hyper-)rectangle of the corresponding manipulation, and activation $\activation_{x,k}$ is an interior point of the space from the dashed boxes.

Since we work with discretised spaces, which is a reasonable assumption for images, we introduce the notion of a \emph{minimal} manipulation. If applying a minimal manipulation, it suffices to check for misclassification just at the end points, that is, $\activation_{x,k}$ and $\manipulation_k(\activation_{x,k})$. This allows an exhaustive, albeit impractical, exploration of the region in unit steps.

A manipulation $\manipulation_k^1(\activation_{y,k})$ is \emph{finer than} $\manipulation_k^2(\activation_{x,k})$, written as $\manipulation_k^1(\activation_{y,k}) \leq  \manipulation_k^2(\activation_{x,k})$, if 
any activation in the hyper-rectangle of the former is also in the hyper-rectangle of the latter. It is implied in this definition that $\activation_{y,k}$ is an activation in the hyper-rectangle of $\manipulation_k^2(\activation_{x,k})$. Moreover, we write $\manipulation_{k,k'}(\activation_{x,k})$ for $\phi_{k'}(...\phi_{k+1}(\manipulation_k(\activation_{x,k})))$, representing the corresponding activation in layer $k'\geq k$ after applying manipulation $\manipulation_{k}$ on the activation $\activation_{x,k}$, where $\manipulation_{k,k}(\activation_{x,k})=\manipulation_{k}(\activation_{x,k})$.
\begin{definition}
A manipulation $\manipulation_k$ on an activation $\activation_{x,k}$ is \emph{minimal} if there does not exist manipulations $\manipulation_k^1$ and $\manipulation_k^2$ and an activation $\activation_{y,k}$ such that $\manipulation_k^1(\activation_{x,k})\leq \manipulation_k(\activation_{x,k})$, $\activation_{y,k} = \manipulation_k^1(\activation_{x,k})$, $ \manipulation_k(\activation_{x,k}) = \manipulation_k^2(\activation_{y,k})$, and $\activation_{y,n} \neq \activation_{x,n}$ and $\activation_{y,n}\neq  \manipulation_{k,n}(\activation_{x,k})$.
\end{definition}
Intuitively, a minimal manipulation does not have a finer manipulation that results in a different classification. However, it is possible to have different classifications before and after applying the minimal manipulation, i.e., it is possible that $\manipulation_{k,n}(\activation_{x,k})\neq \activation_{x,n}$. 
It is not hard to see that the minimality of a manipulation implies that the class change in its associated hyper-rectangle can be detected by checking the class of the end points $\activation_{x,k}$ and $\manipulation_k(\activation_{x,k})$.

\paragraph{\bf Bounded Variation}\label{sec:boundedVariation}

\begin{figure}
\center
\includegraphics[width=10cm,height=6cm]{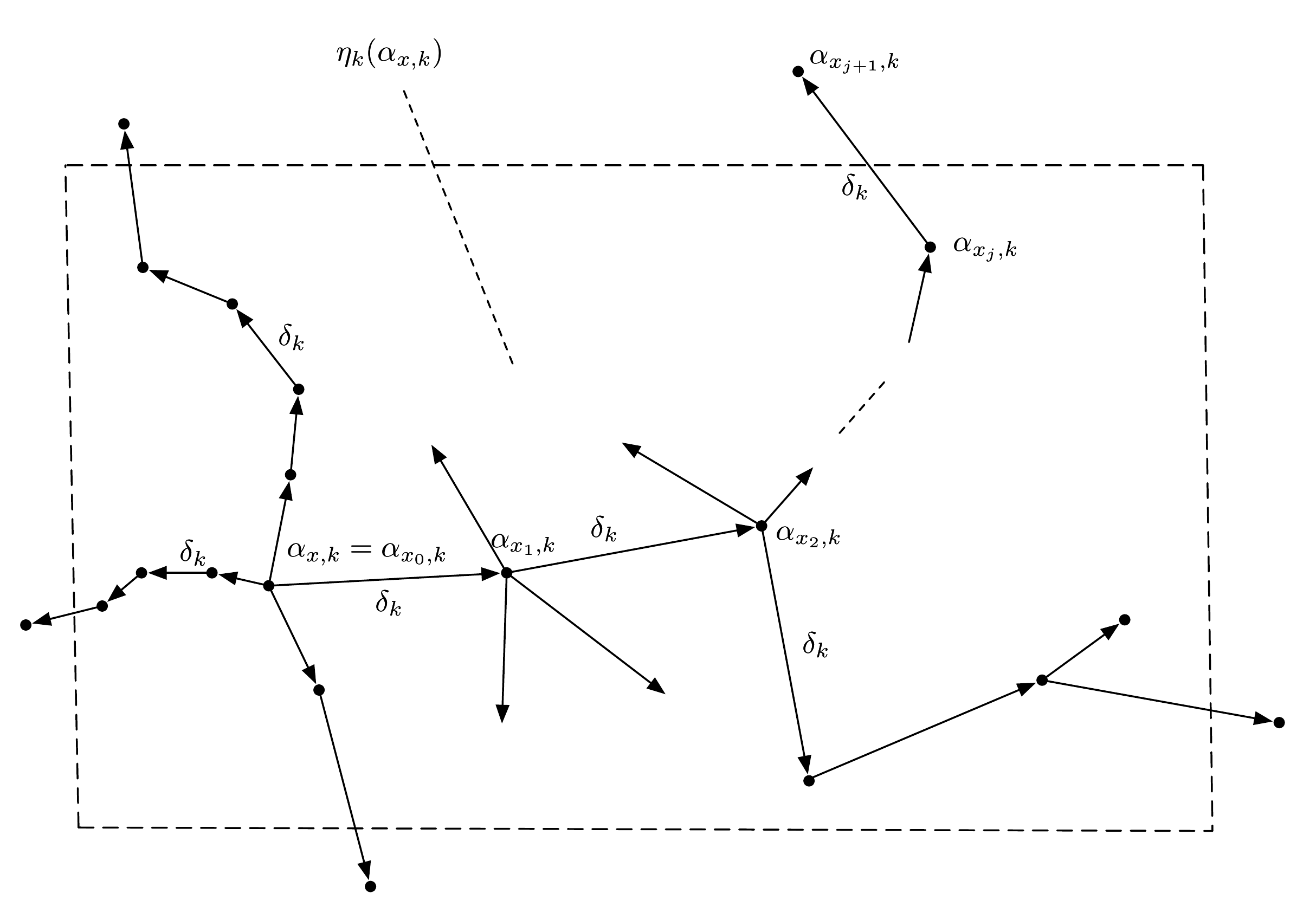}
\caption{Examples of ladders in region $\eta_k(\activation_{x,k})$. Starting from $\activation_{x,k}=\activation_{x_0,k}$, the activations $\activation_{x_1,k}... \activation_{x_j,k}$ form a ladder such that each consecutive activation results from some valid manipulation $\manipulation_k$ applied to a previous activation, and the final activation $\activation_{x_j,k}$ is outside the region $\eta_k(\activation_{x,k})$.}
\label{fig:ladder}
\end{figure}

Recall that we apply manipulations in layer $L_k$, but check the classification decisions in the output layer.
To ensure \emph{finite, exhaustive} coverage of the region, we
introduce a continuity assumption on the mapping from space $D_{L_k}$ to the output space $D_{L_n}$, adapted from the concept of bounded variation~\cite{AFP2000}.
Given an activation $\activation_{x,k}$ with its associated region $\eta_k(\activation_{x,k})$,
we define a ``ladder'' on $\eta_k(\activation_{x,k})$ to be a set $\ladder$ of activations containing $\activation_{x,k}$ and finitely many, possibly zero, activations from $\eta_k(\activation_{x,k})$. The activations in a ladder can be arranged into an increasing  order $\activation_{x,k}=\activation_{x_0,k}<\activation_{x_1,k}<...<\activation_{x_j,k}$ such that every activation $\activation_{x_t,k}\in \ladder$ appears once and has a successor $\activation_{x_{t+1},k}$ such that $\activation_{x_{t+1},k}=\manipulation_k(\activation_{x_t,k})$ for some manipulation $\manipulation_k\in \legal(\activation_{x_t,k})$.
For the greatest element $\activation_{x_j,k}$, its successor should be outside the region $\eta_k(\activation_{x,k})$, i.e., $\activation_{x_{j+1},k}\notin \eta_k(\activation_{x,k})$. Given a ladder $\ladder$, we write $\ladder(t)$ for its $t+1$-th activation, $\ladder[0..t]$ for the prefix of $\ladder$ up to the $t+1$-th activation, and $last(\ladder)$ for the greatest element of $\ladder$.
Figure~\ref{fig:ladder} gives a diagrammatic explanation on the ladders.

\begin{definition}
Let $\ladderset(\eta_k(\activation_{x,k}))$ be the set of ladders in $\eta_k(\activation_{x,k})$. Then the \emph{total variation} of the region $\eta_k(\activation_{x,k})$ on the neural network with respect to $\ladderset(\eta_k(\activation_{x,k}))$ is
$$
V(N;\eta_k(\activation_{x,k})) = \sup_{\ladder\in \ladderset(\eta_k(\activation_{x,k}))} \sum_{\activation_{x_t,k}\in \ladder\setminus \{last(\ladder)\}} \mathrm{diff}_n(\activation_{x_t,n}, \activation_{x_{t+1},n})
$$
where $\mathrm{diff}_n:D_{L_n}\times D_{L_n}\rightarrow \{0,1\}$ is given by $\mathrm{diff}_n(\activation_{x,n}, \activation_{y,n}) = 0$ if $\activation_{x,n} = \activation_{y,n}$ and 1 otherwise. We say that the region $\eta_k(\activation_{x,k})$ is a \emph{bounded variation} if $V(N;\eta_k(\activation_{x,k})) < \infty$, and are particularly interested in the case when $V(N;r_k(\activation_{y,k})) =0$, which is called a \emph{0-variation}.
\end{definition}

The set $\ladderset(\eta_k(\activation_{x,k}))$ is {\em complete} if, for any ladder $\ladder\in \ladderset(\eta_k(\activation_{x,k}))$ of $j+1$ activations, any element $\ladder(t)$ for $0\leq t\leq j$, and any manipulation $\manipulation_k\in \legal(\ladder(t))$,
 there exists a ladder $\ladder'\in \ladderset(\eta_k(\activation_{x,k}))$ such that $\ladder'[0..t]=\ladder[0..t]$ and $\ladder'(t+1)=\manipulation_k(\ladder(t))$.
Intuitively, a complete ladder is a complete tree, on which each node represents an activation and each branch of a node corresponds to a valid manipulation. From the root $\activation_{x,k}$, every path of the tree leading to a leaf is a ladder. Moreover, the set $\ladderset(\eta_k(\activation_{x,k}))$ is {\em covering} if the polyhedra of all activations in it cover the region $\eta_k(\activation_{x,k})$, i.e.,
\begin{equation}\label{equ:covering}
 \eta_k(\activation_{x,k}) \subseteq  \bigcup_{\ladder\in \ladderset(\eta_k(\activation_{x,k}))} \bigcup_{\activation_{x_t,k}\in \ladder\setminus \{last(\ladder)\}}rec(\legal(\activation_{x_t,k}),\activation_{x_t,k}).
\end{equation}

Based on the above, we have the following definition of safety with respect to a set of manipulations. Intuitively, we \emph{iteratively} and \emph{nondeterministically} apply manipulations to explore the region $\eta_k(\activation_{x,k})$, and safety means that no class change is observed by successive application of such manipulations.

\begin{definition}\label{def:finite}
[Safety wrt Manipulations]
Given a neural network $N$, an input $x$ and a set $\manipulationset_k$ of manipulations, we say that $N$ is \emph{safe for input} $x$ \emph{with respect to the region} $\eta_k$ \emph{and manipulations} $\manipulationset_k$, written as $N,\eta_k,\manipulationset_k\models x$, if the region $\eta_k(\activation_{x,k})$ is a 0-variation for the set $\ladderset(\eta_k(\activation_{x,k}))$ of its ladders, which is complete and covering.
\end{definition}

It is straightforward to note that general safety in the sense of Definition~\ref{def:general} implies safety wrt manipulations, in the sense of Definition~\ref{def:finite}.

\begin{theorem}\label{thm:feature}
Given a neural network $N$, an input $x$, and a region $\eta_k$, we have
that
 $N,\eta_k\models x$ implies $N,\eta_k,\manipulationset_k\models x$ for any set of manipulations $\manipulationset_k$.
\end{theorem}

In the opposite direction, we require the minimality assumption on manipulations.

\begin{theorem}\label{thm:enhanced}
Given a neural network $N$, an input $x$, a region $\eta_k(\activation_{x,k})$ and  a set $\manipulationset_k$ of manipulations, we have that $N,\eta_k,\manipulationset_k\models x$ implies $N, \eta_k\models x$ if  the manipulations in $\manipulationset_k$ are minimal.
\end{theorem}

Theorem~\ref{thm:enhanced} means that, under the minimality assumption over the manipulations, an \emph{exhaustive} search through the complete and covering ladder tree from $\ladderset(\eta_k(\activation_{x,k}))$ can find adversarial examples, if any, and enable us to conclude that the network is safe at a given point if none are found. Though computing minimal manipulations is not practical, in discrete spaces by iterating over increasingly
\emph{refined} manipulations
we are able to rule out the existence of adversarial examples in the region.
This contrasts with \emph{partial} exploration according to, e.g.,~\cite{fross-pract,constraints}; for comparison see Section~\ref{sec:related}.

\section{The Verification Framework}
In this section we propose a novel framework for automated verification of safety of classification decisions, which is based on search for an adversarial misclassification within a given region. The key distinctive distinctive features of our framework compared to existing work are: a \emph{guarantee} that a misclassification is found if it exists; the propagation of 
the analysis \emph{layer by layer}; and working with \emph{hidden} layers, in addition to input and output layers. 
Since we reduce verification to a search for adversarial examples, we can achieve safety \emph{verification} (if no misclassifications are found for all layers) or \emph{falsification} (in which case the adversarial examples can be used to fine-tune the network or shown to a human tester).

\subsection{\bf Layer-by-Layer Analysis}
\begin{figure}
\center
\includegraphics[width=10cm,height=5cm]{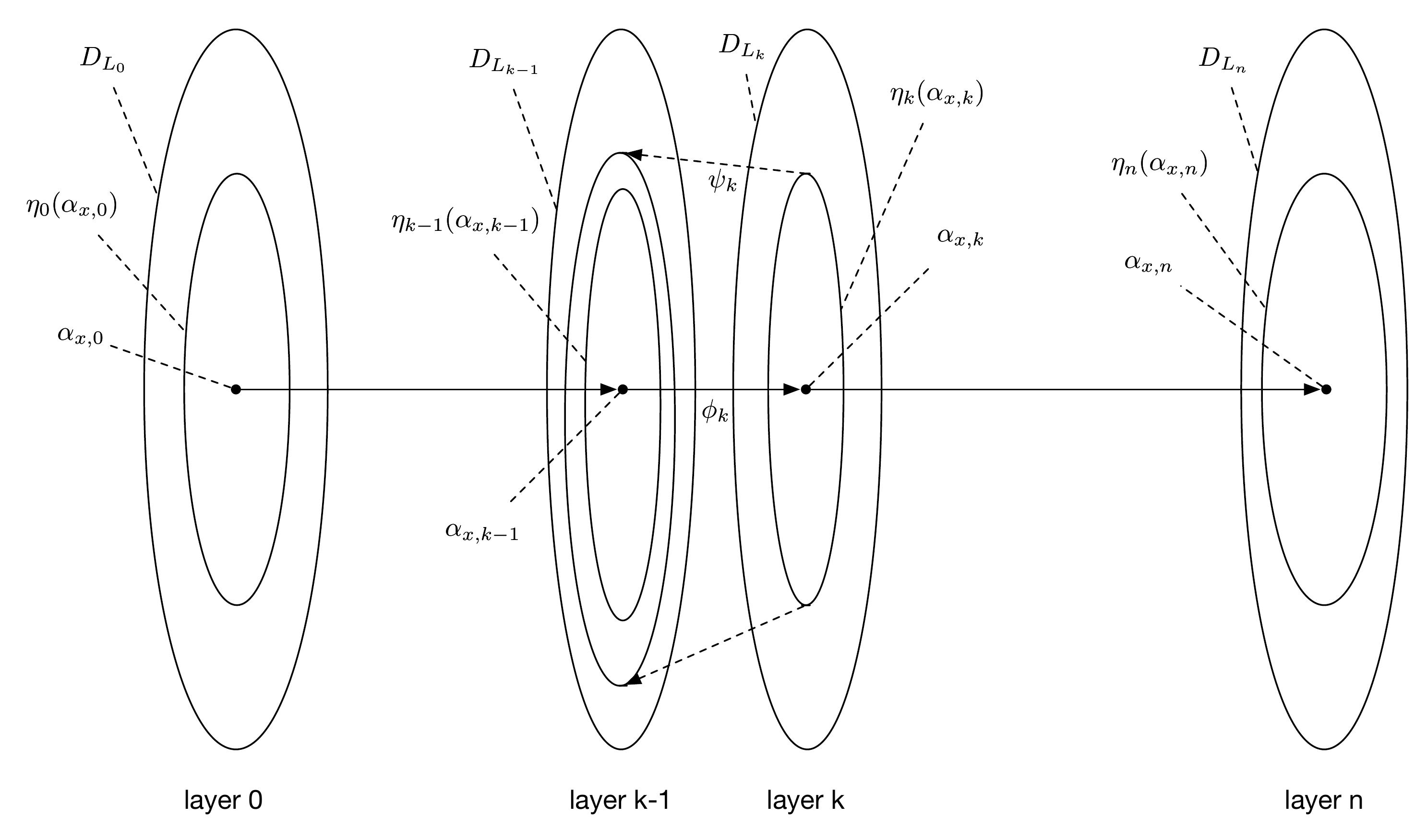}
\caption{Layer by layer analysis according to Definition~\ref{def:ek}}
\label{fig:definition3}
\end{figure}

We first consider how to propagate the analysis layer by layer, which will involve \emph{refining} manipulations through the hidden layers.
To facilitate such analysis, in addition to the activation function $\phi_k:D_{L_{k-1}}\rightarrow D_{L_{k}}$ we also require a mapping $\psi_k:D_{L_{k}}\rightarrow D_{L_{k-1}}$ in the opposite direction,
to represent how a manipulated activation of layer $L_k$ affects the activations of layer $L_{k-1}$. We can simply take $\psi_k$ as the inverse function of $\phi_k$.
In order to propagate safety of regions $\eta_k(\activation_{x,k}$) at a point $x$ into deeper layers, we assume the existence of functions $\eta_k$ that map activations to regions, and
 impose the following restrictions on the functions $\phi_k$ and $\psi_k$, shown diagrammatically in Figure~\ref{fig:definition3}. \begin{definition}\label{def:ek}
The functions $\{\eta_0,\eta_1,...,\eta_n\}$ and $\{\psi_1,...,\psi_n\}$ mapping activations to regions are such that
\begin{enumerate}
\item $\eta_k(\activation_{x,k}) \subseteq D_{L_k}$, for $k=0,...,n$,
\item $\activation_{x,k}\in \eta_k(\activation_{x,k})$, for $k=0,...,n$, and
\item $\eta_{k-1}(\activation_{i,k-1})\subseteq \psi_k(\eta_k(\activation_{x,k}))$ for all $k=1,...,n$.
\end{enumerate}
\end{definition}
Intuitively, the first two conditions state that each function $\eta_k$ assigns a region around the activation $\activation_{x,k}$, and the last condition that mapping the region $\eta_k$ from layer $L_k$ to $L_{k-1}$ via $\psi_k$ should cover the region $\eta_{k-1}$.
The aim is to compute functions $\eta_{k+1},...,\eta_n$ based on $\eta_k$ and the neural network.

The size and complexity of a deep neural network generally means that determining whether a given set $\manipulationset_k$ of manipulations is minimal is intractable. To partially counter this, we define a \emph{refinement} relation between safety wrt manipulations for consecutive layers in the sense that $N,\eta_k,\manipulationset_k\models x$ is a refinement of $N,\eta_{k-1},\manipulationset_{k-1}\models x$ if all manipulations $\manipulation_{k-1}$ in $\manipulationset_{k-1}$ are refined by a sequence of manipulations $\manipulation_{k}$ from the set $\manipulationset_{k}$. Therefore, although we cannot theoretically confirm the minimality of $\manipulationset_k$, they are refined layer by layer and, in discrete settings, this process can be bounded from below by the unit step.
Moreover, we can work gradually from a specific layer inwards until an adversarial example is found, finishing processing when reaching the output layer. 

\begin{figure}
\includegraphics[width=12cm,height=2cm]{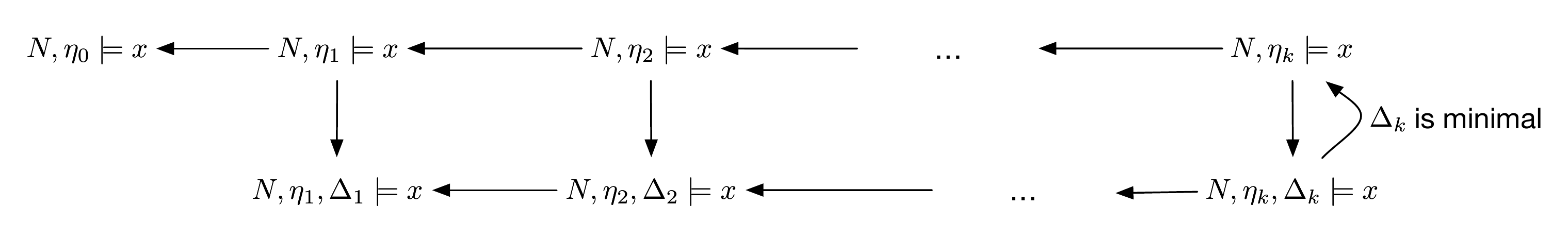}
\caption{Refinement framework}
\label{fig:refinement}
\end{figure}

The refinement framework is given in Figure~\ref{fig:refinement}.
The arrows represent the implication relations between the safety notions and are labelled with conditions if needed. The goal of the refinements is to find a chain of implications to justify $N,\eta_0\models x$. The fact that $N,\eta_k\models x$ implies $N,\eta_{k-1}\models x$ is due to the constraints in Definition~\ref{def:ek} when $\psi_k=\phi_k^{-1}$. The fact that $N,\eta_k\models x$ implies $N,\eta_{k}, \manipulationset_k\models x$ follows from Theorem~\ref{thm:feature}. The implication from $N,\eta_{k}, \manipulationset_k\models x$ to  $N,\eta_k\models x$ under the condition that $\manipulationset_k$ is minimal is due to Theorem~\ref{thm:enhanced}.

We now define the notion of \emph{refinability} of manipulations between layers. Intuitively, a manipulation in layer $L_{k-1}$ is refinable in layer $L_k$ if there exists a sequence of manipulations in layer $L_k$ that implements the manipulation in layer $L_{k-1}$.
\begin{definition}\label{def:layerRefinement}
A manipulation $\manipulation_{k-1}(\activation_{y,k-1})$ is \emph{refinable} in layer $L_k$ if there exist activations $\activation_{x_0,k},...,\activation_{x_j,k}\in D_{L_k}$ and valid manipulations $\manipulation_k^1\in \legal(\activation_{x_0,k}),...,\manipulation_k^j\in \legal(\activation_{x_{j-1},k})$ such that $\activation_{y,k}=\activation_{x_0,k}$, $\manipulation_{k-1,k}(\activation_{y,k-1})=\activation_{x_j,k}$, and $\activation_{x_t,k}=\manipulation_k^t(\activation_{x_{t-1},k})$ for $1\leq t\leq j$.
Given a neural network $N$ and an input $x$, the manipulations $\manipulationset_k$ are a \emph{refinement by layer} of $\eta_{k-1}, \manipulationset_{k-1}$ and $\eta_k$ if,
for all $\activation_{y,k-1}\in \eta_{k-1}(\activation_{z,k-1}) $, all its valid manipulations $\manipulation_{k-1}(\activation_{y,k-1})$ are refinable in layer $L_k$.
\end{definition}

We have the following theorem stating that the refinement of safety notions is implied by the ``refinement by layer'' relation.
\begin{theorem}\label{thm:gradually}
Assume a neural network $N$ and an input $x$. For all layers $k\geq 1$, if manipulations $\manipulationset_k$ are refinement by layer of $\eta_{k-1}, \manipulationset_{k-1}$ and $\eta_k$, then we have that
$N,\eta_k,\manipulationset_k\models x$ implies $N,\eta_{k-1},\manipulationset_{k-1}\models x$.
\end{theorem}

We note that any adversarial example of safety wrt manipulations $N,\eta_k,\manipulationset_k\models x$ is also an adversarial example for general safety $N,\eta_k\models x$.
However, an adversarial example $\activation_{x,k}$ for $N,\eta_k\models x$ at layer $k$ needs to be checked to see if it is an adversarial example of $N,\eta_0\models x$, i.e. for the input layer. Recall that $Pre_{k'}(\activation_{x,k})$ is not necessarily unique. This is equivalent to checking the emptiness of $Pre_{0}(\activation_{x,k})\cap \eta_0(\activation_{x,0})$. If we start the analysis with a hidden layer $k>0$ and there is no specification for $\eta_0$, we can instead consider checking the emptiness of $\{\activation_{y,0}\in Pre_{0}(\activation_{x,k})~|~\activation_{y,n} \neq \activation_{x,n}\}$.

\subsection{The Verification Method}\label{sec:impl}

We summarise the theory developed thus far as a search-based recursive verification procedure given below.
The method is parameterised by the region $\eta_k$ around a given point and a family of manipulations $\manipulationset_k$. The
manipulations are specified by the user for the classification problem at hand, or alternatively can be selected automatically, as described in Section~\ref{sec:selection}. 
The vector norm to identify the region can also be specified by the user and can vary by layer. The method can start in any layer, with analysis propagated into deeper layers, and terminates when a misclassification is found. If an adversarial example is found by manipulating a hidden layer, it can be mapped back to the input layer, see Section~\ref{pre-image}.

\begin{algorithm}\label{alg:main}
Given a neural network $N$ and an input $x$, recursively perform the following steps, starting from some layer $l\geq 0$. Let $k\geq l$ be the current layer under consideration.
\begin{enumerate}
\item determine a region $\eta_k$ such that if $k>l$ then $\eta_k$ and $\eta_{k-1}$ 
satisfy Definition~\ref{def:ek}; 
\item determine a manipulation set $\manipulationset_k$ such that if $k> l$ then $\manipulationset_k$ is a refinement by layer of $\eta_{k-1},\manipulationset_{k-1}$ and $\eta_k$ according to Definition~\ref{def:layerRefinement}; 
\item verify whether $N,\eta_k,\manipulationset_k\models x$,
\begin{enumerate}
\item if $N,\eta_k,\manipulationset_k\models x$ then
\begin{enumerate}
\item report that $N$ is safe at $x$  with respect to $\eta_k(\activation_{x,k})$ and $\manipulationset_k$, and
\item continue to layer $k+1$;
\end{enumerate}
\item if $N,\eta_k,\manipulationset_k\not\models x$, then report an adversarial example.
\end{enumerate}
\end{enumerate}
\end{algorithm}

We implement Algorithm~\ref{alg:main} by utilising satisfiability modulo theory (SMT) solvers. The SMT problem is a decision problem for logical formulas with respect to combinations of background theories expressed in classical first-order logic with equality. 
For checking refinement by layer, we use the theory of linear real arithmetic with existential and universal quantifiers, and for verification within a layer (0-variation)
we use the same theory but without universal quantification.
The details of the encoding and the approach taken to compute the regions and manipulations are included in Section~\ref{sec:selection}.
To enable practical verification of deep neural networks, we employ a number of heuristics described in the remainder of this section.

\subsection{Feature Decomposition and Discovery}\label{sec:features}

While Theorem~\ref{thm:feature} and \ref{thm:enhanced} provide a {\em finite} way to verify safety of neural network classification decisions,  
the high-dimensionality of the region $\eta_k(\activation_{x,k})$ can make any computational approach impractical. We therefore use the concept of a {\em feature} to partition the region $\eta_k(\activation_{x,k})$ into a set of features, and exploit their independence and low-dimensionality.
This allows us to work with state-of-the-art networks that have hundreds, and even thousands, of dimensions.  

Intuitively, a feature defines for each point in the high-dimensional space $D_{L_k}$ the most explicit salient feature it has, e.g., the red-coloured frame of a street sign in Figure~\ref{fig:streetsign}. 
Formally, for each layer $L_k$, a feature function $f_k:D_{L_k}\rightarrow \powerset{D_{L_k}}$ assigns a small region for each activation $\activation_{x,k}$ in the space $D_{L_k}$, where $ \powerset{D_{L_k}}$ is the set of subspaces of $D_{L_k}$. The region $f_k(\activation_{x,k})$ may have lower dimension than that of $D_k$. 
It has been argued, in e.g.~\cite{CISZ2008} for natural images, that natural data, for example natural images and sound, forms a high-dimensional manifold, which embeds tangled manifolds to represent their features. Feature manifolds usually have lower dimension than the data manifold, and  a classification algorithm is to separate a set of tangled manifolds.
By assuming that the appearance of features is independent, we can manipulate them one by one regardless of the manipulation order, and thus reduce the
problem of size $O(2^{d_1+...+d_m})$ into a %
set of smaller problems of size $O(2^{d_1}),...,O(2^{d_m})$.

The analysis of activations in hidden layers, as performed by our method, provides an opportunity to {\em discover the features  automatically}.
Moreover, defining the feature $f_k$ on each activation as a single region corresponding to a specific feature  is without loss of generality: although an activation may include multiple features, the independence relation between features suggests the existence of a total relation between these features. The function $f_k$ essentially defines for each activation one particular feature, subject to certain criteria such as explicit knowledge, but features can also be explored in parallel.

\newcommand{\partition}{\pi}

Every feature $f_k(\activation_{y,k})$ is identified by a pre-specified number $dims_{k,f}$ of dimensions. 
Let
$dims_k(f_k(\activation_{y,k}))$  be the set of dimensions selected according to some heuristic. Then we have that
\begin{equation}
f_k(\activation_{y,k})(p)= \left\{
\begin{array}{ll}
\eta_k(\activation_{x,k})(p), & \text{ if }p\in dims_k(f_k(\activation_{y,k}))\\
\lbrack\activation_{y,k}(p),\activation_{y,k}(p)\rbrack & \text{ otherwise.}
\end{array}
\right.
\label{equ:feature}
\end{equation}
Moreover, we need a set of features  to partition the region $\eta_k(\activation_{x,k})$ as follows.
\begin{definition}\label{def:partition}
A set $\{f_1,...,f_m\}$ of regions is a partition of $\eta_k(\activation_{x,k})$, written as $\partition(\eta_k(\activation_{x,k}))$, if $dims_{k,f}(f_i)\cap dims_{k,f}(f_j)=\emptyset$ for $i,j\in \{1,...,m\}$ and $\eta_k(\activation_{x,k})= \times_{i=1}^m f_i$.
\end{definition}

Given such a partition $\partition(\eta_k(\activation_{x,k}))$, we define a function $\activations(x,k)$ by
\begin{equation}
\label{equ:acts}
\activations(x,k) = \{ \activation_{y,k} \in x ~|~x \in \partition(\eta_k(\activation_{x,k}))\}
\end{equation}
which contains one point for each feature.
Then, we reduce the checking of 0-variation of a region $\eta_k(\activation_{x,k})$ to the following problems:
\begin{itemize}
\item checking whether the points in $\activations(x,k)$ have the same class as $\activation_{x,k}$, and
\item checking the 0-variation of all features in $\partition(\eta_k(\activation_{x,k}))$.
\end{itemize}

In the above procedure, the checking of points in $\activations(x,k)$ can be conducted either  by  following a pre-specified sequential order (\emph{single-path} search) or by exhaustively searching all possible orders (\emph{multi-path} search).  In Section~\ref{sec:experiments} we demonstrate that single-path search according to the prominence of features can enable us to find adversarial examples, while multi-path search may find other examples whose distance to the original input image is smaller.

\subsection{Selection of Regions and Manipulations}\label{sec:selection}

The procedure summarised in Algorithm~\ref{alg:main} is typically invoked for a given image in the input layer, but, providing insight about hidden layers is available, it can start from any layer $L_l$ in the network. The selection of regions can be automated, as described below.

For the first layer to be considered, i.e., $k=l$, the region $\eta_k(\activation_{x,k})$ is defined by first selecting the subset of $dims_k$ dimensions from $P_k$ whose activation values are furthest away from the average activation value of the layer\footnote{We also considered other approaches, including computing derivatives up to several layers, but for the experiments we conduct they are less effective.}. Intuitively, the knowledge represented by these activations is  more explicit than the knowledge represented by the other dimensions, and manipulations over more explicit knowledge are more likely to result in a class change.
Let $avg_k=(\sum_{p\in P_k}\activation_{x,k}(p))/n_k$ be the average activation value of layer $L_k$. We let  $dims_k(\eta_k(\activation_{x,k}))$ be the first $dims_k$ dimensions $p\in P_k$ with the greatest values $|\activation_{x,k}(p)-avg|$ among all dimensions, and then define
\begin{equation}\label{equ:region}
\eta_k(\activation_{x,k})=\times_{p\in dims_k(\eta_k(\activation_{x,k}))}[\activation_{x,k}(p)-s_p*m_p, \activation_{x,k}(p)+s_p*m_p]
\end{equation}
i.e., a $dims_k$-polytope containing the activation $\activation_{x,k}$, where $s_p$ represents a small span and $m_p$ represents the number of such spans. Let $V_k=\{s_p, m_p~|~p\in dims_k(\eta_k(\activation_{x,k}))\}$ be a set of variables.

Let $d$ be a function mapping from $dims_k(\eta_k(\activation_{x,k}))$ to $\{-1,0,+1\}$ such that $\{d(p)\neq 0~|~p\in dims_k(\eta_k(\activation_{x,k}))\} \neq \emptyset$, and $D(dims_k(\eta_k(\activation_{x,k})))$ be the set of such functions.
Let a manipulation $\manipulation_k^d$ be
\begin{equation}\label{equ:manipulation}
\manipulation_k^d(\activation_{y,k})(p) = \left\{
\begin{array}{ll}
 \activation_{y,k}(p)-s_p & \text{    if }d(p) = -1  \\
\activation_{y,k}(p) & \text{    if }d(p) = 0  \\
\activation_{y,k}(p) + s_p & \text{    if }d(p) = +1  \\

\end{array}
\right.
\end{equation}
for activation $\activation_{y,k}\in \eta_k(\activation_{x,k})$. That is, each manipulation changes a subset of the dimensions by the span $s_p$, according to the directions given in $d$. The set  $\manipulationset_k$ is defined by collecting the set of all such manipulations. Based on this, we can define a set $\ladderset(\eta_k(\activation_{x,k}))$ of ladders, which is complete and covering.

\subsubsection{Determining the region $\eta_k$ according to $\eta_{k-1}$}

Given $\eta_{k-1}(\activation_{x,k-1})$ and the  functions $\phi_k$ and $\psi_k$, we can automatically determine a region $\eta_k(\activation_{x,k})$ satisfying Definition~\ref{def:ek} using the following approach.
According to the function $\phi_k$, the activation value $\activation_{x,k}(p)$ of perceptron $p\in P_k$ is computed from activation values of  a subset of perceptrons in $P_{k-1}$. We let $Vars(p) \subseteq P_{k-1}$ be such a set of perceptrons.
The selection of dimensions in $dims_k(\eta_k(\activation_{x,k}))$ depends on $dims_{k-1}(\eta_{k-1}(\activation_{x,k-1}))$ and $\phi_k$, by requiring that,
for every $p'\in dims_{k-1}(\eta_{k-1}(\activation_{x,k-1}))$, there is at least one dimension $p\in dims_k(\eta_k(\activation_{x,k}))$ such that $p'\in Vars(p)$.
We let
\begin{equation}
dims_k(\eta_k(\activation_{x,k})) = \{ \argmax_{p\in P_k}\{~|\activation_{x,k}(p)-avg_k|~|~p'\in Vars(p) \} ~|~ p' \in dims_{k-1}(\eta_{k-1}(\activation_{x,k-1}))\}
\end{equation}
Therefore, the restriction of Definition~\ref{def:ek} can be expressed with the following formula:
\begin{equation}\label{equ:def3}
\forall \activation_{y,k-1}\in \eta_k(\activation_{x,k-1}): \activation_{y,k-1} \in \psi_k(\eta_k(\activation_{x,k})).
\end{equation}
We omit the details of rewriting $\activation_{y,k-1}\in \eta_k(\activation_{x,k-1})$ and $\activation_{y,k-1} \in \psi_k(\eta_k(\activation_{x,k}))$ into Boolean expressions, which follow from standard techniques.
Note that this expression includes variables in $V_k,V_{k-1}$ and $\activation_{y,k-1}$. The variables in $V_{k-1}$ are fixed for a given $\eta_{k-1}(\activation_{x,k-1})$. Because such a region  $\eta_k(\activation_{x,k})$ always exists, a simple iterative procedure can be invoked to gradually increase the size of the region represented with variables in $V_k$ to eventually satisfy the expression.


\subsubsection{Determining the manipulation set $\manipulationset_k$ according to $\eta_k(\activation_{x,k})$, $\eta_{k-1}(\activation_{x,k-1})$, and $\manipulationset_{k-1}$}

The values of the variables $V_k$ obtained from the satisfiability of Eqn (\ref{equ:def3}) yield a definition of manipulations using Eqn (\ref{equ:manipulation}). However, the obtained values for span variables $s_p$ do not necessarily satisfy the ``refinement by layer'' relation as defined in Definition~\ref{def:layerRefinement}.
Therefore, we need to adapt the values for the variables $V_k$ while, at the same time, retaining the region $\eta_k(\activation_{x,k})$.
To do so, we could rewrite the constraint in Definition~\ref{def:layerRefinement} into a formula, which can then be solved by an SMT solver. But, in practice, we notice that such {\em precise} computations easily lead to overly small spans $s_p$, which in turn result in an unacceptable amount of computation needed to verify the relation $N,\eta_k,\manipulationset_k\models x$.

To reduce computational cost, we work with a weaker ``refinable in layer $L_k$'' notion, parameterised with respect to precision $\varepsilon$.
Given two activations $\activation_{y,k}$ and $\activation_{m,k}$, we use $dist(\activation_{y,k},\activation_{m,k})$ to represent their 
distance.

\begin{definition}\label{def:layerRefinementwithError}
A manipulation $\manipulation_{k-1}(\activation_{y,k-1})$ is \emph{refinable} in layer $L_k$ with precision $\varepsilon > 0$ if there exists a sequence of activations $\activation_{x_0,k},...,\activation_{x_j,k}\in D_{L_k}$ and valid manipulations $\manipulation_k^1\in \legal(\activation_{x_0,k}),...,\manipulation_k^d\in \legal(\activation_{x_{j-1},k})$ such that $\activation_{y,k}=\activation_{x_0,k}$, $\manipulation_{k-1,k}(\activation_{y,k-1})\in rec(\activation_{x_{j-1},k},\activation_{x_j,k})$, $dist(\activation_{x_{j-1},k},\activation_{x_j,k}) \leq \epsilon $, and $\activation_{x_t,k}=\manipulation_k^t(\activation_{x_{t-1},k})$ for $1\leq t\leq j$.
Given a neural network $N$ and an input $x$, the manipulations $\manipulationset_k$ are a \emph{refinement by layer} of $\eta_k, \eta_{k-1}, \manipulationset_{k-1}$ with precision $\varepsilon$ if,
for all $\activation_{y,k-1}\in \eta_{k-1}(\activation_{x,k-1}) $, all its legal manipulations $\manipulation_{k-1}(\activation_{y,k-1})$ are refinable in layer $L_k$ with precision $\varepsilon$.
\end{definition}

Comparing with Definition~\ref{def:layerRefinement}, the above definition  replaces  $\manipulation_{k-1,k}(\activation_{y,k-1})=\activation_{x_j,k}$ with $\manipulation_{k-1,k}(\activation_{y,k-1})\in rec(\activation_{x_{j-1},k},\activation_{x_j,k})$ and $dist(\activation_{x_{j-1},k},\activation_{x_j,k}) \leq \varepsilon $. Intuitively, instead of requiring a manipulation to reach the activation $\manipulation_{k-1,k}(\activation_{y,k-1})$ precisely, this definition allows for each $\manipulation_{k-1,k}(\activation_{y,k-1})$ to be within the hyper-rectangle $rec(\activation_{x_{j-1},k},\activation_{x_j,k})$.
To find suitable values for $V_k$ according to  the approximate ``refinement-by-layer'' relation, 
we use a variable $h$ to represent the maximal number of manipulations of layer $L_k$ used to express a manipulation in layer $k-1$. The value of $h$ (and variables $s_p$ and $n_p$ in $V_k$) are automatically adapted to ensure the satisfiability
 of  the following formula, which expresses the constraints of Definition~\ref{def:layerRefinementwithError}:
\begin{equation}\label{equ:manipulationset}
\begin{array}{l}
\forall \activation_{y,k-1}\in \eta_k(\activation_{x,k-1})\forall d\in D(dims_k(\eta_k(\activation_{x,k-1})))\forall \manipulation_{k-1}^d\in \legal_{k-1}(\activation_{y,k-1})
\\
\exists \activation_{y_0,k},...,\activation_{y_h,k} \in \eta_k(\activation_{x,k}): \activation_{y_0,k}=\activation_{y,k}\land  \bigwedge_{t=0}^{h-1}
\activation_{y_{t+1},k}=\manipulation_k^d(\activation_{y_t,k})\land 
\\
 \bigvee_{t=0}^{h-1}(\manipulation_{k-1,k}^d(\activation_{y,k})\in rec(\activation_{y_t,k},\activation_{y_{t+1},k}) \land dist(\activation_{y_{t},k},\activation_{y_{t+1},k}) \leq \varepsilon).
\end{array}
\end{equation}
It is noted that $s_p$ and $m_p$ for $p\in  dims_k(\eta_k(\activation_{x,k}))$ are employed when expressing $\manipulation_k^d$. The manipulation $\manipulation_k^d$ is obtained from $\manipulation_{k-1}^d$ by considering the corresponding relation between dimensions in $dims_k(\eta_k(\activation_{x,k}))$ and $dims_{k-1}(\eta_{k-1}(\activation_{x,k-1}))$.  

Adversarial examples shown in Figures \ref{fig:mnist}, \ref{fig:illustrative}, and \ref{fig:streetsign} were found using single-path search and automatic selection of regions and manipulations.

\subsection{Mapping Back to Input Layer}\label{pre-image}

When manipulating the hidden layers, we may need to map back an activation in layer $k$ to the input layer to obtain an input image that resulted in misclassification, which involves 
computation of $Pre_{0}(\activation_{y,k})$ described next.
To check the 0-variation of a region $\eta_k(\activation_{x,k})$, we need to compute $\mathrm{diff}_n(\activation_{x,n}, \activation_{y,n})$ for many points $\activation_{y,x}$ in $\eta_k(\activation_{x,k})$, where  $\mathrm{diff}_n:D_{L_n}\times D_{L_n}\rightarrow \{0,1\}$ is given by $\mathrm{diff}_n(\activation_{x,n}, \activation_{y,n}) = 0$ if $\activation_{x,n} = \activation_{y,n}$ and 1 otherwise. Because $\activation_{x,n}$ is known, we only need to compute $\activation_{y,n}$. We can compute $\activation_{y,n}$ by finding a point $\activation_{y,0}\in Pre_0(\activation_{y,k})$ and then using the neural network to predict the value $\activation_{y,n}$. It should be noted that, although $Pre_0(\activation_{y,k})$ may include more than one point, all points  have the same class, so any point in $Pre_0(\activation_{y,k})$ is sufficient for our purpose.

To compute $\activation_{y,0}$ from $\activation_{y,k}$, we use functions $\psi_k, \psi_{k-1},...,\psi_1$ and compute points $\activation_{y,k-1}, \activation_{y,k-2},...,\activation_{y,0}$ such that 
$$\activation_{y,j-1} = \psi_{j}(\activation_{y,j}) \land \activation_{y,j-1} \in \eta_{j-1}(\activation_{x,j-1})$$ 
for $1\leq j\leq k$. The computation relies on an SMT solver to encode the functions $\psi_k, \psi_{k-1},...,\psi_1$ if they are piecewise linear functions, and by taking the corresponding inverse functions directly if they are sigmoid functions. It is possible that, for some $1\leq j\leq k$, no point can be found by SMT solver, which means that the point $\activation_{y,k}$ does not have any corresponding point in the input layer. We can safely discard these points. The maxpooling function $\psi_j$ selects from every $m*m$ dimensions the maximal element for some $m>0$. The computation of the maxpooling layer $\psi_{j-1}$ is combined with the computation of the next layer $\psi_j$, that is, finding $\activation_{y,j-2}$ with the following expression
$$ \exists \activation_{x,j-1}: \activation_{y,j-2} = \psi_{j-1}(\psi_{j}(\activation_{y,j})) \land \activation_{y,j-1} \in \eta_{j-1}(\activation_{x,j-1}) \land \activation_{y,j-2} \in \eta_{j-2}(\activation_{x,j-2}) $$
This is to ensure that in the expression $\activation_{y,j-2} = \psi_{j-1}(\psi_{j}(\activation_{y,j}))$ we can reuse $m*m-1$ elements in $\activation_{x,j-2}$ and only need to replace the maximal element. 

Figures \ref{fig:mnist}, \ref{fig:illustrative}, and \ref{fig:streetsign} show images obtained by mapping back from the first hidden layer to the input layer.

\section{Experimental Results}\label{sec:experiments}

The proposed framework has been implemented as a software tool called DLV (Deep Learning Verification)~\cite{DLV}  written in Python, see Appendix of \cite{HKWW2016} for details of input parameters and how to use the tool. The SMT solver we employ is Z3~\cite{z3}, which has Python APIs. The neural networks are built from a widely-used neural networks library Keras~\cite{keras} with a deep learning package Theano~\cite{theano} as its backend. 

We validate DLV on a set of experiments performed for neural networks trained for classification based on a predefined multi-dimensional surface (small size networks), as well as image classification (medium size networks). These networks respectively use two representative types of layers: fully connected layers and convolutional layers. They may also use other types of layers, e.g., the ReLU layer, the pooling layer, the zero-padding layer, and the dropout layer.
The first three demonstrate the single-path search functionality on the Euclidean ($L^2$) norm, whereas the fourth (GTSRB) multi-path search for the $L^1$ and $L^2$ norms. 

The experiments are conducted on a MacBook Pro laptop, with 2.7 GHz Intel Core i5 CPU and 8 GB memory.

\paragraph{\bf Two-Dimensional Point Classification Network}

\begin{figure}
\center
\parbox{5cm}{
\includegraphics[width=5cm,height=5cm]{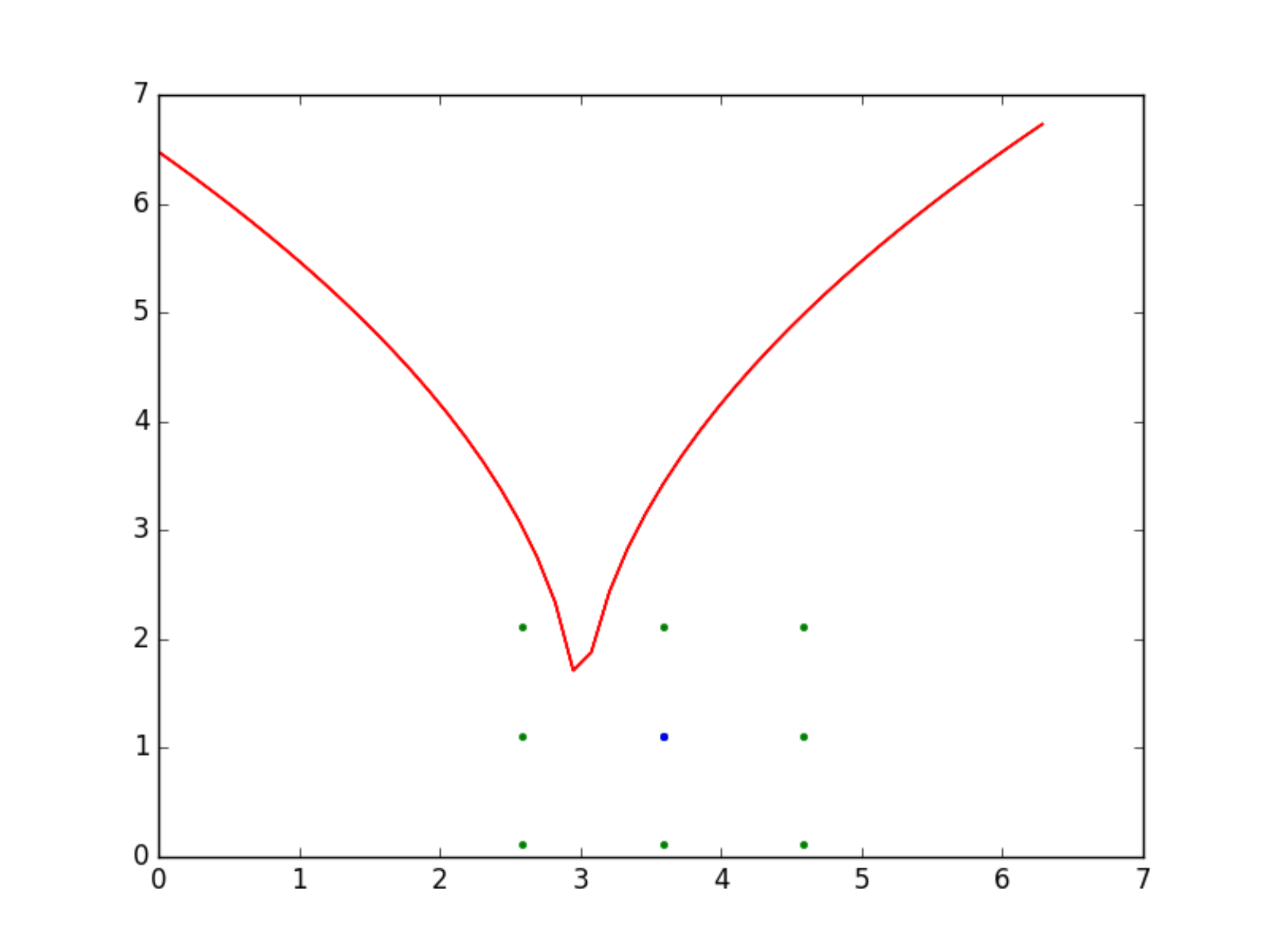}
\caption{Input layer\label{fig:curve1}}
}
\parbox{5cm}{
\includegraphics[width=5cm,height=5cm]{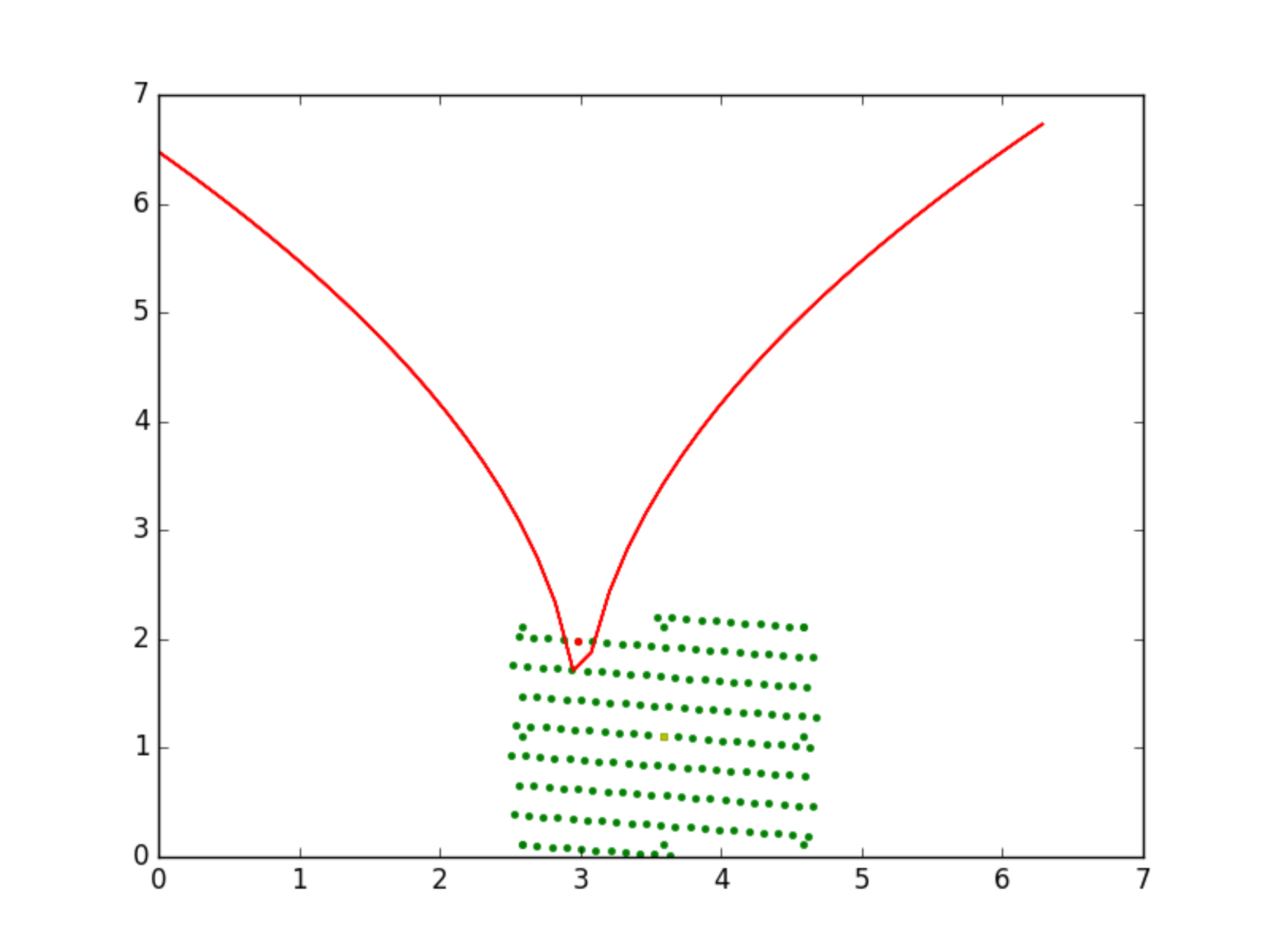}
\caption{First hidden layer\label{fig:curve2}}
}
\end{figure}

To demonstrate exhaustive verification facilitated by our framework, we consider a neural network trained for
classifying points above and below
a two-dimensional curve shown in red in Figure~\ref{fig:curve1} and Figure~\ref{fig:curve2}. The network has three fully-connected hidden layers with the ReLU activation function. The input layer has two perceptrons, every hidden layer has 20 perceptrons, and the output layer has two perceptrons. The network is trained with 5,000 points sampled from the provided two-dimensional space, and has an accuracy of more than 99\%.

For a given input $x=(3.59,1.11)$, we start from the input layer and define a region around this point by taking unit steps in both directions
$$\eta_0(\activation_{x,0})=[3.59-1.0,3.59+1.0]\times [1.11-1.0,1.11+1.0]=[2.59,4.59]\times [0.11,2.11]$$
The manipulation set $\manipulationset_0$ is shown in Figure~\ref{fig:curve1}: there are 9 points, of which the point in the middle represents the activation $\activation_{x,0}$ and the other 8 points represent the activations resulting from applying one of the manipulations in $\manipulationset_0$ on $\activation_{x,0}$. Note that, although there are class changes in the region $\eta_0(\activation_{x,0})$, the manipulation set $\manipulationset_0$ is not able to detect such changes. Therefore, we have that $N,\eta_0,\manipulationset_0\models x$.

Now consider layer $k=1$. To obtain the region $\eta_1(\activation_{x,1})$, the tool selects two dimensions $p_{1,17},p_{1,19}\in P_1$ in layer $L_1$ with indices 17 and 19
and computes
$$\eta_1(\activation_{x,1})=[\activation_{x,1}(p_{1,17})-3.6, \activation_{x,1}(p_{1,17})+3.6]\times [\activation_{x,1}(p_{1,19})-3.52, \activation_{x,1}(p_{1,19})+3.52]$$
The manipulation set $\manipulationset_1$, after mapping back to the input layer with function $\psi_1$, is given as Figure~\ref{fig:curve2}. Note that $\eta_1$ and $\eta_0$ satisfy Definition~\ref{def:ek}, and $\manipulationset_1$ is a refinement by layer of $\eta_0,\manipulationset_0$ and $\eta_1$. We can see that a class change can be detected (represented as the red coloured point). Therefore, we have that $N,\eta_1,\manipulationset_1\not\models x$.

\paragraph{\bf Image Classification Network for the MNIST Handwritten Image Dataset}

The well-known MNIST image dataset contains images of size $28\times 28$ and one channel and the network is trained with the source code given in~\cite{mnistNetwork}. The trained network is of medium size with 600,810 parameters, has an accuracy of more than 99\%, and is state-of-the-art. It has 12 layers, within which there are 2 convolutional layers, as well as layers such as ReLU, dropout, fully-connected layers and a softmax layer. The images are preprocessed to make the value of each pixel within the bound $[0,1]$.

\begin{figure}
\centering
\parbox{2.3cm}{
\includegraphics[width=1.1cm,height=1.1cm]{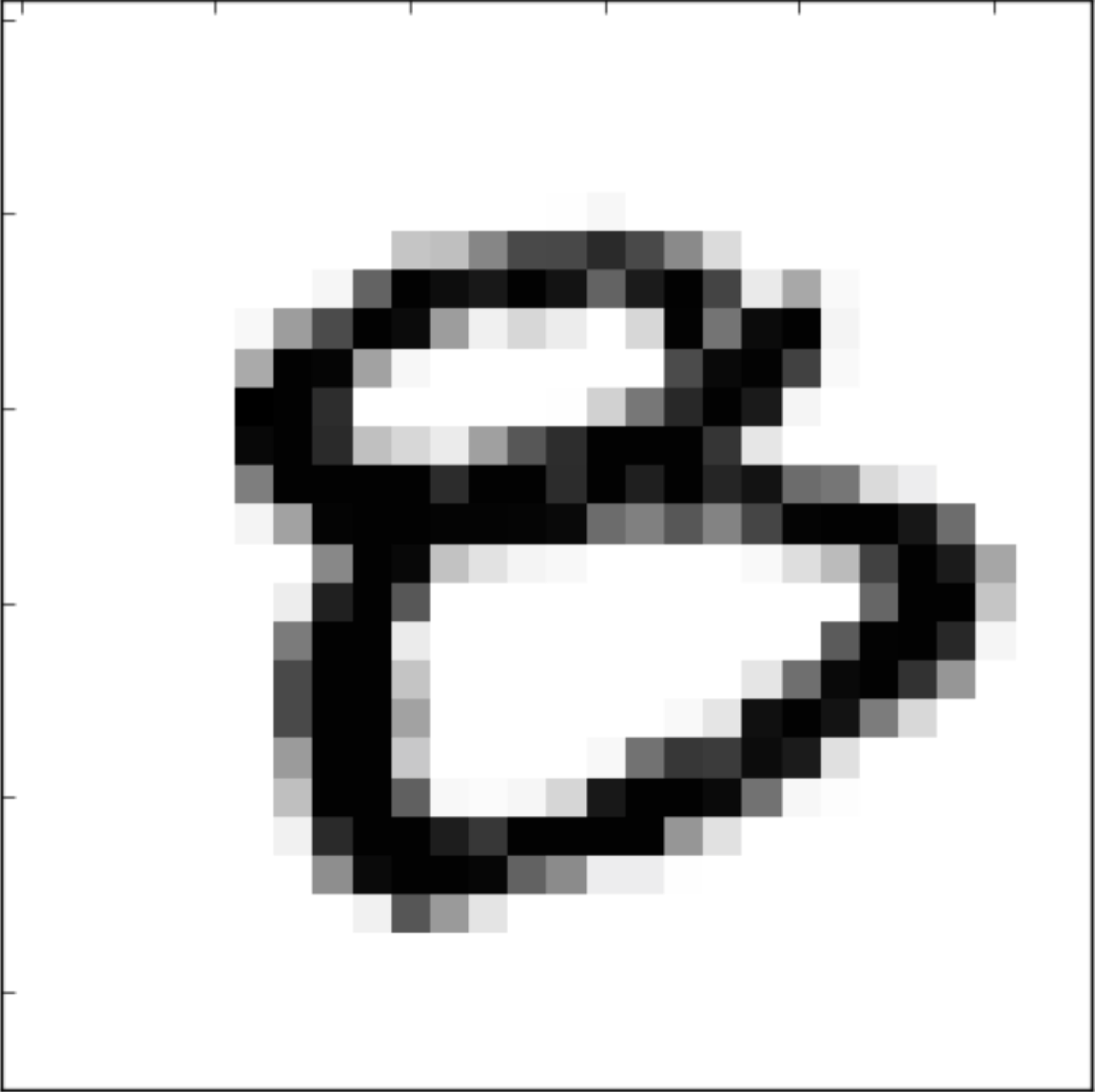}
\includegraphics[width=1.1cm,height=1.1cm]{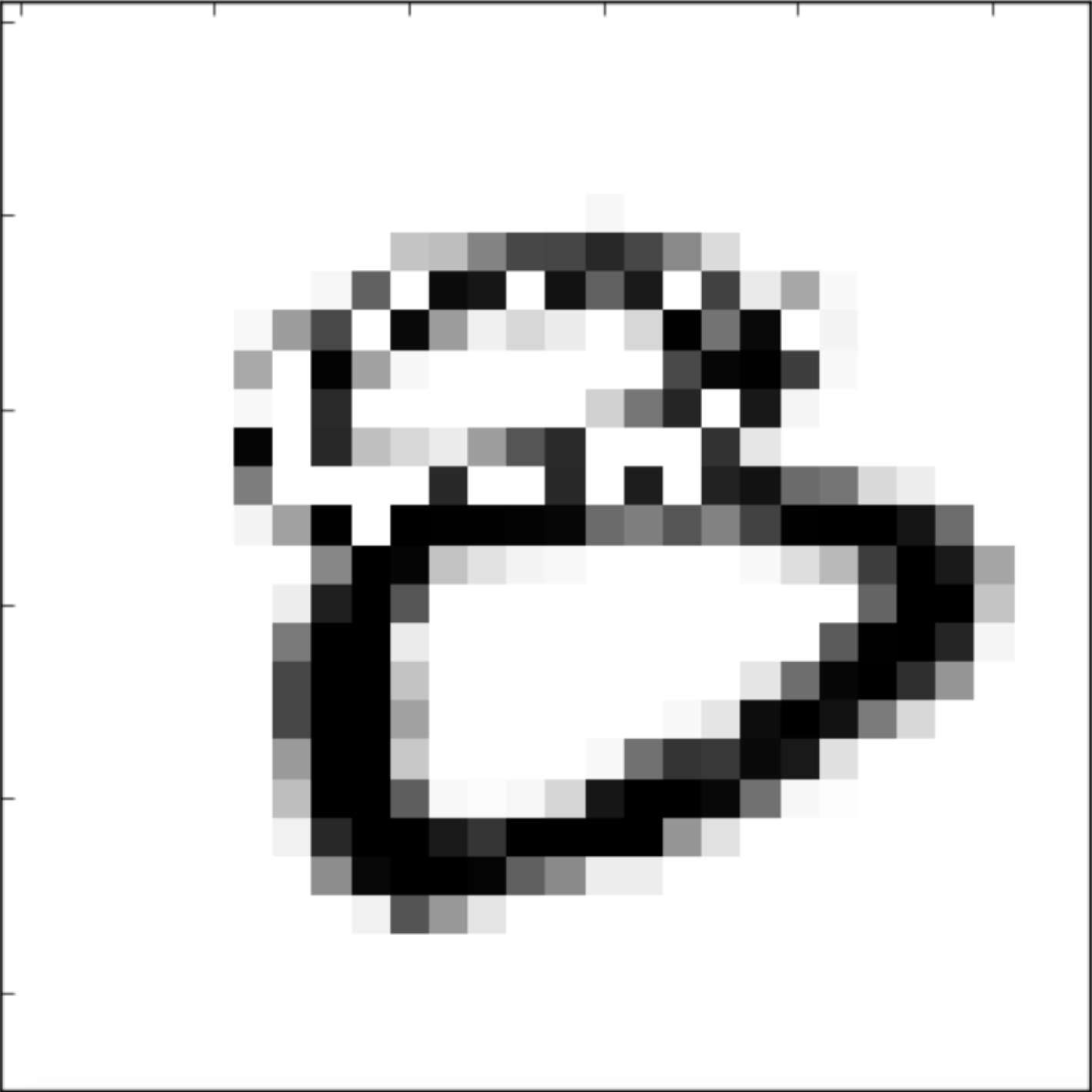}
\text{\hspace{0.8cm}8 to 0}
}
\parbox{2.3cm}{
\includegraphics[width=1.1cm,height=1.1cm]{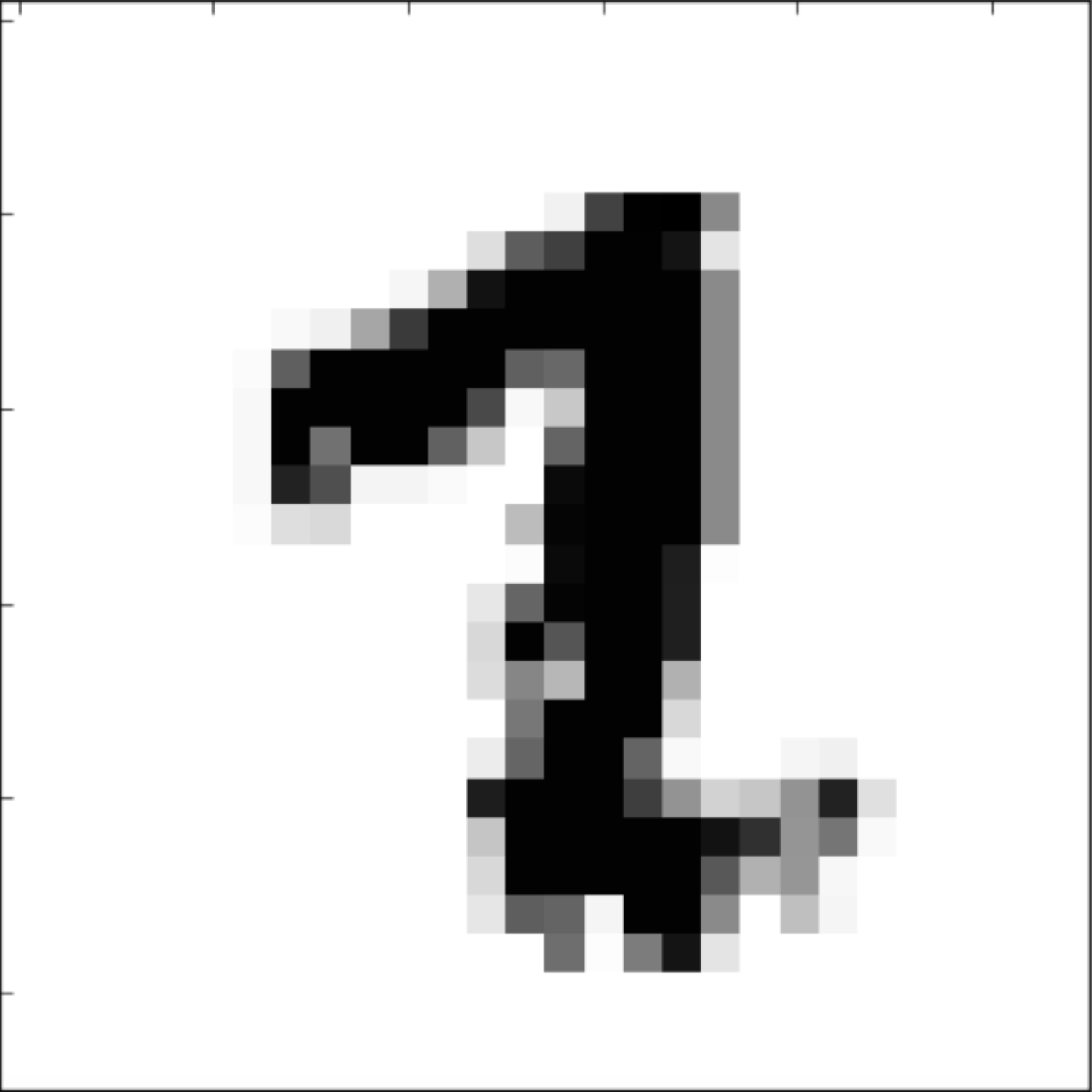}
\includegraphics[width=1.1cm,height=1.1cm]{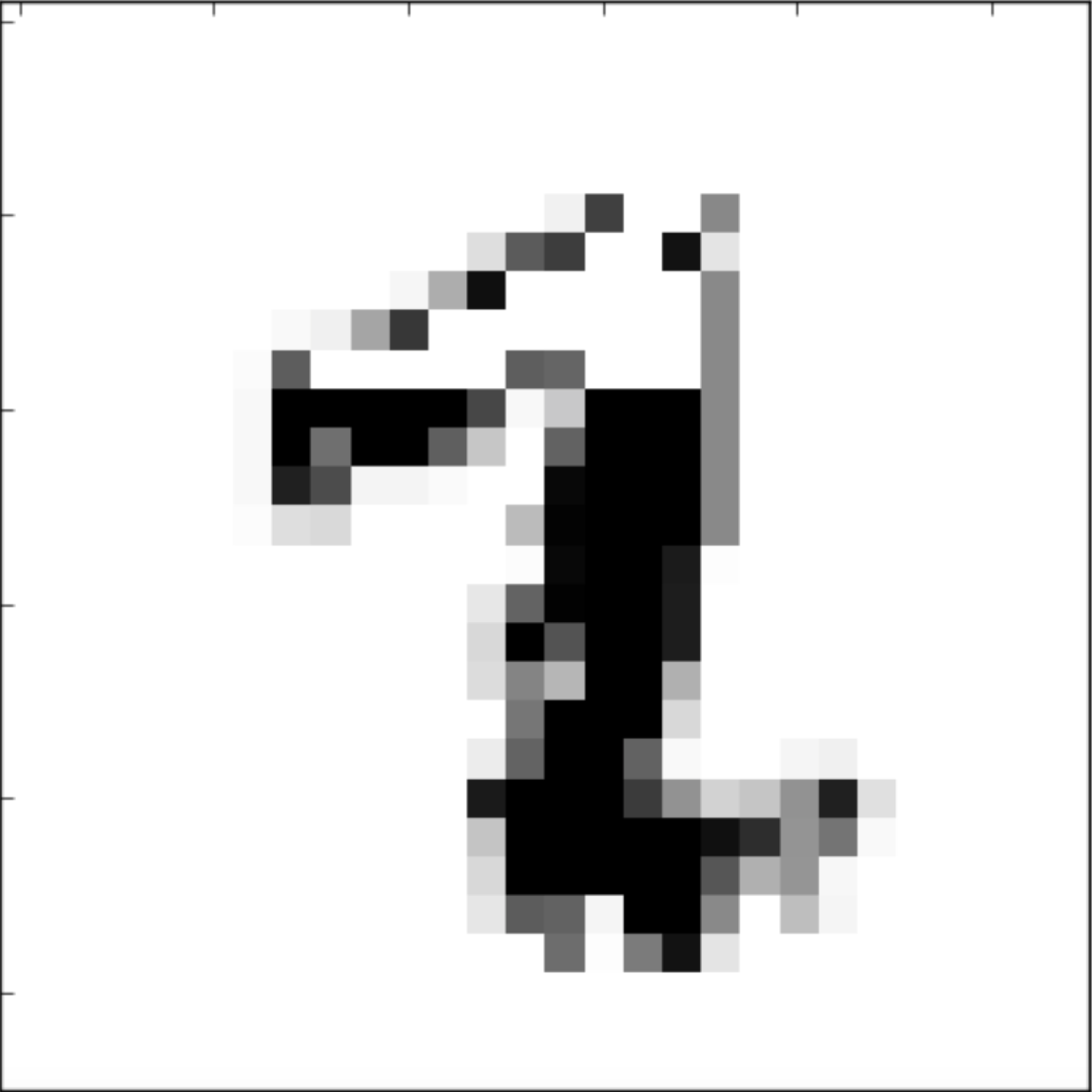}
\text{\hspace{0.8cm}2 to 1}
}
\parbox{2.3cm}{
\includegraphics[width=1.1cm,height=1.1cm]{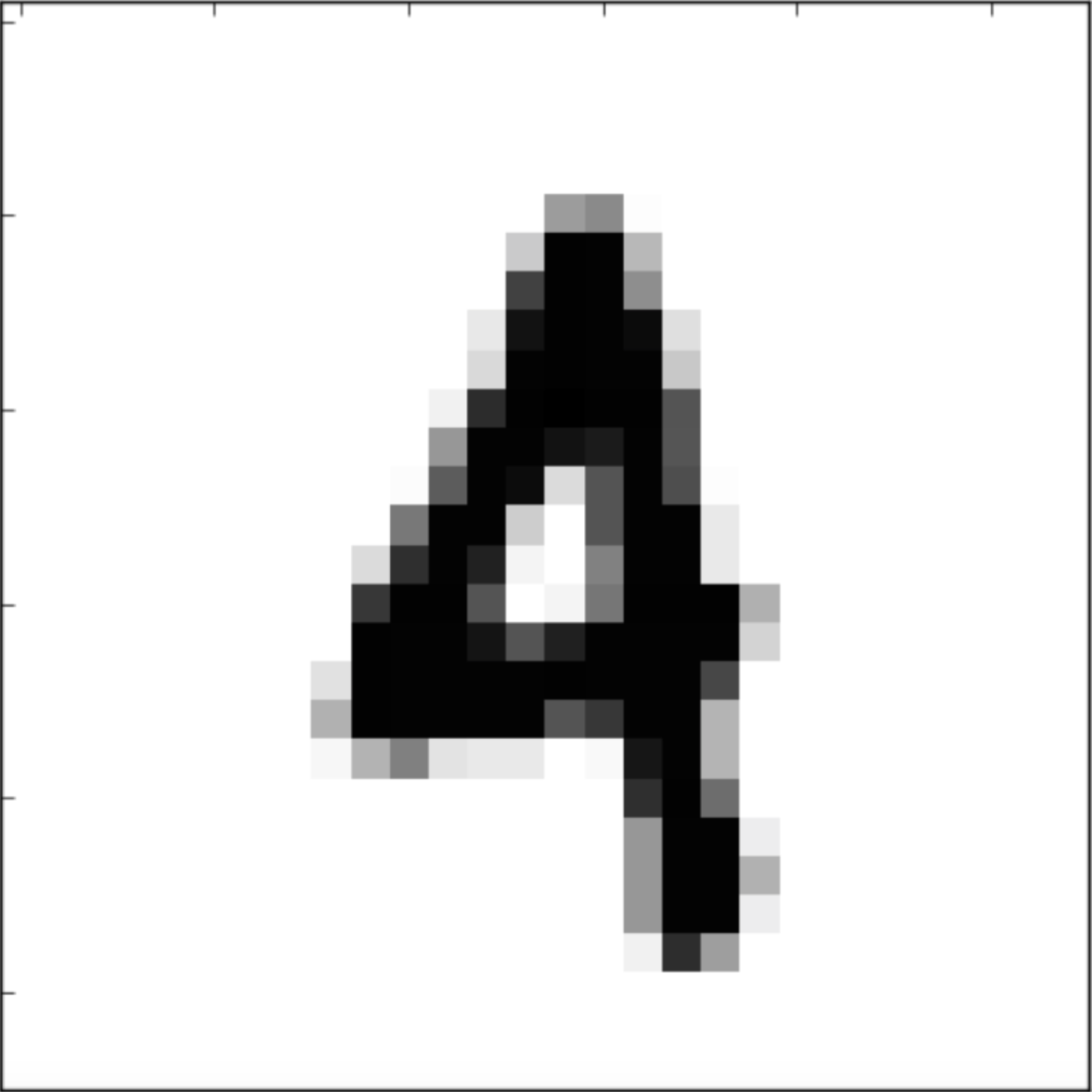}
\includegraphics[width=1.1cm,height=1.1cm]{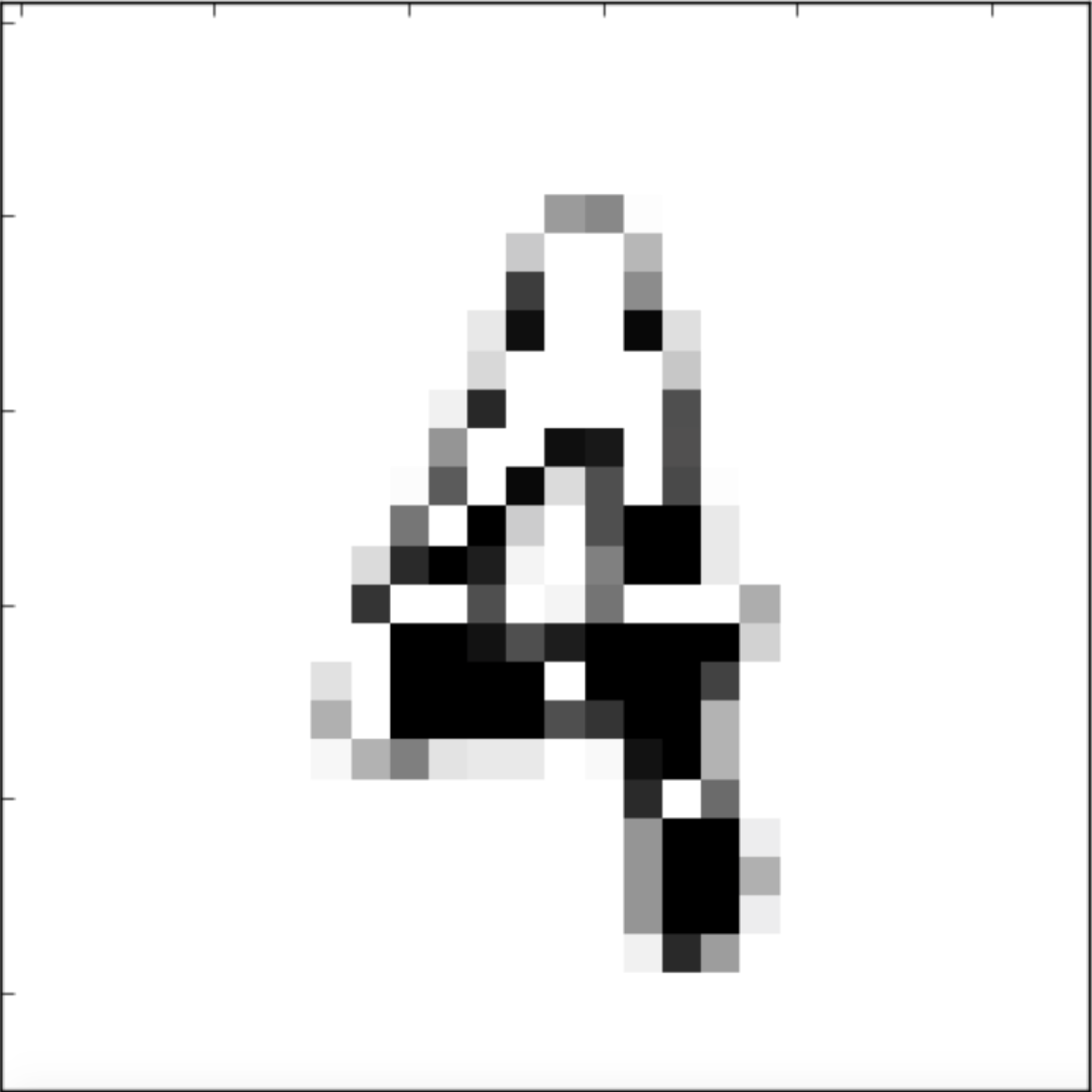}
\text{\hspace{0.8cm}4 to 2}
}
\parbox{2.3cm}{
\includegraphics[width=1.1cm,height=1.1cm]{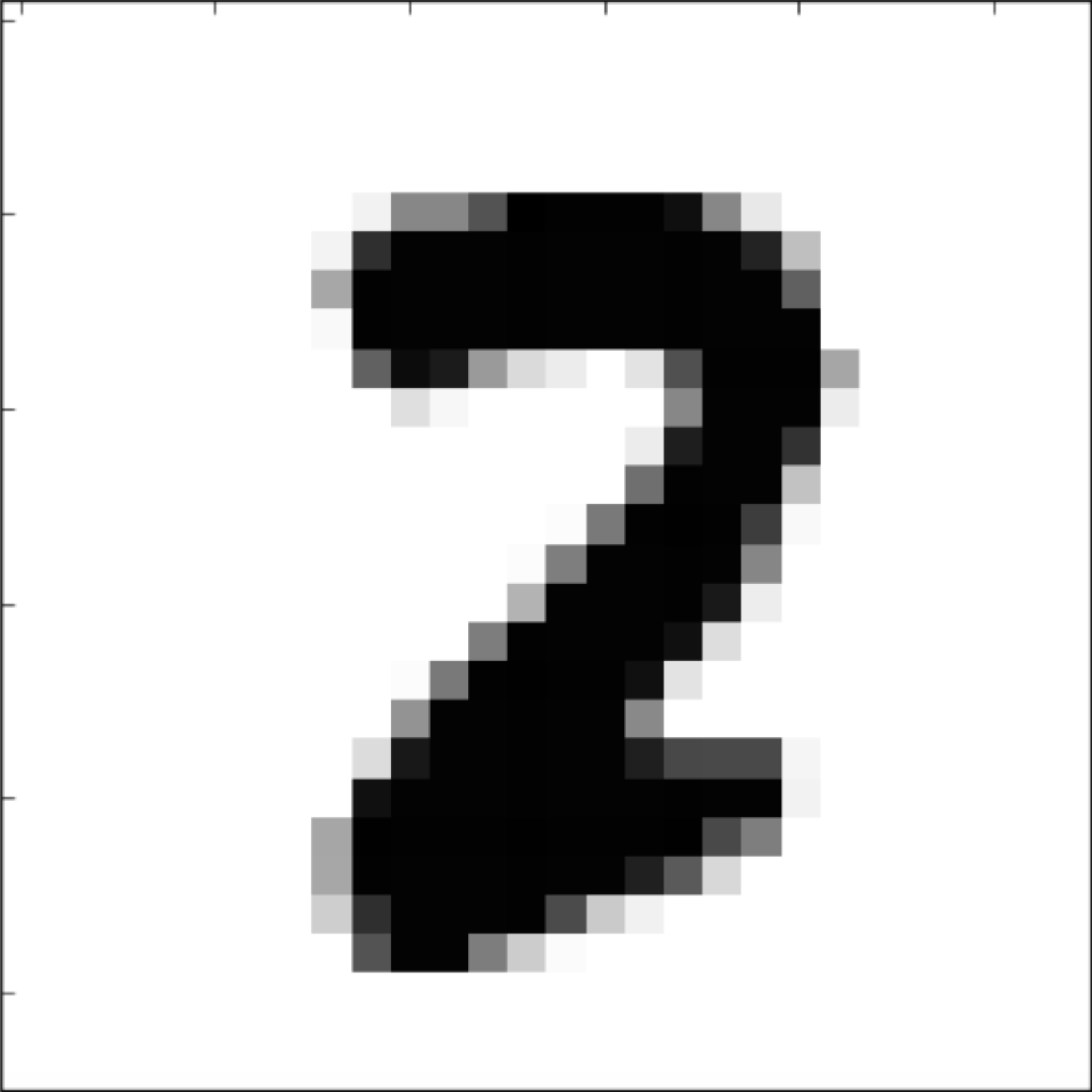}
\includegraphics[width=1.1cm,height=1.1cm]{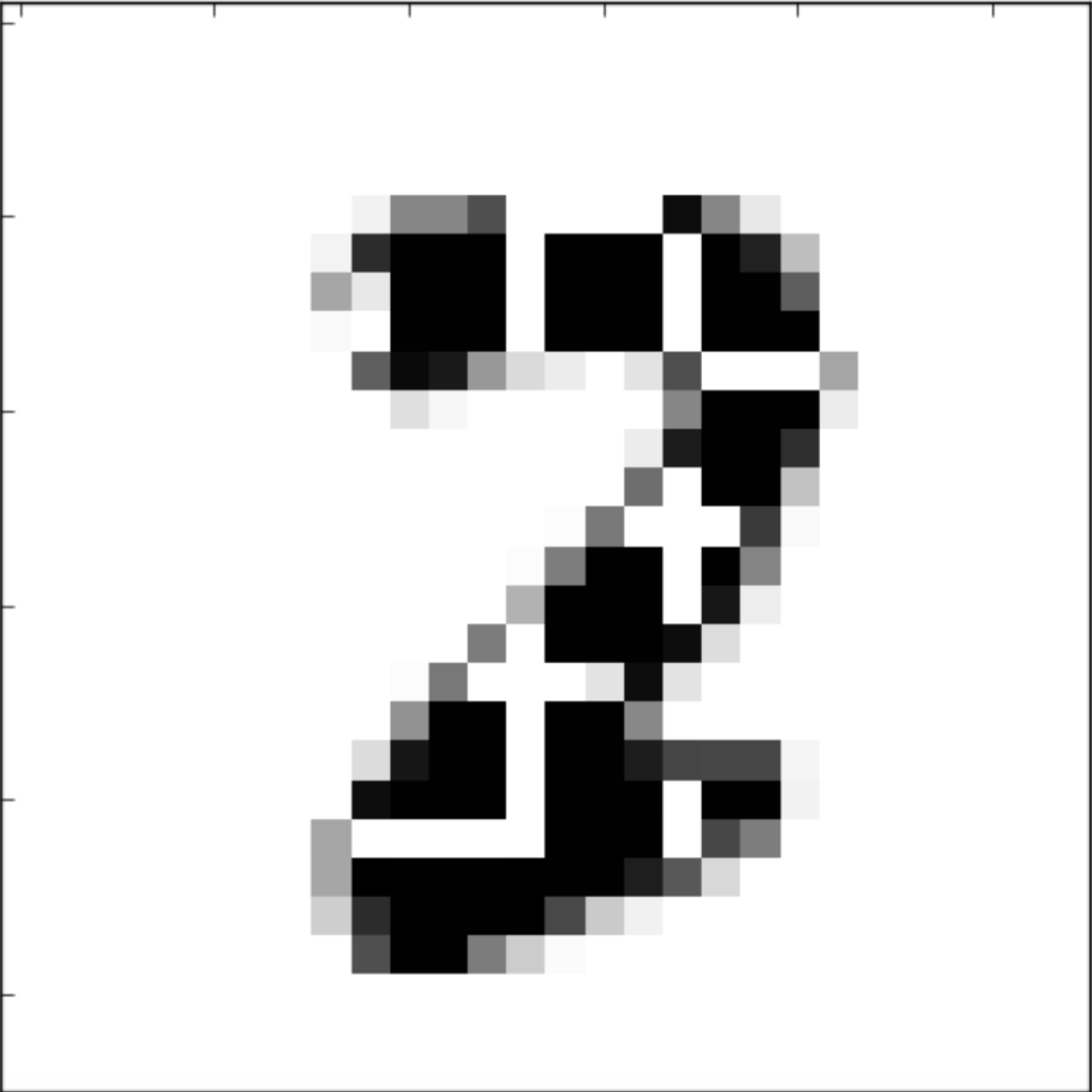}
\text{\hspace{0.8cm}2 to 3}
}
\parbox{2.3cm}{
\includegraphics[width=1.1cm,height=1.1cm]{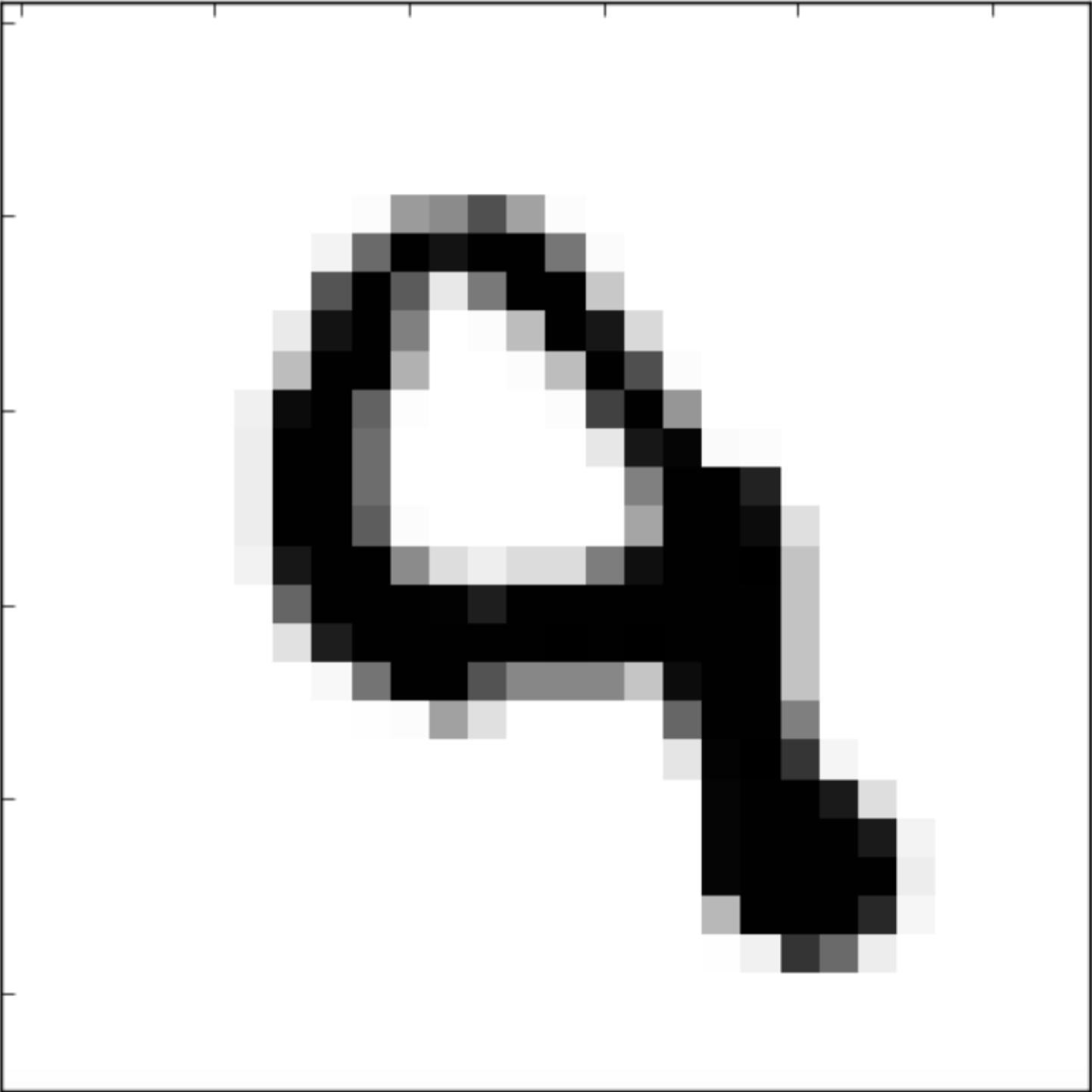}
\includegraphics[width=1.1cm,height=1.1cm]{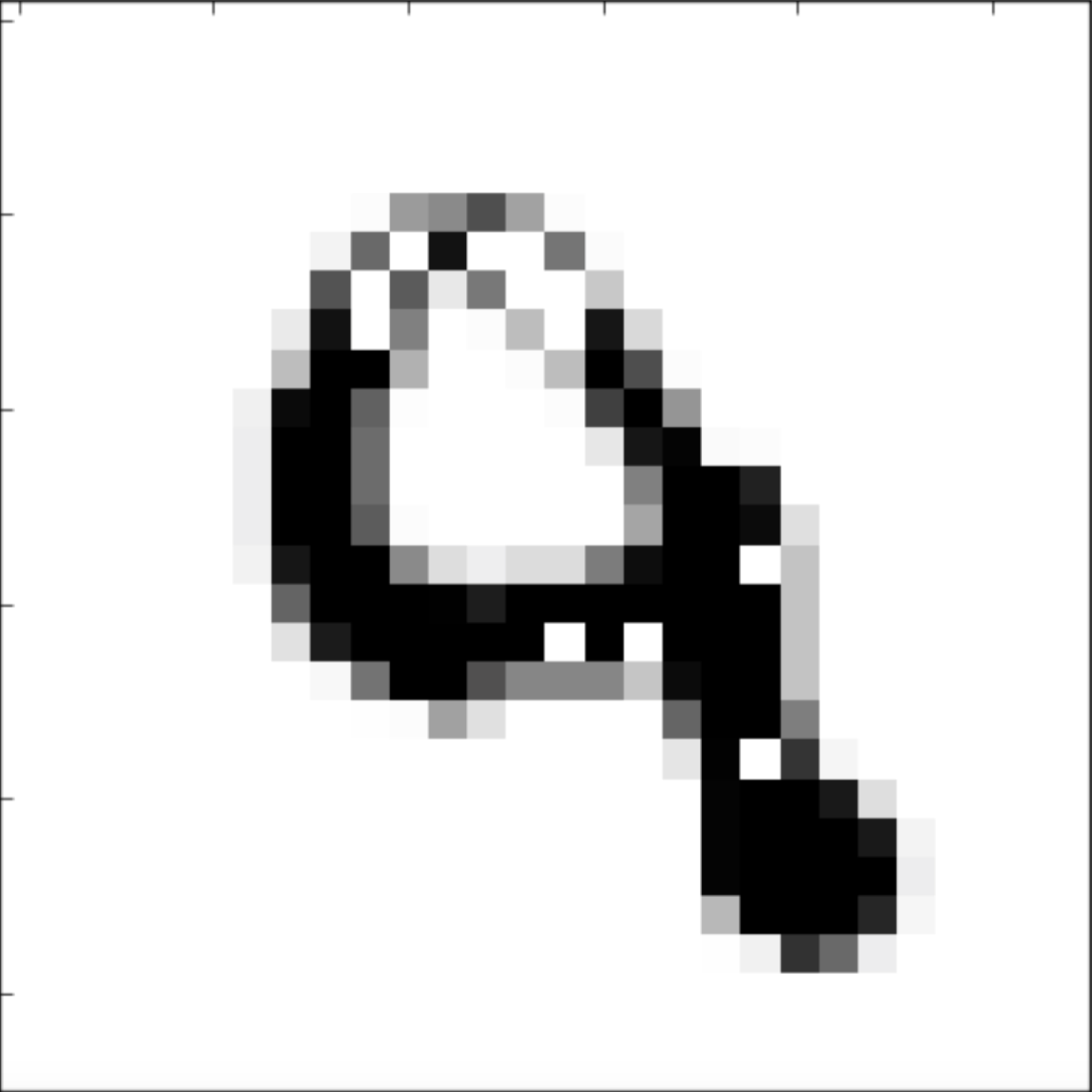}
\text{\hspace{0.8cm}9 to 4}
}\\
\vspace{4pt}
\parbox{2.3cm}{
\includegraphics[width=1.1cm,height=1.1cm]{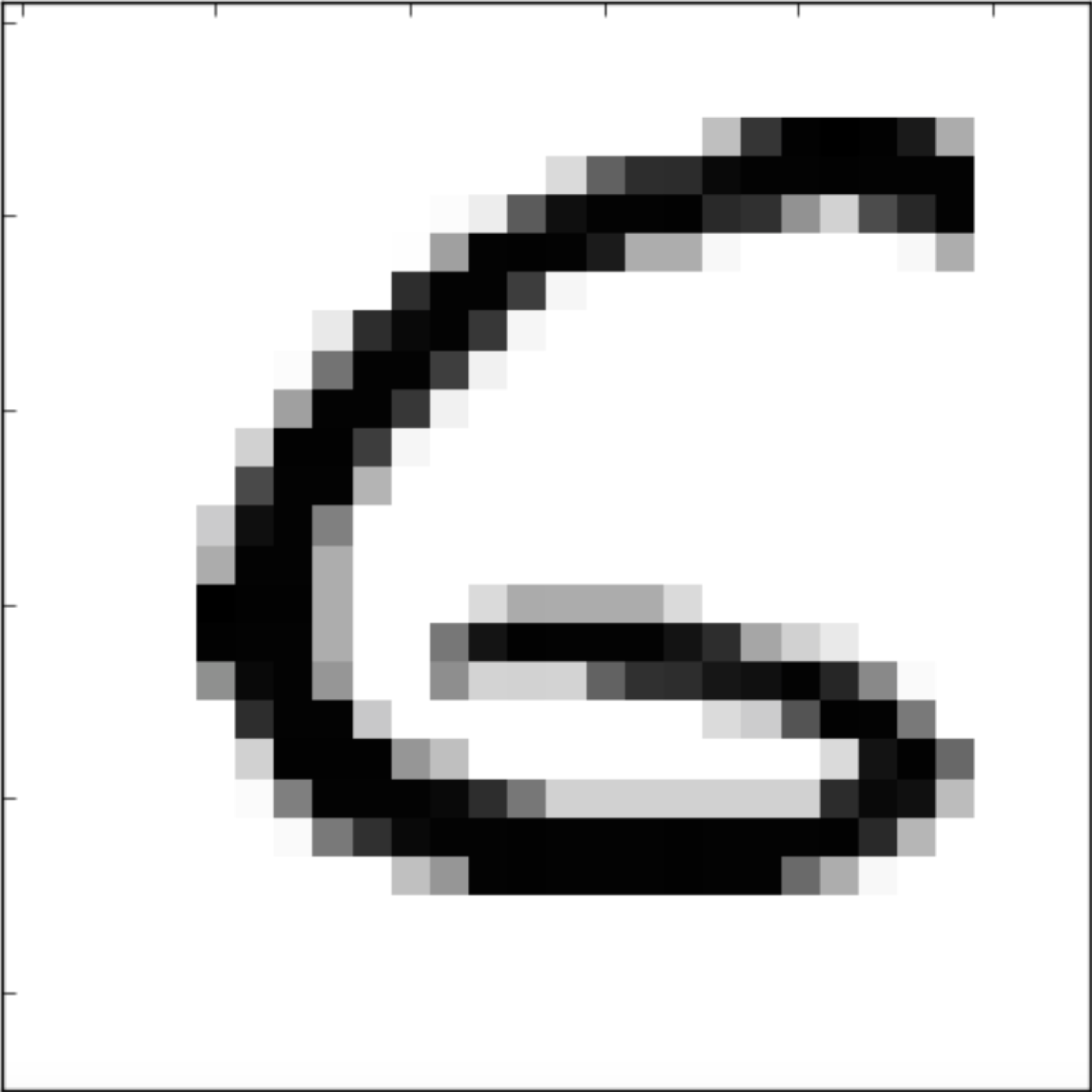}
\includegraphics[width=1.1cm,height=1.1cm]{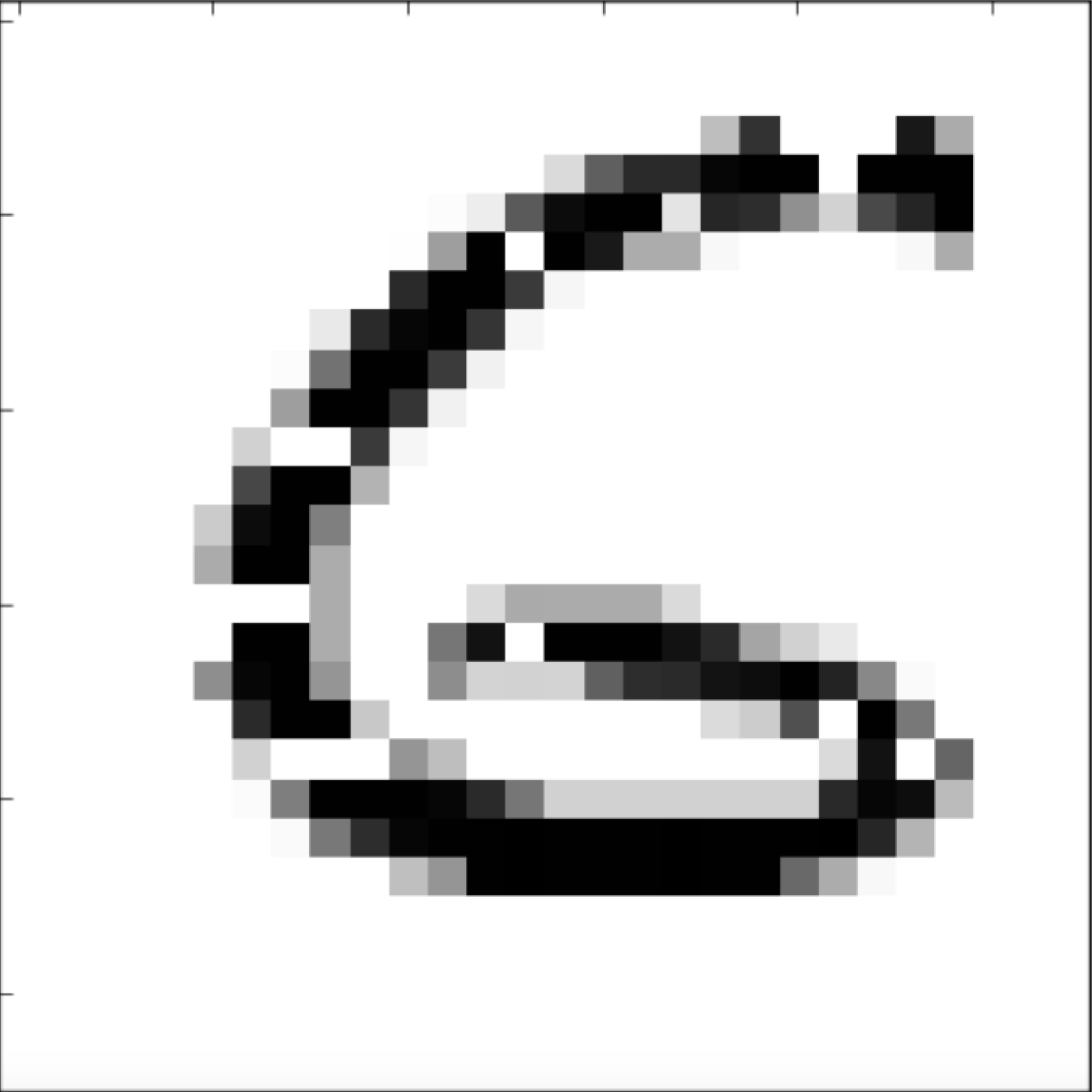}
\text{\hspace{0.8cm}6 to 5}
}
\parbox{2.3cm}{
\includegraphics[width=1.1cm,height=1.1cm]{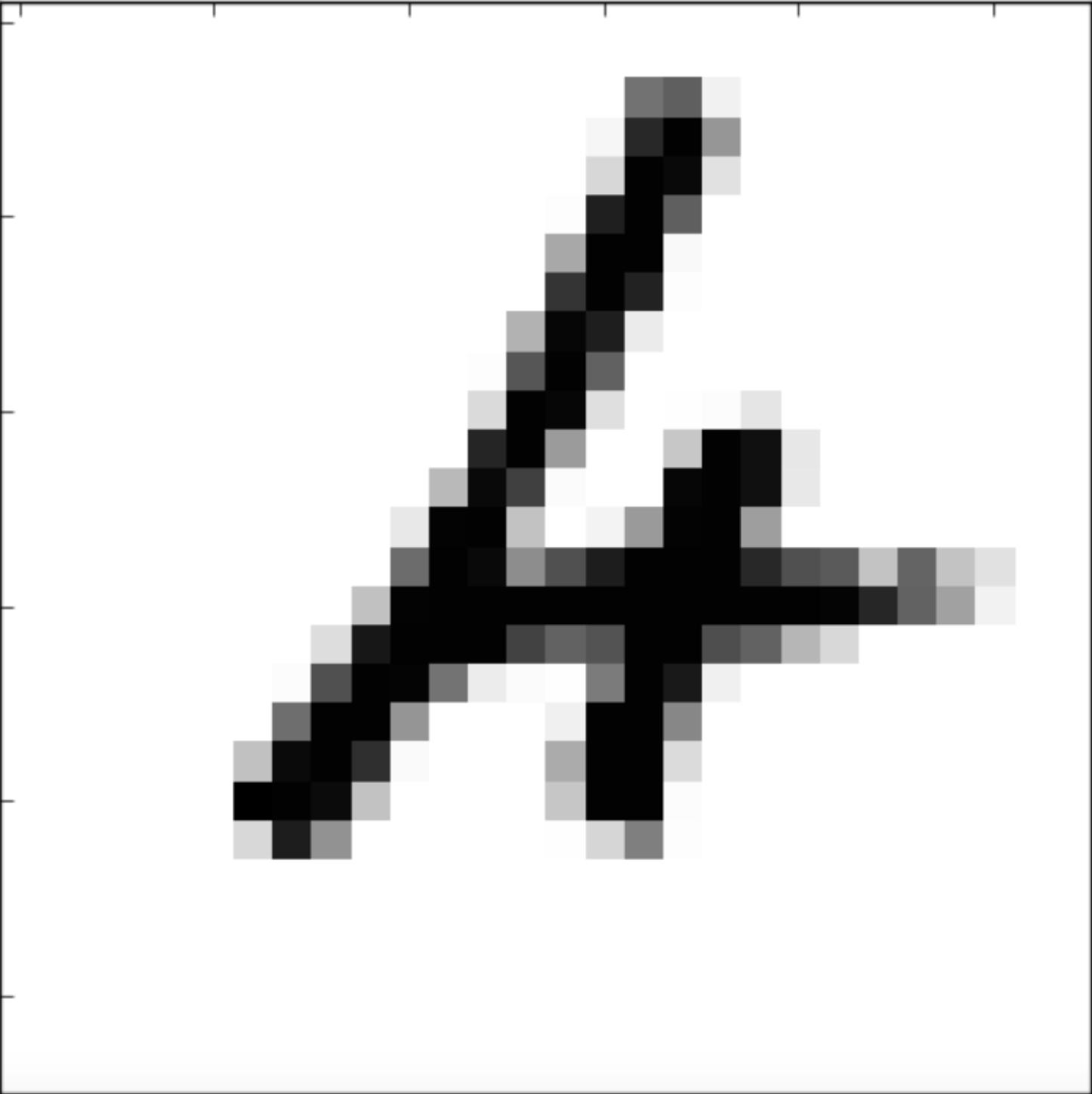}
\includegraphics[width=1.1cm,height=1.1cm]{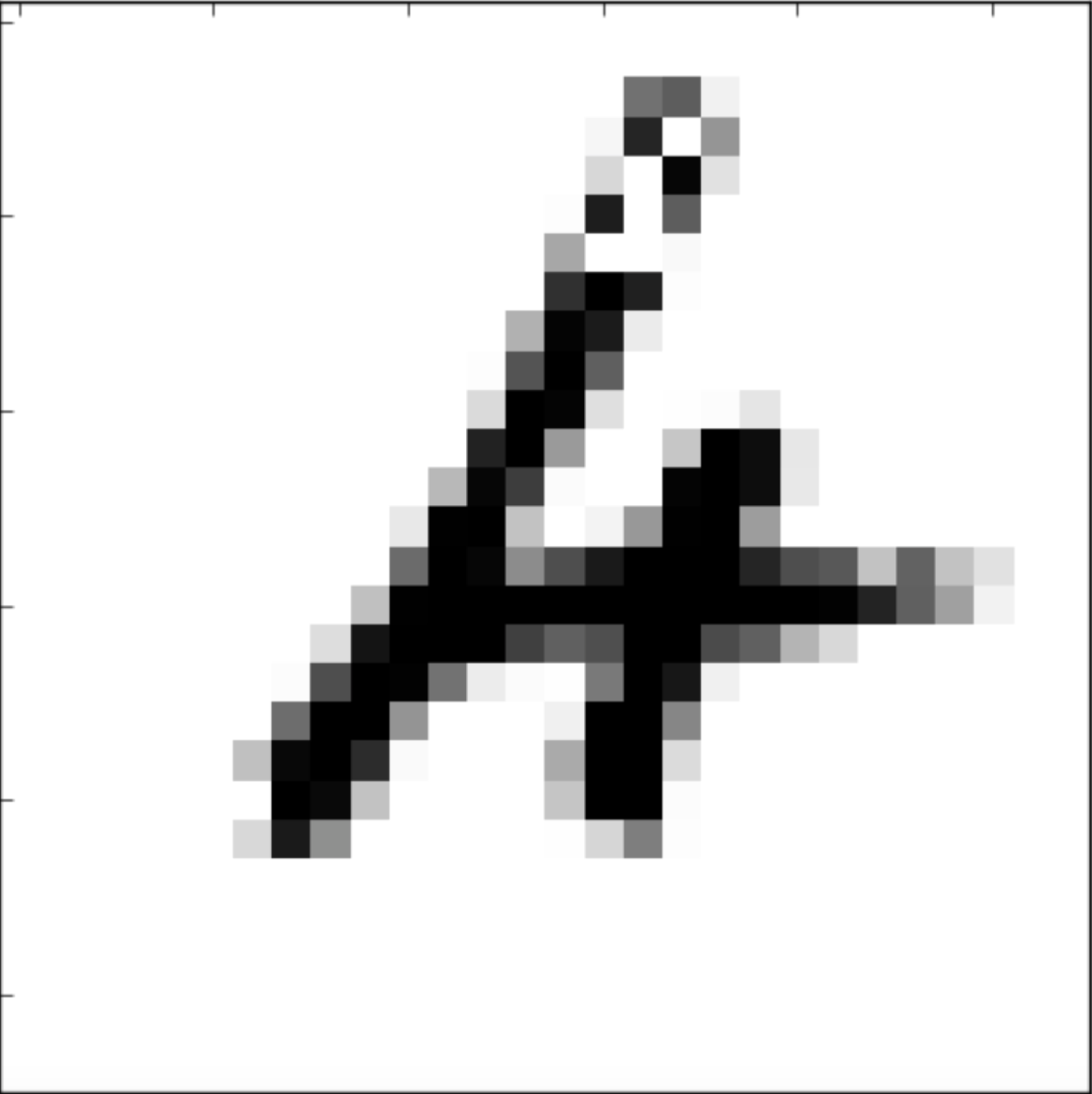}
\text{\hspace{0.8cm}4 to 6}
}
\parbox{2.3cm}{
\includegraphics[width=1.1cm,height=1.1cm]{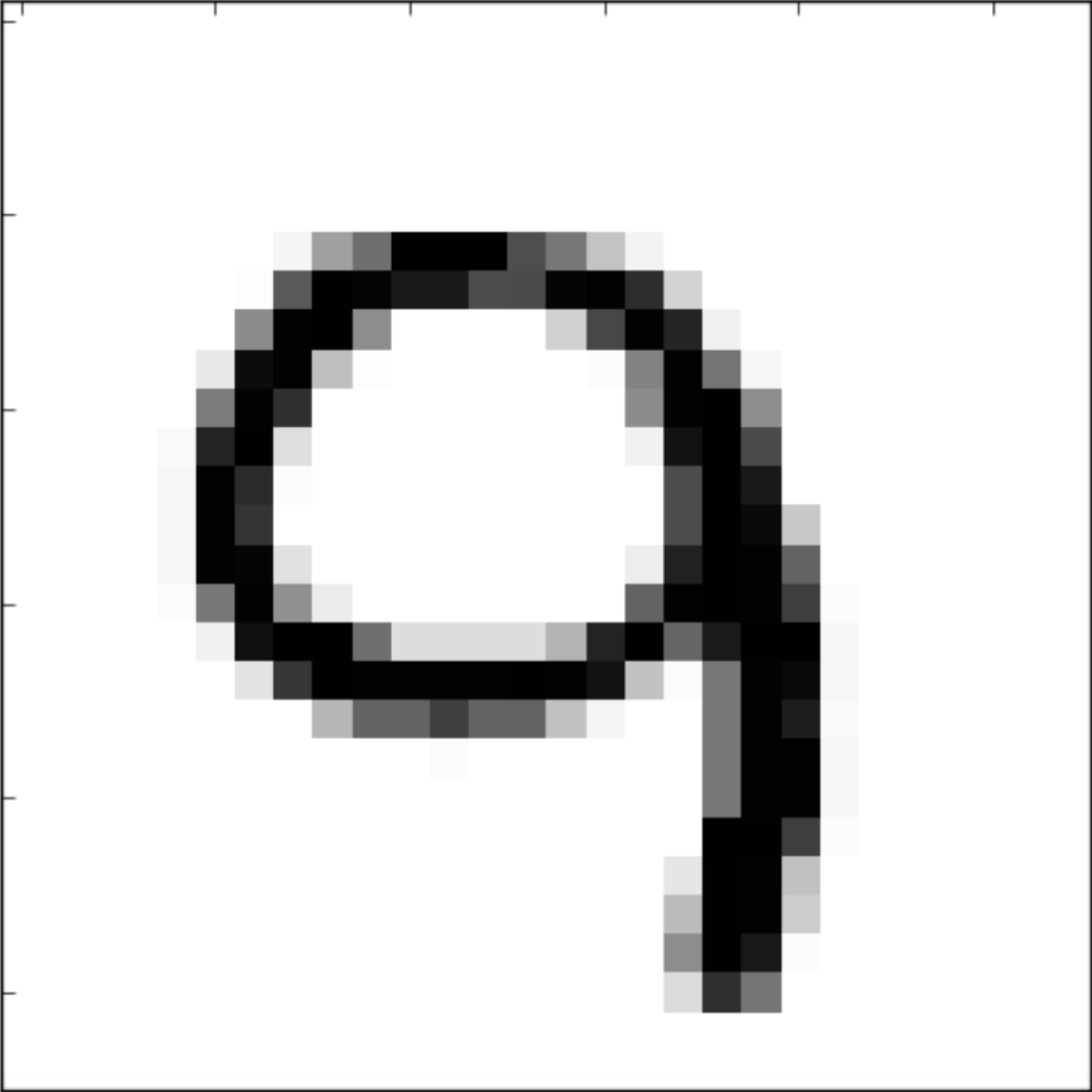}
\includegraphics[width=1.1cm,height=1.1cm]{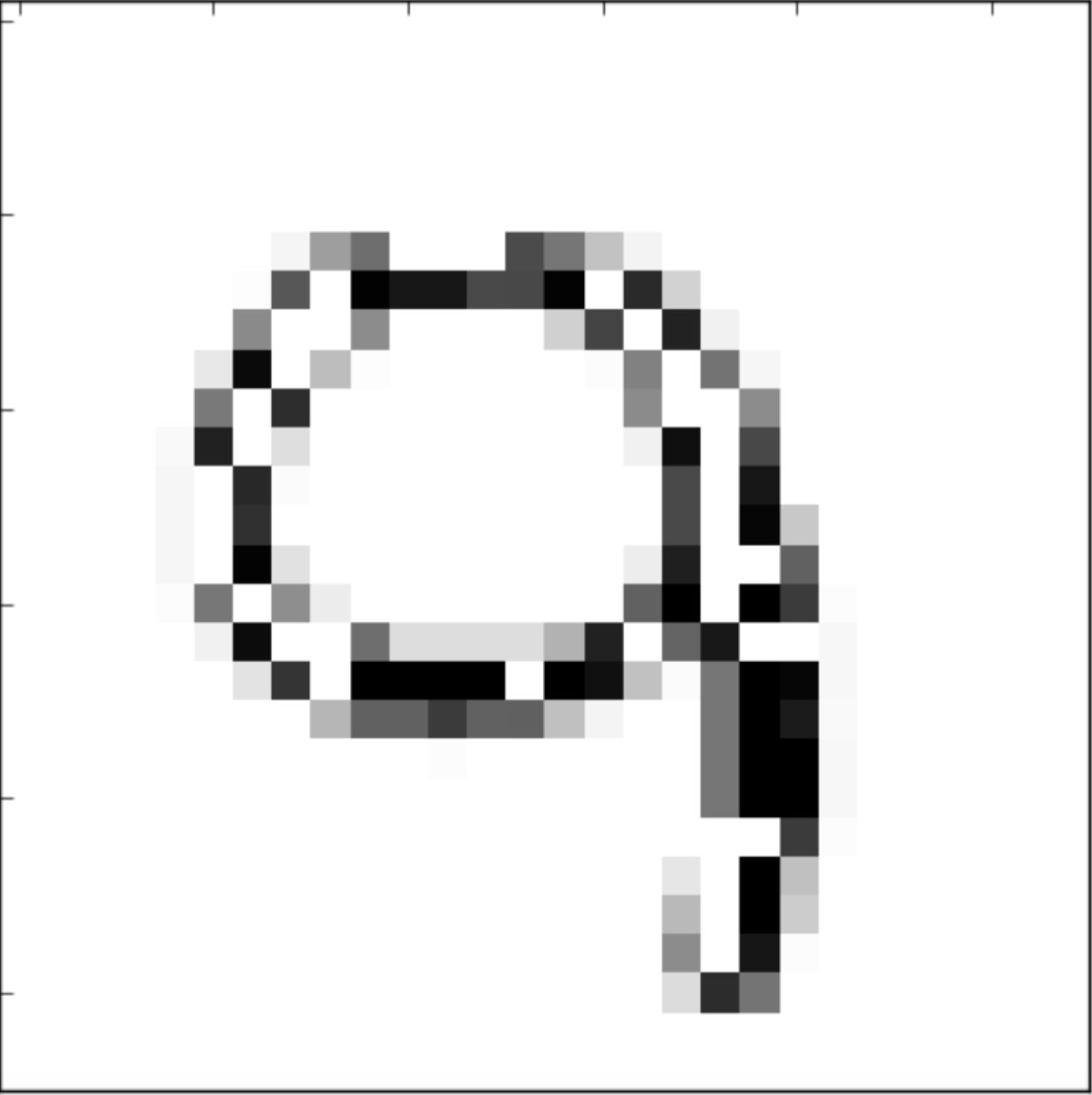}
\text{\hspace{0.8cm}9 to 7}
}
\parbox{2.3cm}{
\includegraphics[width=1.1cm,height=1.1cm]{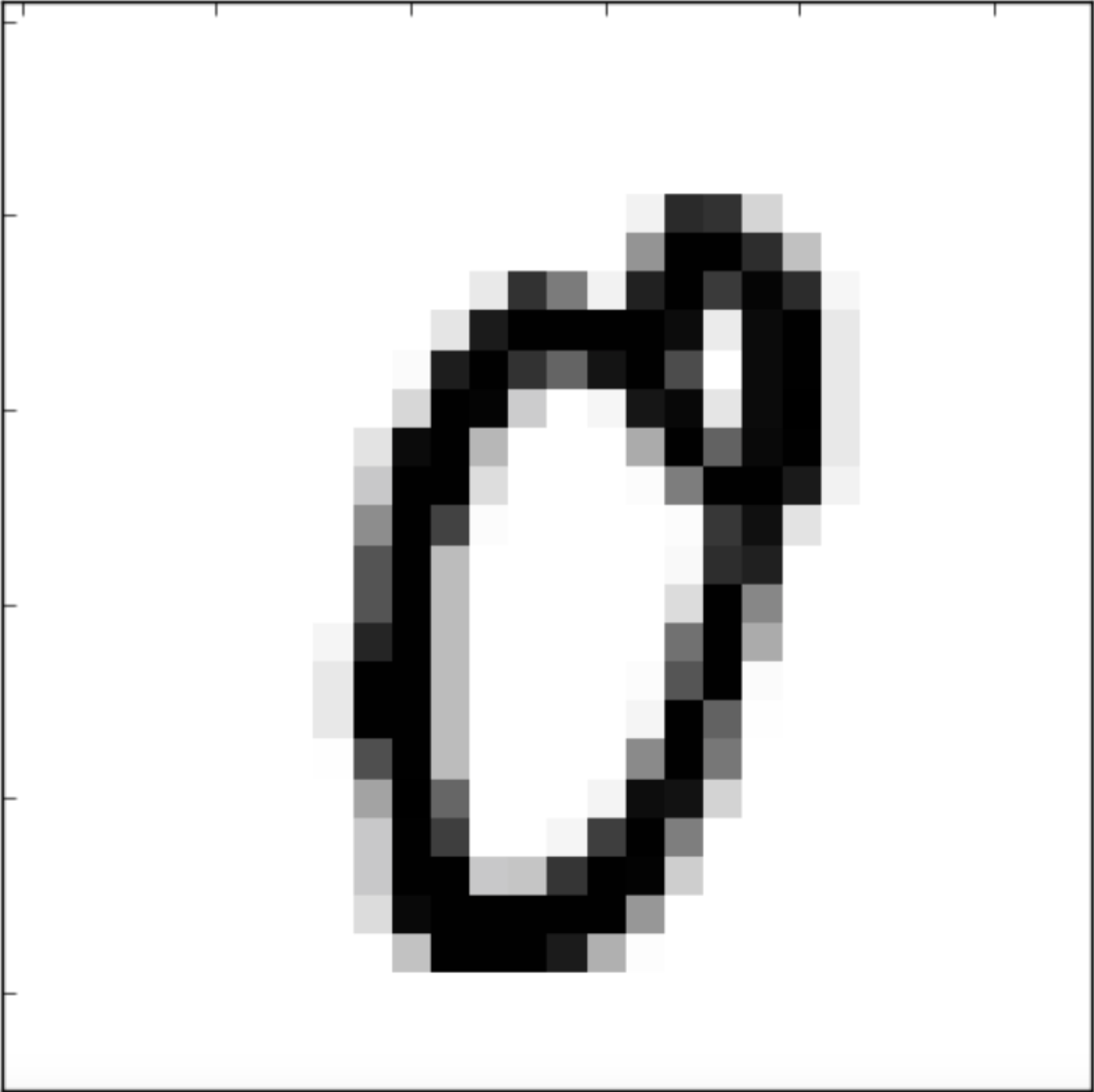}
\includegraphics[width=1.1cm,height=1.1cm]{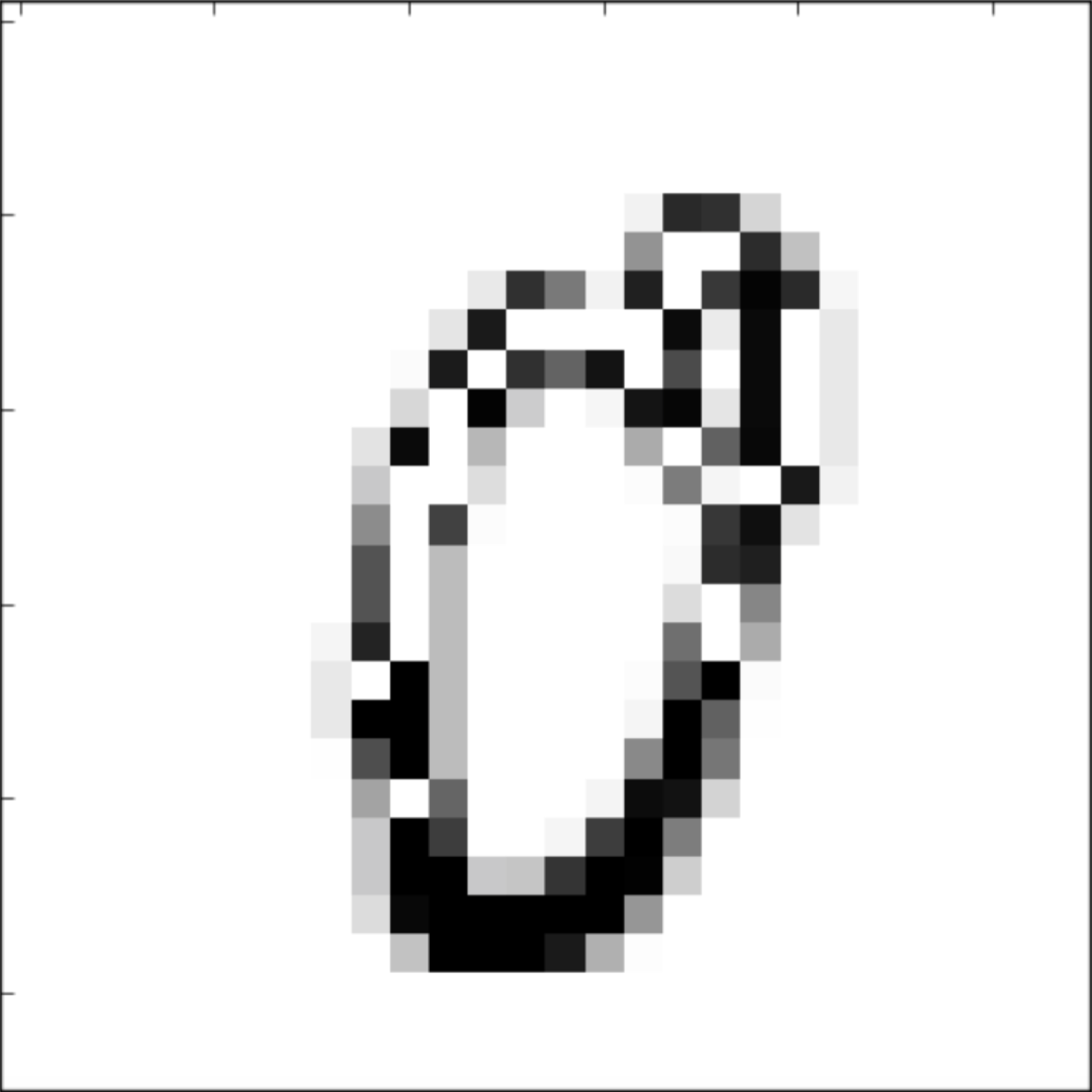}
\text{\hspace{0.8cm}0 to 8}
}
\parbox{2.3cm}{
\includegraphics[width=1.1cm,height=1.1cm]{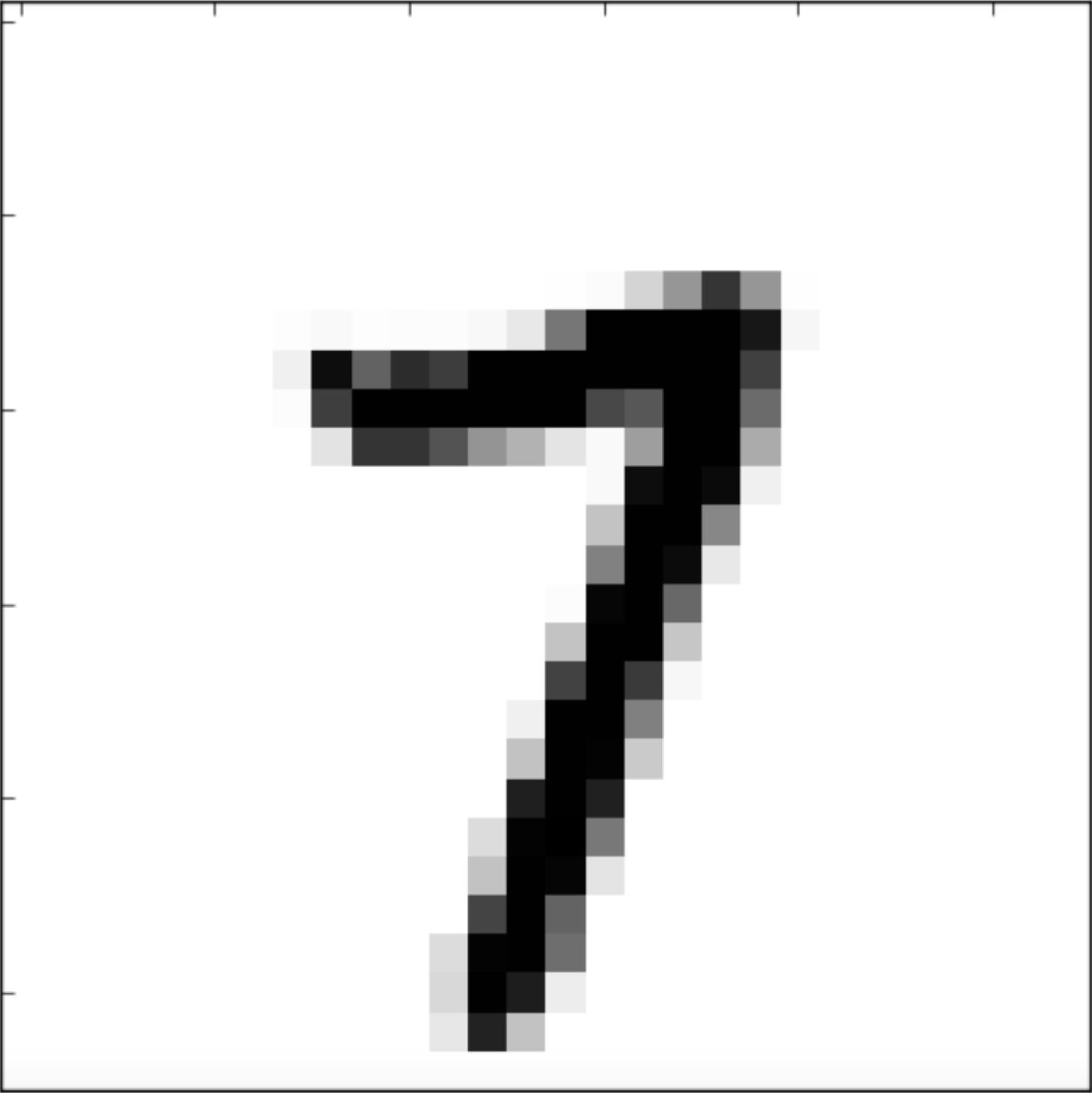}
\includegraphics[width=1.1cm,height=1.1cm]{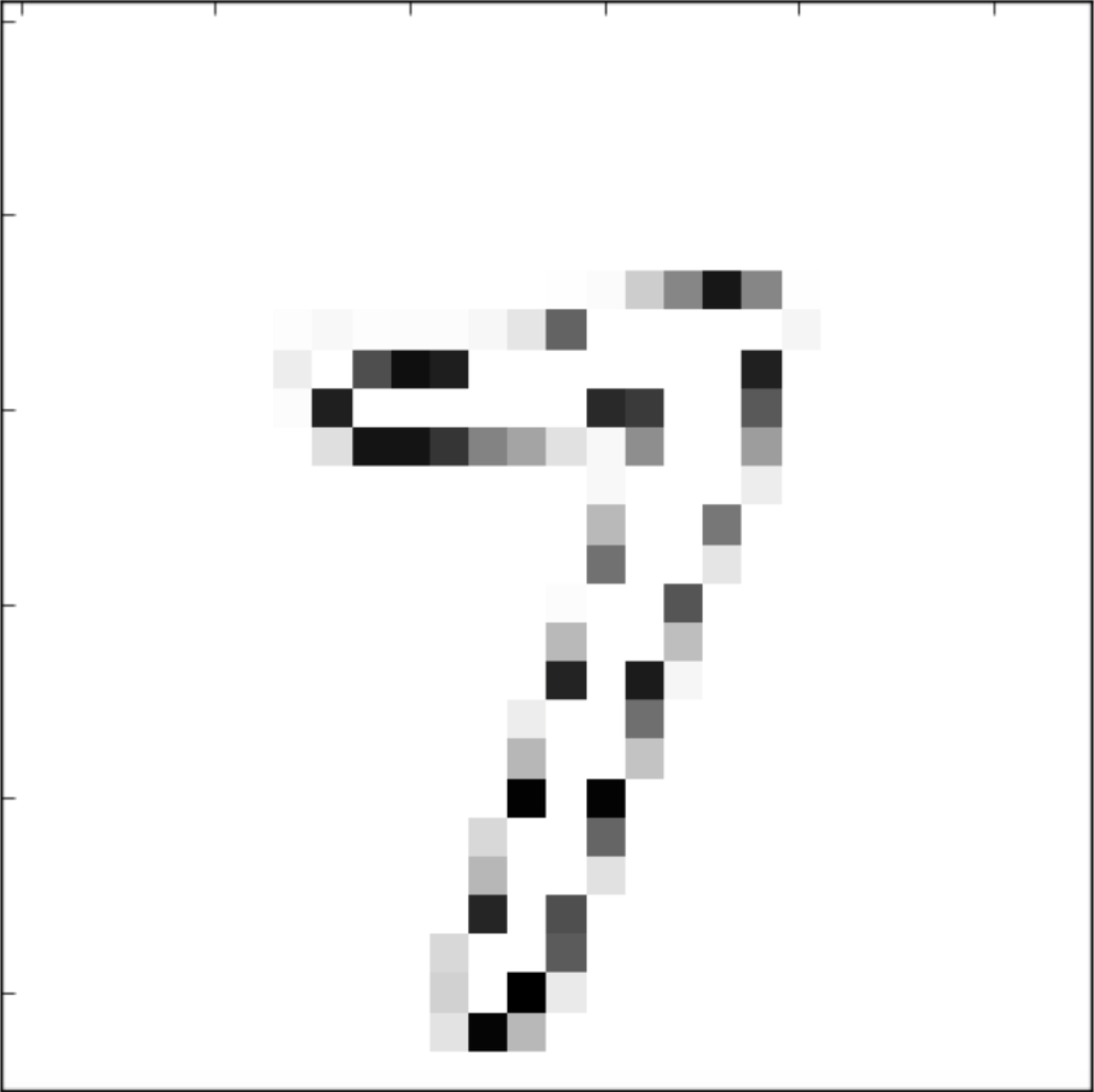}
\text{\hspace{0.8cm}7 to 9}
}
\caption{Adversarial examples for a neural network trained on MNIST}
\label{fig:mnist}
\end{figure}

Given an image $x$, we start with layer $k=1$ and the parameter set to at most 150 dimensions (there are 21632 dimensions in layer $L_1$). All $\eta_k,\manipulationset_k$ for $k\geq 2$ are computed according to the simple heuristic mentioned in Section~\ref{sec:impl} and satisfy Definition~\ref{def:ek} and Definition~\ref{def:layerRefinement}.
For the region $\eta_1(\activation_{x,1})$, we allow changes to the activation value of each selected dimension that are within [-1,1]. The set $\manipulationset_1$ includes manipulations that can change the activation value for a subset of the 150 dimensions, by incrementing or decrementing the value for each dimension by 1.
The experimental results show that for most of the examples we can find a class change within 100 dimensional changes in layer $L_1$, by comparing the number of pixels that have changed, and some of them can have less than 30 dimensional changes. Figure~\ref{fig:mnist} presents examples of such class changes for layer $L_1$. We also experiment on images with up to 40 dimensional changes in layer $L_1$; the tool is able to check the entire network, reaching the output layer and claiming that $N,\eta_k,\manipulationset_k\models x$ for all $k\geq 1$.  
While training of the network takes half an hour, finding an adversarial example takes up to several minutes.

\paragraph{\bf Image Classification Network for the CIFAR-10 Small Image Dataset}

We work with a medium size neural network, trained with the source code from \cite{cifar10} for more than 12 hours on the well-known CIFAR10 dataset. The inputs to the network are images of size $32\times 32$ with three channels. The trained network has 1,250,858 real-valued parameters and includes convolutional layers, ReLU layers, max-pooling layers, dropout layers, fully-connected layers, and a softmax layer.

\begin{figure}
\parbox{2.9cm}{
\includegraphics[width=1.4cm,height=1.4cm]{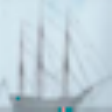}
\includegraphics[width=1.4cm,height=1.4cm]{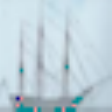}
}
\parbox{2.9cm}{
\includegraphics[width=1.4cm,height=1.4cm]{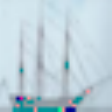}
\includegraphics[width=1.4cm,height=1.4cm]{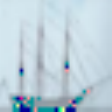}
}
\parbox{2.9cm}{
\includegraphics[width=1.4cm,height=1.4cm]{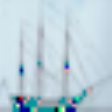}
\includegraphics[width=1.4cm,height=1.4cm]{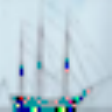}
}
\parbox{2.9cm}{
\includegraphics[width=1.4cm,height=1.4cm]{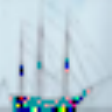}
\includegraphics[width=1.4cm,height=1.4cm]{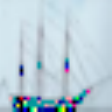}
}
\caption{An illustrative example of mapping back to input layer from the Cifar-10 mataset: the last image classifies as a truck. }
\label{fig:illustrative}
\end{figure}

As an illustration of the type of perturbations that we are investigating, consider the images in Figure~\ref{fig:illustrative}, which correspond to the parameter setting of up to 25, 45, 65, 85, 105, 125, 145 dimensions, respectively, for layer $k=1$. The manipulations change the activation values of these dimensions. Each image is obtained by mapping back from the first hidden layer and represents a point close to the boundary of the corresponding region. 
The relation $N,\eta_1,\manipulationset_1\models x$ holds for the first 7 images, but fails for the last one and the image is classified as a truck. 
Intuitively, our choice of the region $\eta_1(\activation_{x,1})$ identifies the subset of dimensions with most extreme activations, taking advantage of the analytical capability of the first hidden layer.
A higher number of selected dimensions implies a larger region 
in which we apply manipulations, and, more importantly, suggests a more dramatic change to the knowledge represented by the activations when moving to the boundary of the region.

We also work with 500 dimensions and otherwise the same experimental parameters as for MNIST.
Figure~\ref{fig:cifar10} in Appendix of~\cite{HKWW2016} gives 16 pairs of original images (classified correctly) and perturbed images (classified wrongly). We found that, while the manipulations lead to human-recognisable modifications to the images, the perturbed images can be  classified wrongly by the network. For each image, finding an adversarial example ranges from seconds to 20 minutes.

\paragraph{\bf Image Classification Network for the ImageNet Dataset}

We also conduct experiments on a large image classification network trained on the popular ImageNet dataset. The images are of size $224\times 224$ and have three channels. The network is the model of the 16-layer network~\cite{SZ2014}, called VGG16, used by the VGG team in the ILSVRC-2014 competition, downloaded from~\cite{VGG16}. The trained network has 138,357,544 real-valued parameters and includes convolutional layers, ReLU layers, zero-padding layers, dropout layers, max-pooling layers, fully-connected layers, and a softmax layer. The experimental parameters are the same as for the previous two experiments, except that we work with 20,000 dimensions.

Several additional pairs of original and perturbed images are included in Figure~\ref{fig:imageNet} in Appendix of \cite{HKWW2016}.
In Figure~\ref{fig:streetsign} we also give two examples of street sign images. The image on the left is reported unsafe for the second layer with 6346 dimensional changes (0.2\% of the 3,211,264 dimensions of layer $L_2$). 
The one on the right is reported safe for
20,000 dimensional changes of layer $L_2$. 
It appears that more complex manipulations, involving more dimensions (perceptrons), are needed in this case to cause a class change.

\begin{figure}
\parbox{6.2cm}{
\includegraphics[width=3cm,height=3cm]{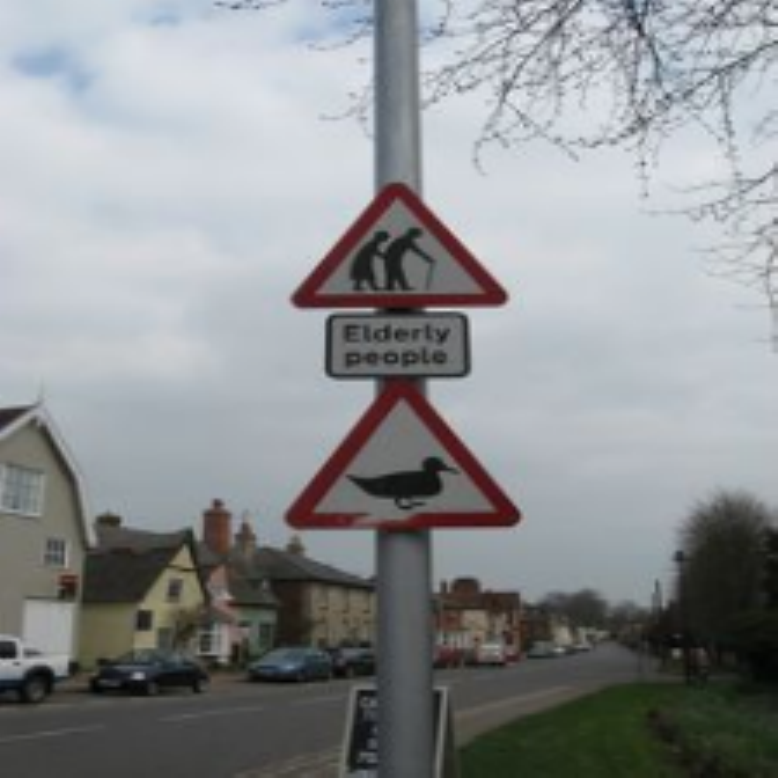}
\includegraphics[width=3cm,height=3cm]{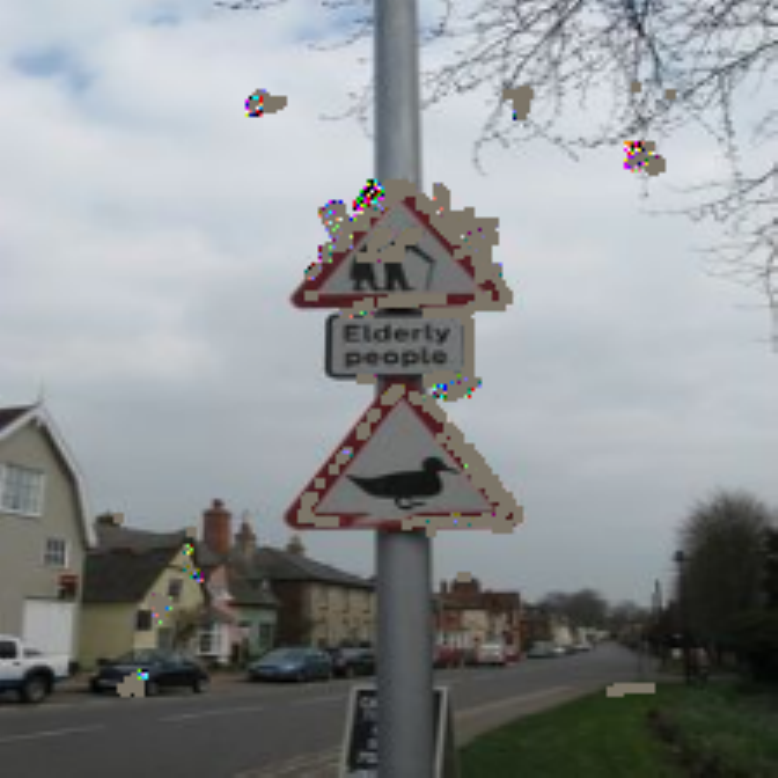}
}
\parbox{6.2cm}{
\includegraphics[width=3cm,height=3cm]{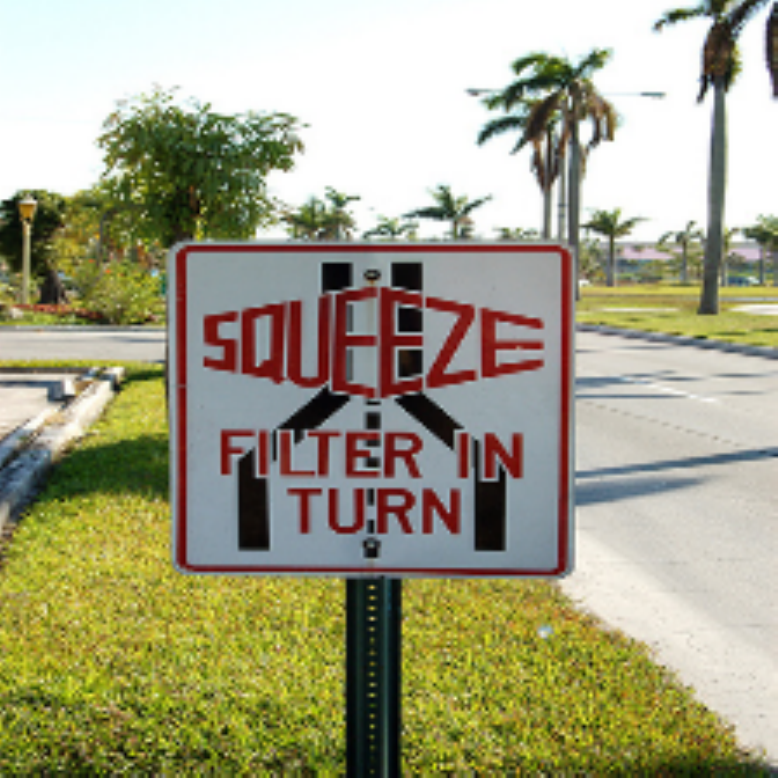}
\includegraphics[width=3cm,height=3cm]{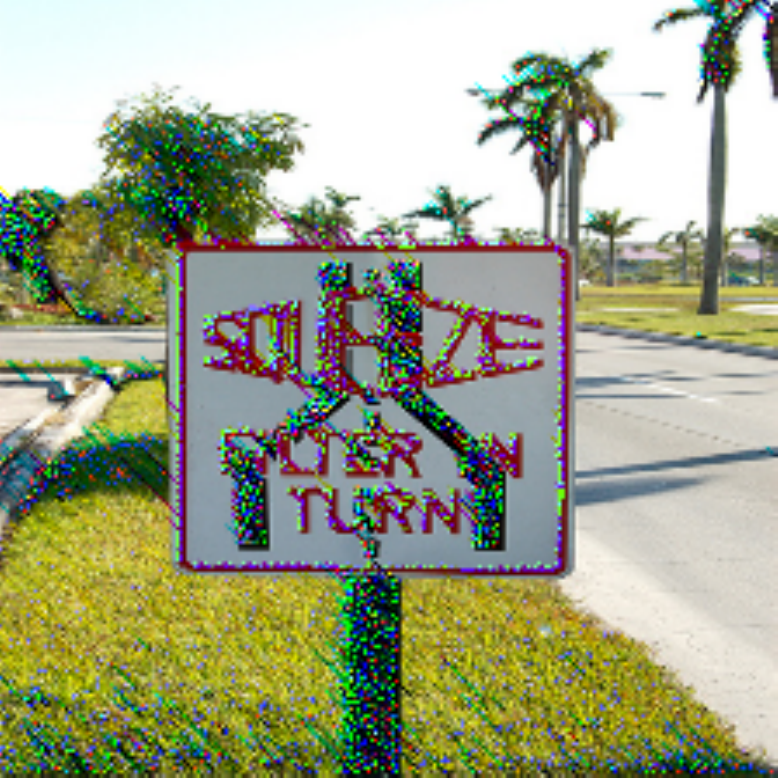}
}
\caption{Street sign images. Found an adversarial example for the left image (class changed into bird house), but cannot find an adversarial example for the right image for 20,000 dimensions.}
\label{fig:streetsign}
\end{figure}

\subsection{The German Traffic Sign Recognition Benchmark (GTSRB)}

\begin{figure}
\centering
\parbox{4cm}{
\includegraphics[width=1.9cm]{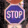}
\includegraphics[width=1.9cm]{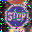}
\center{``stop'' \\ to ``30m speed limit''}
}
\parbox{4cm}{
\includegraphics[width=1.9cm]{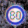}
\includegraphics[width=1.9cm]{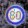}
\center{``80m speed limit'' \\ to ``30m speed limit''}
}
\parbox{4cm}{
\includegraphics[width=1.9cm]{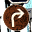}
\includegraphics[width=1.9cm]{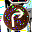}
\center{``go right'' \\ to ``go straight''}
}
\caption{Adversarial examples for the network trained on the GTSRB dataset by multi-path search}
\label{fig:gtsrbMultiPath}
\end{figure}

We evaluate DLV on the GTSRB dataset (by resizing images into size 32*32), which has 43 classes. Figure~\ref{fig:gtsrbMultiPath} presents the results for the multi-path search.
The first case (approx. 20 minutes to manipulate) is a stop sign (confidence 1.0) changed into a speed limit of 30 miles, with an $L_1$ distance of 0.045 and $L_2$ distance of 0.19. The confidence of the manipulated image is 0.79. The second, easy, case (seconds to manipulate) is a speed limit of 80 miles (confidence 0.999964) changed into a speed limit of 30 miles, with an $L_1$ distance of 0.004 and $L_2$ distance of 0.06. The confidence of the manipulated image is 0.99 (a very high confidence of misclassification). Also, a ``go right'' sign can be easily manipulated into a sign classified as ``go straight''.  

Figure~\ref{fig:moreGtsrb} in \cite{HKWW2016} presents additional adversarial examples obtained when selecting single-path search.

\section{Comparison}
\label{sec:comparison}

We compare our approach with two existing approaches for finding adversarial examples, i.e., fast gradient sign method (FGSM)~\cite{SZSBEGF2014} and Jacobian saliency map algorithm (JSMA)~\cite{PMJFCS2015}. 
FGSM calculates the optimal attack for a linear approximation of the network cost, whereas DLV explores a proportion of dimensions in the feature space in the input or hidden layers. JSMA finds a set of dimensions in the input layer to manipulate, according to the linear approximation (by computing the Jacobian matrix) of the model from current output to a nominated target output. Intuitively, the difference between DLV's manipulation and JSMA is that DLV manipulates over features discovered in the activations of the hidden layer, while JSMA manipulates according to the partial derivatives, which depend on the parameters of the network. 

\begin{figure}
\includegraphics[width=1.2cm]{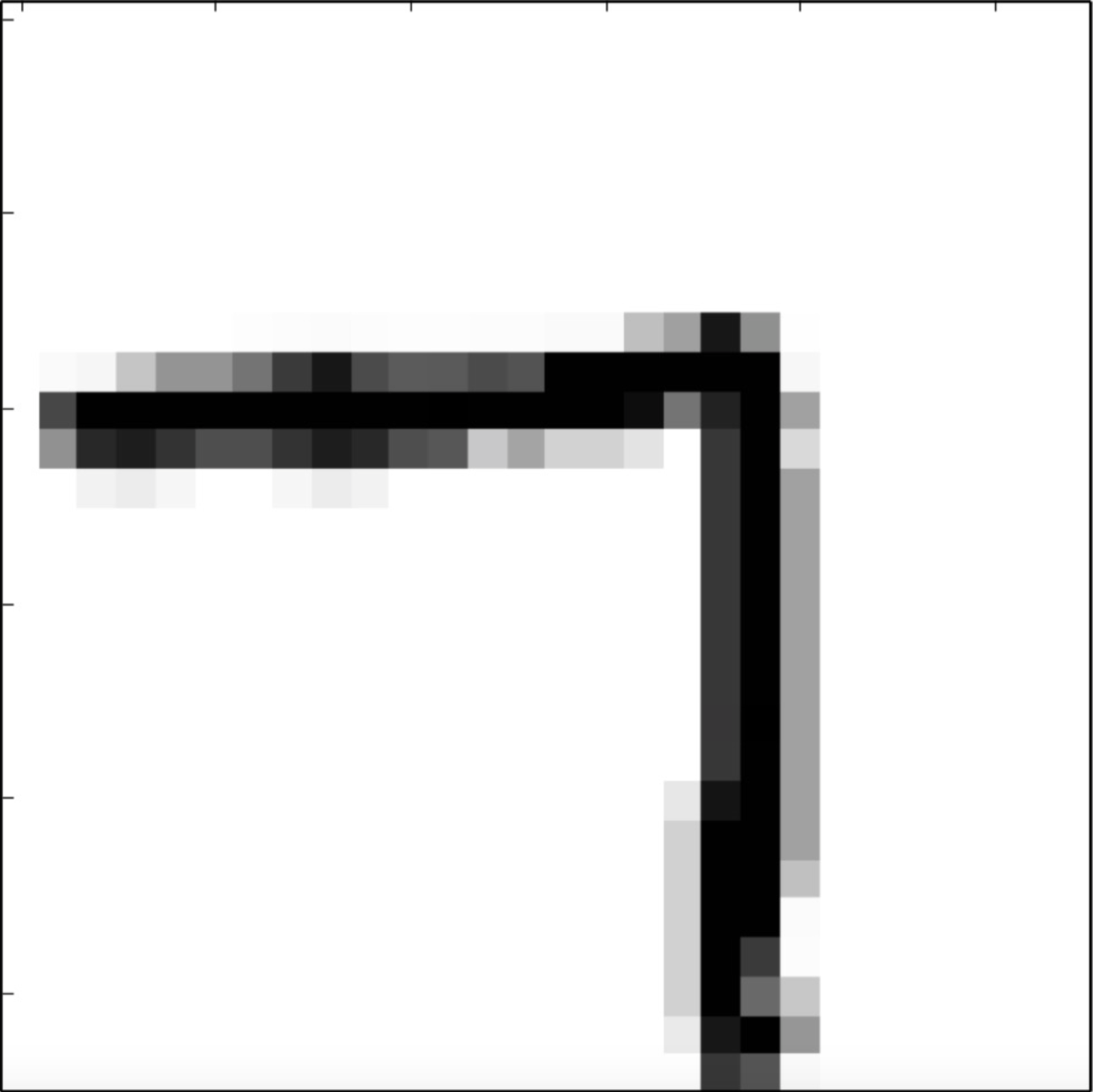}
\includegraphics[width=1.2cm]{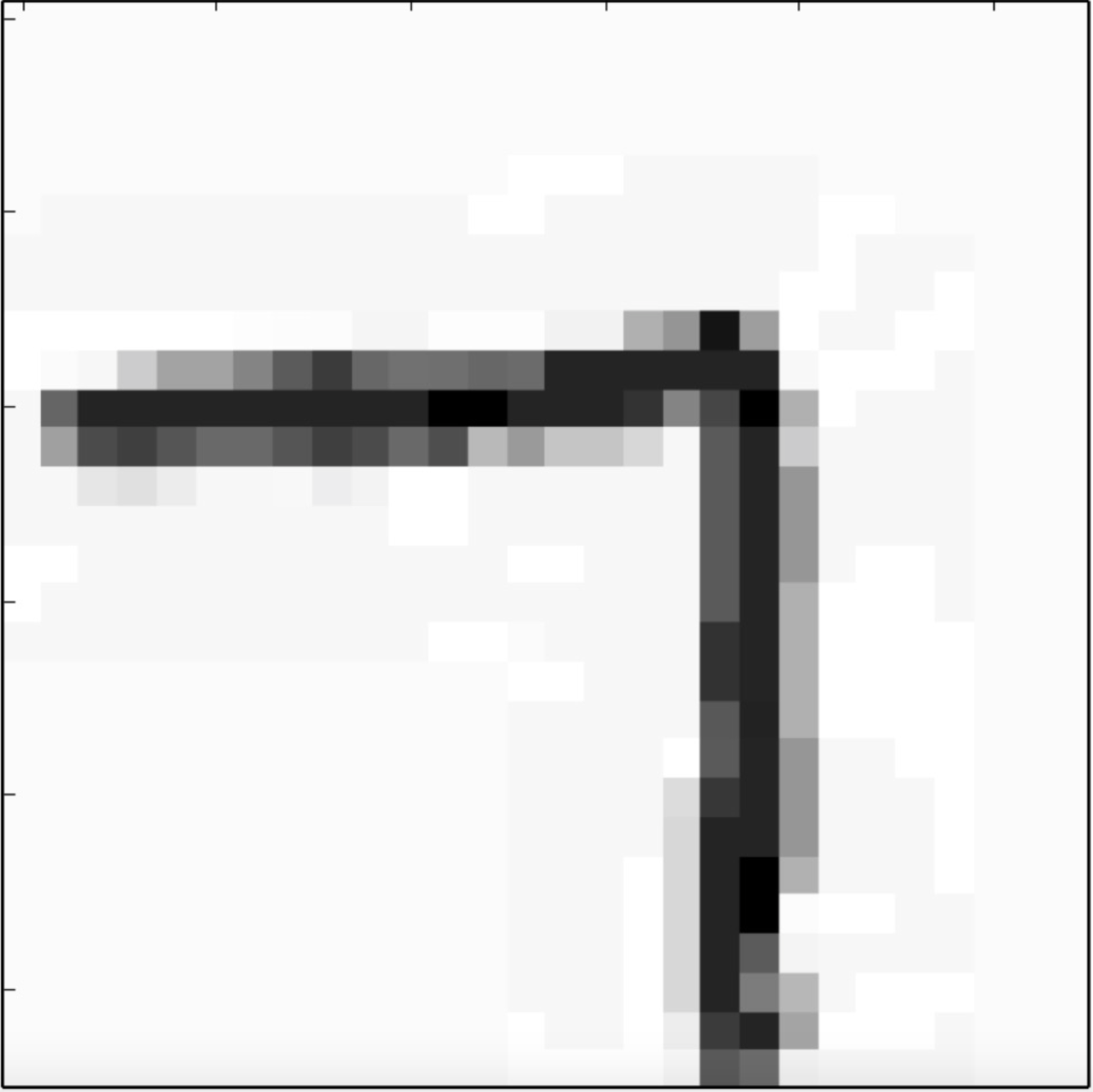}
\includegraphics[width=1.2cm]{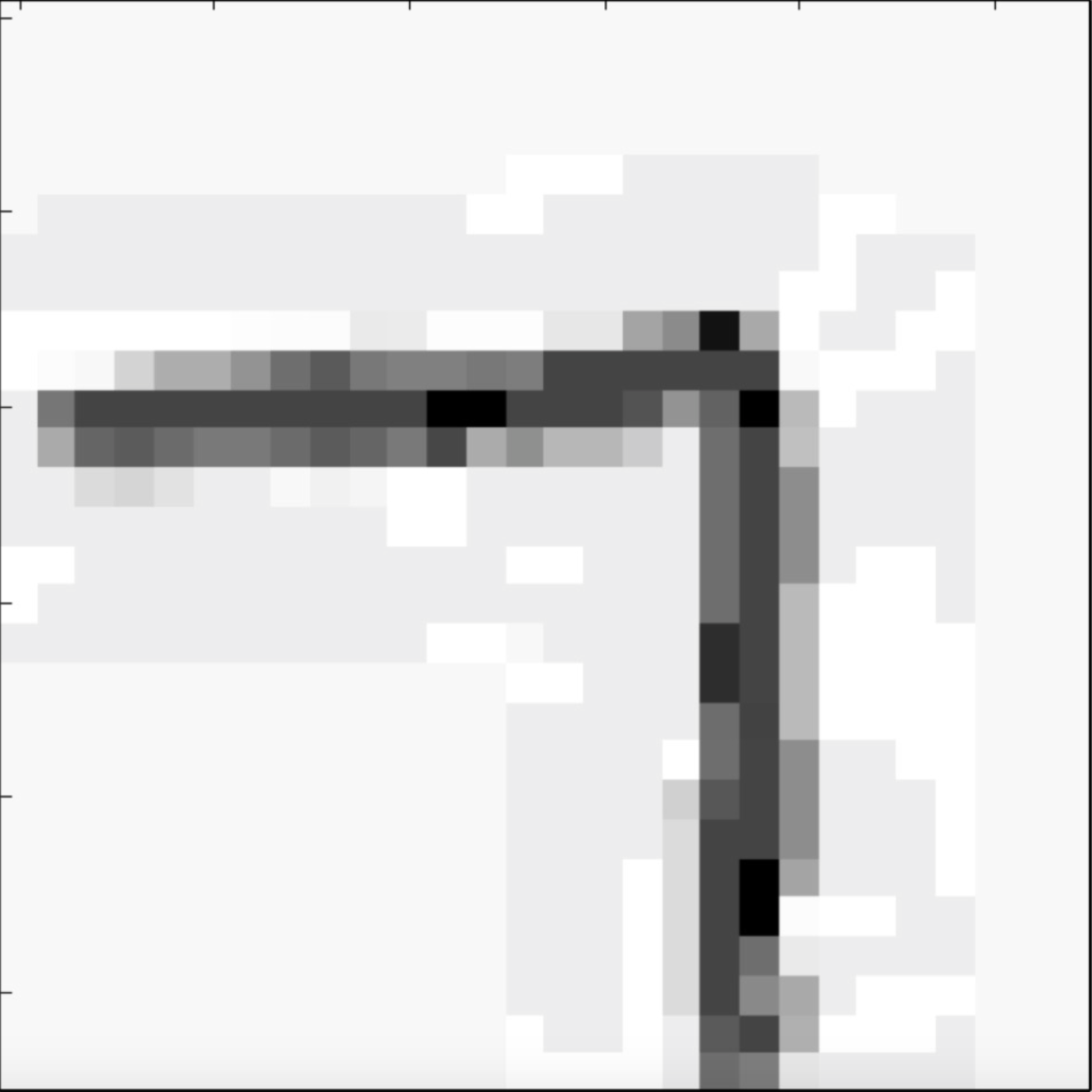}
\includegraphics[width=1.2cm]{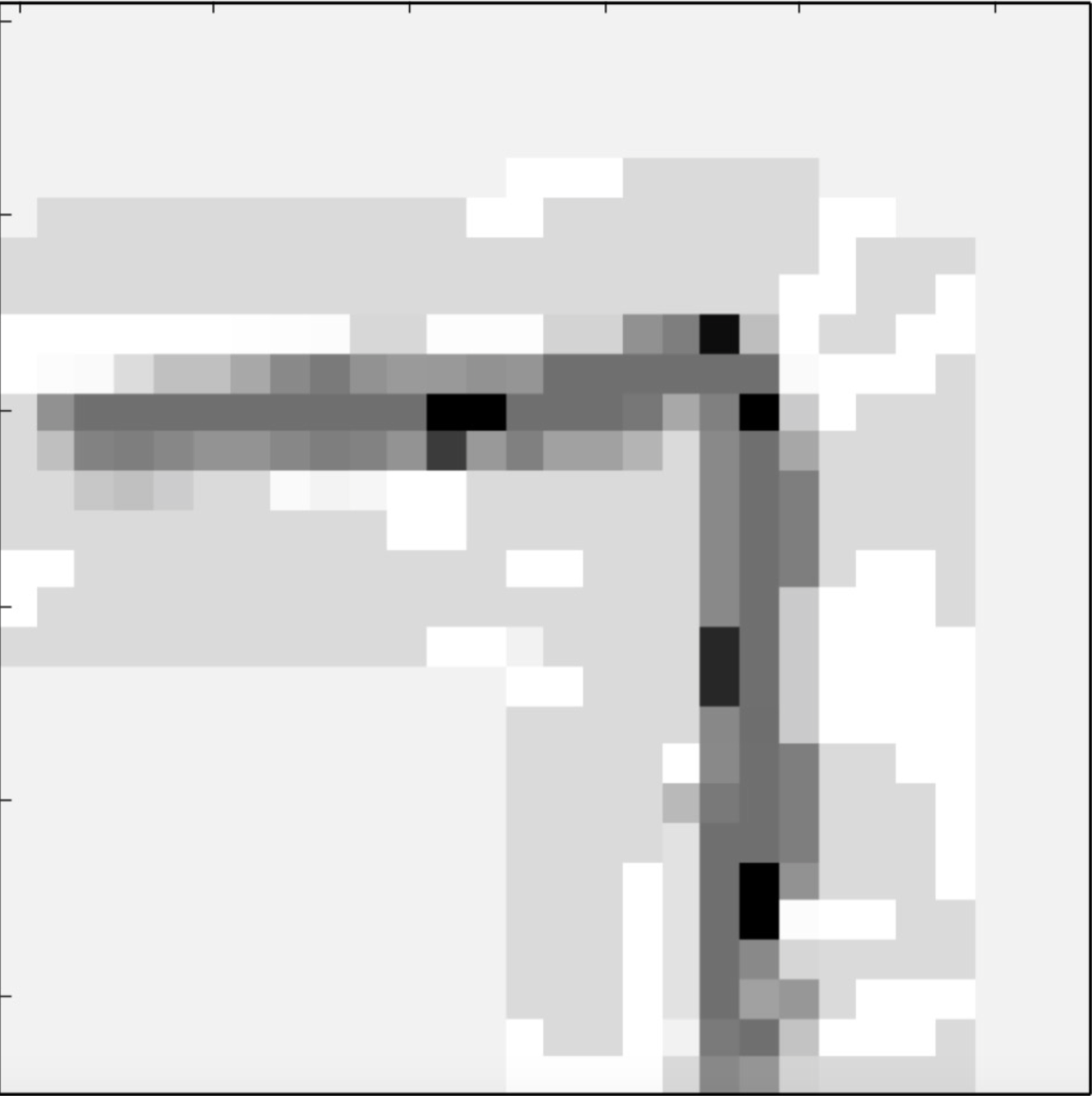}
\includegraphics[width=1.2cm]{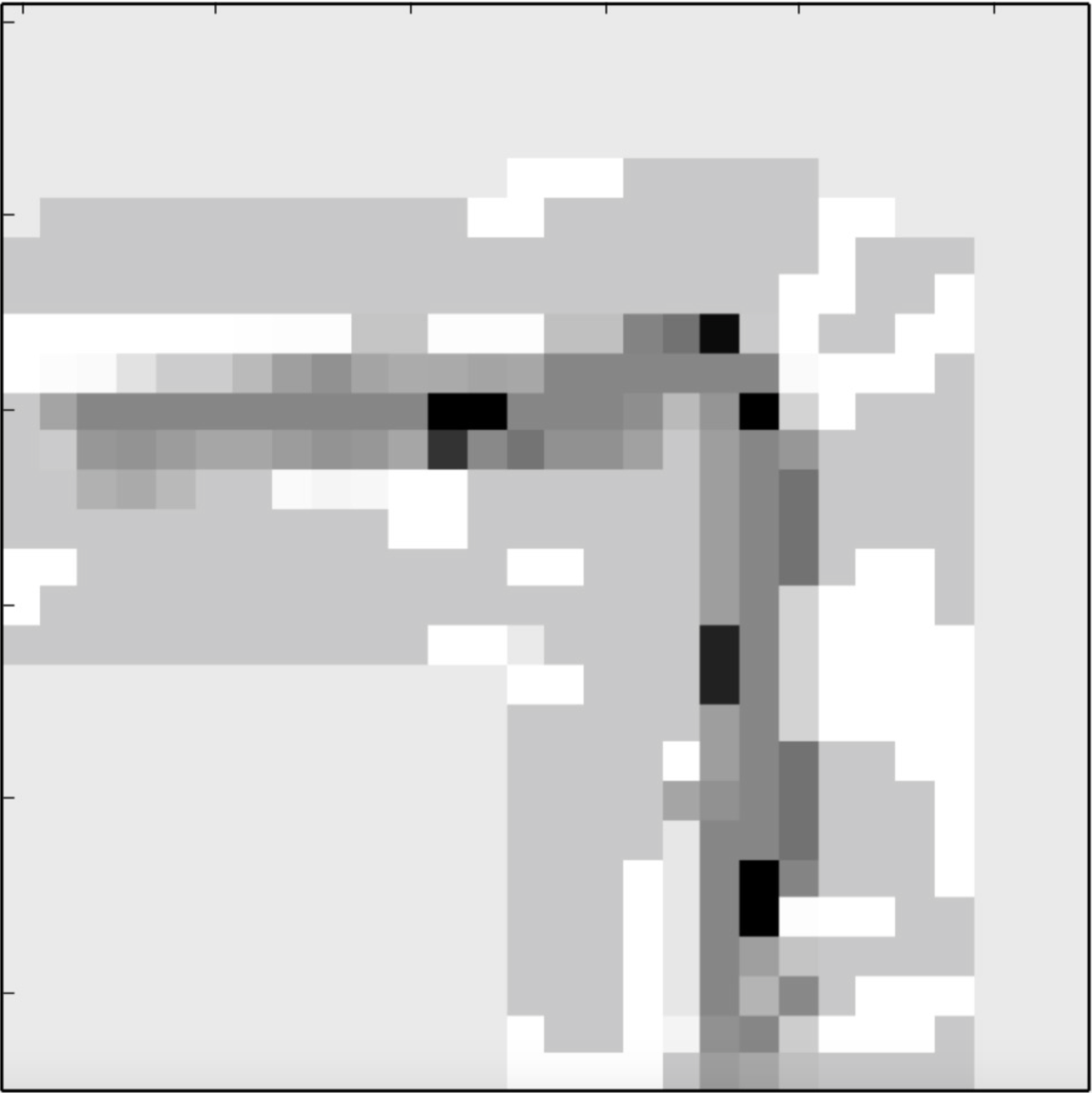}
\includegraphics[width=1.2cm]{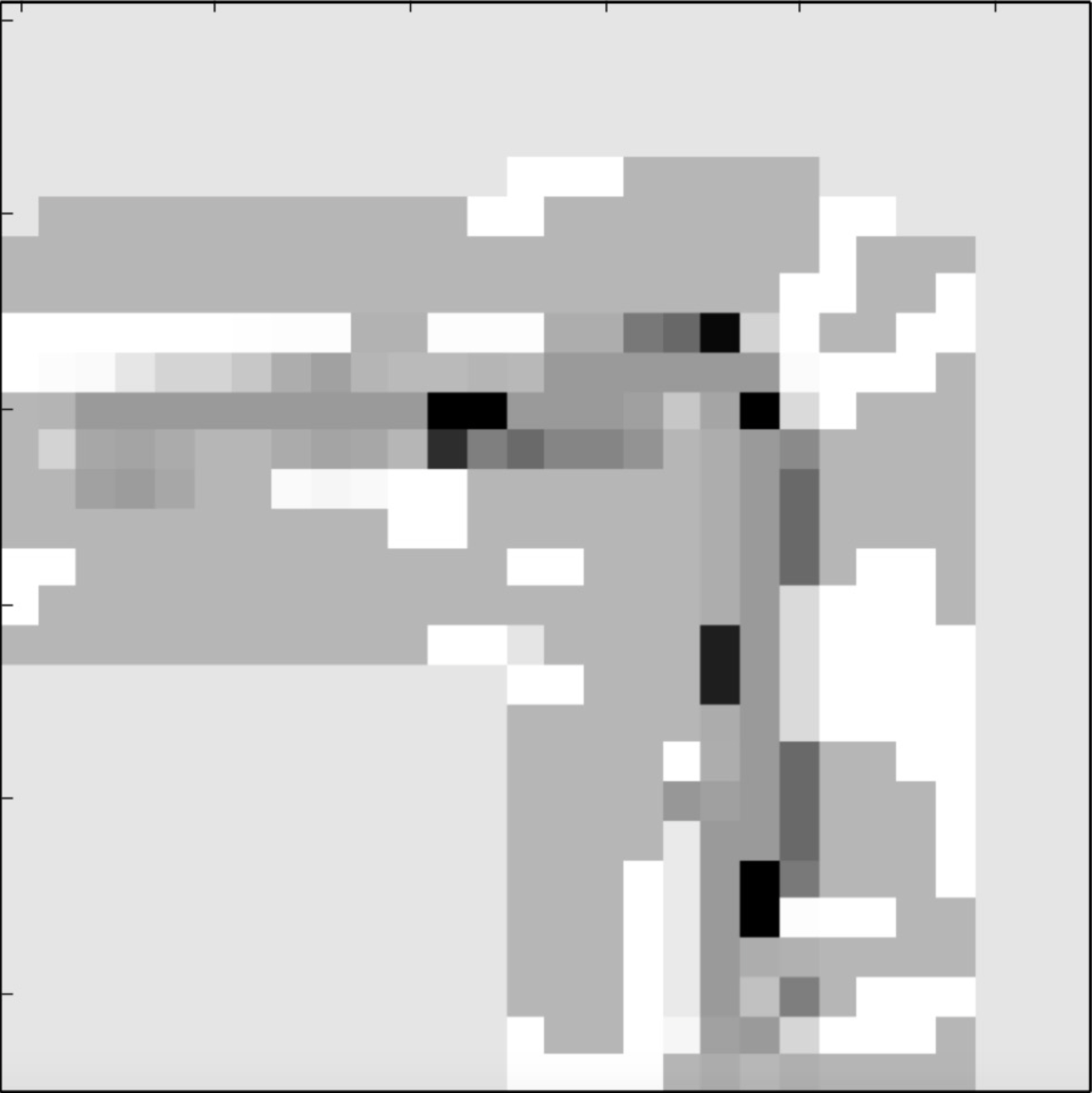}

\includegraphics[width=1.2cm]{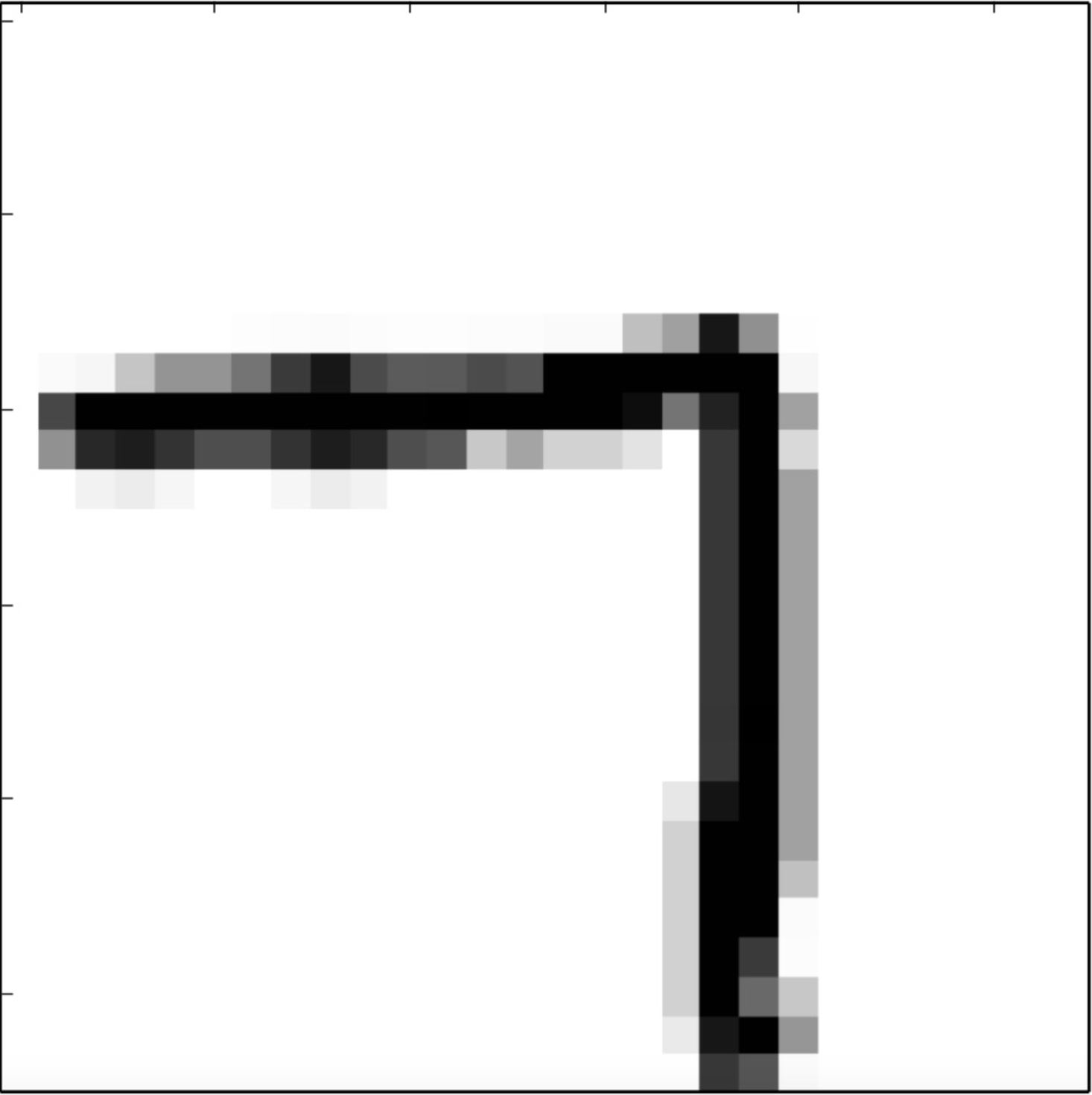}
\includegraphics[width=1.2cm]{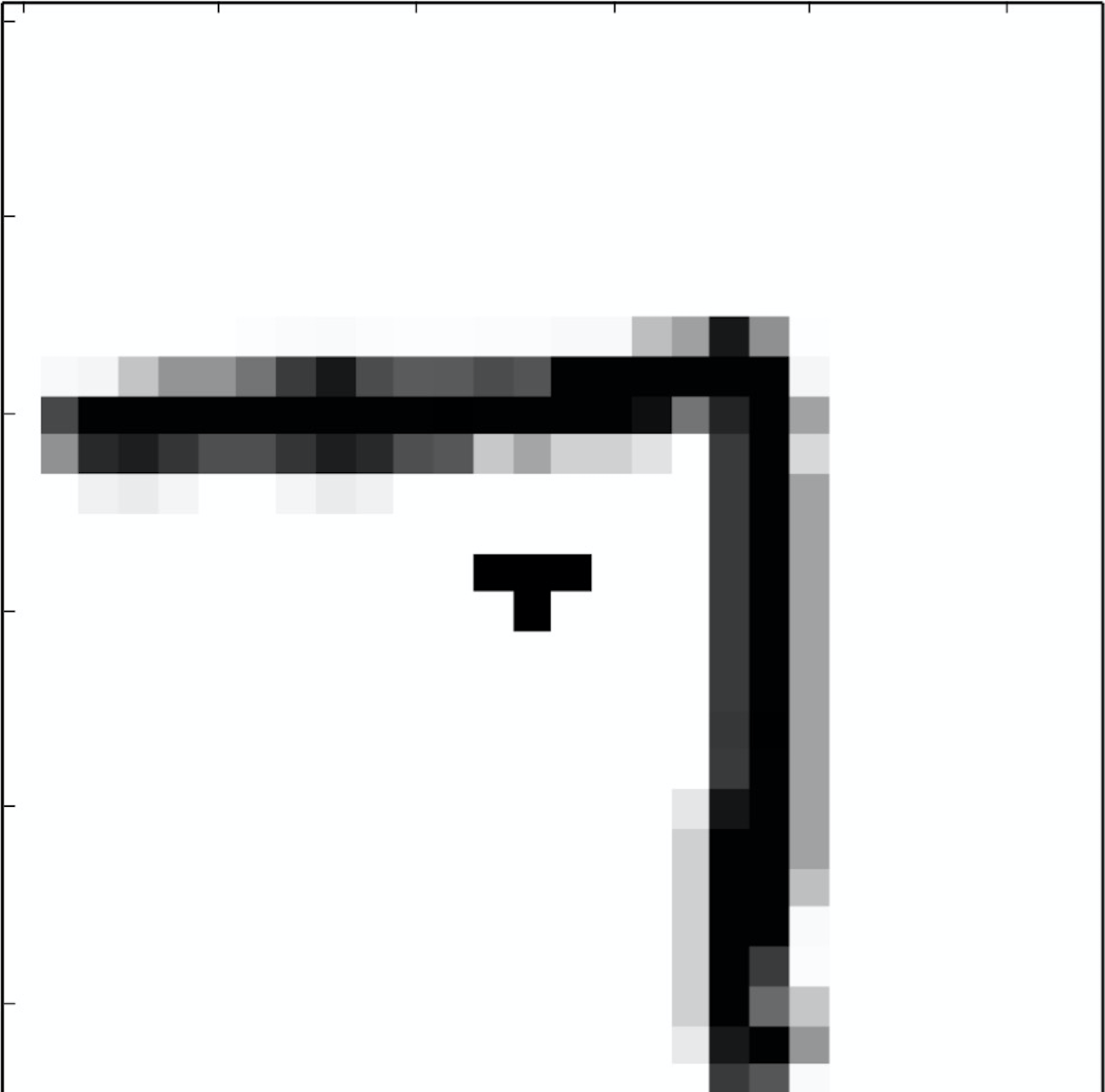}
\includegraphics[width=1.2cm]{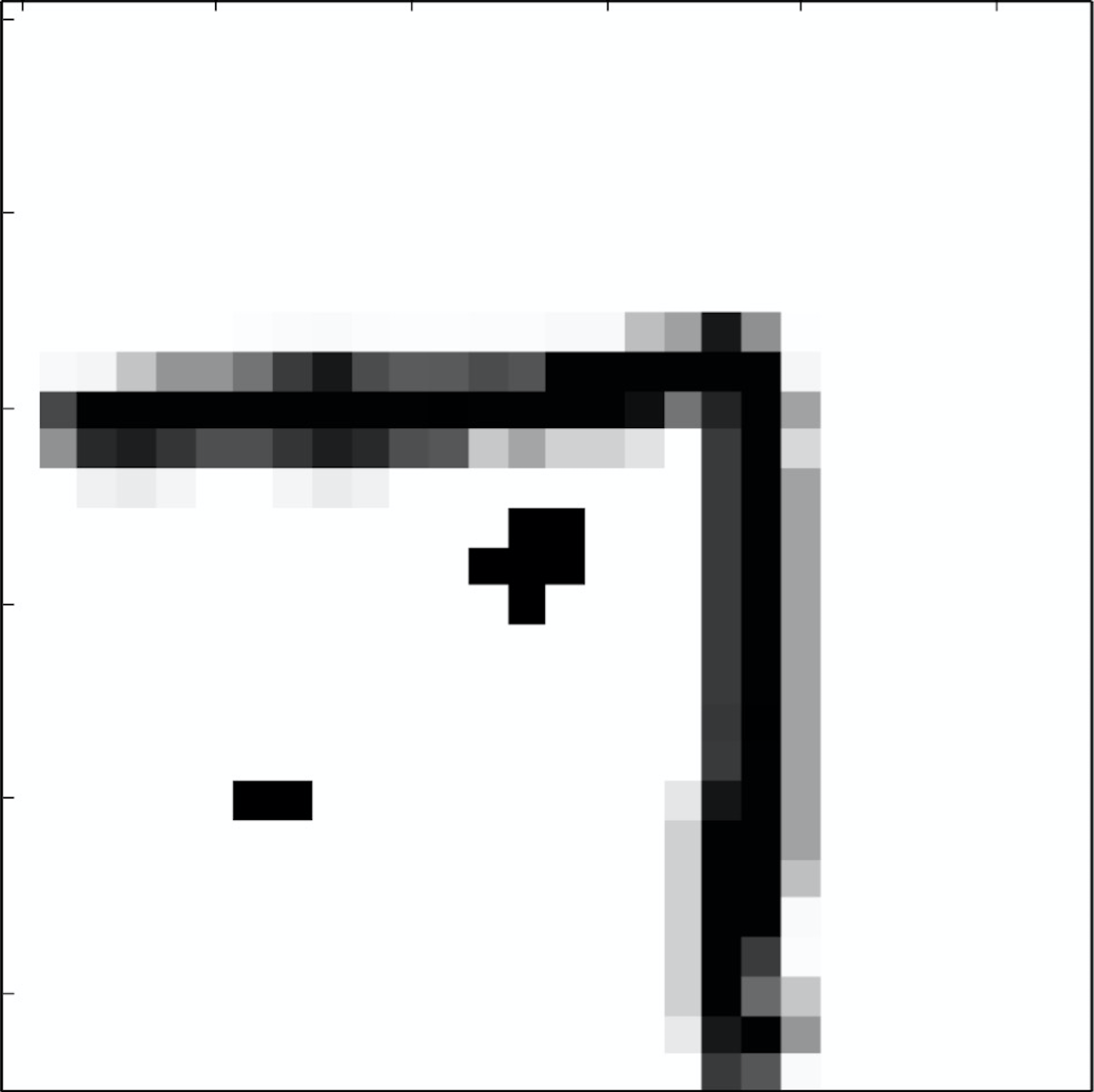}
\includegraphics[width=1.2cm]{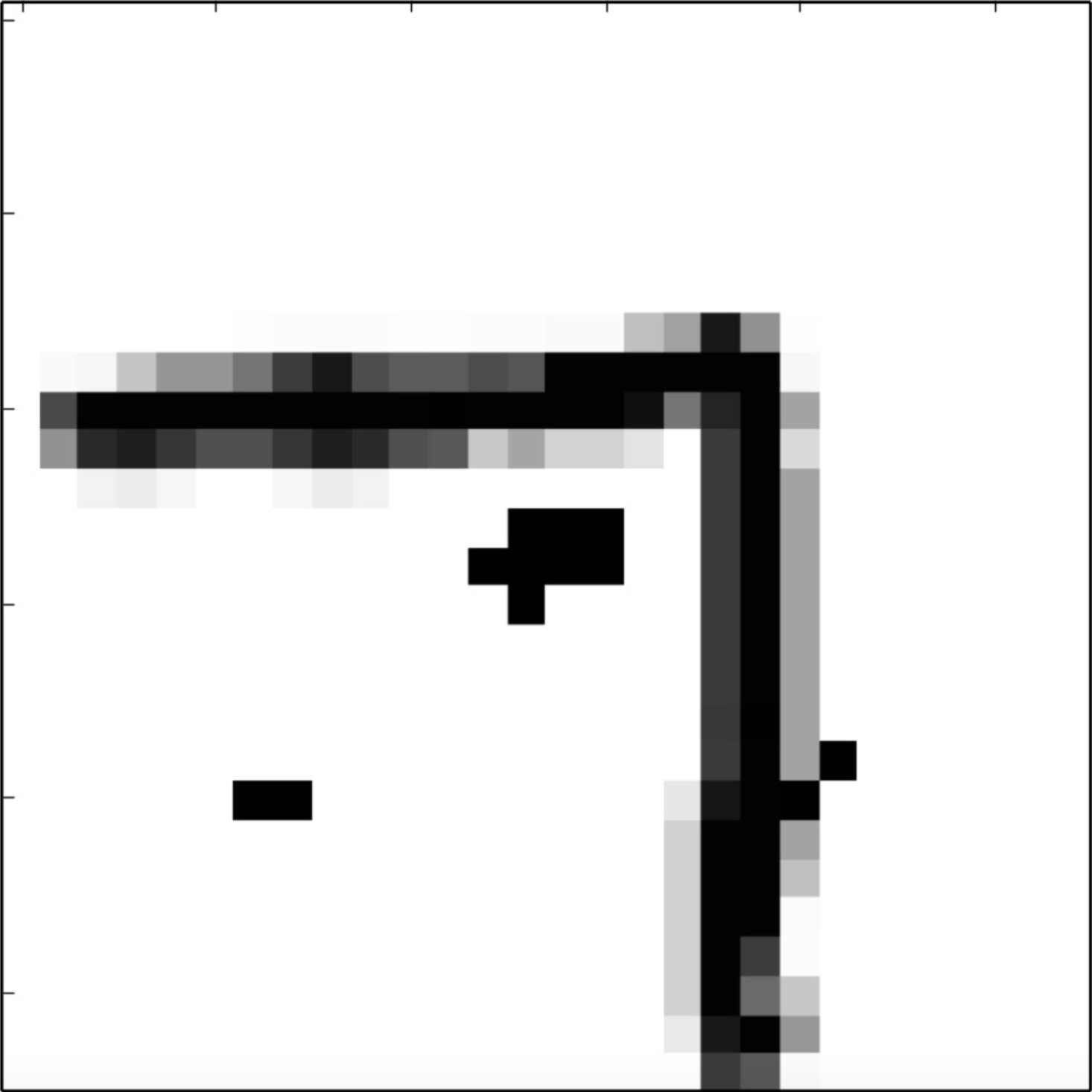}
\includegraphics[width=1.2cm]{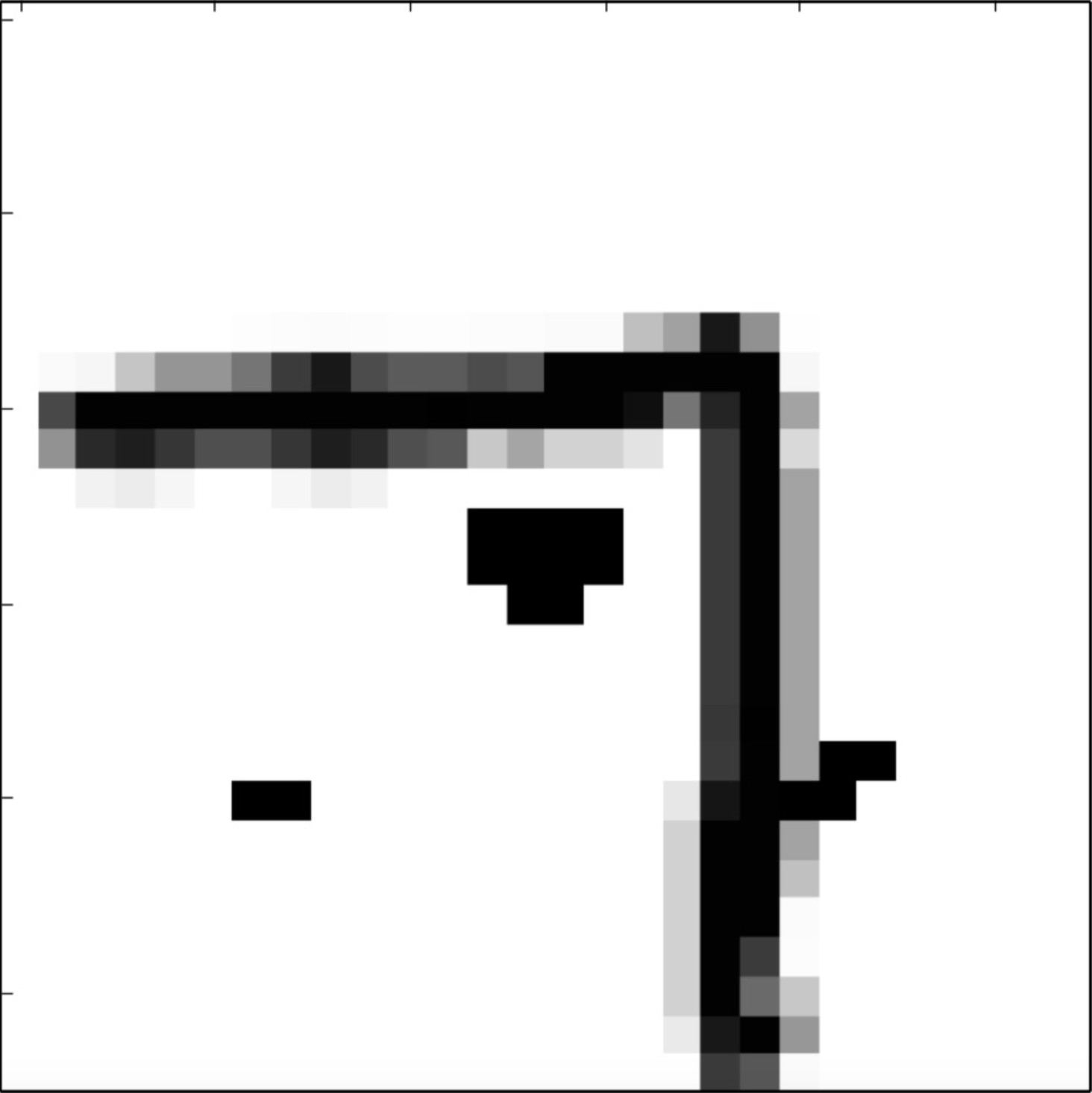}
\includegraphics[width=1.2cm]{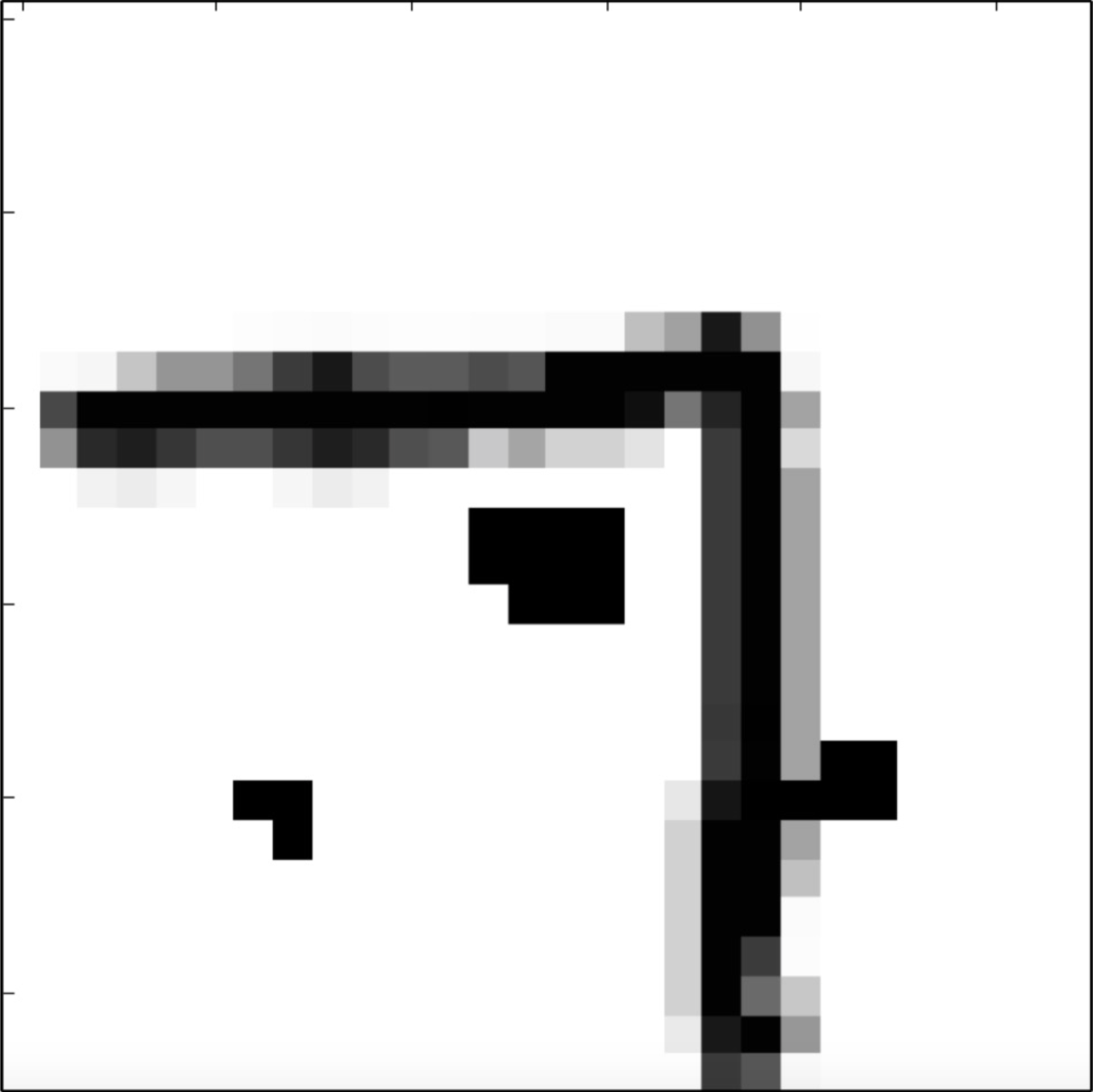}
\includegraphics[width=1.2cm]{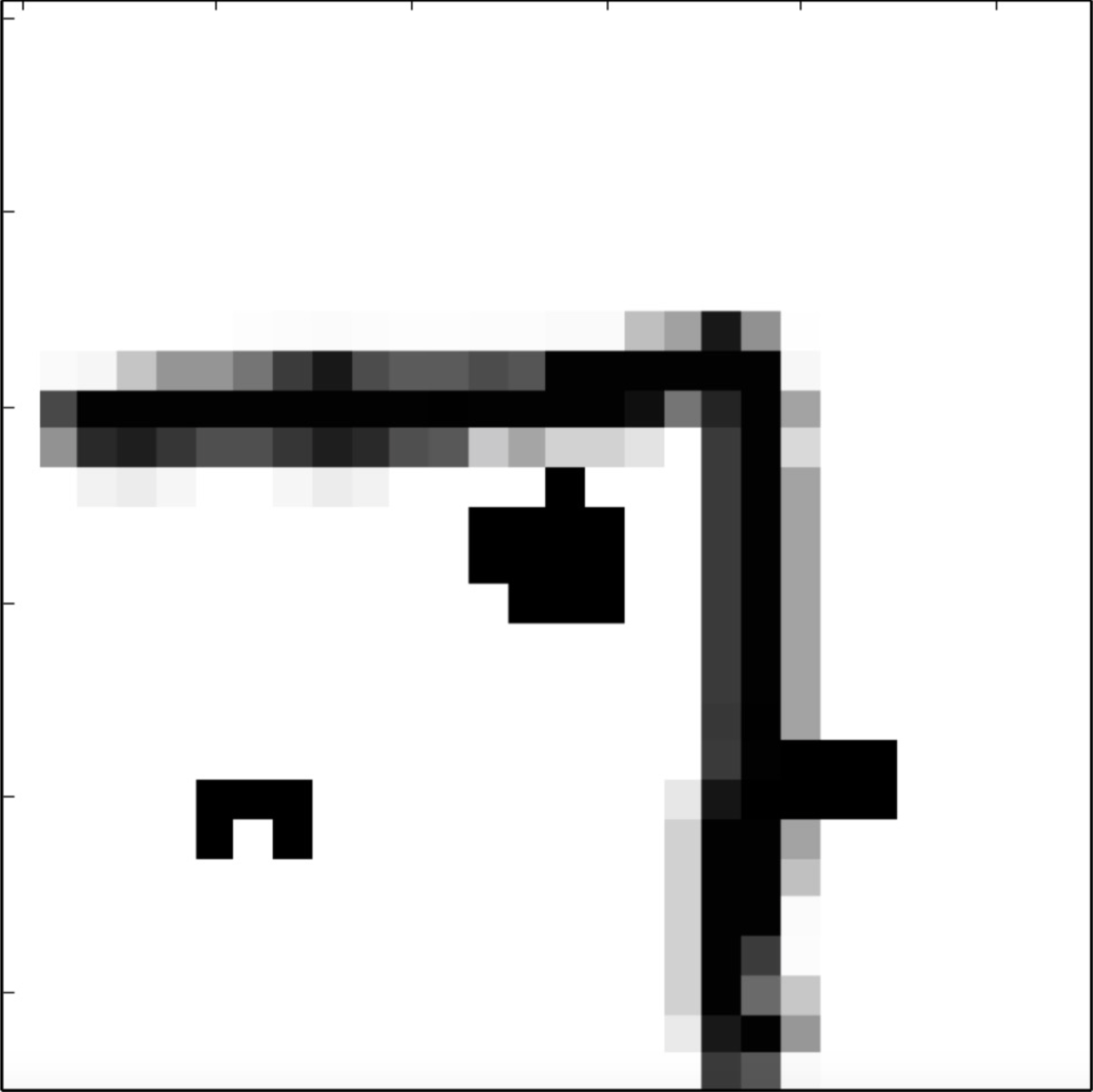}

\includegraphics[width=1.2cm]{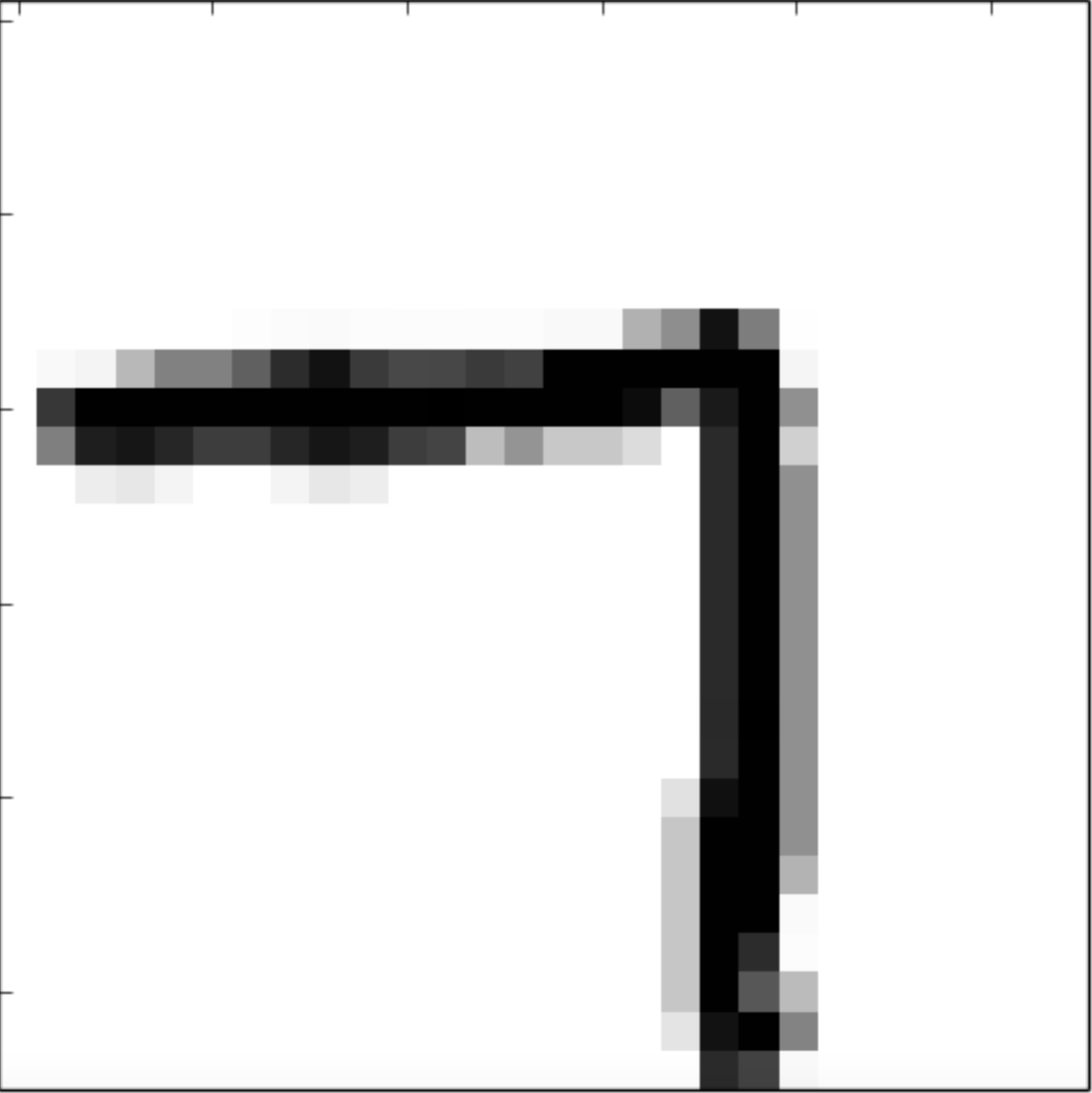}
\includegraphics[width=1.2cm]{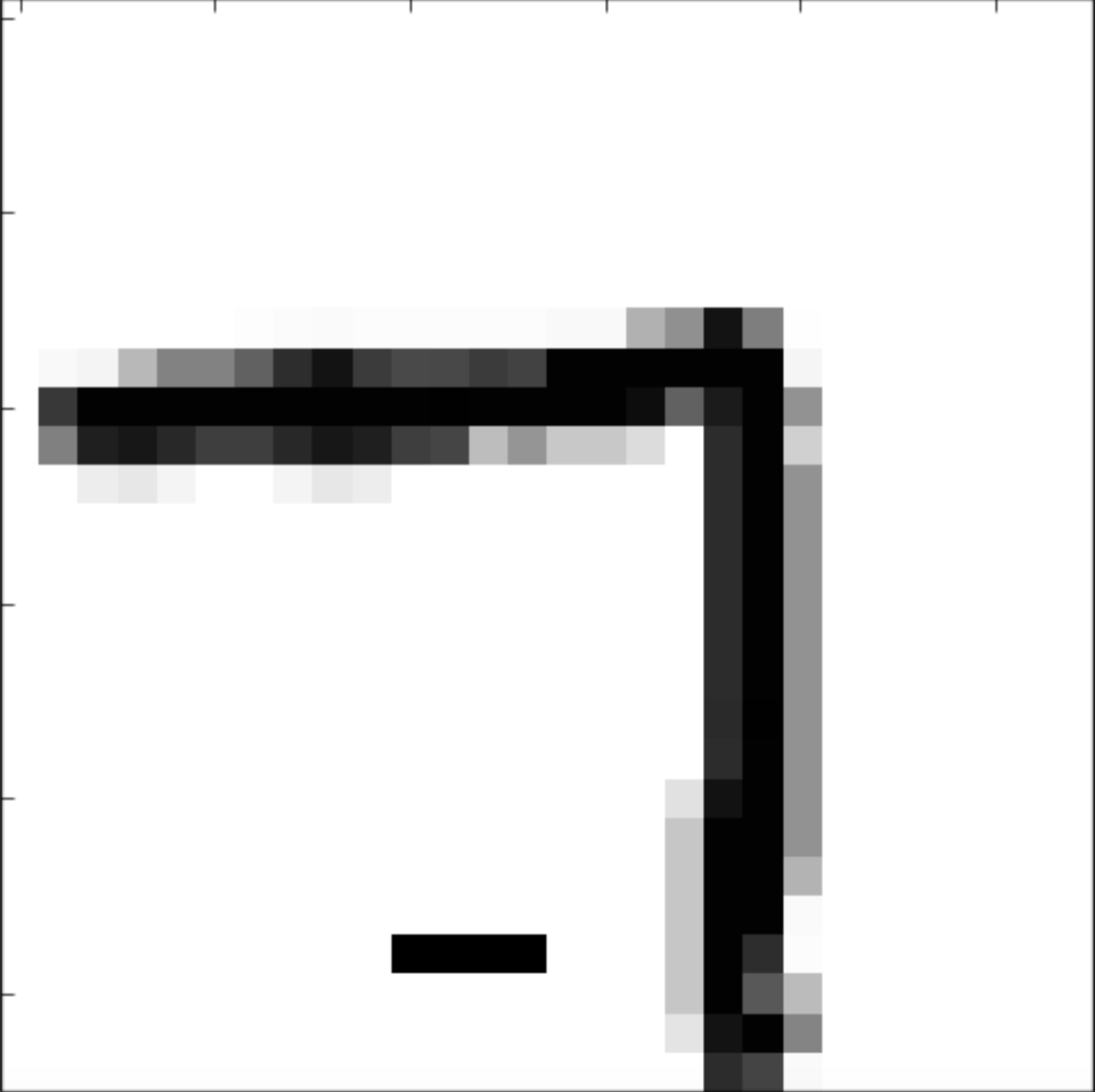}
\includegraphics[width=1.2cm]{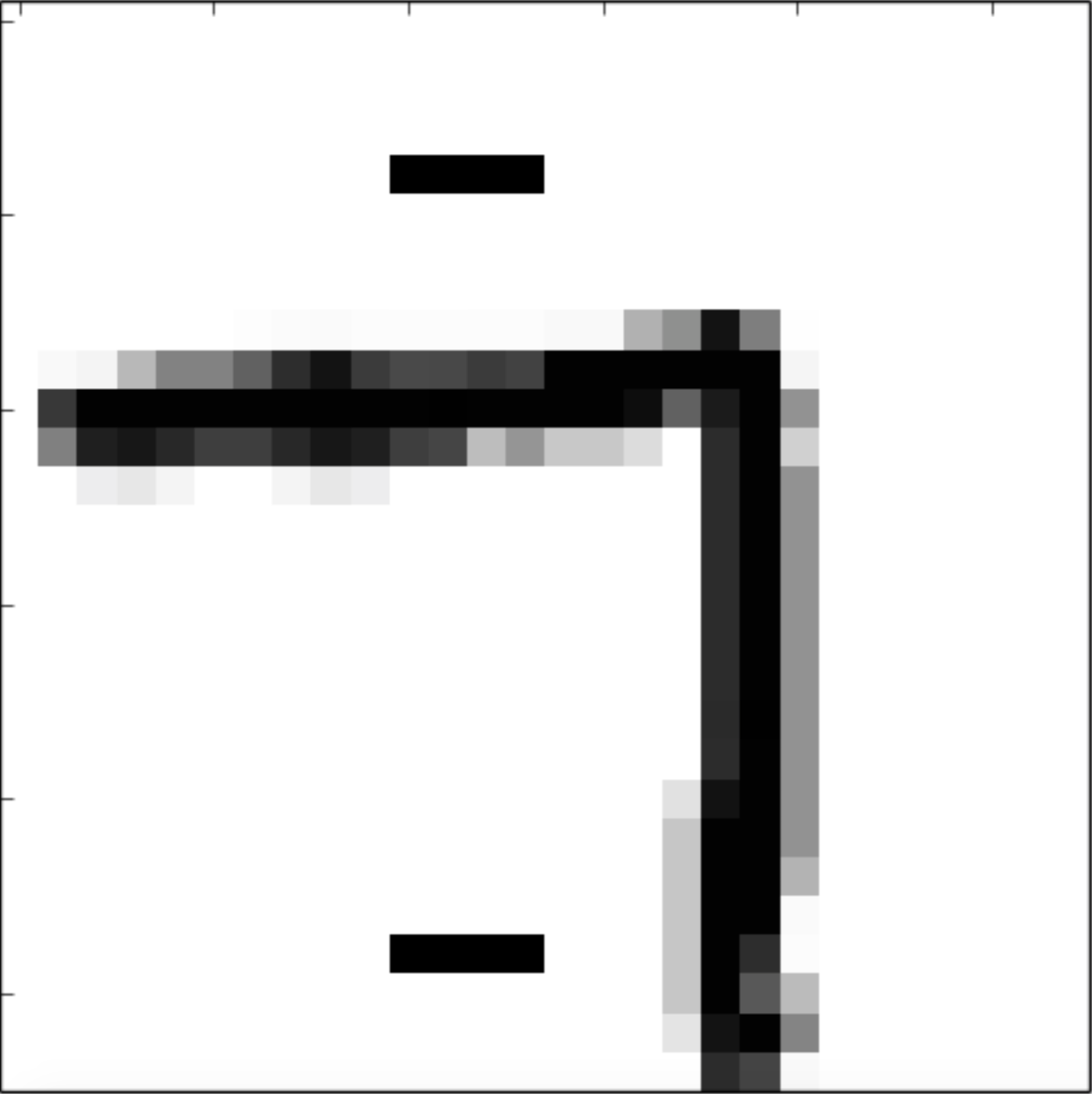}
\includegraphics[width=1.2cm]{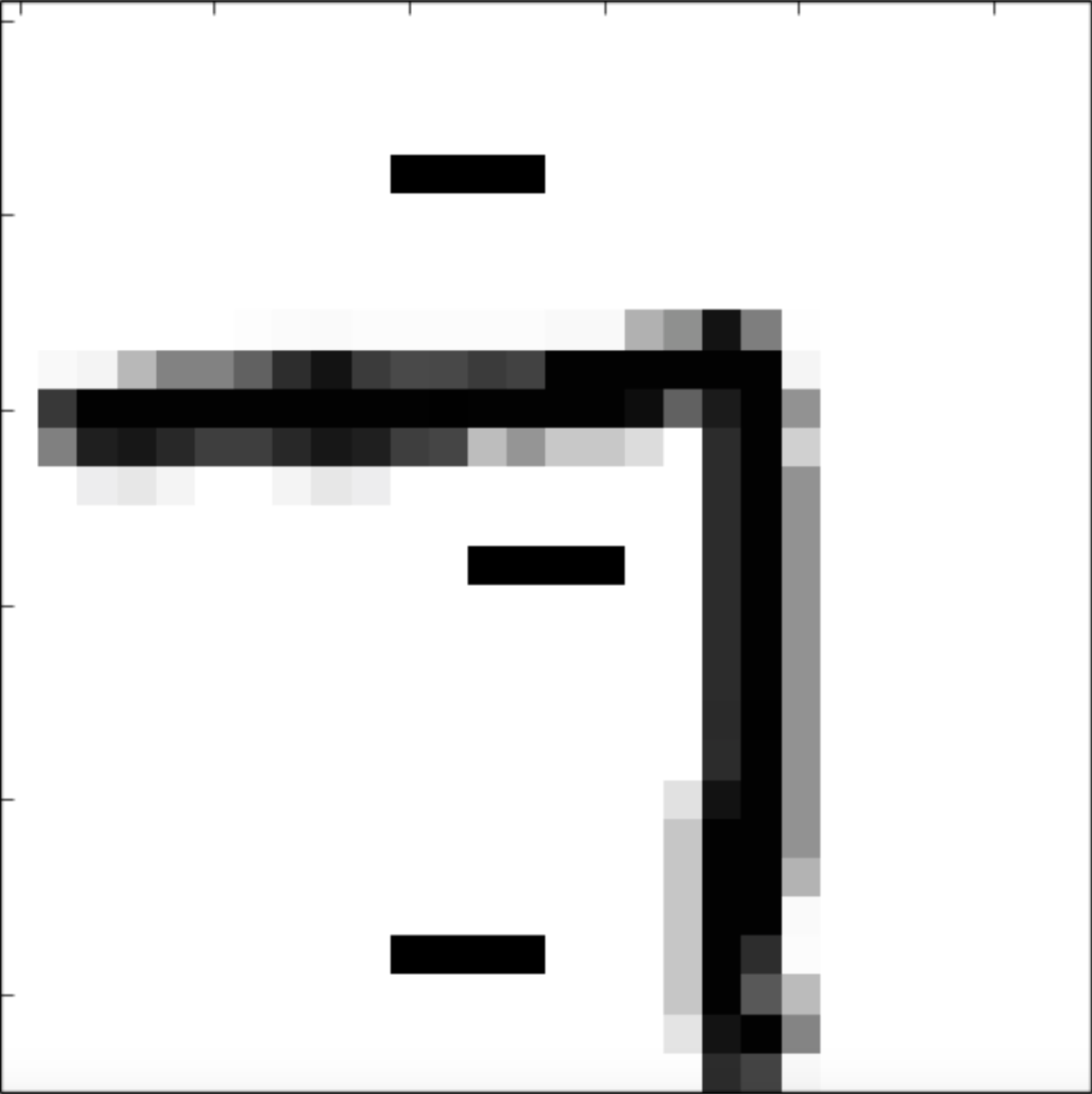}

\caption{FGSM vs. JSMA vs. DLV, where FGSM and JSMA search a single path and DLV multiple paths. 
Top row: Original image (7) perturbed deterministically by FGSM with $\epsilon=0.05, 0.1, 0.2, 0.3, 0.4$, with the final image (i.e., $\epsilon=0.4$) misclassified as 9.
Middle row: Original image (7) perturbed deterministically by JSMA   
with $\epsilon=0.1$ and $\theta=1.0$. We show even numbered images of the 12 produced by JSMA, with the final image misclassified as 3.
Bottom row: Original image (7) perturbed nondeterministically by DLV, for the same manipulation on a single pixel as that of JSMA (i.e., $s_p*m_p = 1.0$) and working in the input layer, with the final image misclassified as 3.
}
\label{fig:manipulations}
\end{figure}

{\bf Experiment 1.}
We randomly select an image from the MNIST dataset. Figure~\ref{fig:manipulations} shows some intermediate and final images obtained by running the three approaches: FGSM, JSMA and DLV. 
FGSM has a single parameter, $\epsilon$, where a greater $\epsilon$ represents a greater perturbation along the gradient of cost function. Given an $\epsilon$, for each input example a perturbed example is returned  and we test whether it is an adversarial example by checking for misclassification against the original image. We gradually increase the parameter $\epsilon=0.05, 0.1, 0.2, 0.3, 0.4$, with the last image (i.e., $\epsilon=0.4$) witnessing a class change, see the images in the top row of Figure~\ref{fig:manipulations}. FGSM can efficiently manipulate a set of images, but it requires a relatively large manipulation to find a misclassification.

For the JSMA approach, we conduct the experiment on a setting with  parameters $\epsilon=0.1$
and $\theta=1.0$.
The parameter $\epsilon=0.1$ means that we only consider adversarial examples changing no more than 10\% of all the pixels, which is sufficient here. As stated in \cite{practical-blackbox}, the parameter $\theta=1.0$, which allows a maximum change to every pixel, can ensure that fewer pixels need to be changed. The approach takes a series of manipulations to gradually lead to a misclassification, see the images in the middle row of Figure~\ref{fig:manipulations}. 
The misclassified image has an 
$L^2$ (Euclidean) distance of 0.17 and an  $L^1$ (Manhattan) distance of 0.03 from the original image. While JSMA can find adversarial examples with smaller distance from the original image, it takes longer to manipulate a set of images. 

Both FGSM and JSMA follow their specific heuristics to deterministically explore the space of images. However, in some cases, the heuristics may omit better adversarial examples. In the experiment for DLV, instead of giving features a specific order and manipulating them sequentially, we allow the program to nondeterministically choose features. This is currently done by MCTS (Monte Carlo Tree Search), which has a theoretical guarantee of convergence for infinite sampling. Therefore, the high-dimensional space is explored by following many different paths. By taking the same manipulation on a single pixel as that of JSMA (i.e., $s_p*m_p = 1.0$) and working on the input layer, DLV is able to find another perturbed image that is also classified as 3 but has a smaller distance ($L^2$ distance is 0.14 and $L^1$ distance is 0.02) from the original image, see the images in the last row of Figure~\ref{fig:manipulations}.  In terms of the time taken to find an adversarial example, DLV may take longer than JSMA, since it searches over many different paths.  

\begin{table}
\center
\begin{tabular}{|c|c|c|c|c|c|c|c|c|}
\hline
 & FGSM ($\epsilon = 0.1$)   &   ($0.2$)  &   ($0.4$)  & DLV $(dims_l=75)$ & $(150)$  & $(450)$  & JSMA ($\theta = 0.1$) & ($0.4$)\\
\hline
$L^2$ & 0.08  &  0.15    & 0.32  & 0.19 & 0.22   & 0.27  & 0.11 & 0.11 \\
$L^1$ & 0.06    &  0.12   & 0.25  & 0.04 & 0.06   & 0.09  & 0.02 & 0.02\\
\% & 17.5\%   &   70.9\%    & 97.2\%  & 52.3\% & 79\%  & 98\%   & 92\% & 99\%\\

\hline
\end{tabular}
\caption{FGSM vs. DLV (on a single path) vs. JSMA}
\label{tab:fgsmdlv}
\label{tab:complexity}
\end{table}

{\bf Experiment 2.}
Table~\ref{tab:fgsmdlv} gives a comparison of robustness evaluation of the three appraoches on the MNIST dataset. 
For FGSM, we vary the input parameter $\epsilon$ according to the values
$\{0.1,0.2,0.4\}$. 
For DLV, we select regions as defined in Section~\ref{sec:selection}  
on a single path (by defining a specific order on the features and manipulating them sequentially) for the first hidden layer. The experiment is parameterised by varying the maximal number of dimensions to be changed, i.e., 
$dims_l\in \{75,150,450\}$. 
For each input image, an adversarial example is returned, if found, by manipulating fewer than the maximal number of dimensions. When the maximal number has been reached, DLV will report failure and return the last perturbed example.
For JSMA, the experiment is conducted by letting $\theta$ take the value in the set $\{0.1,0.4\}$ and setting $\epsilon$ to $1.0$.

We collect three statistics, i.e., the average $L^1$ distance over the adversarial examples, the average $L^2$ distance over the  adversarial examples, and the success rate of finding adversary examples. Let $L^d(x,\manipulation(x))$ for $d\in \{1,2\}$ be the distance between an input $x$ and the returned perturbed image $\manipulation(x)$, and $\mathrm{diff}(x,\manipulation(x))\in \{0,1\}$ be a Boolean value representing whether $x$ and $\manipulation(x)$ have different classes. We let 
$$L^d = \displaystyle \frac{\sum_{\text{x in test set}} \mathrm{diff}(x,\manipulation(x)) \times L^d(x,\manipulation(x))}{\sum_{\text{x in test set} }\mathrm{diff}(x,\manipulation(x))}$$
and 
$$
\% = \displaystyle \frac{\sum_{\text{x in test set}} \mathrm{diff}(x,\manipulation(x))}{\text{the number of examples in test set}}
$$
We note that the approaches yield different perturbed examples $\manipulation(x)$. 

The test set size is 500 images selected randomly. DLV takes 1-2 minutes to manipulate each input image in MNIST. JSMA takes about 10 minutes for each image, but it works for 10  classes, so the running time is similar to that of DLV. FGSM works with a set of images, so it is the fastest per image.

For the case when the success rates are very high, i.e., 97.2\% for FGSM with $\epsilon=0.4$, 98\% for DLV with $dims_l=450$, and 99\% for JSMA with $\theta = 0.4$, JSMA has the smallest average distances, followed by DLV, which has smaller average distances than FGSM on both $L^1$ and $L^2$ distances. 

We mention that a smaller distance leading to a misclassification may result in a lower rate of transferability~\cite{practical-blackbox}, meaning 
that a misclassification can be harder to witness on another model trained on the same (or a small subset of) data-set.

\section{Related Work}\label{sec:related}
AI safety is recognised an an important problem, see e.g.,~\cite{DBLP:journals/corr/SeshiaS16,DBLP:journals/corr/AmodeiOSCSM16}.
An early verification approach
for neural networks was proposed in~\cite{PT2010}, where, using the notation of this paper, safety is defined as the existence, for all inputs in a region $\eta_0\in D_{L_0}$, of a corresponding output in another region $\eta_n\subseteq D_{L_n}$. They encode the entire network as a set of constraints, approximating the sigmoid using constraints, which can then be solved by a SAT solver, but their approach only works with 6 neurons (3 hidden neurons). A similar idea is presented in \cite{SWWB2015}. In contrast, we work layer by layer and obtain much greater scalability. Since the first version of this paper appeared~\cite{HKWW2016}, another constraint-based method has been proposed in~\cite{KBDJK2017} which improves on \cite{PT2010}. While they consider more general correctness properties than this paper, they can only handle the ReLU activation functions, by extending the Simplex method 
to work with the piecewise linear ReLU functions that cannot be expressed using linear programming. 
This necessitates a search tree (instead of a search path as in Simplex), for which a heuristic search is proposed and shown to be complete. The approach is demonstrated on networks with 300 ReLU nodes, but as it encodes the full network it is unclear whether it can be scaled to work with practical deep neural networks: for example, the MNIST network has 630,016 ReLU nodes. They also handle continuous spaces directly without discretisation, the benefits of which are not yet clear, since it is argued in~\cite{DBLP:journals/corr/GoodfellowSS14} that linear behaviour in high-dimensional spaces is sufficient to cause adversarial examples.  

Concerns about the instability of neural networks to adversarial examples were first raised in~\cite{Biggio2013,SZSBEGF2014}, where
optimisation is used to identify misclassifications. A method for computing the perturbations is also proposed, which is based on box-constrained optimisation and is approximate in view of non-convexity of the search space.
This work is followed by \cite{DBLP:journals/corr/GoodfellowSS14}, which introduced the much faster FGSM method, and \cite{KGB2016}, which employed a compromise between the two (iterative, but with a smaller number of iterations than \cite{SZSBEGF2014}).
In our notation, \cite{DBLP:journals/corr/GoodfellowSS14} 
uses a \emph{deterministic, iterative} manipulation
$
\manipulation(x) = x + \epsilon sign(\triangledown_xJ(x,\activation_{x,n})),
$
where $x$ is an image in matrix representation, $\epsilon$ is a hyper-parameter that can be tuned to get different manipulated images, and $J(x,\activation_{x,n})$ is the cross-entropy cost function of the neural network on input $x$ and class $\activation_{x,n}$. Therefore, their approach will test a set of discrete points in the region $\eta_0(\activation_{x,0})$ of the input layer.
Therefore these manipulations
will test a lasso-type ladder tree (i.e., a  ladder tree without branches) $\ladderset(\eta_k(\activation_{x,k}))$, which does not satisfy the covering property. %
In \cite{NYC2015}, instead of working with a single image, an evolutionary algorithm is employed for a population of images.
For each individual image in the current population, the manipulation is the mutation and/or crossover. While mutations can be \emph{nondeterministic}, the manipulations of an individual image are also following a lasso-type ladder tree which is not covering. We also mention that \cite{ZSLG2016} uses several  distortions such as JPEG compression, thumbnail resizing, random cropping, etc, to test the robustness of the trained network. These distortions can be understood as manipulations.
All these attacks do not leverage any specific properties of the model family, and do not guarantee that they will find a misclassified image in the constraint region, even if such an image exists.

The notion of robustness studied in \cite{fross-theory} 
has some similarities to our definition of safety, except that the authors work with values \emph{averaged} over the input distribution $\mu$, which is difficult to estimate accurately in high dimensions. 
As in \cite{SZSBEGF2014,KGB2016}, they use optimisation without convergence guarantees, as a result computing only an approximation to the minimal perturbation. 
In~\cite{constraints} pointwise robustness is adopted, which corresponds to our general safety; they also use a constraint solver but represent the full constraint system 
by reduction to a convex LP problem, and only verify an approximation of the property.
In contrast, we work directly with activations rather than an encoding of activation functions, and our method 
\emph{exhaustively} searches through the complete ladder tree for an adversarial example by iterative and nondeterministic application of manipulations. Further, our definition of a manipulation is more flexible, since it allows us to select a \emph{subset} of dimensions, and each such subset can have a different region diameter computed with respect to a different norm.

\section{Conclusions}

This paper presents an automated verification framework for checking safety of deep neural networks that is based on a systematic exploration of a region around a data point to search for adversarial manipulations of a given type, and propagating the analysis into deeper layers. Though we focus on the classification task, the approach also generalises to other types of networks.
We have implemented the approach using SMT and validated it on several state-of-the-art neural network classifiers for realistic images. The results are encouraging, with adversarial examples found in some cases in a matter of seconds when working with few dimensions, but the verification process itself is exponential in the number of features and has prohibitive complexity for larger images. The performance and scalability of our method can be significantly improved through parallelisation. It would be interesting to see if the notions of regularity suggested in~\cite{Mallat2016} permit a symbolic approach, and whether an abstraction refinement framework can be formulated to improve the scalability and computational performance.

{\bf Acknowledgements}. 
This paper has greatly benefited from discussions with several researchers. We are particularly grateful to Martin Fraenzle, Ian Goodfellow and Nicolas Papernot.

\bibliographystyle{plain}
\bibliography{trust}

\newpage

\appendix

\section{Input Parameters and Experimental Setup}

The DLV tool accepts as input a network $N$ and an image $x$, and has the following input parameters:
\begin{itemize}
\item an integer $l\in [0,n]$ indicating the starting layer $L_l$,
\item an integer $dims_l\geq 1$ indicating the maximal number of dimensions that need to be considered in layer $L_l$,
\item the values of variables $s_p$ and $m_p$  in $V_l$; for simplicity, we ask that, for all dimensions $p$ that will be selected by the automated procedure, $s_p$ and $m_p$ have the same values,
\item the precision $\varepsilon \in [0,\infty)$, 
\item an integer $dims_{k,f}$ indicating the number of dimensions for each feature; for simplicity, we ask that every feature has the same number of dimensions and $dims_{k,f}=dims_{k',f}$ for all layers $k$ and $k'$, and
\item type of search: either heuristic (single-path) or Monte Carlo Tree Search (MCTS) (multi-path). 
\end{itemize}

\subsection{Two-Dimensional Point Classification Network}

\begin{itemize}
\item $l=0$
\item $dims_l = 2$,
\item $s_p=1.0$ and $m_p=1.0$,
\item $\varepsilon =0.1$, and
\item $dims_{k,f} = 2$
\end{itemize}

\subsection{Network for the MNIST Dataset}

\begin{itemize}
\item $l=1$
\item $dims_l = 150$,
\item $s_p=1.0$ and $m_p=1.0$,
\item $\varepsilon = 1.0$, and
\item $dims_{k,f} = 5$
\end{itemize}

\subsection{Network for the CIFAR-10 Dataset}

\begin{itemize}
\item $l=1$
\item $dims_l = 500$,
\item $s_p=1.0$ and $m_p=1.0$,
\item $\varepsilon = 1.0$, and
\item $dims_{k,f} = 5$
\end{itemize}

\subsection{Network for the GTSRB Dataset}

\begin{itemize}
\item $l=1$
\item $dims_l = 1000$,
\item $s_p=1.0$ and $m_p=1.0$,
\item $\varepsilon = 1.0$, and
\item $dims_{k,f} = 5$
\end{itemize}

\subsection{Network for the ImageNet Dataset}

\begin{itemize}
\item $l=2$
\item $dims_l = 20,000$,
\item $s_p=1.0$ and $m_p=1.0$,
\item $\varepsilon = 1.0$, and
\item $dims_{k,f} = 5$
\end{itemize}


\section{Additional Adversarial Examples Found for the CIFAR-10, ImageNet, and MNIST Networks}\label{app:advers}

\begin{figure}
\parbox{2.9cm}{
\includegraphics[width=1.4cm,height=1.4cm]{images/cifar10/390_original_as_automobile.pdf}
\includegraphics[width=1.4cm,height=1.4cm]{images/cifar10/390_automobile_modified_into_bird.pdf}
{automobile to bird}
}
\parbox{2.9cm}{
\includegraphics[width=1.4cm,height=1.4cm]{images/cifar10/204_original_as_automobile.pdf}
\includegraphics[width=1.4cm,height=1.4cm]{images/cifar10/204_automobile_modified_into_frog.pdf}
{automobile to frog }
}
\parbox{2.9cm}{
\includegraphics[width=1.4cm,height=1.4cm]{images/cifar10/201automobile.pdf}
\includegraphics[width=1.4cm,height=1.4cm]{images/cifar10/201automobileToairplane.pdf}
{automobile to airplane }
}
\parbox{2.9cm}{
\includegraphics[width=1.4cm,height=1.4cm]{images/cifar10/193_original_as_automobile.pdf}
\includegraphics[width=1.4cm,height=1.4cm]{images/cifar10/193_automobile_modified_into_horse.pdf}
{automobile to horse }
}\\
\parbox{2.9cm}{
\includegraphics[width=1.4cm,height=1.4cm]{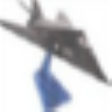}
\includegraphics[width=1.4cm,height=1.4cm]{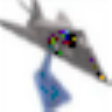}
{airplane to dog }
}
\parbox{2.9cm}{
\includegraphics[width=1.4cm,height=1.4cm]{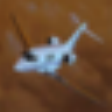}
\includegraphics[width=1.4cm,height=1.4cm]{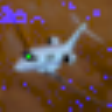}
{airplane to deer }
}
\parbox{2.9cm}{
\includegraphics[width=1.4cm,height=1.4cm]{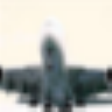}
\includegraphics[width=1.4cm,height=1.4cm]{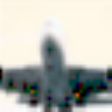}
{airplane to truck }
}
\parbox{2.9cm}{
\includegraphics[width=1.4cm,height=1.4cm]{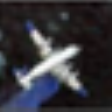}
\includegraphics[width=1.4cm,height=1.4cm]{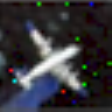}
{airplane to cat }
}\\
\parbox{2.9cm}{
\includegraphics[width=1.4cm,height=1.4cm]{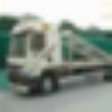}
\includegraphics[width=1.4cm,height=1.4cm]{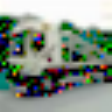}
{truck to frog }
}
\parbox{2.9cm}{
\includegraphics[width=1.4cm,height=1.4cm]{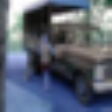}
\includegraphics[width=1.4cm,height=1.4cm]{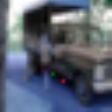}
{truck to cat }
}
\parbox{2.9cm}{
\includegraphics[width=1.4cm,height=1.4cm]{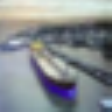}
\includegraphics[width=1.4cm,height=1.4cm]{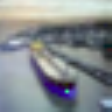}
{ship to bird }
}
\parbox{2.9cm}{
\includegraphics[width=1.4cm,height=1.4cm]{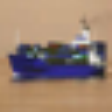}
\includegraphics[width=1.4cm,height=1.4cm]{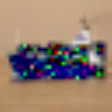}
{ship to airplane }
}\\
\parbox{2.9cm}{
\includegraphics[width=1.4cm,height=1.4cm]{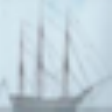}
\includegraphics[width=1.4cm,height=1.4cm]{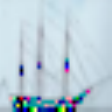}
{ship to truck}
}
\parbox{2.9cm}{
\includegraphics[width=1.4cm,height=1.4cm]{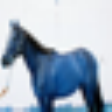}
\includegraphics[width=1.4cm,height=1.4cm]{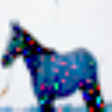}
{horse to cat }
}
\parbox{2.9cm}{
\includegraphics[width=1.4cm,height=1.4cm]{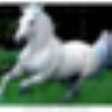}
\includegraphics[width=1.4cm,height=1.4cm]{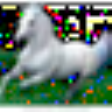}
{horse to automobile}
}
\parbox{2.9cm}{
\includegraphics[width=1.4cm,height=1.4cm]{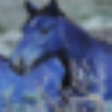}
\includegraphics[width=1.4cm,height=1.4cm]{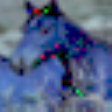}
{horse to truck}
}
\caption{Adversarial examples for a neural network trained on the CIFAR-10 dataset by single-path search}
\label{fig:cifar10}
\end{figure}

\begin{figure}
\parbox{6.2cm}{
\includegraphics[width=3cm,height=3cm]{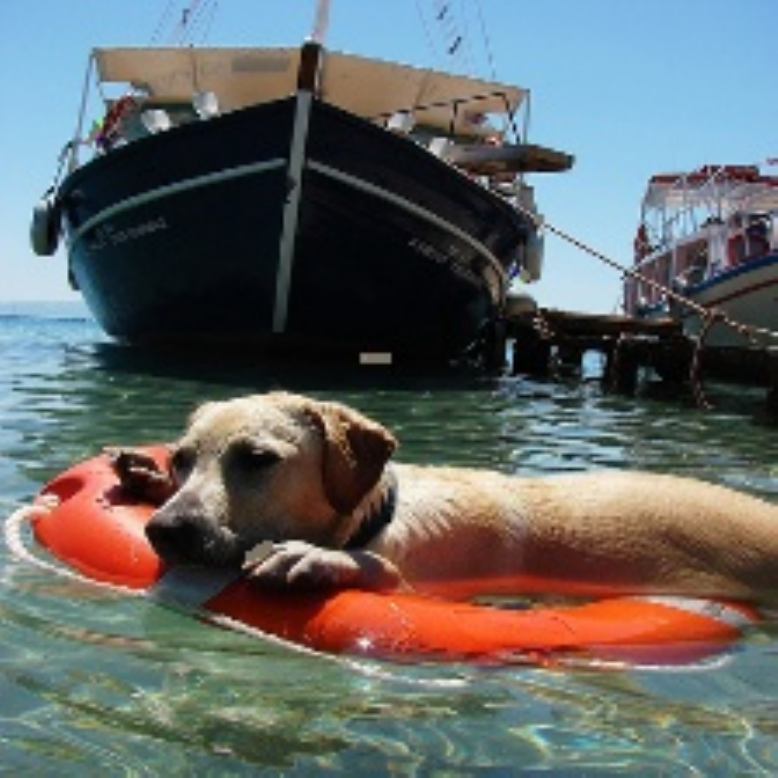}
\includegraphics[width=3cm,height=3cm]{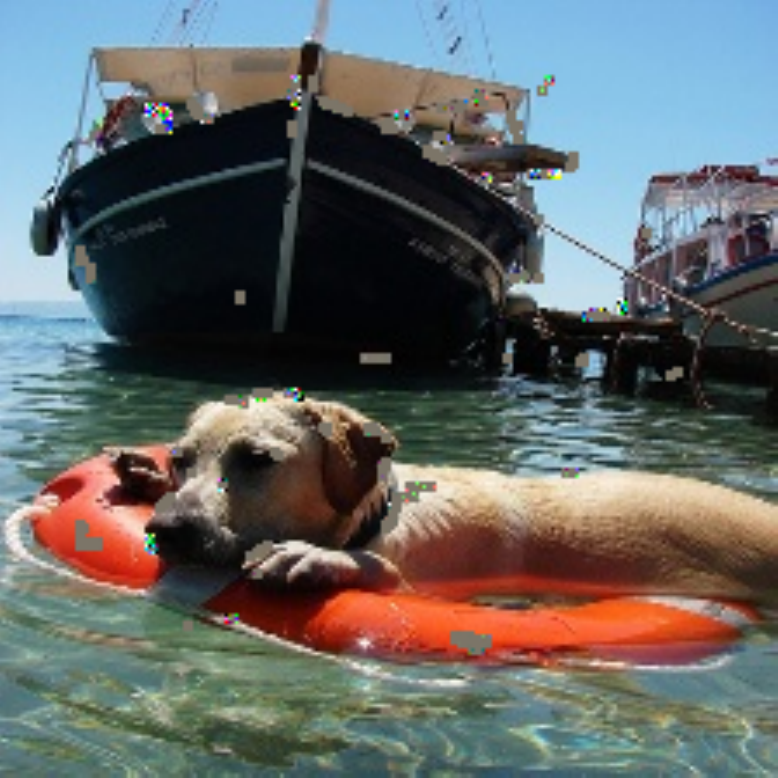}
{labrador to life boat}
}
\parbox{6.2cm}{
\includegraphics[width=3cm,height=3cm]{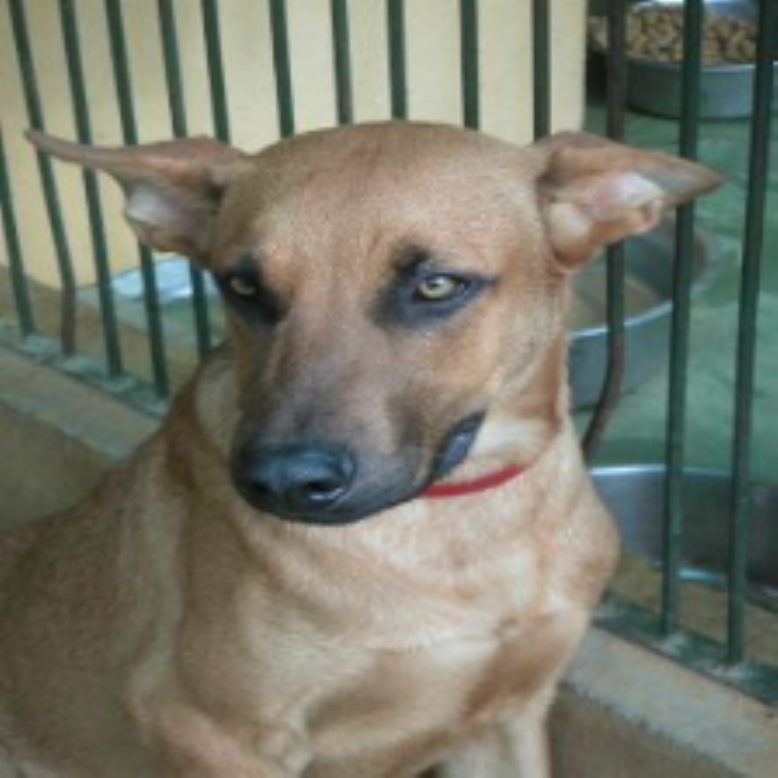}
\includegraphics[width=3cm,height=3cm]{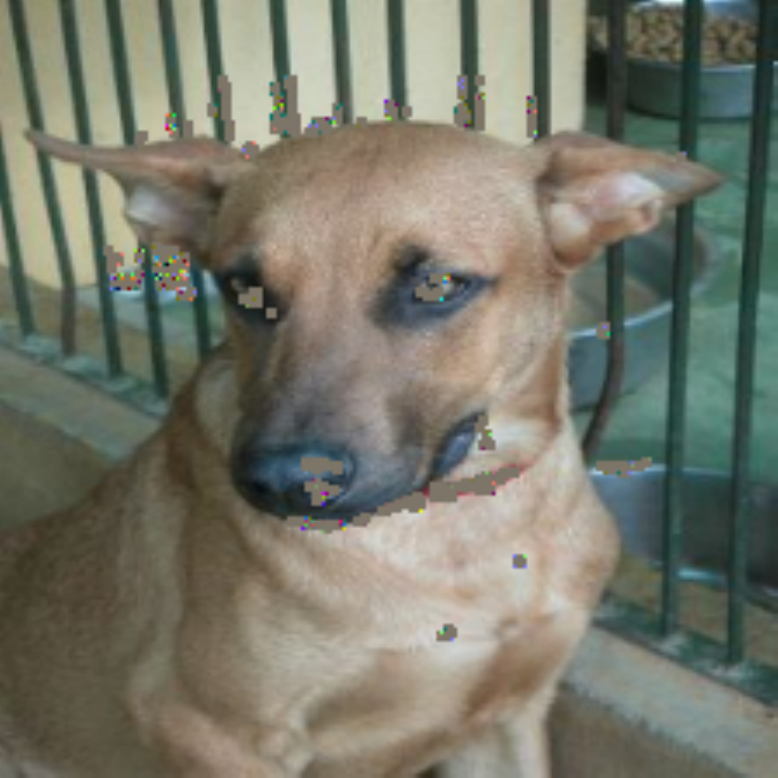}
{rhodesian  ridgeback to malinois}
}

\parbox{6.2cm}{
\includegraphics[width=3cm,height=3cm]{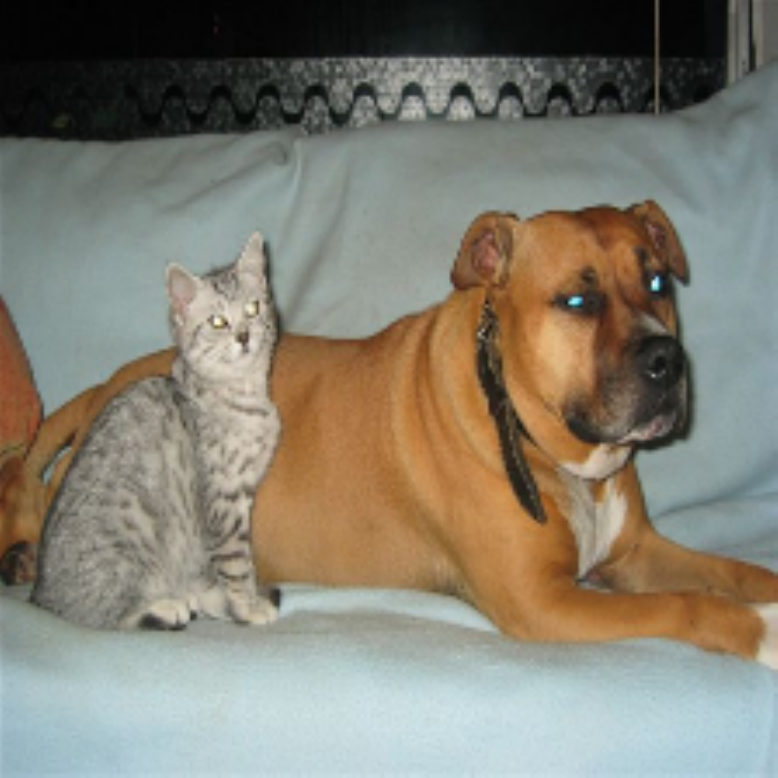}
\includegraphics[width=3cm,height=3cm]{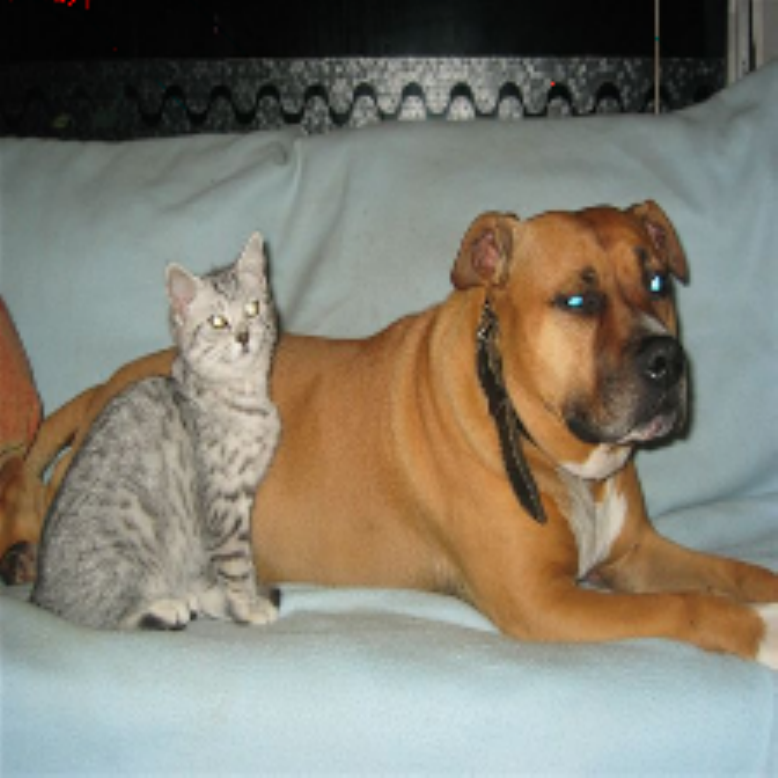}
{boxer to rhodesian ridgeback}
}
\parbox{6.2cm}{
\includegraphics[width=3cm,height=3cm]{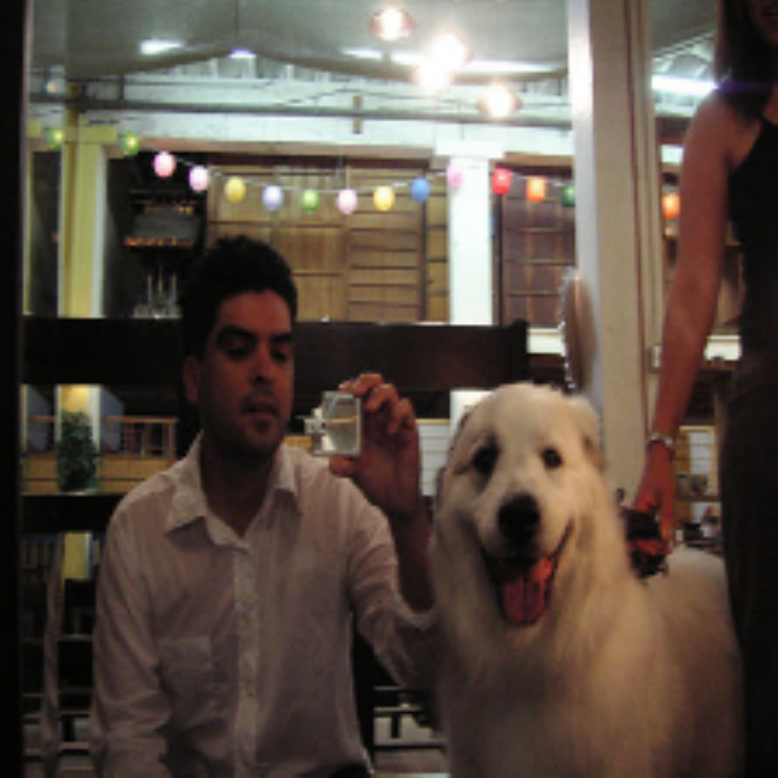}
\includegraphics[width=3cm,height=3cm]{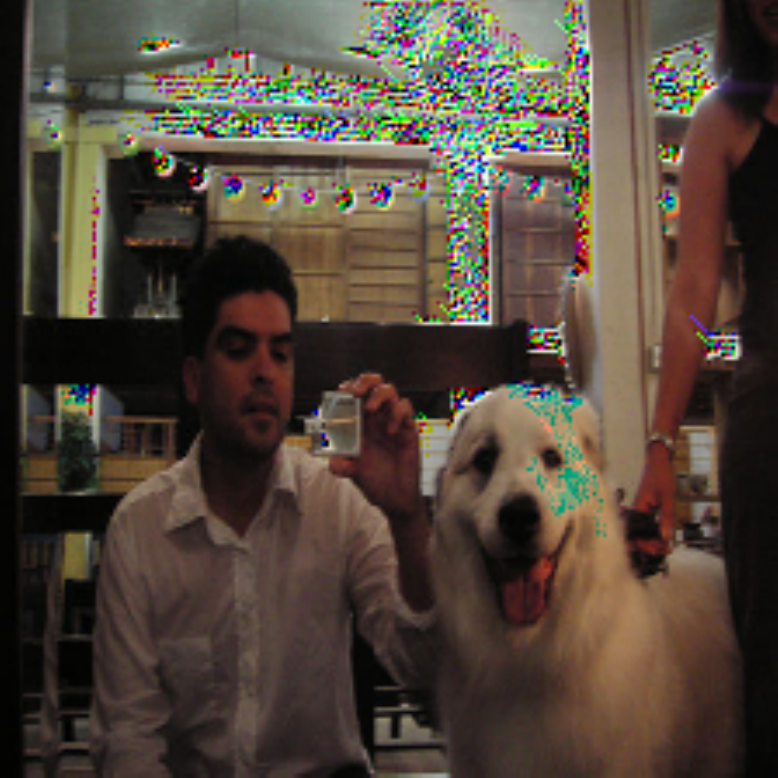}
{great pyrenees to kuvasz}
}
\caption{Adversarial Examples for the VGG16 Network Trained on the imageNet Dataset By Single-Path Search}
\label{fig:imageNet}
\end{figure}

\begin{figure}
\centering
\parbox{2.3cm}{
\includegraphics[width=1.1cm]{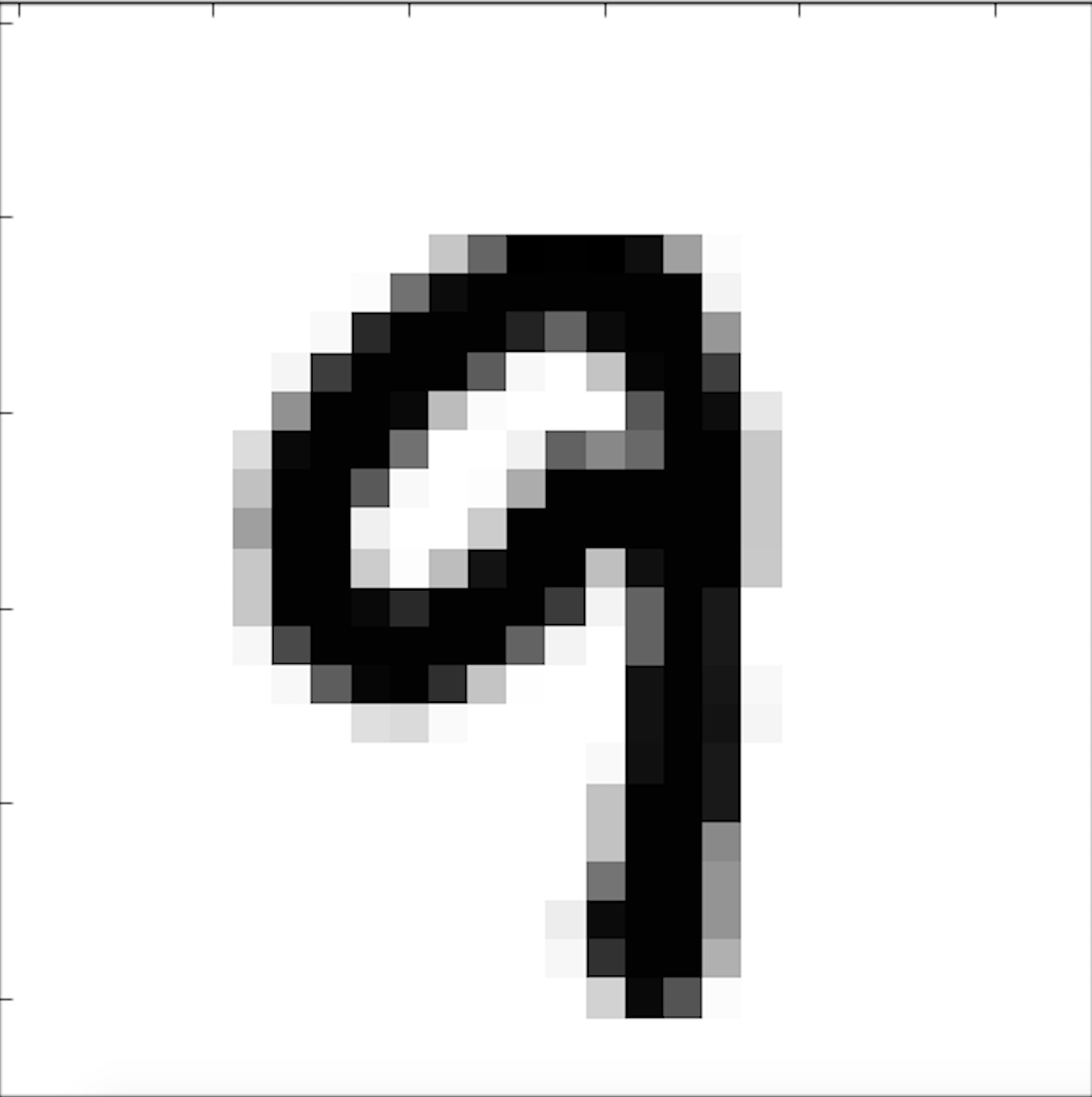}
\includegraphics[width=1.1cm]{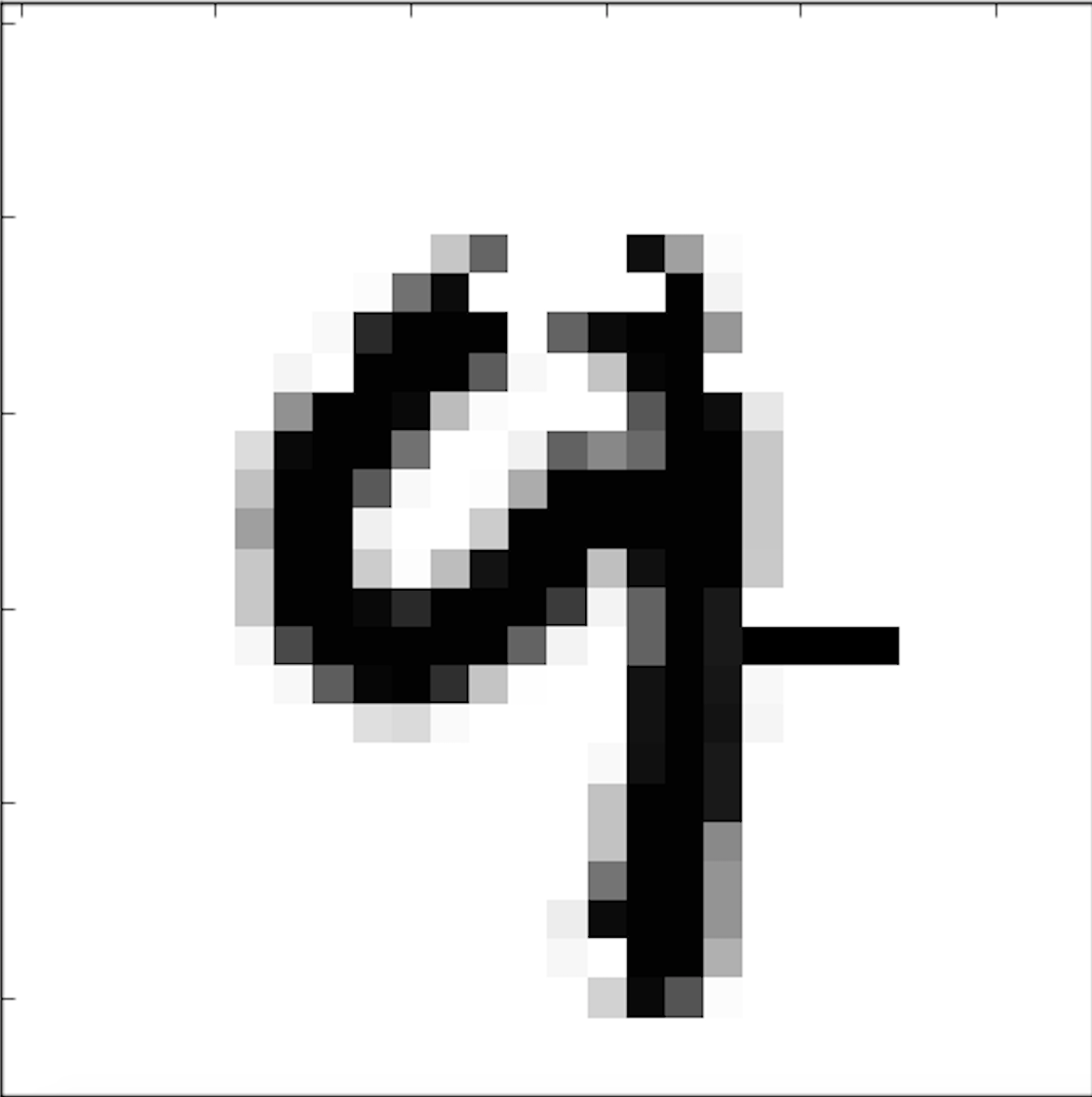}
\text{\hspace{0.8cm}9 to 4}
}
\parbox{2.3cm}{
\includegraphics[width=1.1cm]{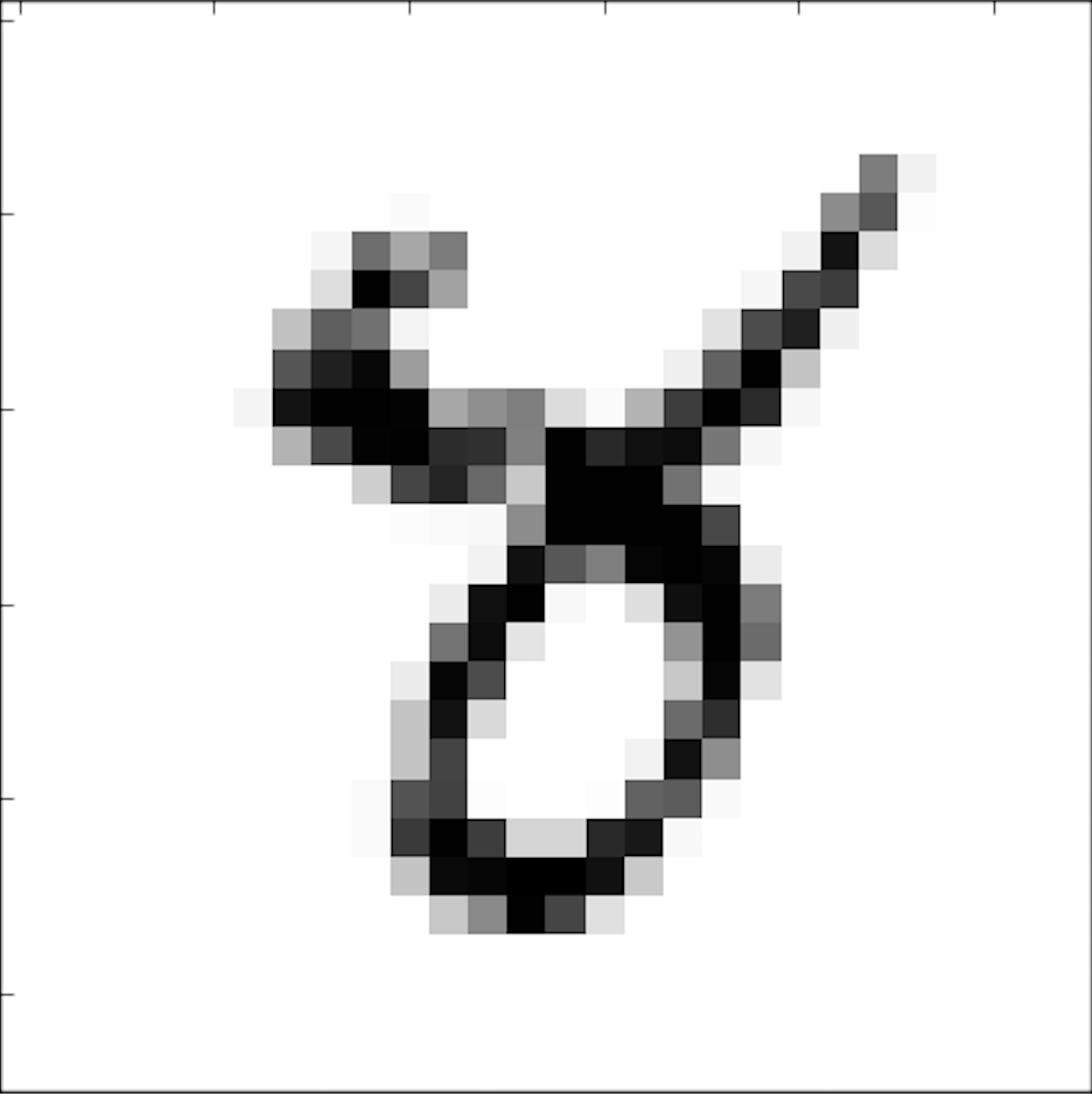}
\includegraphics[width=1.1cm]{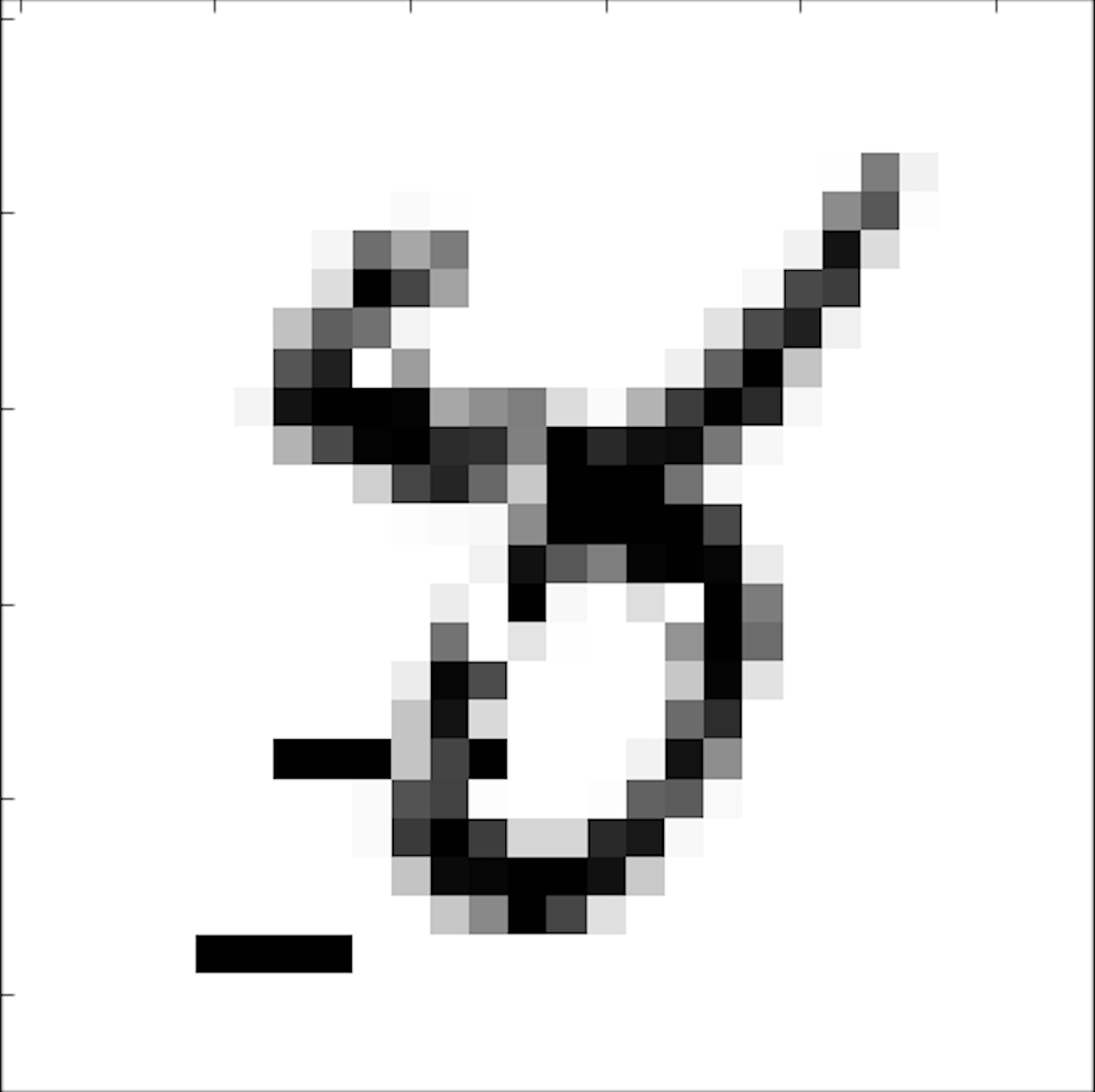}
\text{\hspace{0.8cm}8 to 3}
}
\parbox{2.3cm}{
\includegraphics[width=1.1cm]{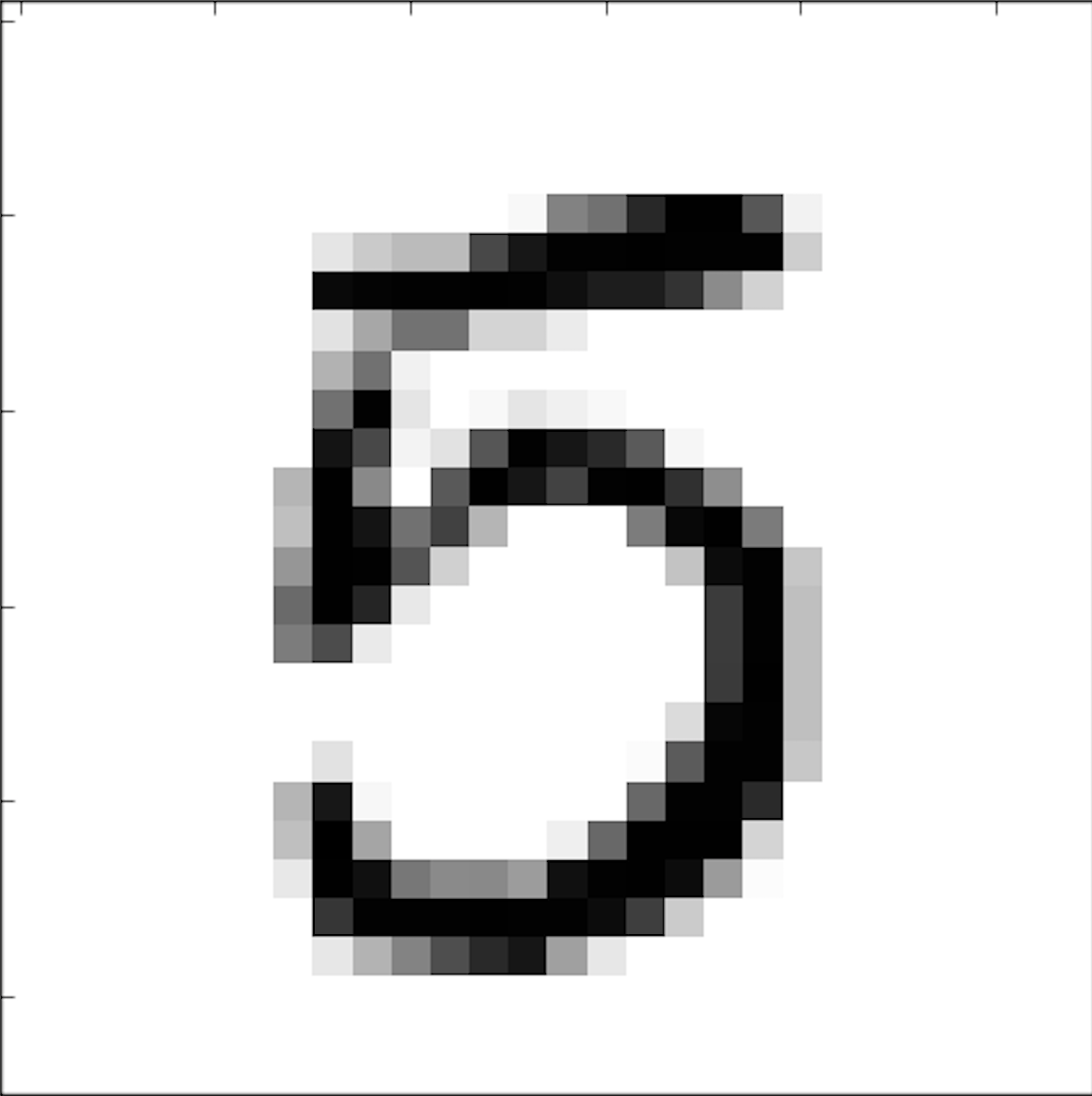}
\includegraphics[width=1.1cm]{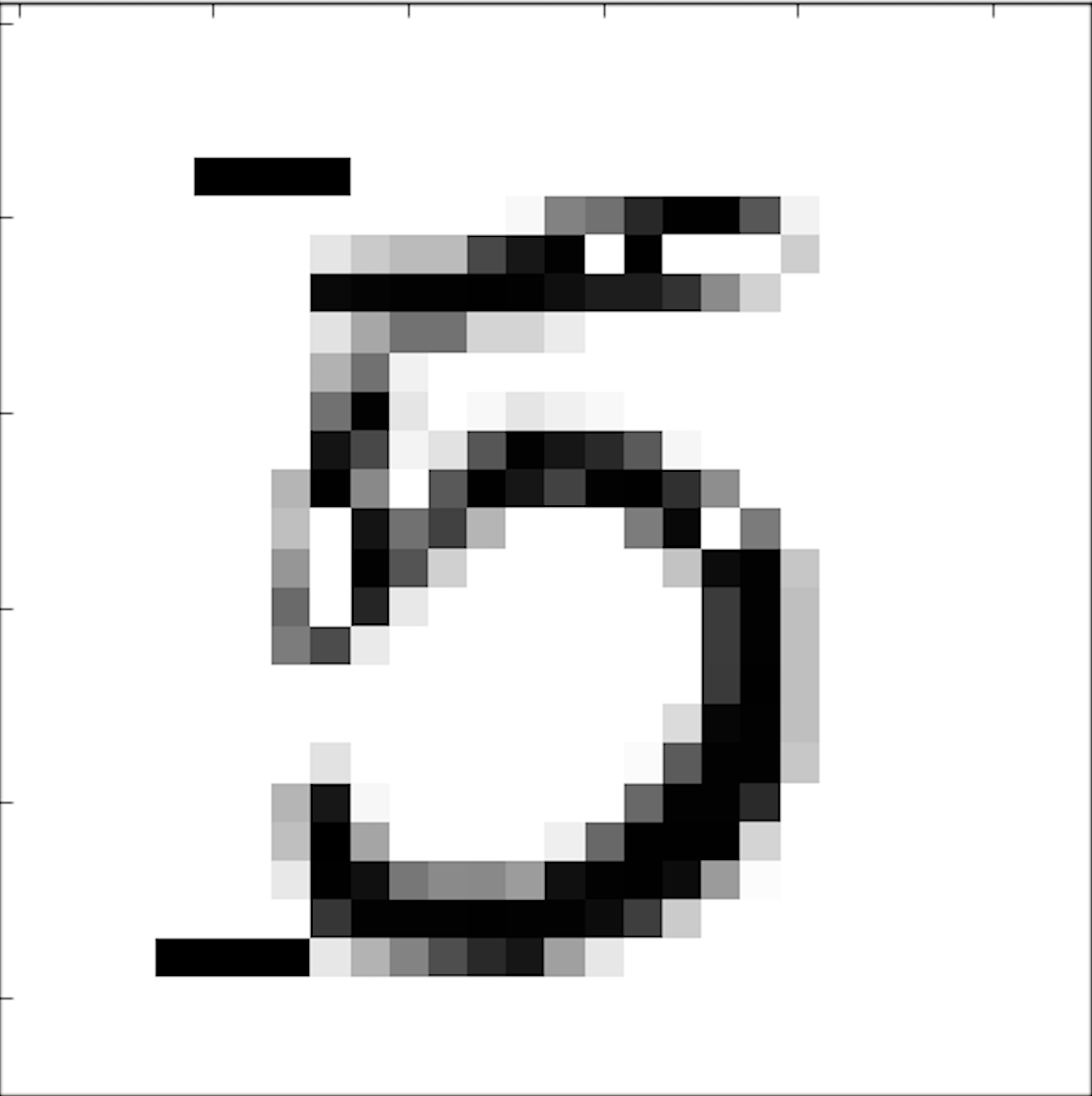}
\text{\hspace{0.8cm}5 to 3}
}
\parbox{2.3cm}{
\includegraphics[width=1.1cm]{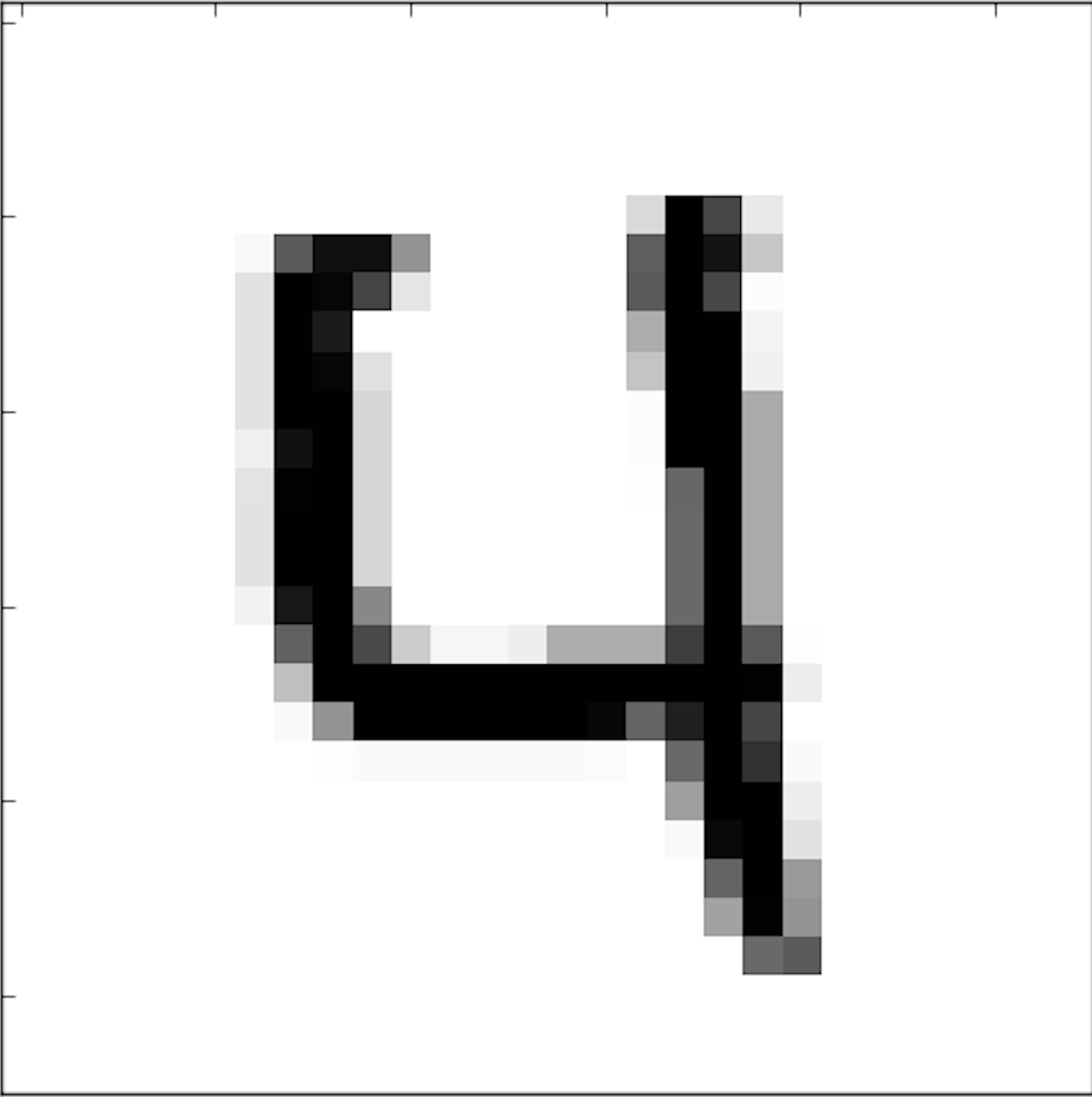}
\includegraphics[width=1.1cm]{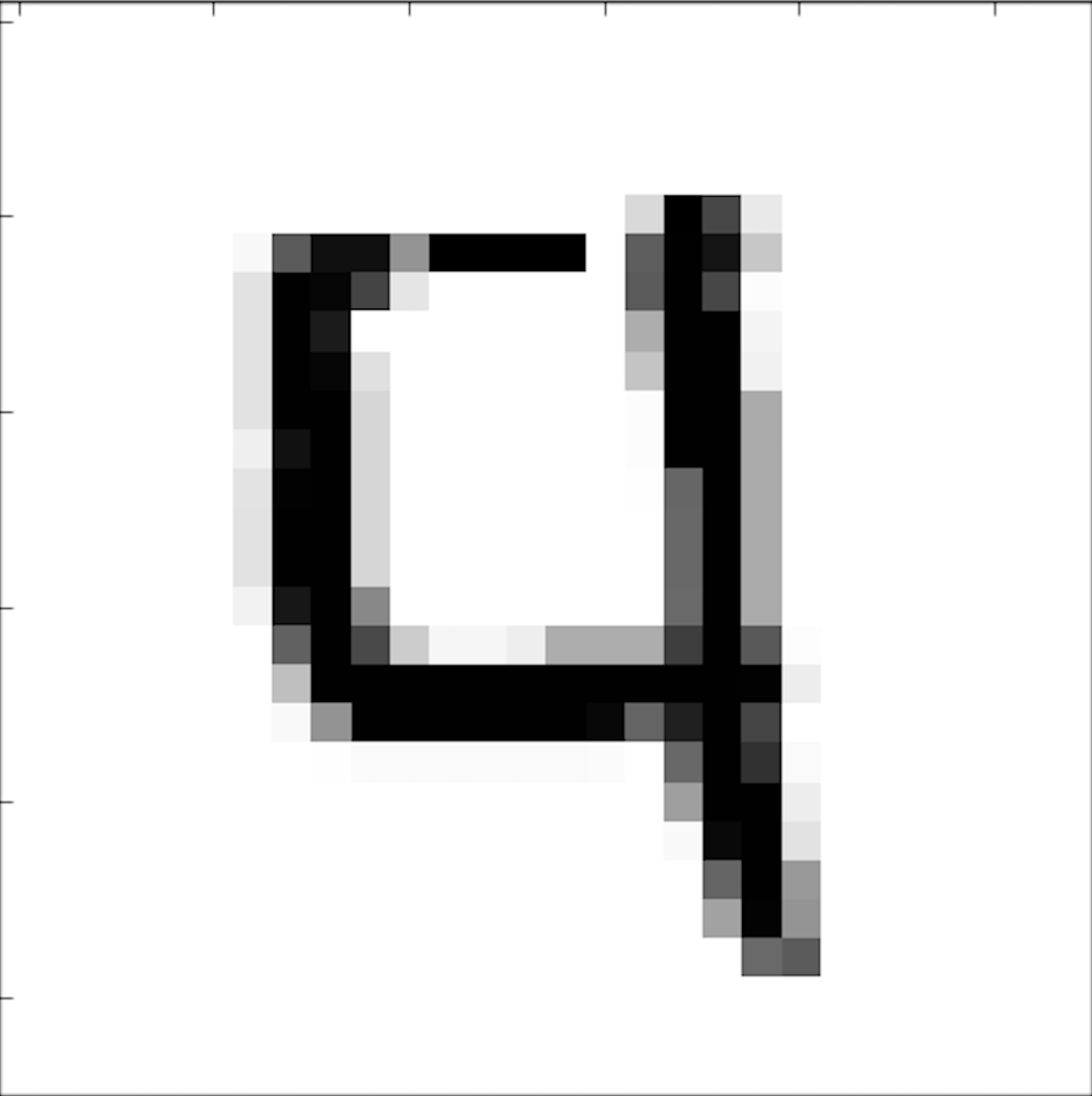}
\text{\hspace{0.8cm}4 to 9}
}
\parbox{2.3cm}{
\includegraphics[width=1.1cm]{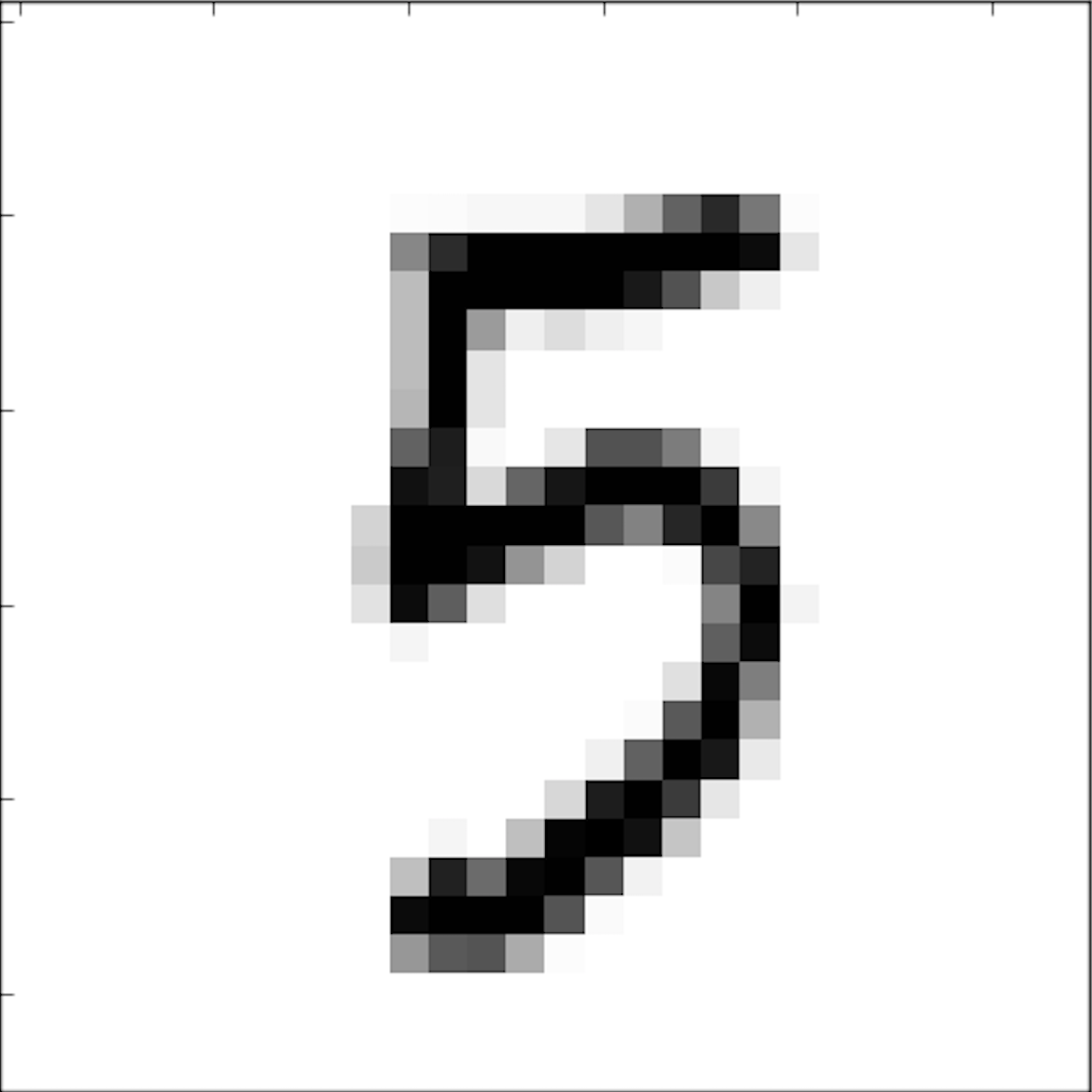}
\includegraphics[width=1.1cm]{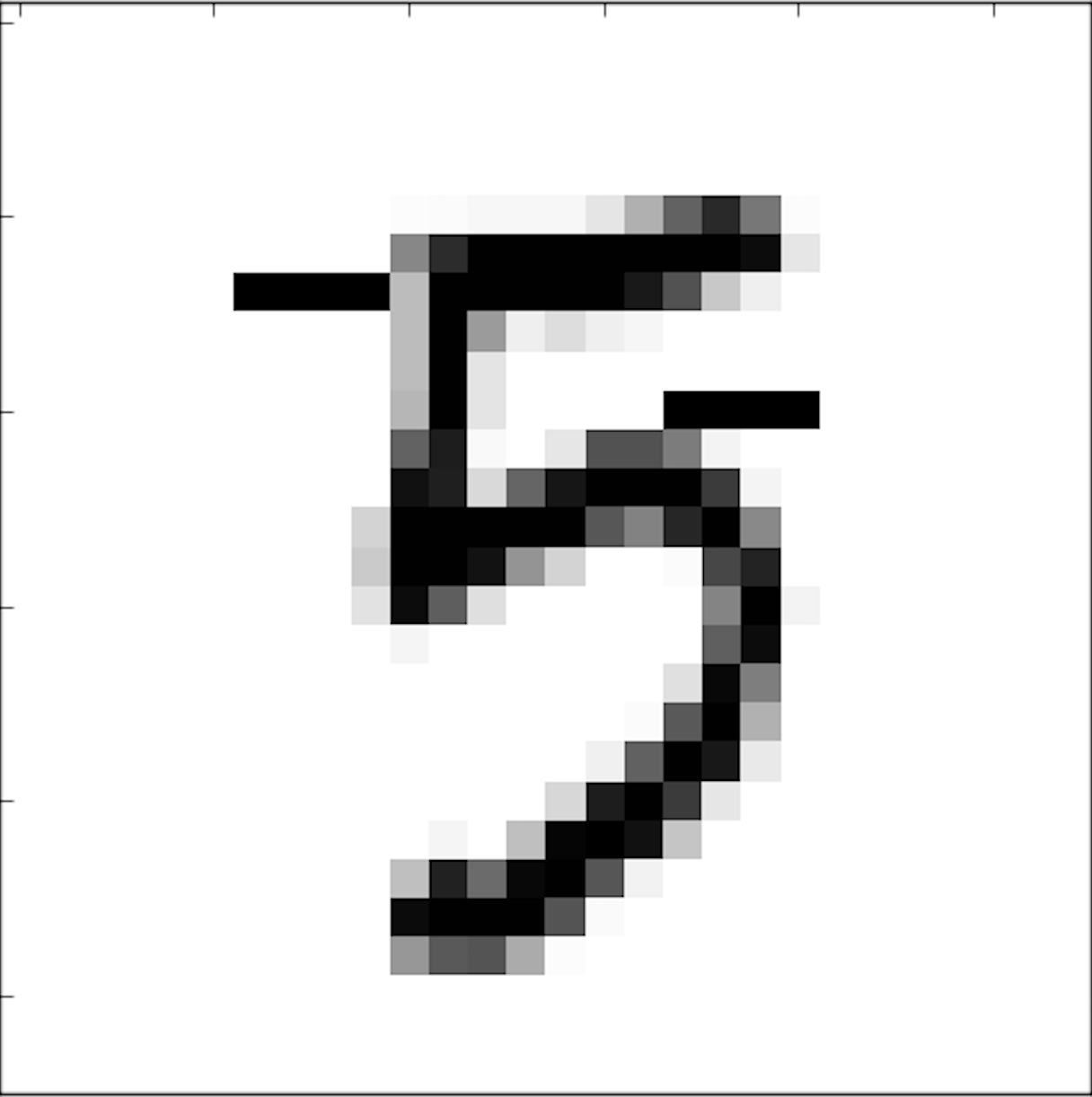}
\text{\hspace{0.8cm}5 to 3}
}\\

\parbox{2.3cm}{
\includegraphics[width=1.1cm]{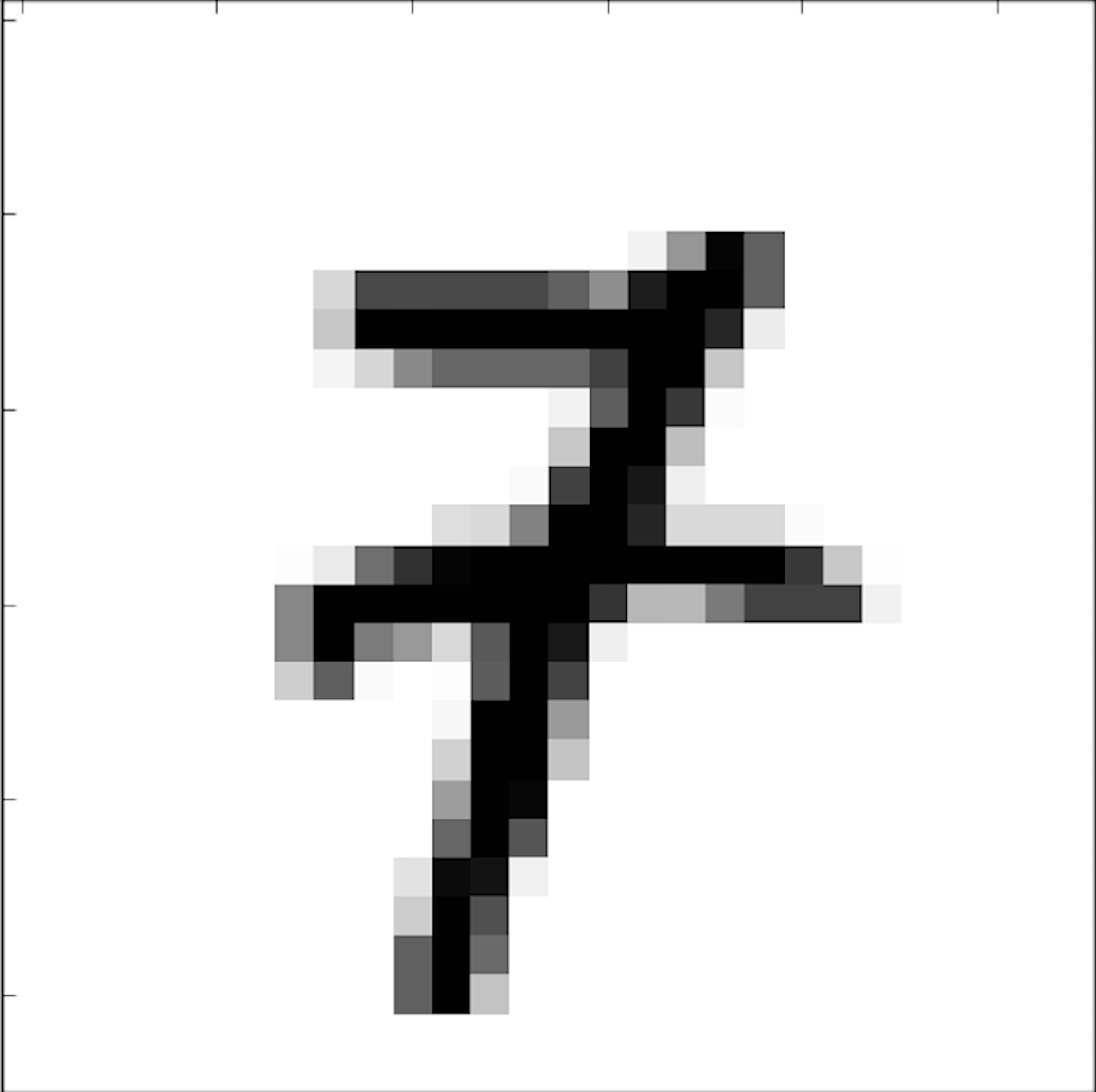}
\includegraphics[width=1.1cm]{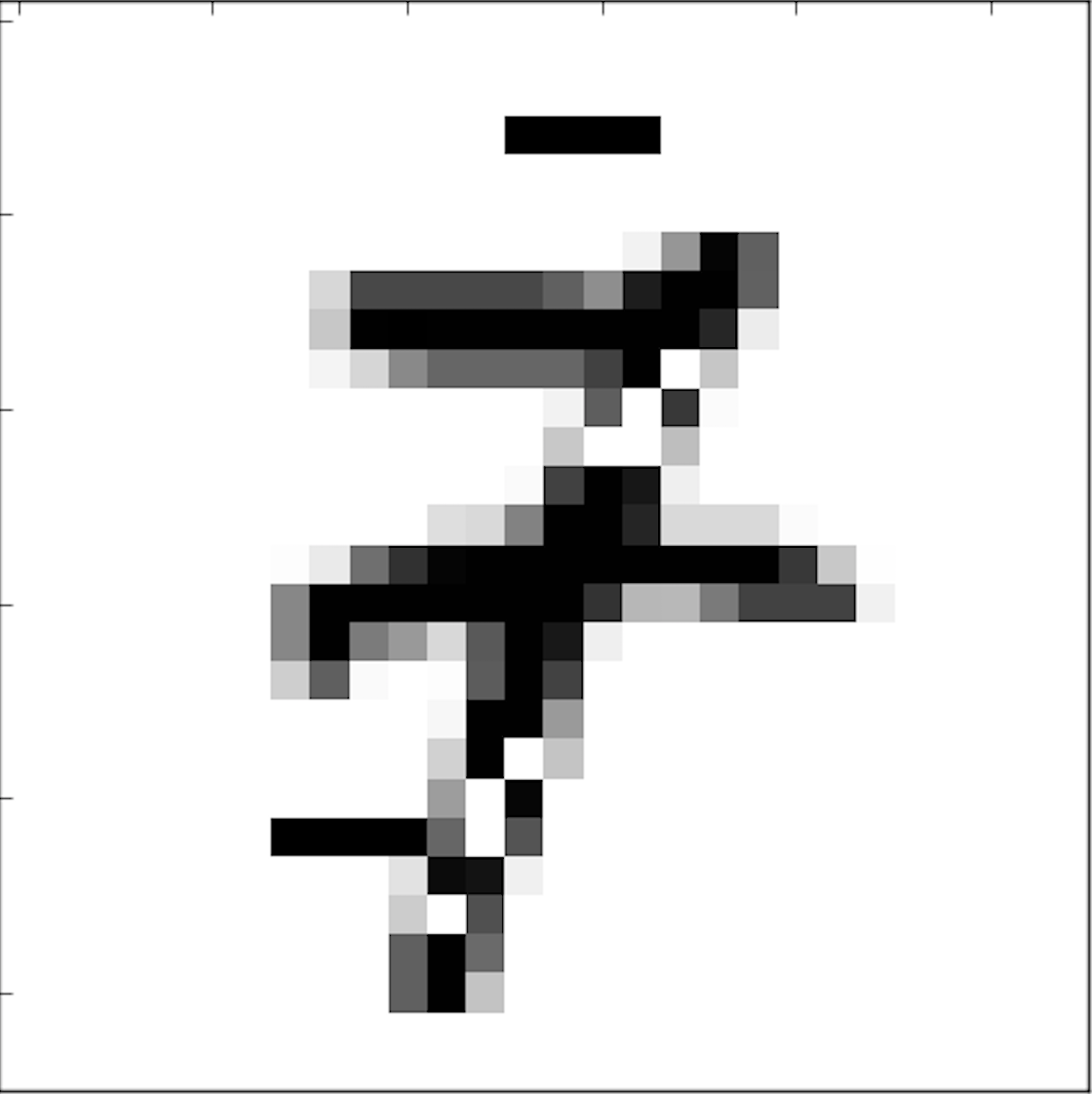}
\text{\hspace{0.8cm}7 to 3}
}
\parbox{2.3cm}{
\includegraphics[width=1.1cm]{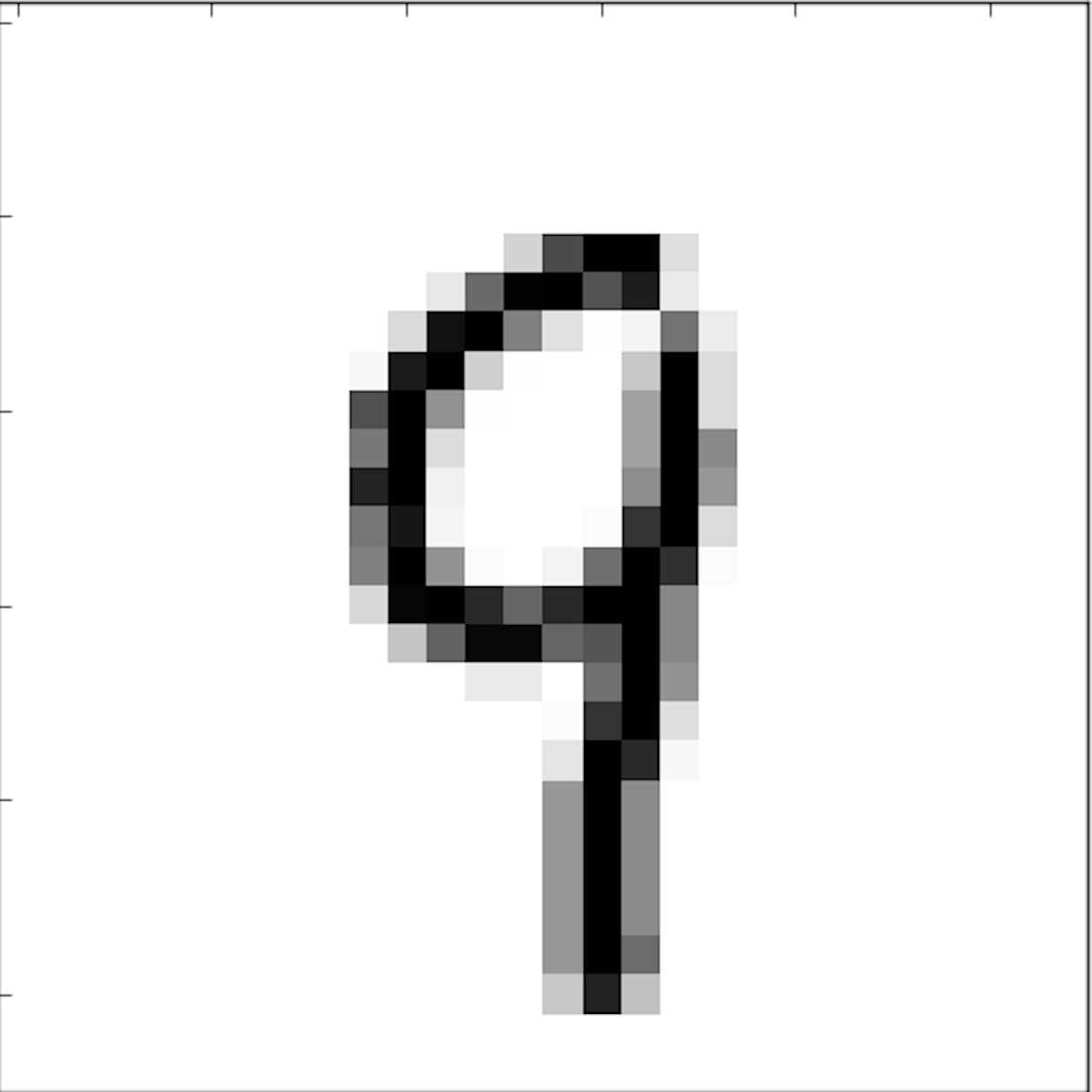}
\includegraphics[width=1.1cm]{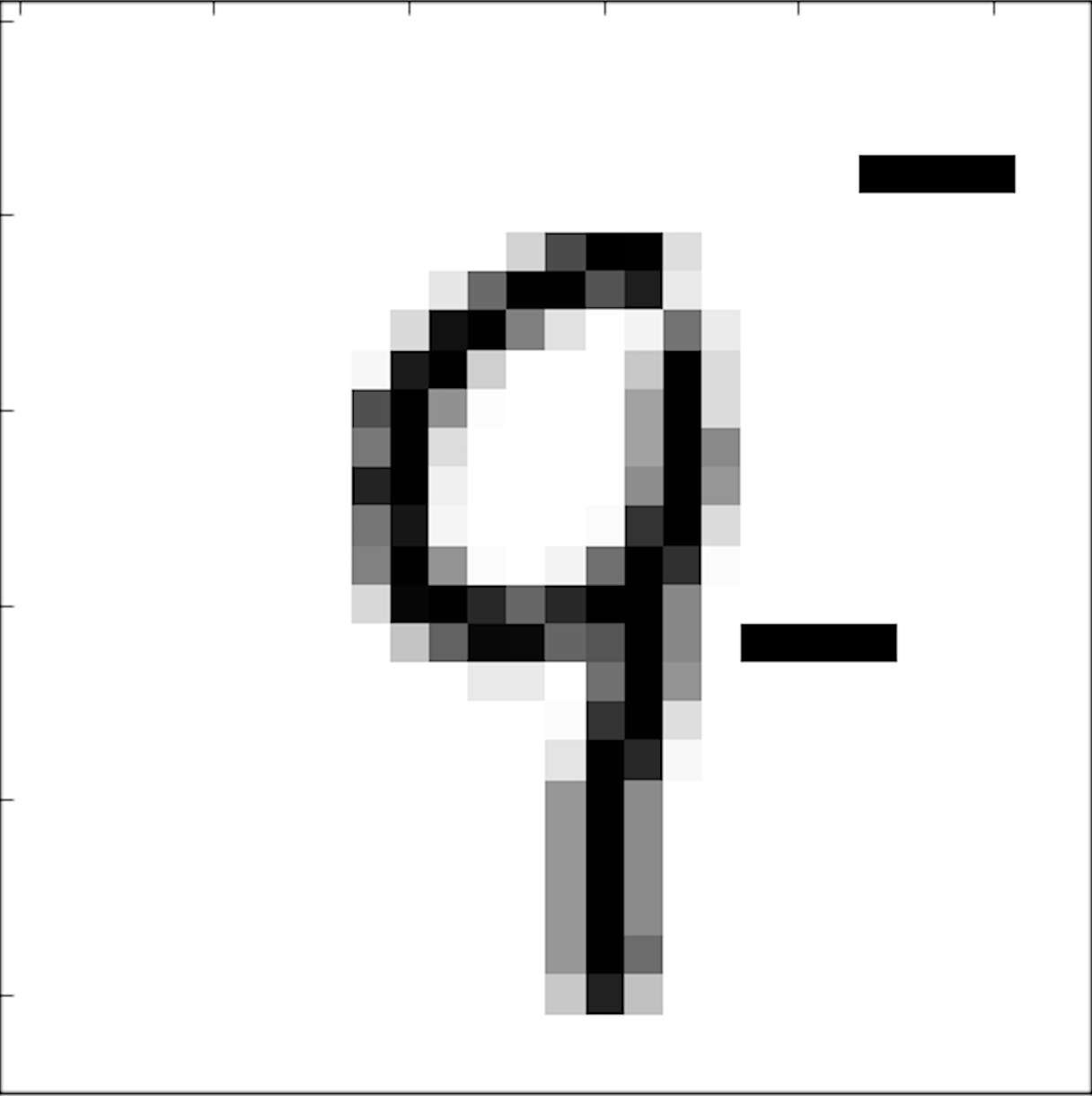}
\text{\hspace{0.8cm}9 to 4}
}
\parbox{2.3cm}{
\includegraphics[width=1.1cm]{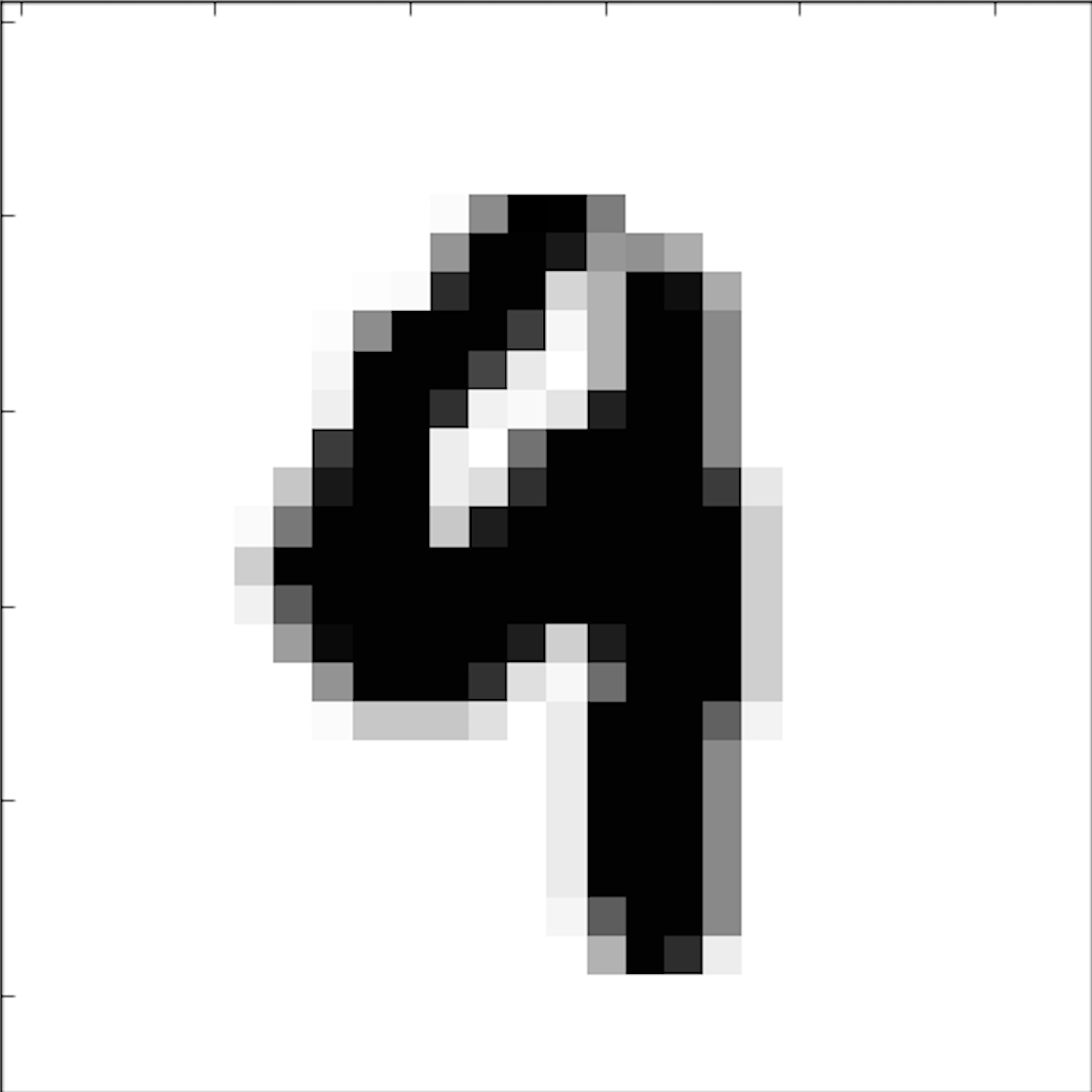}
\includegraphics[width=1.1cm]{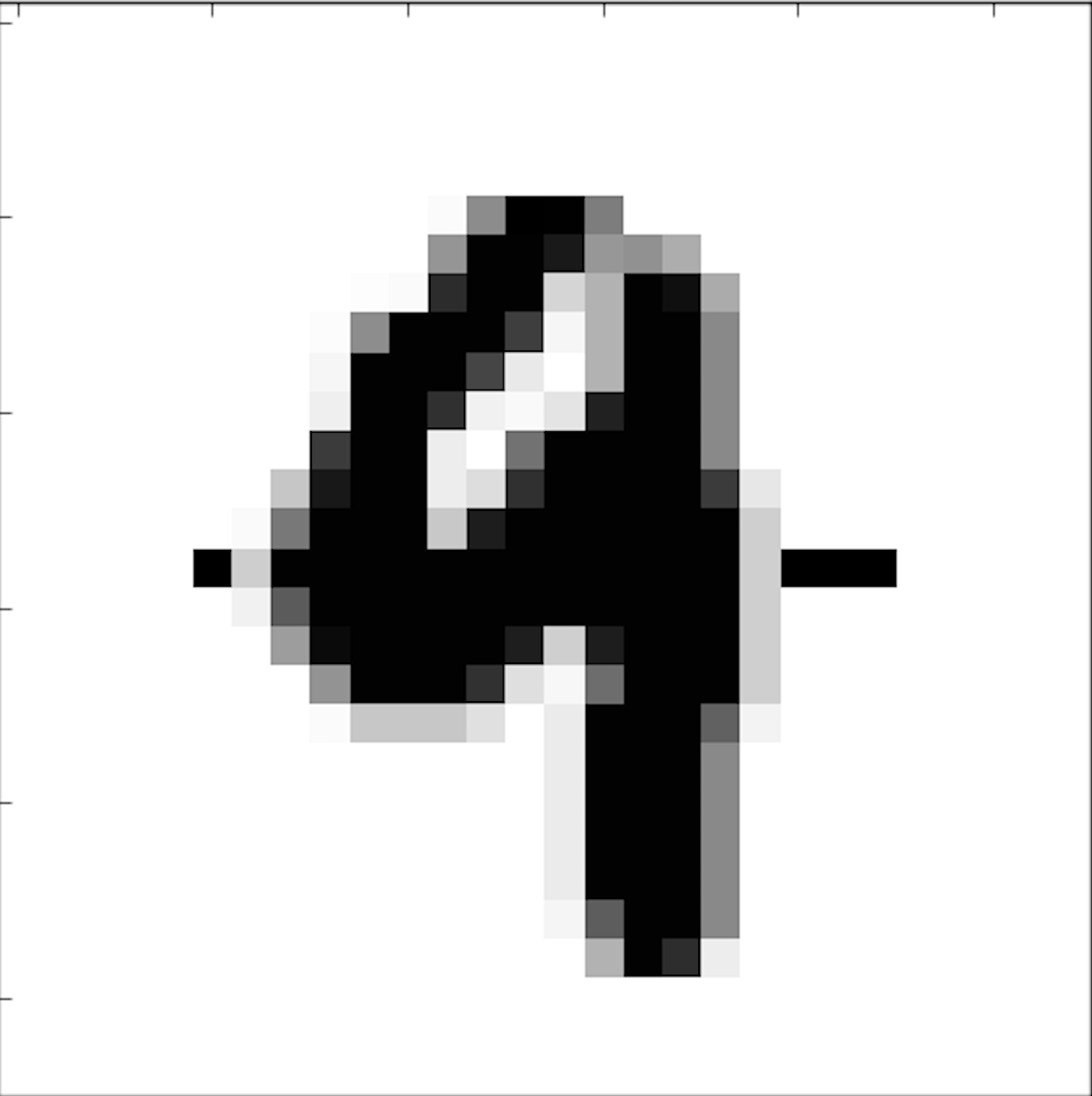}
\text{\hspace{0.8cm}9 to 4}
}
\parbox{2.3cm}{
\includegraphics[width=1.1cm]{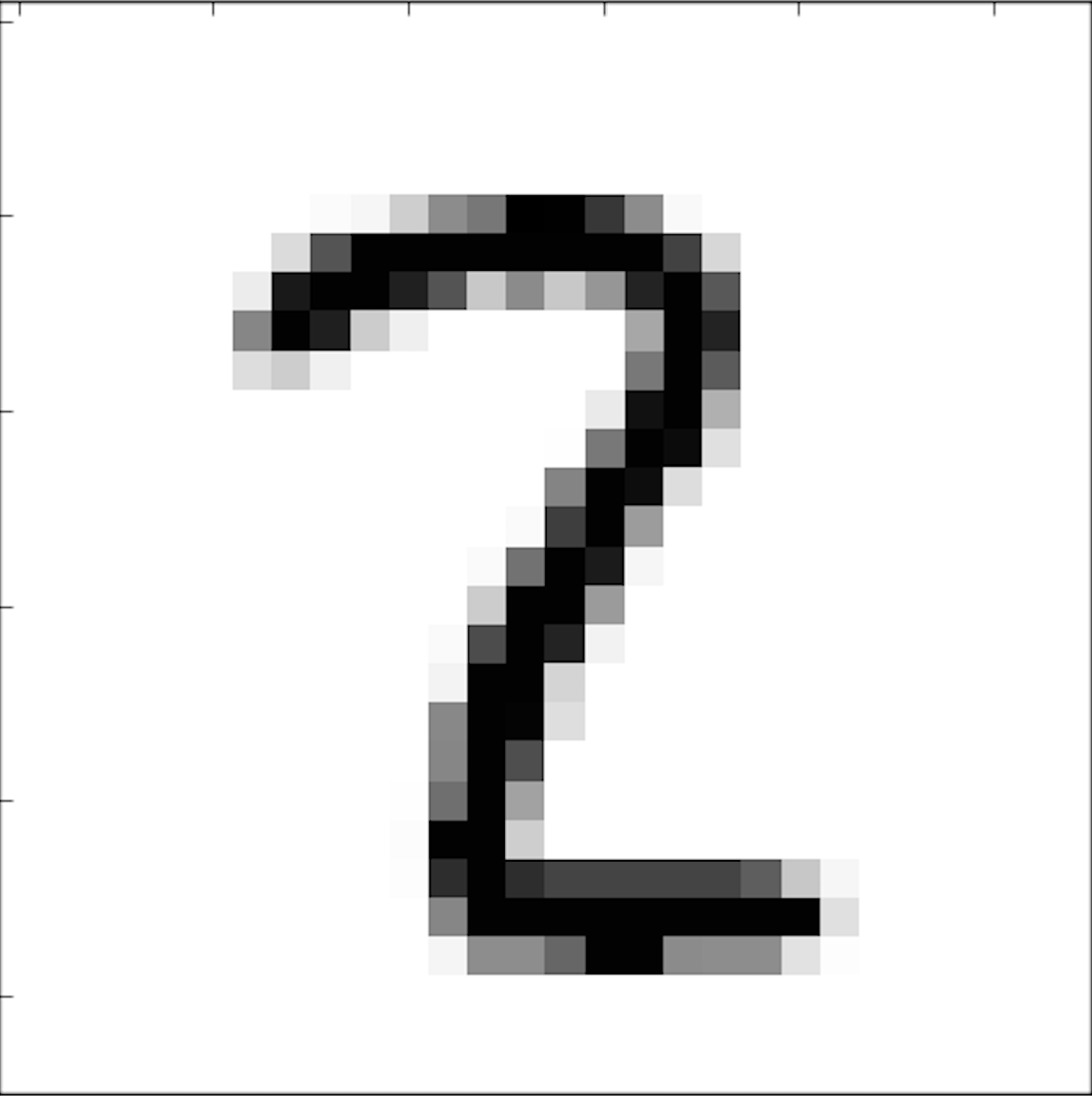}
\includegraphics[width=1.1cm]{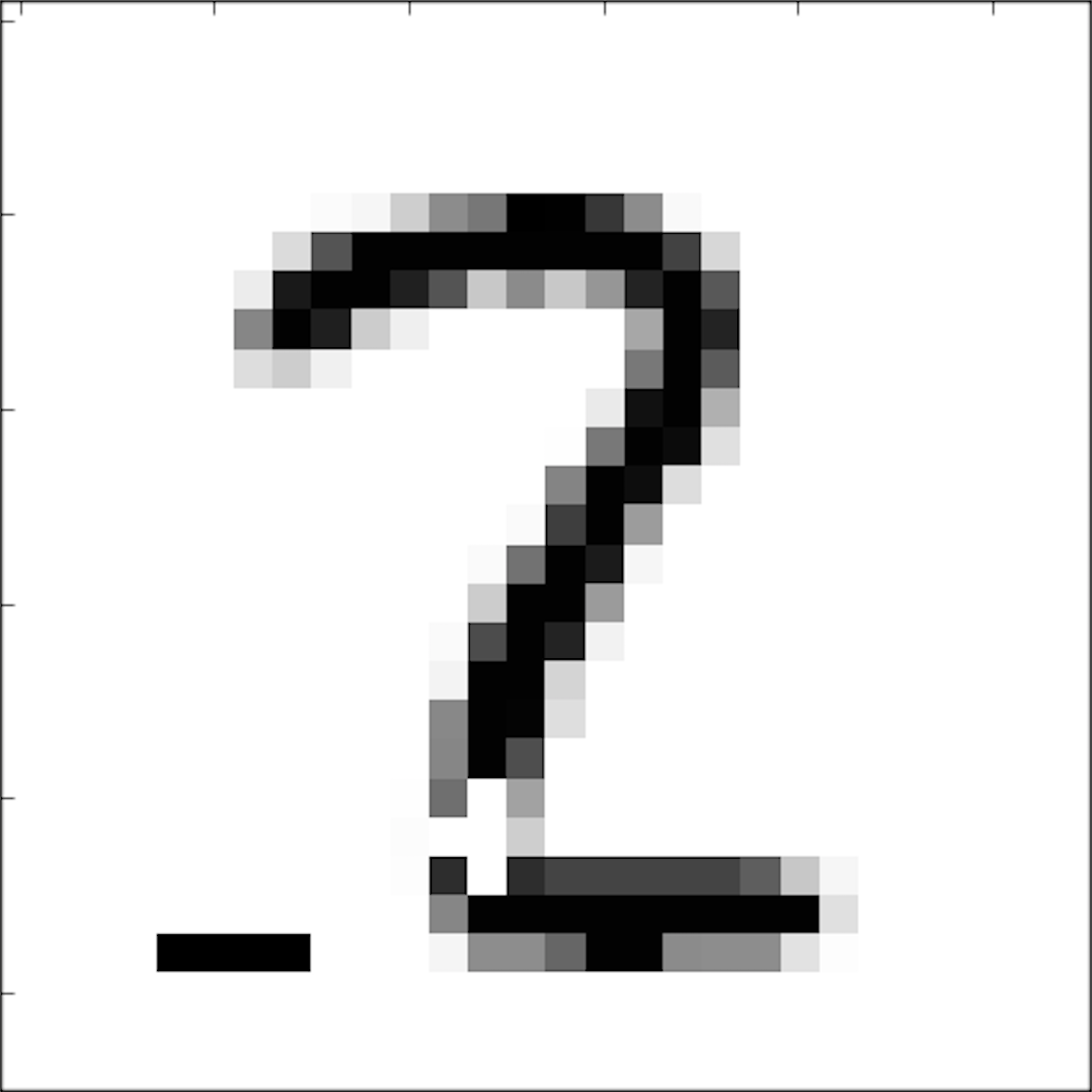}
\text{\hspace{0.8cm}2 to 3}
}
\parbox{2.3cm}{
\includegraphics[width=1.1cm]{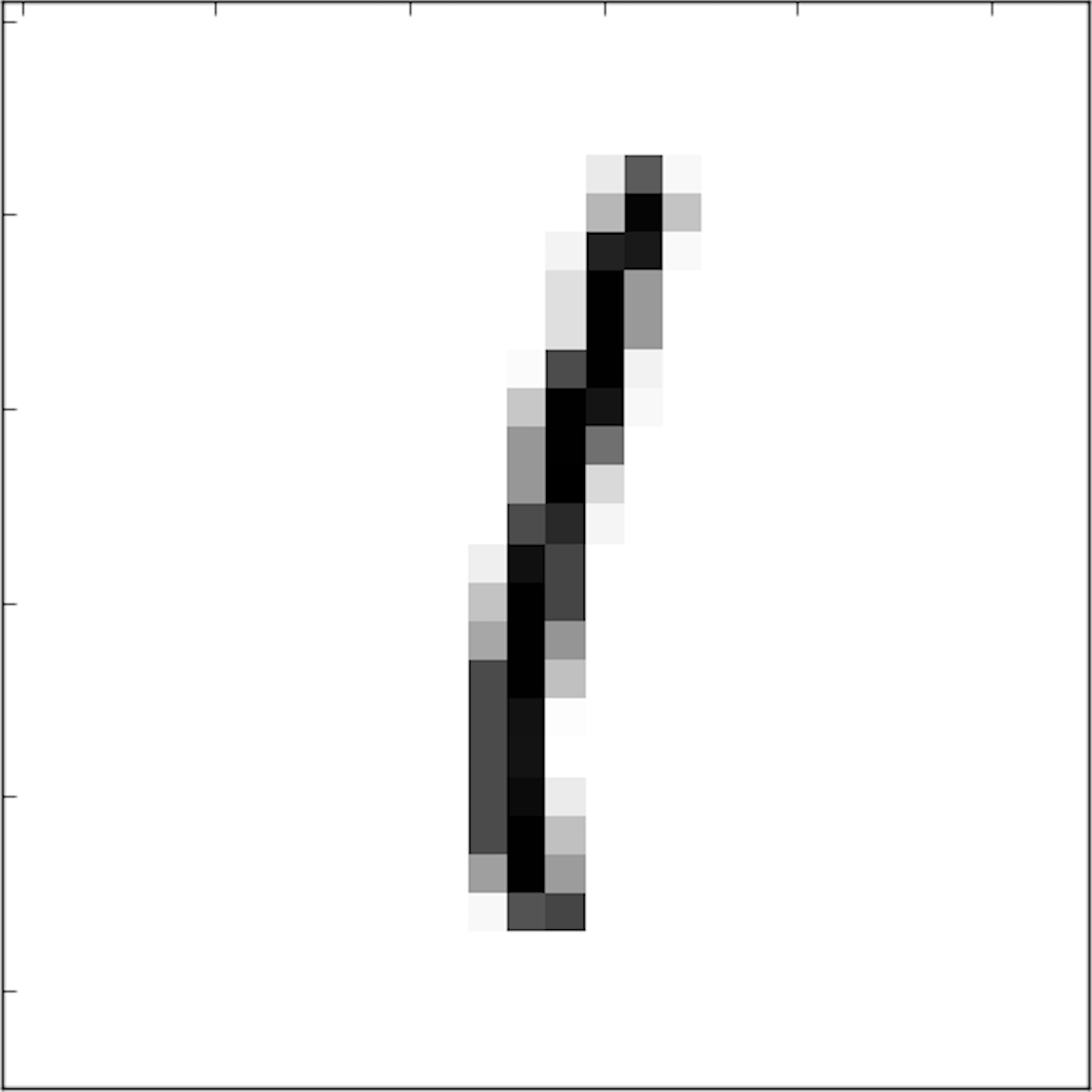}
\includegraphics[width=1.1cm]{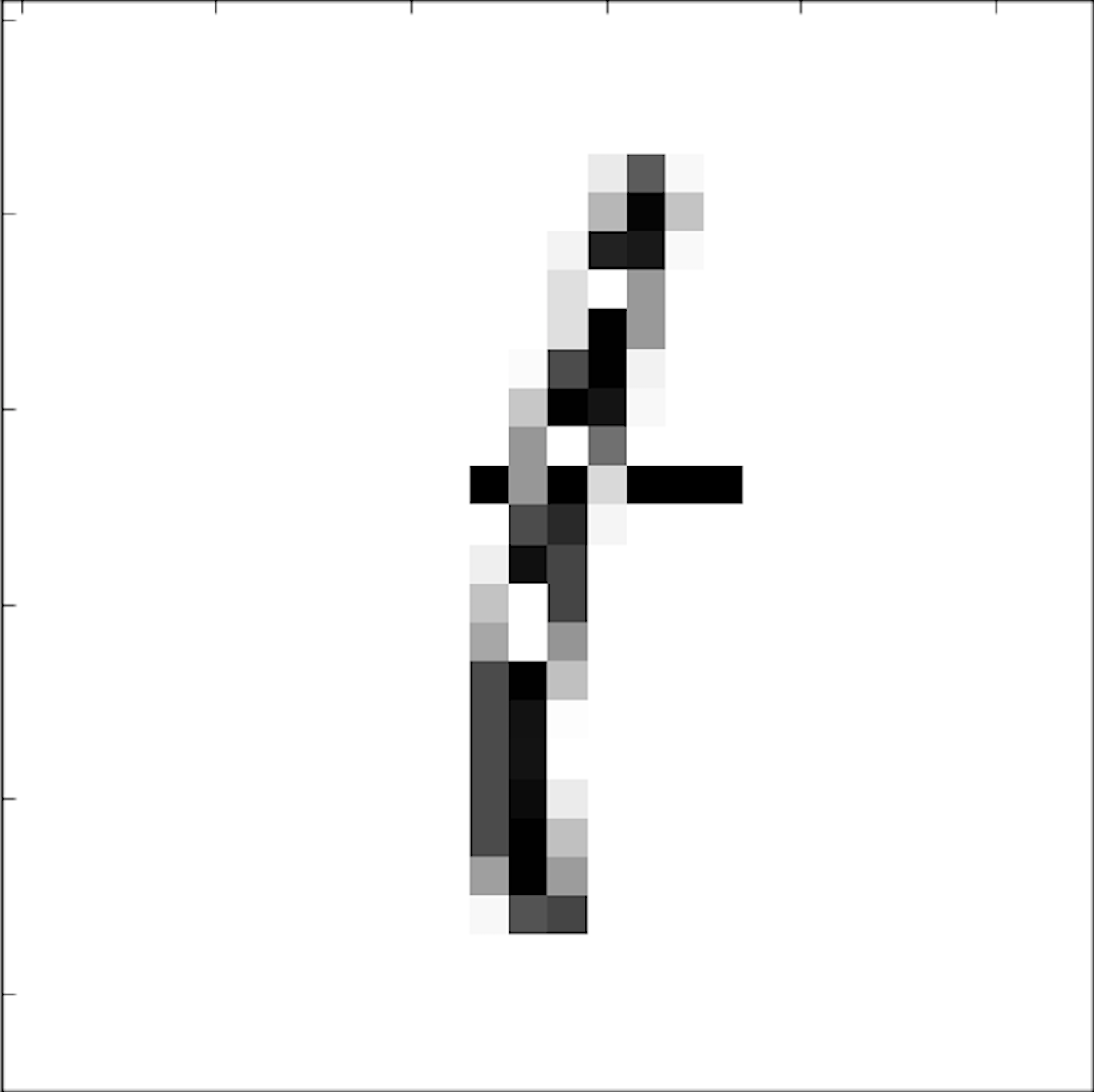}
\text{\hspace{0.8cm}1 to 8}
}\\

\parbox{2.3cm}{
\includegraphics[width=1.1cm]{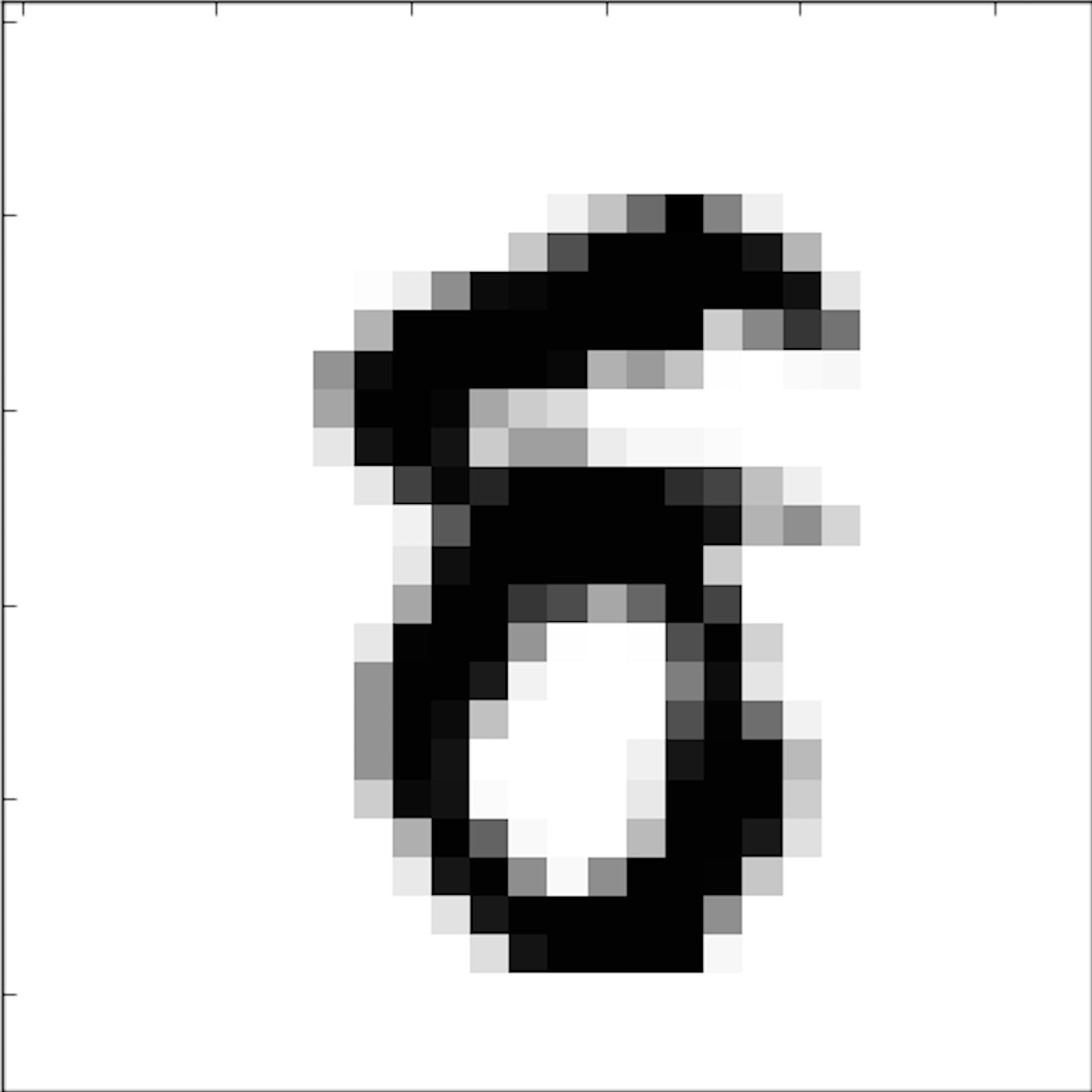}
\includegraphics[width=1.1cm]{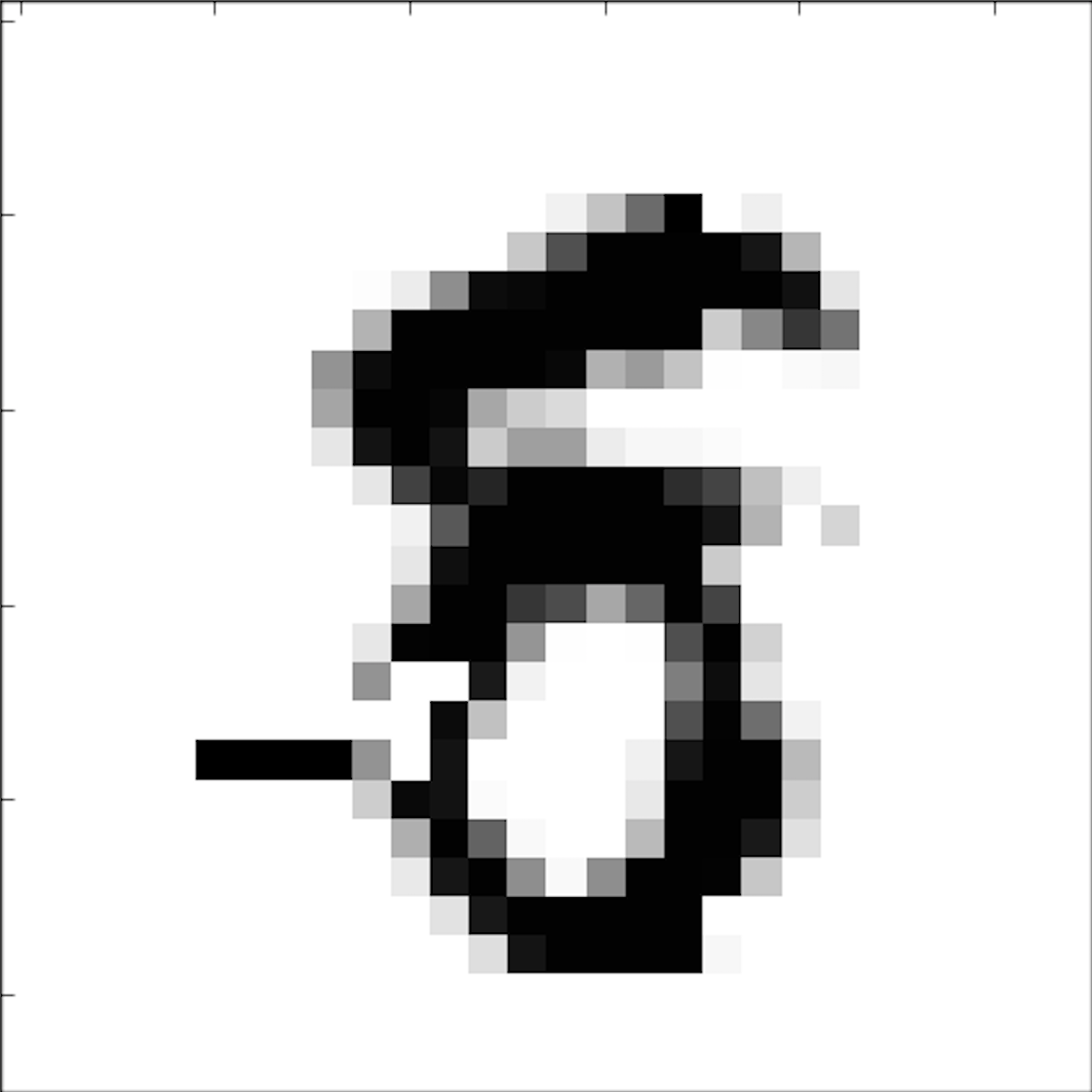}
\text{\hspace{0.8cm}8 to 5}
}
\parbox{2.3cm}{
\includegraphics[width=1.1cm]{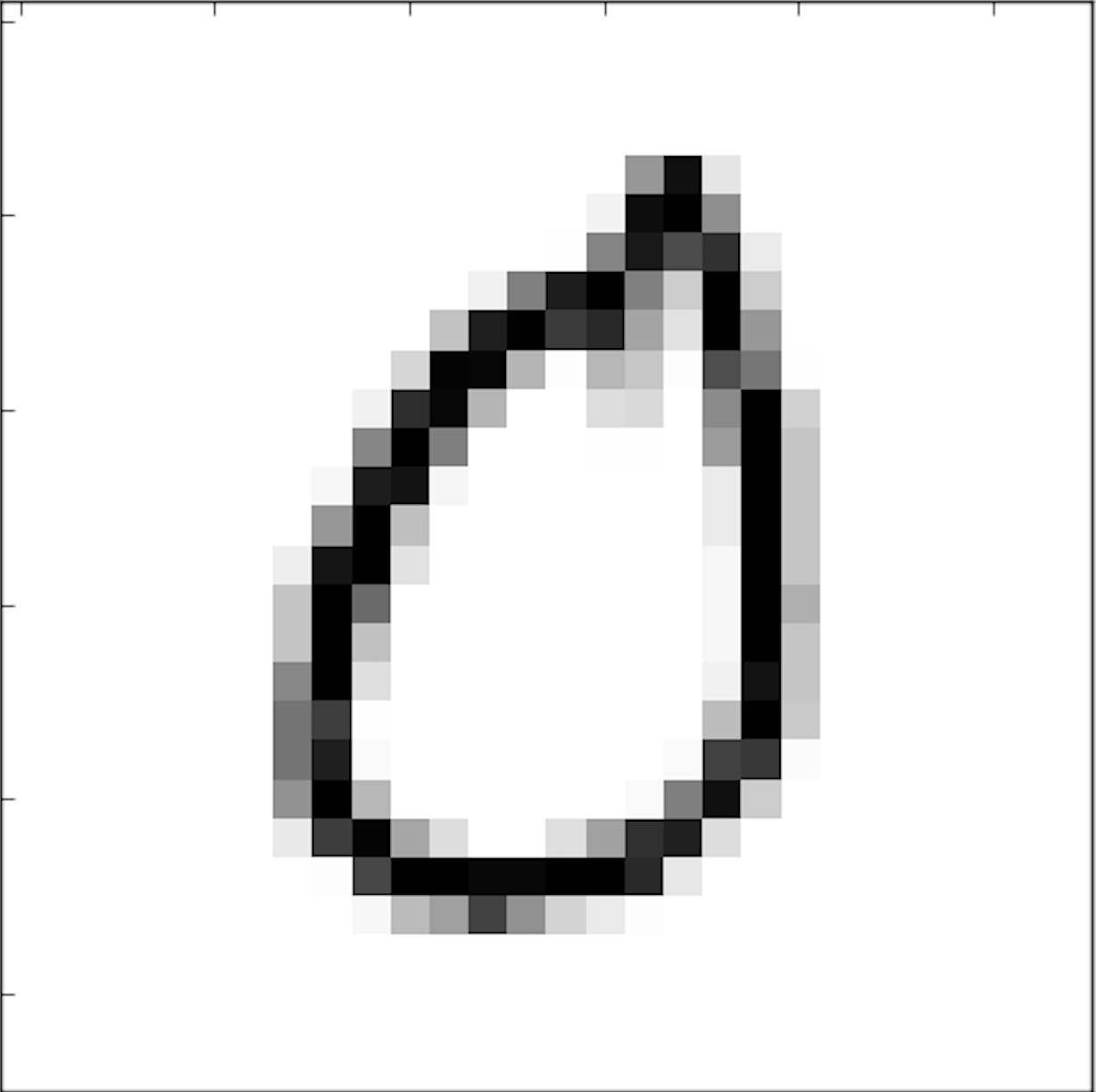}
\includegraphics[width=1.1cm]{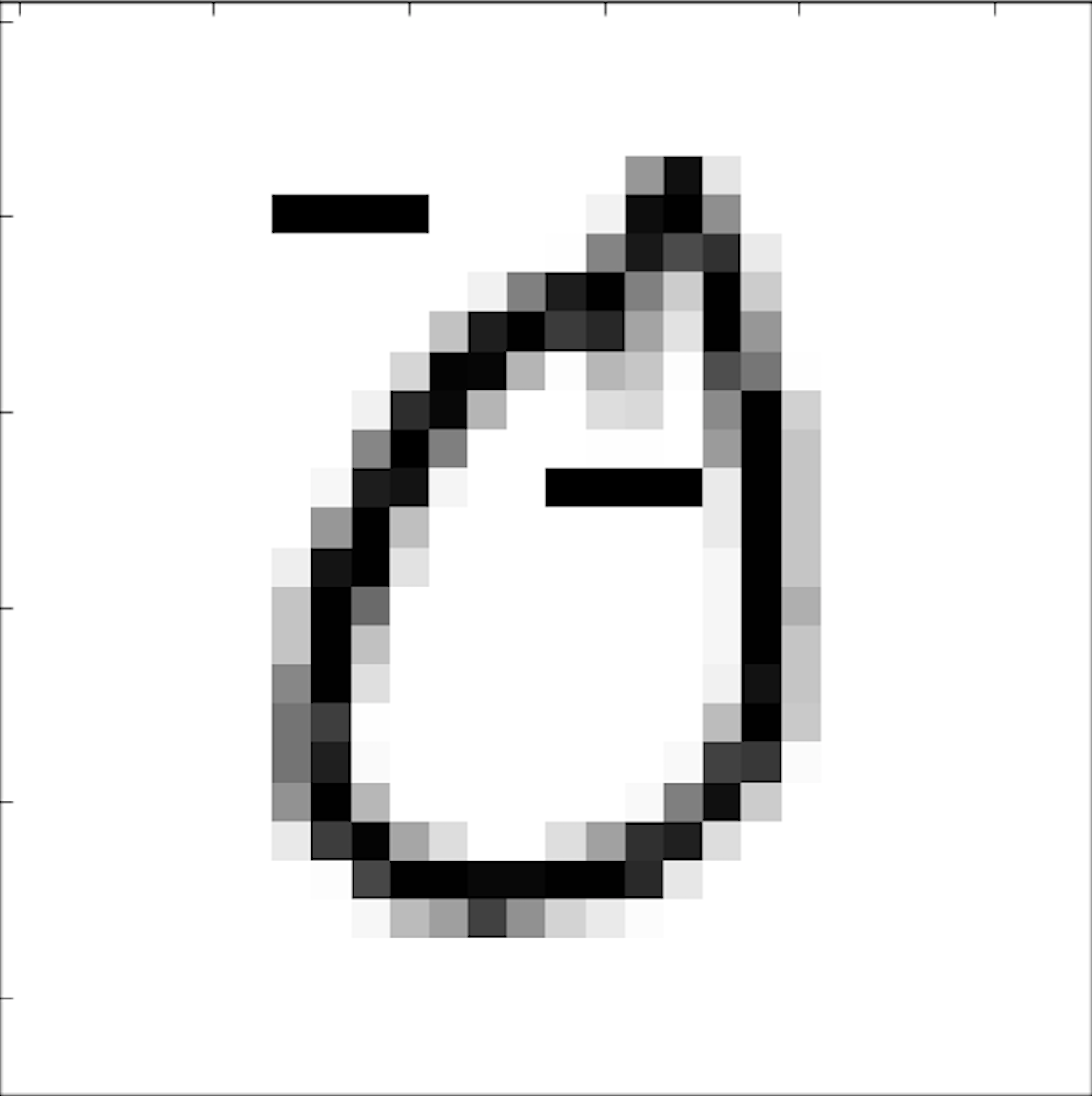}
\text{\hspace{0.8cm}0 to 3}
}
\parbox{2.3cm}{
\includegraphics[width=1.1cm]{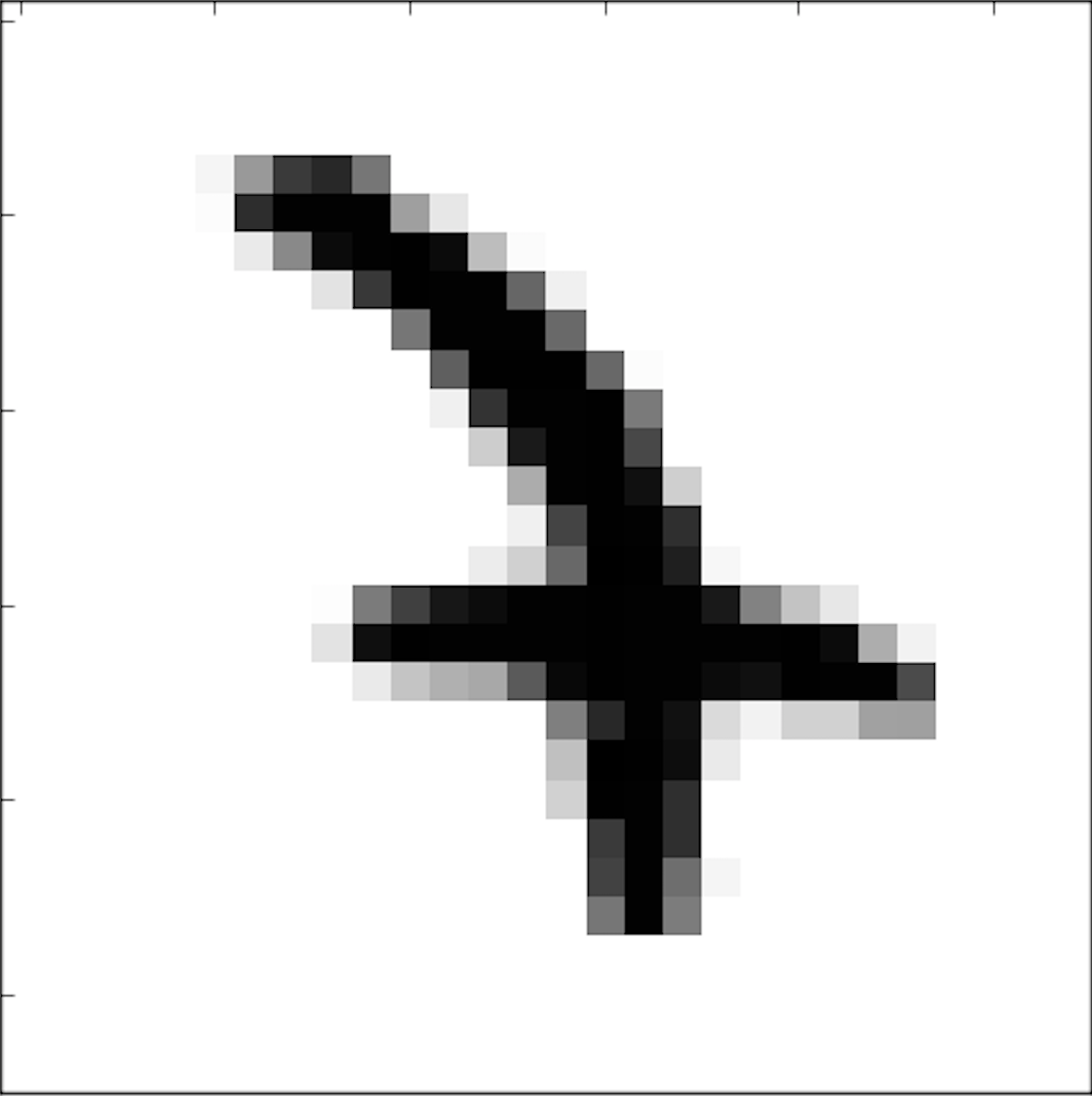}
\includegraphics[width=1.1cm]{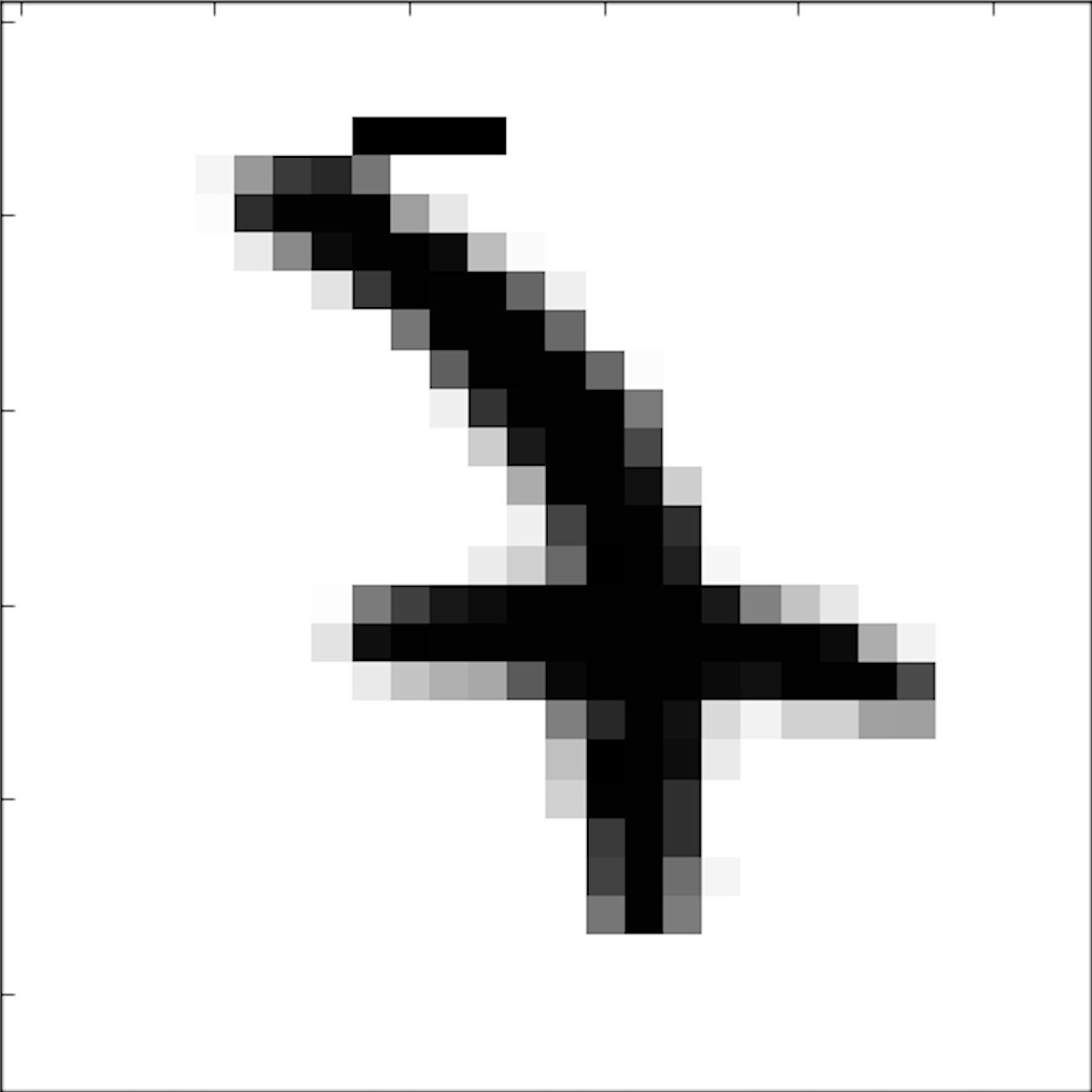}
\text{\hspace{0.8cm}7 to 2}
}
\parbox{2.3cm}{
\includegraphics[width=1.1cm]{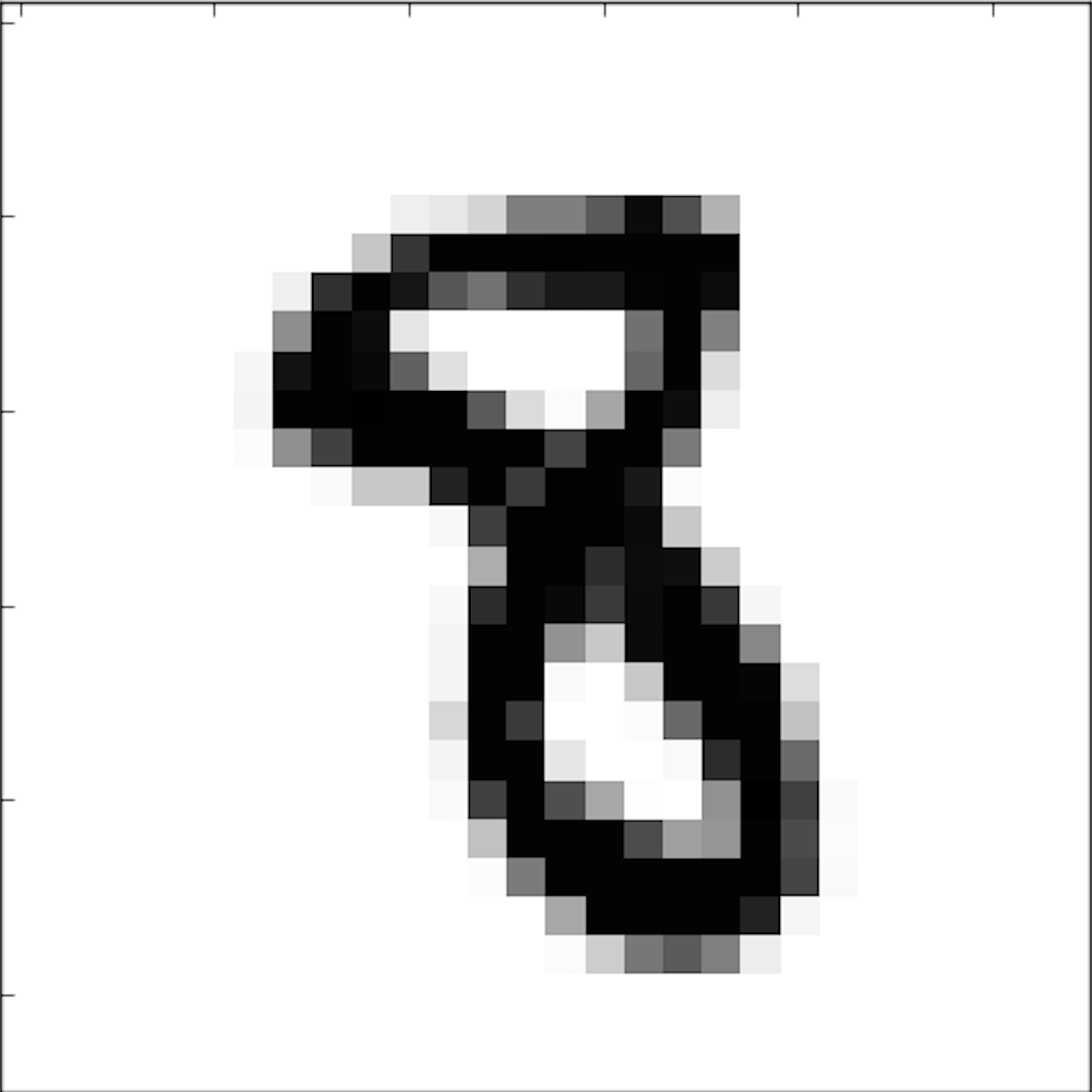}
\includegraphics[width=1.1cm]{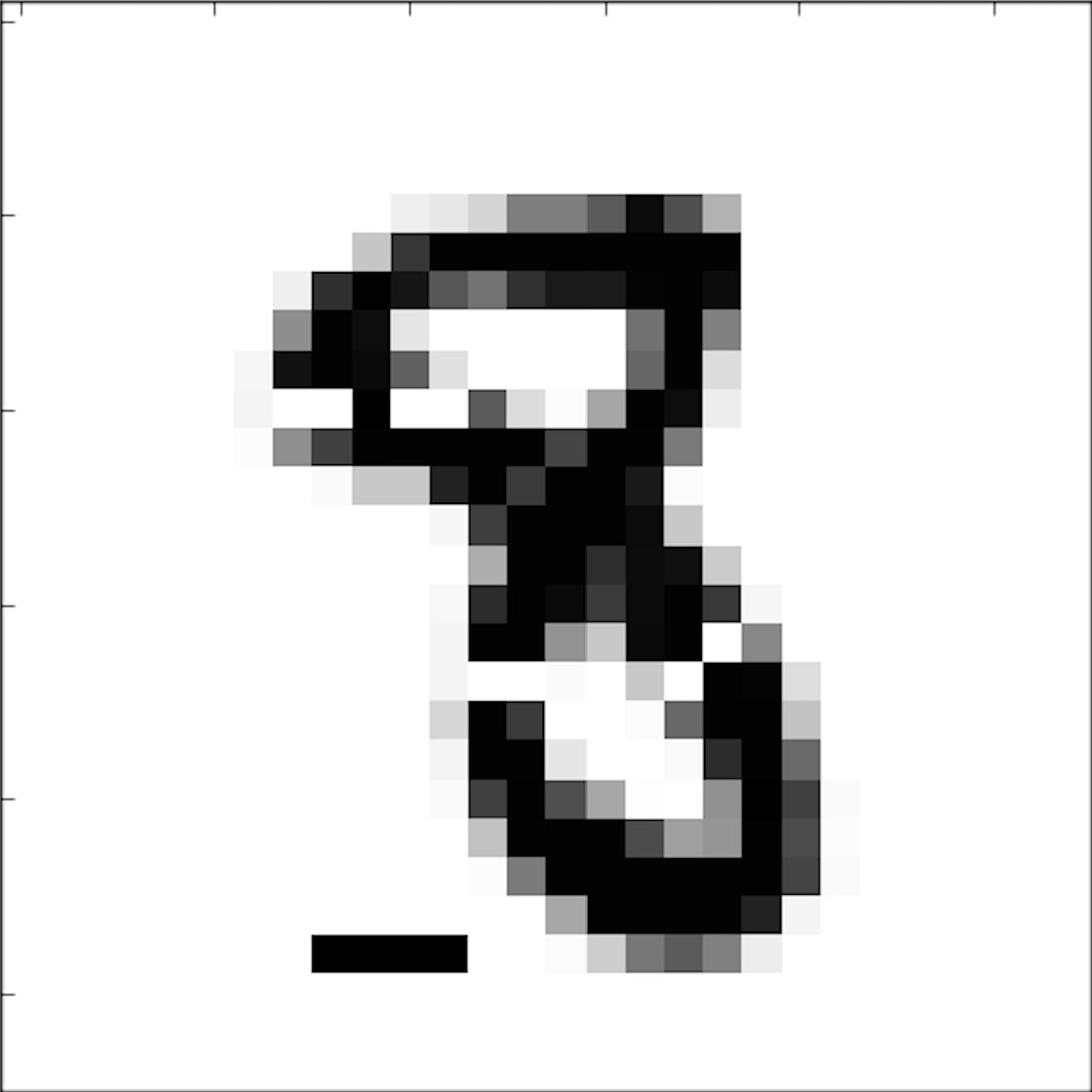}
\text{\hspace{0.8cm}8 to 3}
}
\parbox{2.3cm}{
\includegraphics[width=1.1cm]{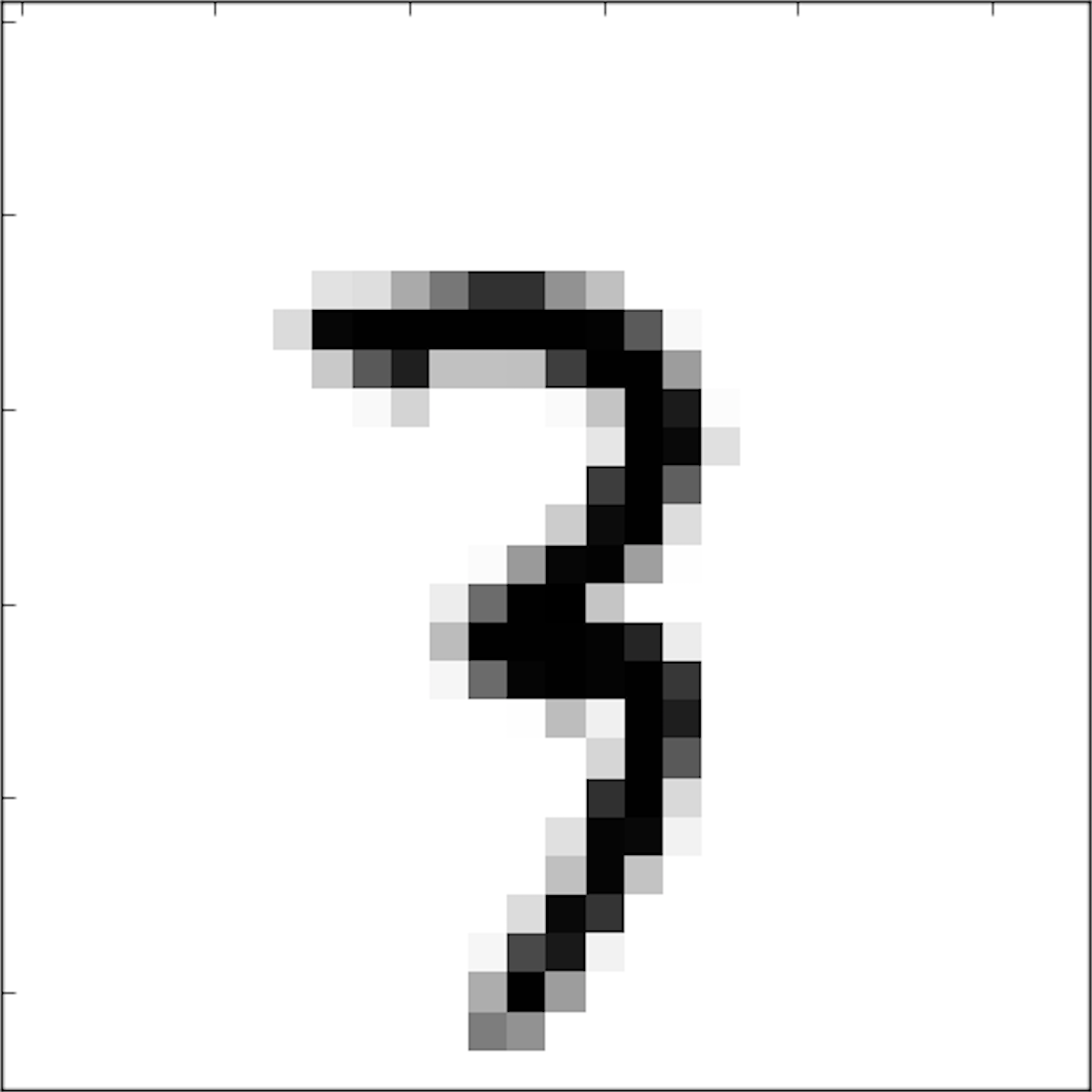}
\includegraphics[width=1.1cm]{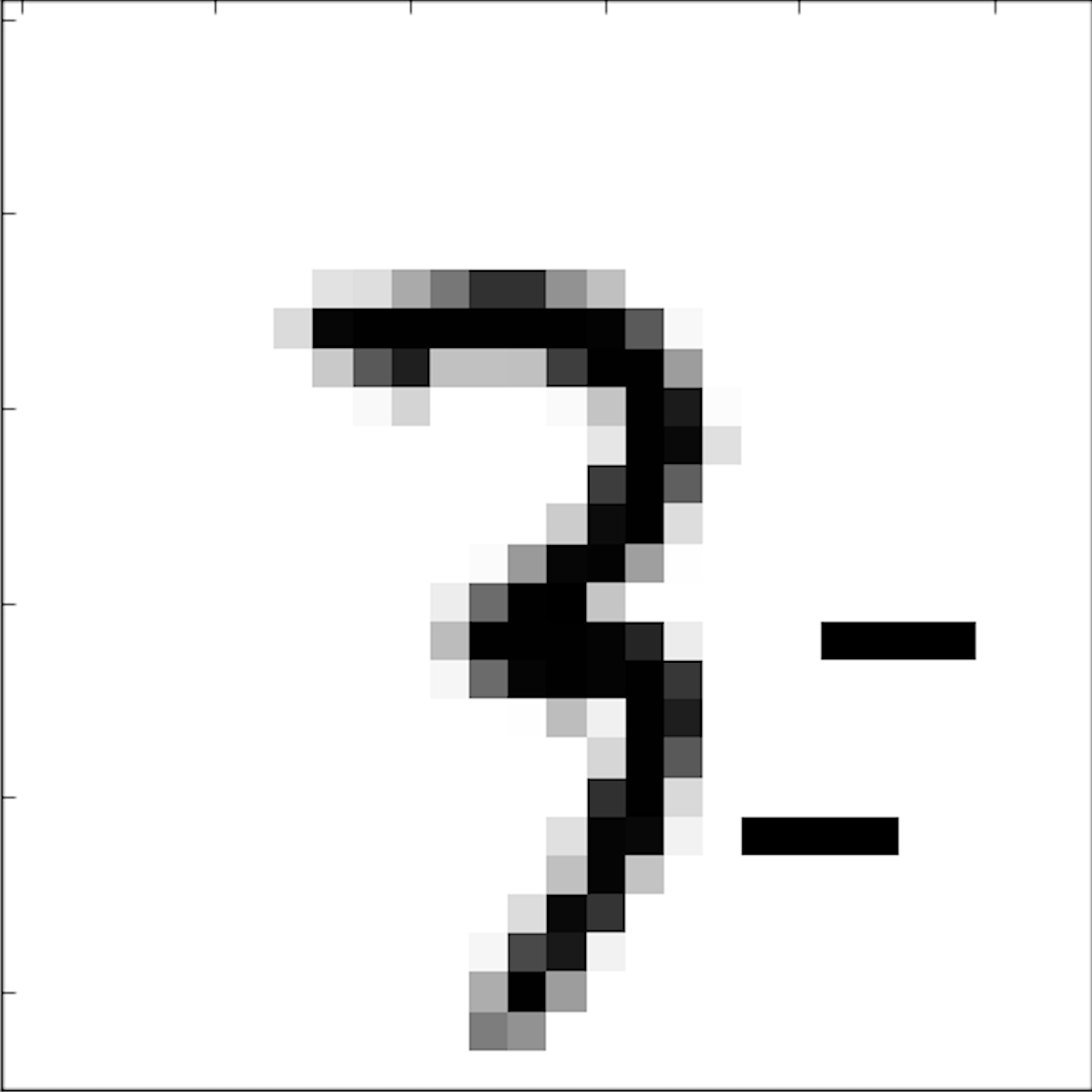}
\text{\hspace{0.8cm}3 to 2}
}\\

\parbox{2.3cm}{
\includegraphics[width=1.1cm]{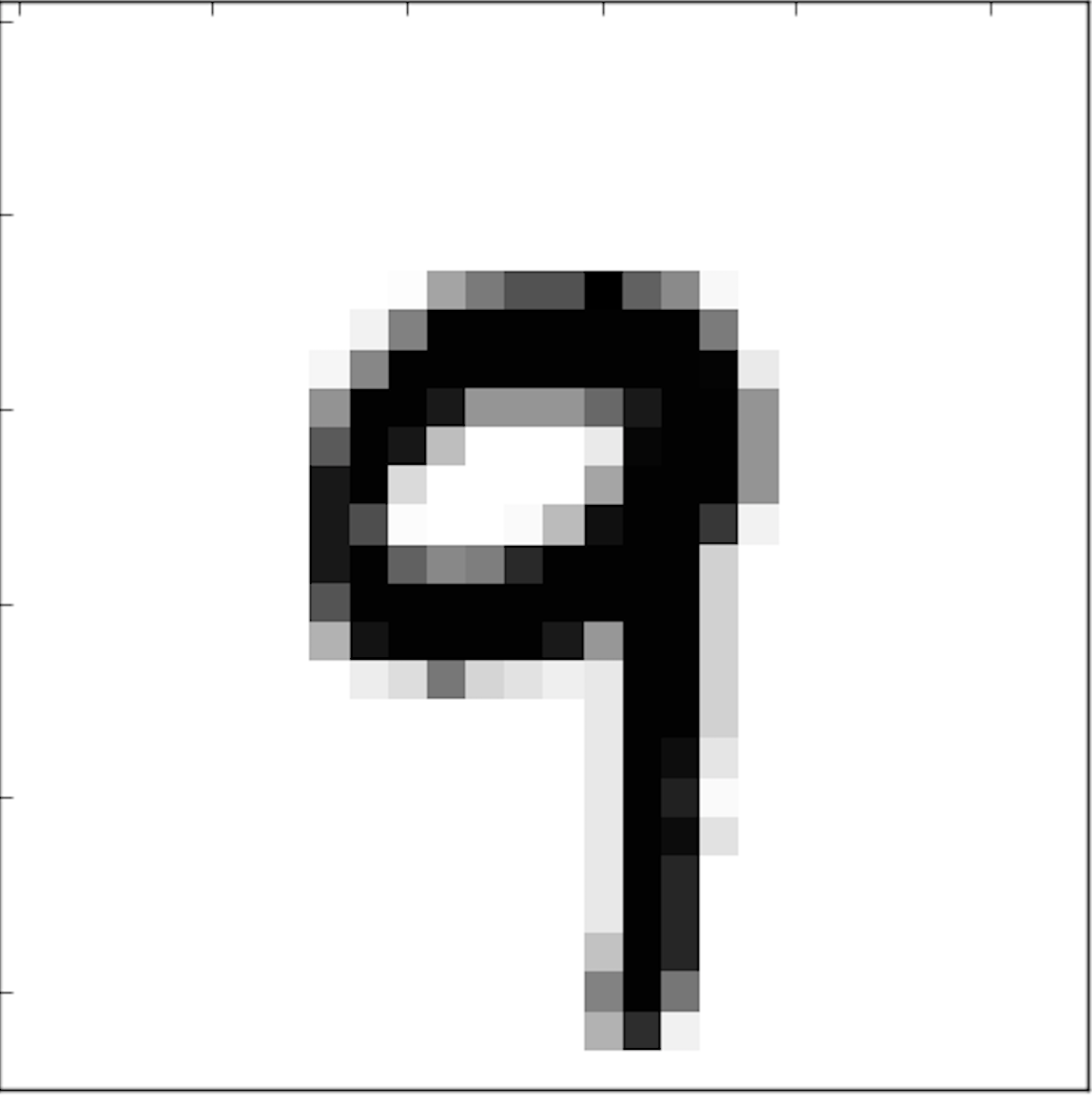}
\includegraphics[width=1.1cm]{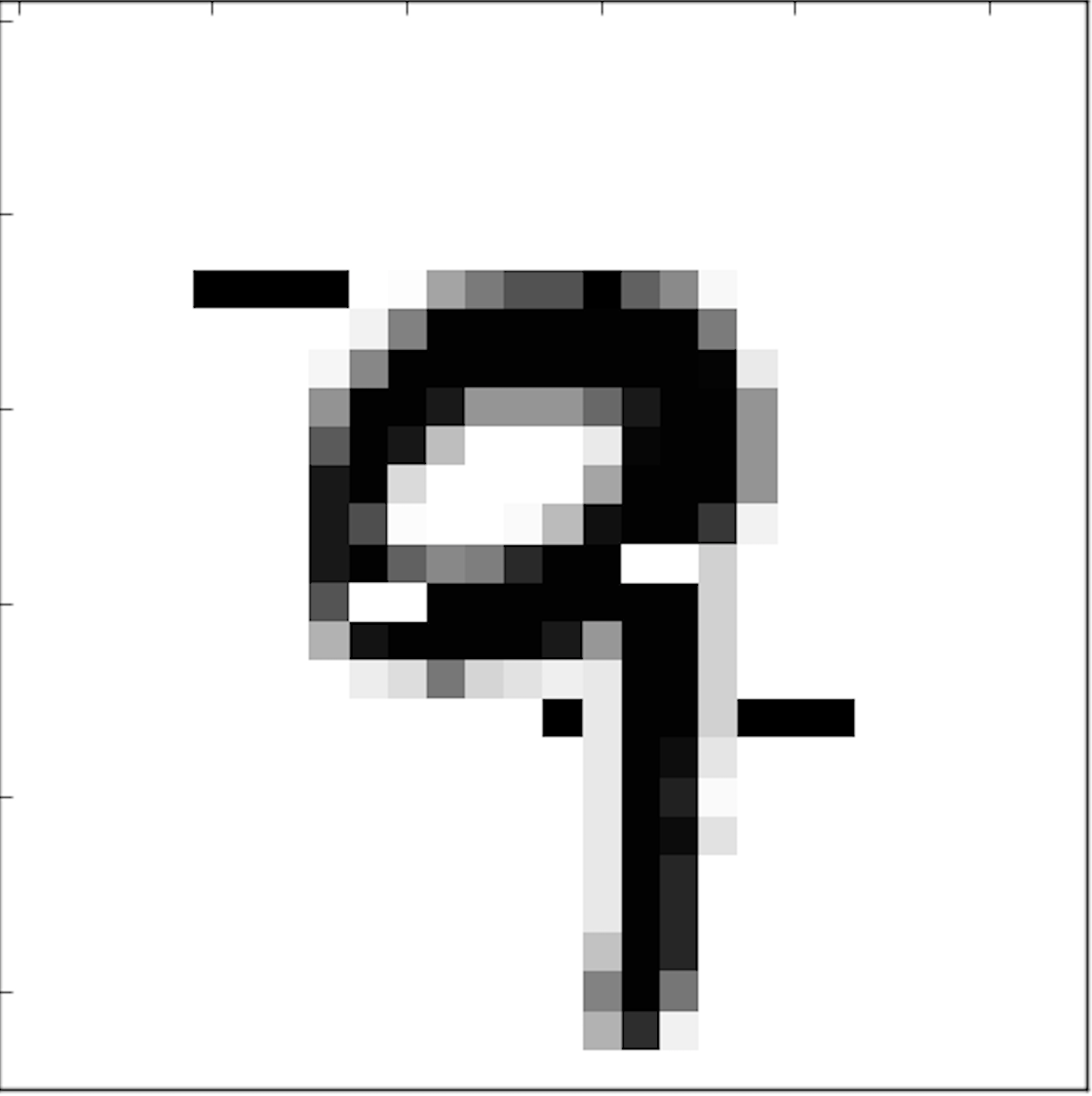}
\text{\hspace{0.8cm}9 to 7}
}
\parbox{2.3cm}{
\includegraphics[width=1.1cm]{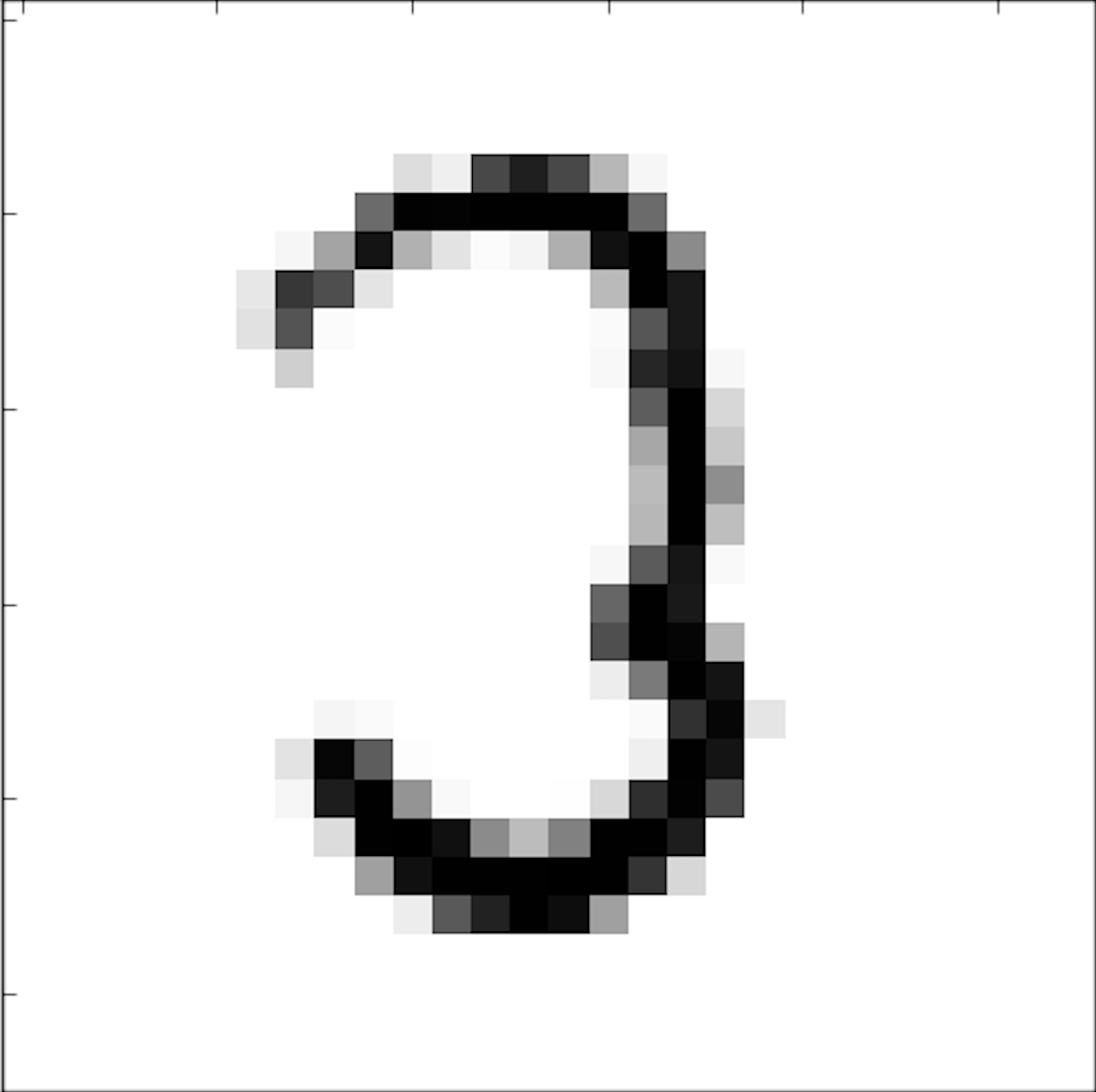}
\includegraphics[width=1.1cm]{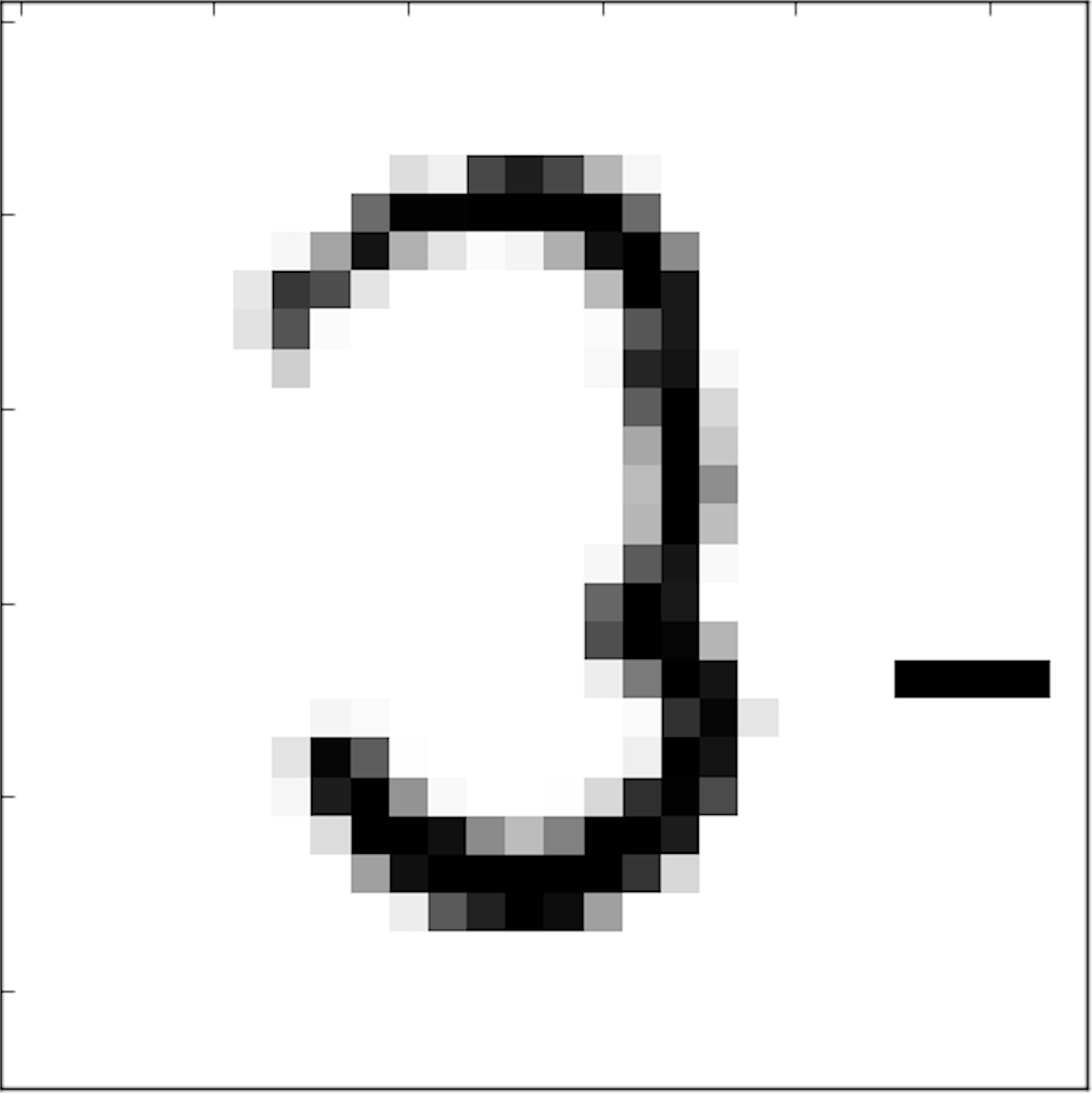}
\text{\hspace{0.8cm}3 to 2}
}
\parbox{2.3cm}{
\includegraphics[width=1.1cm]{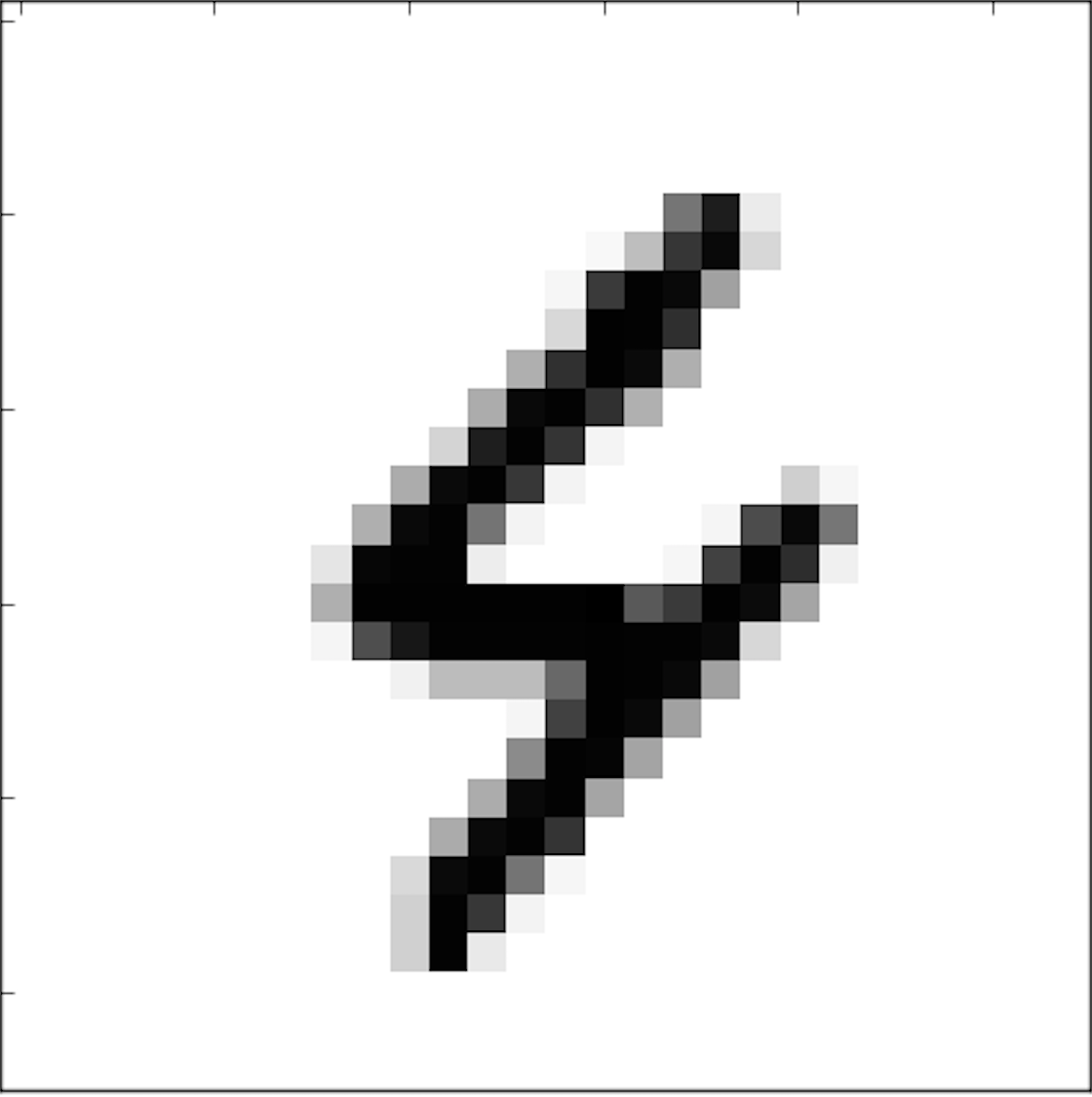}
\includegraphics[width=1.1cm]{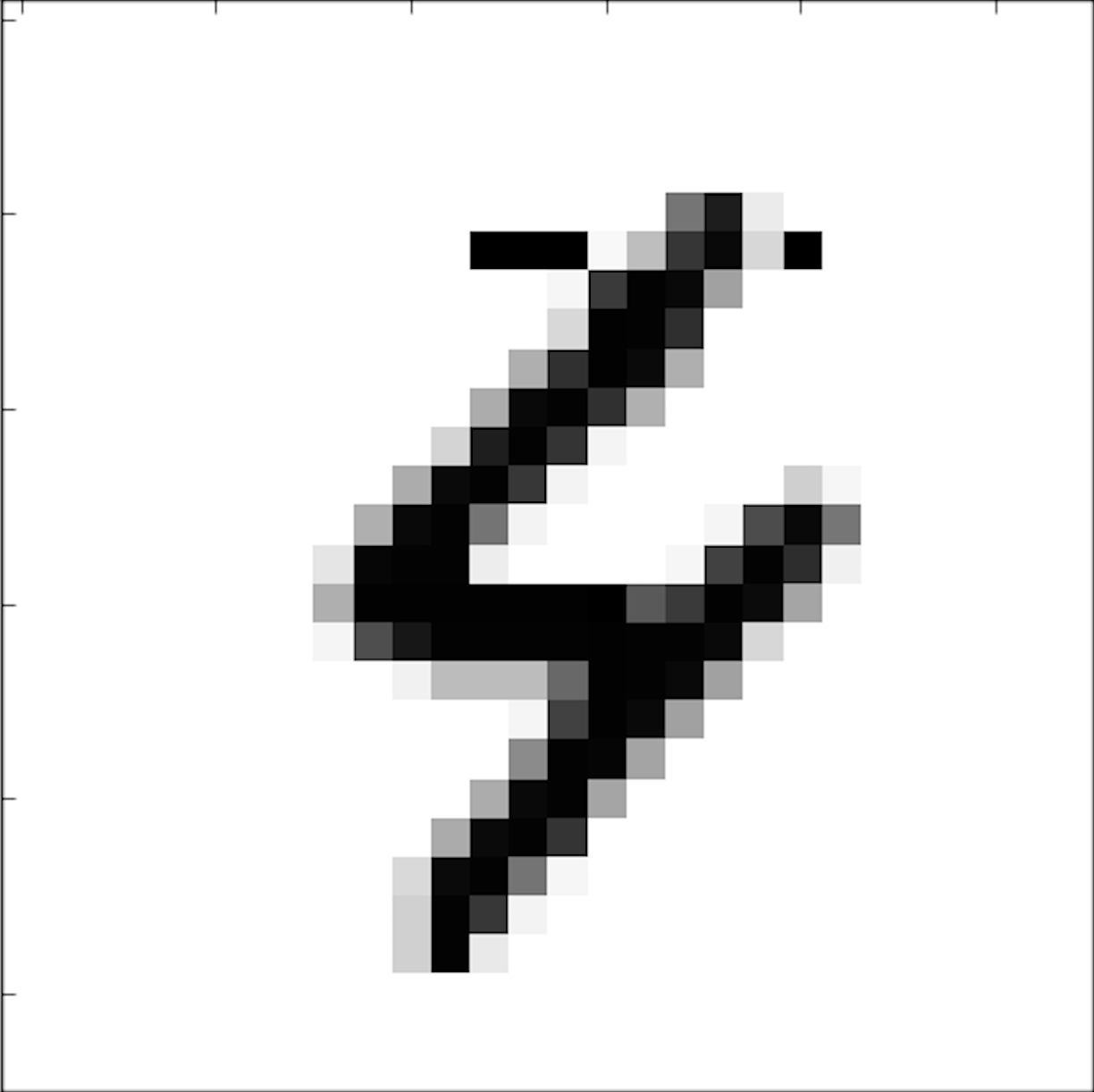}
\text{\hspace{0.8cm}4 to 9}
}
\parbox{2.3cm}{
\includegraphics[width=1.1cm]{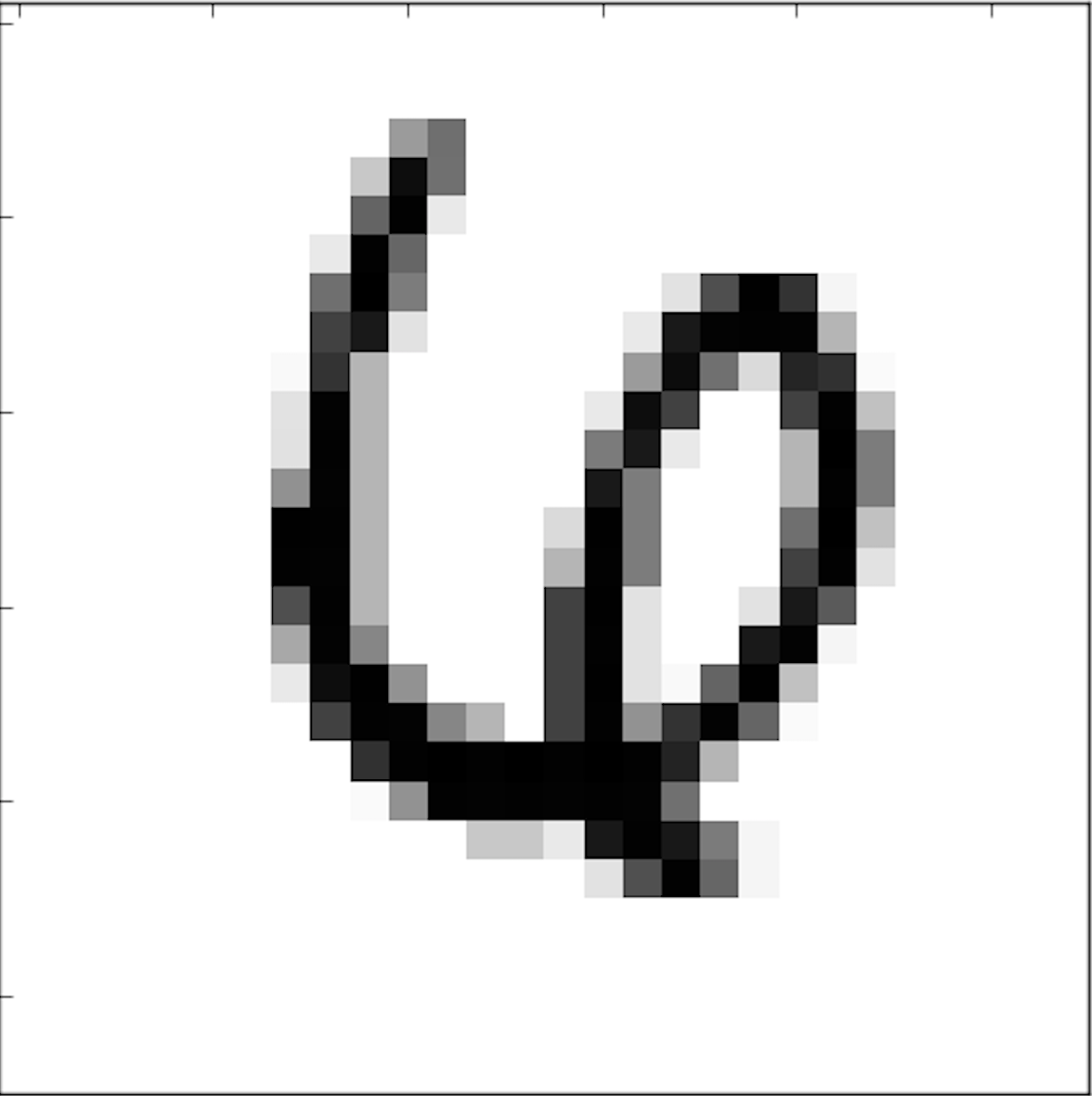}
\includegraphics[width=1.1cm]{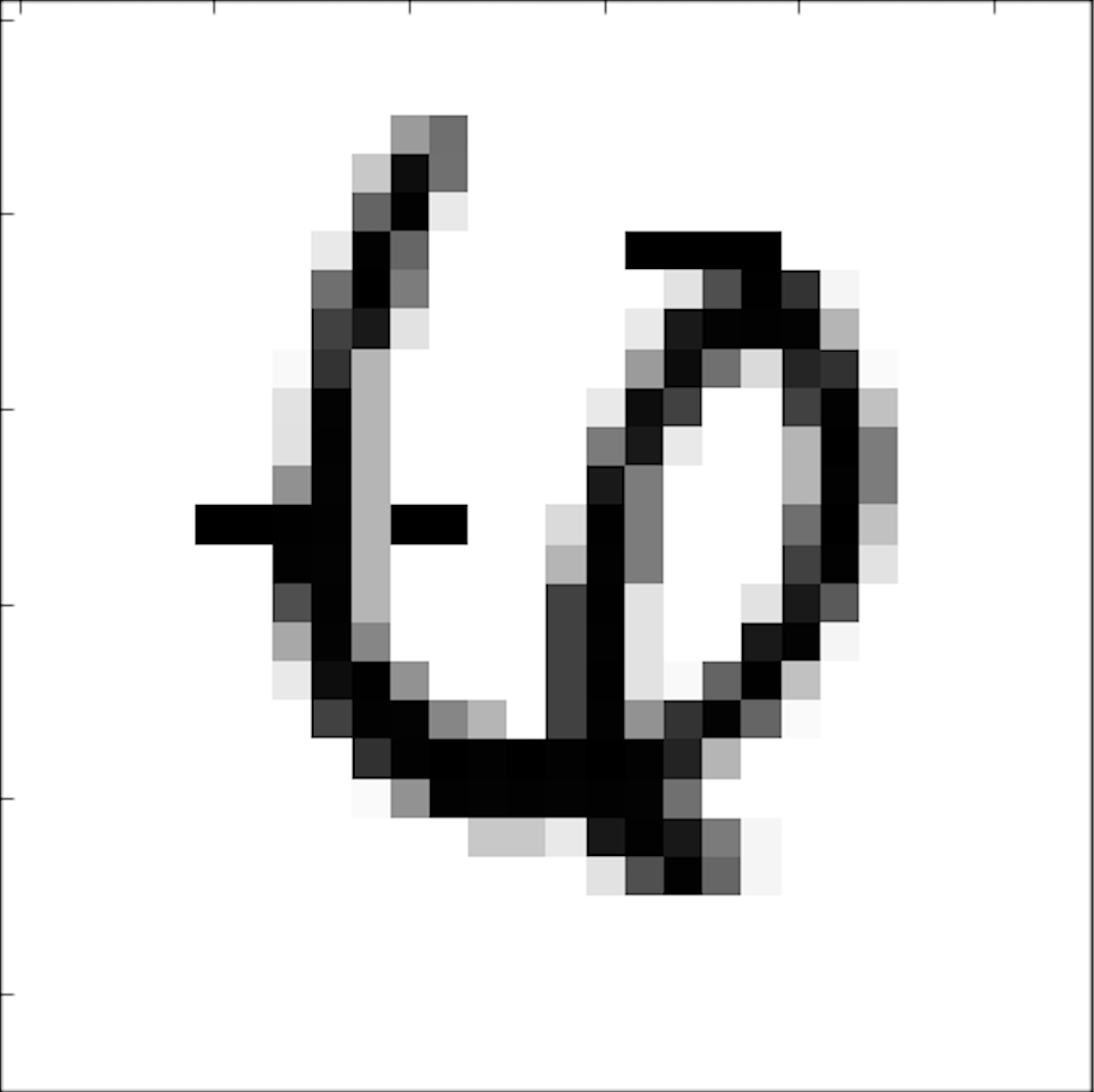}
\text{\hspace{0.8cm}6 to 4}
}
\parbox{2.3cm}{
\includegraphics[width=1.1cm]{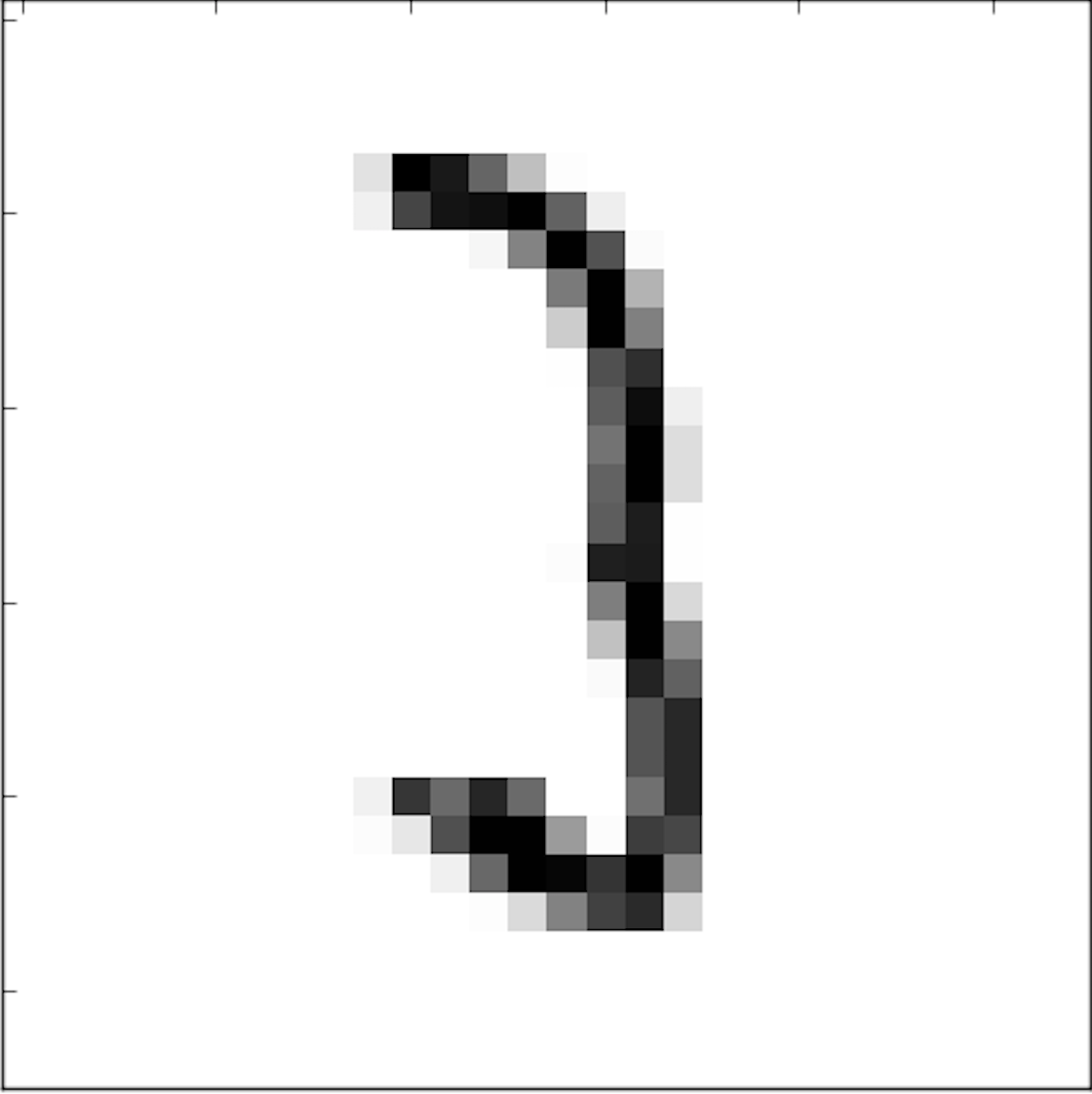}
\includegraphics[width=1.1cm]{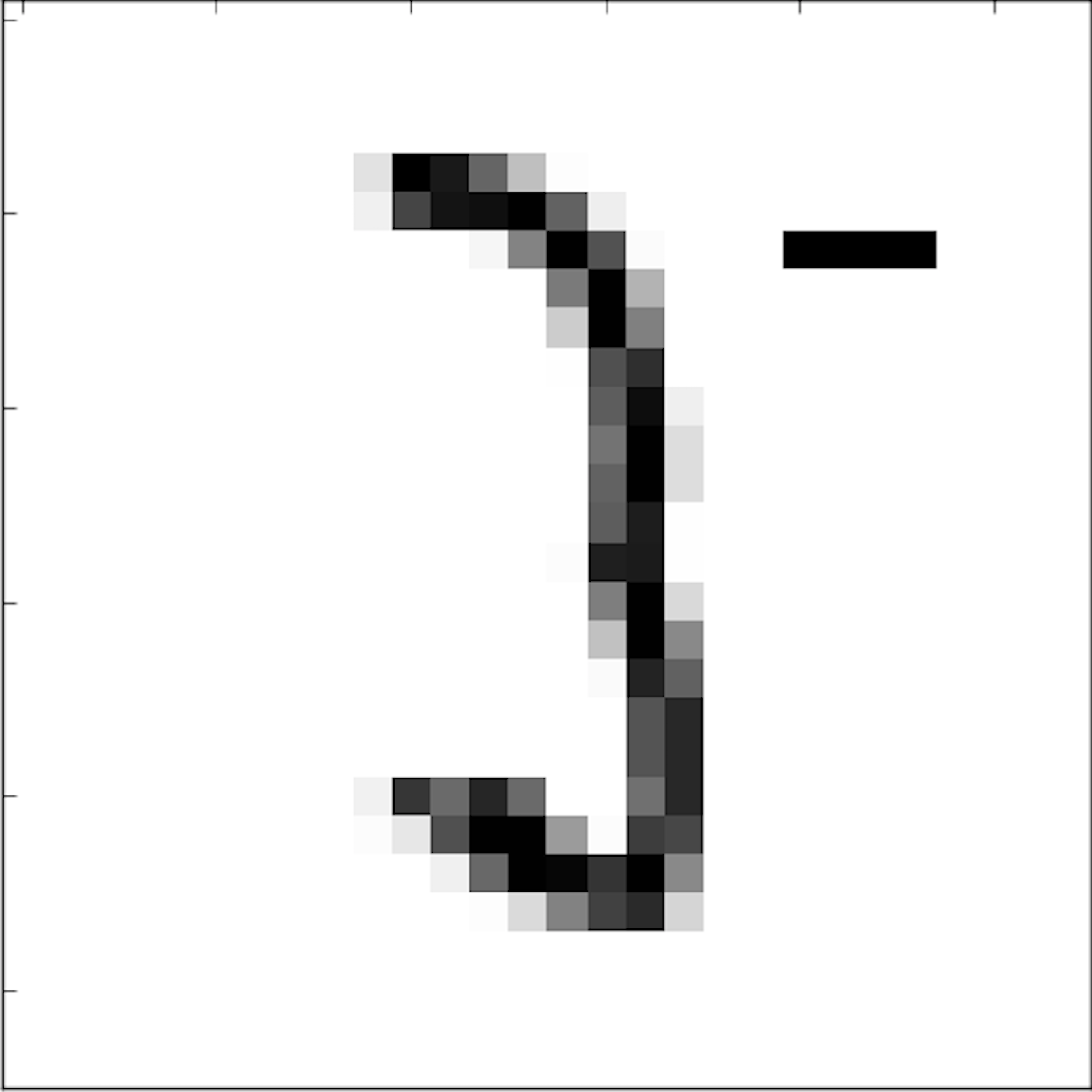}
\text{\hspace{0.8cm}3 to 5}
}\\

\parbox{2.3cm}{
\includegraphics[width=1.1cm]{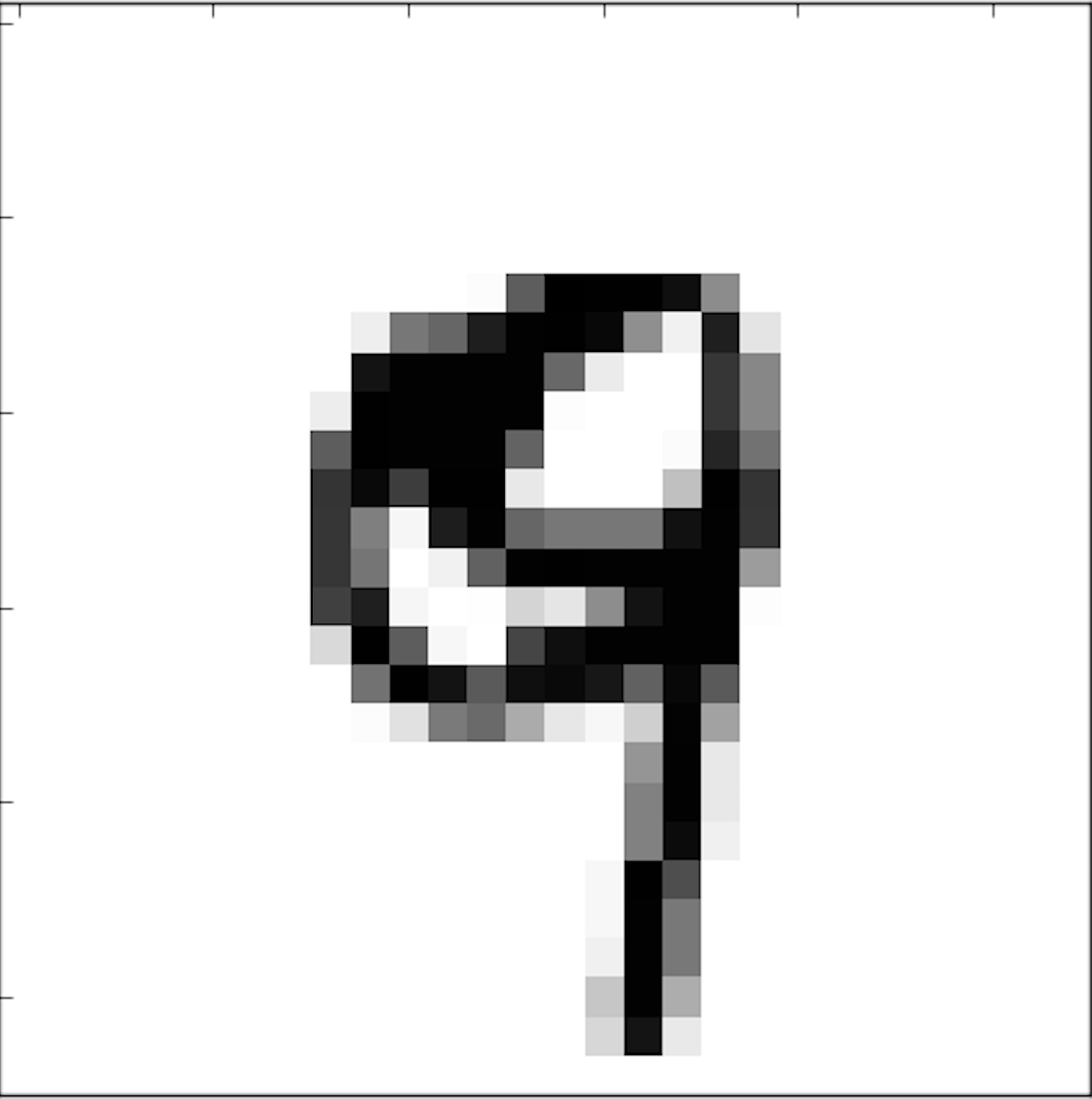}
\includegraphics[width=1.1cm]{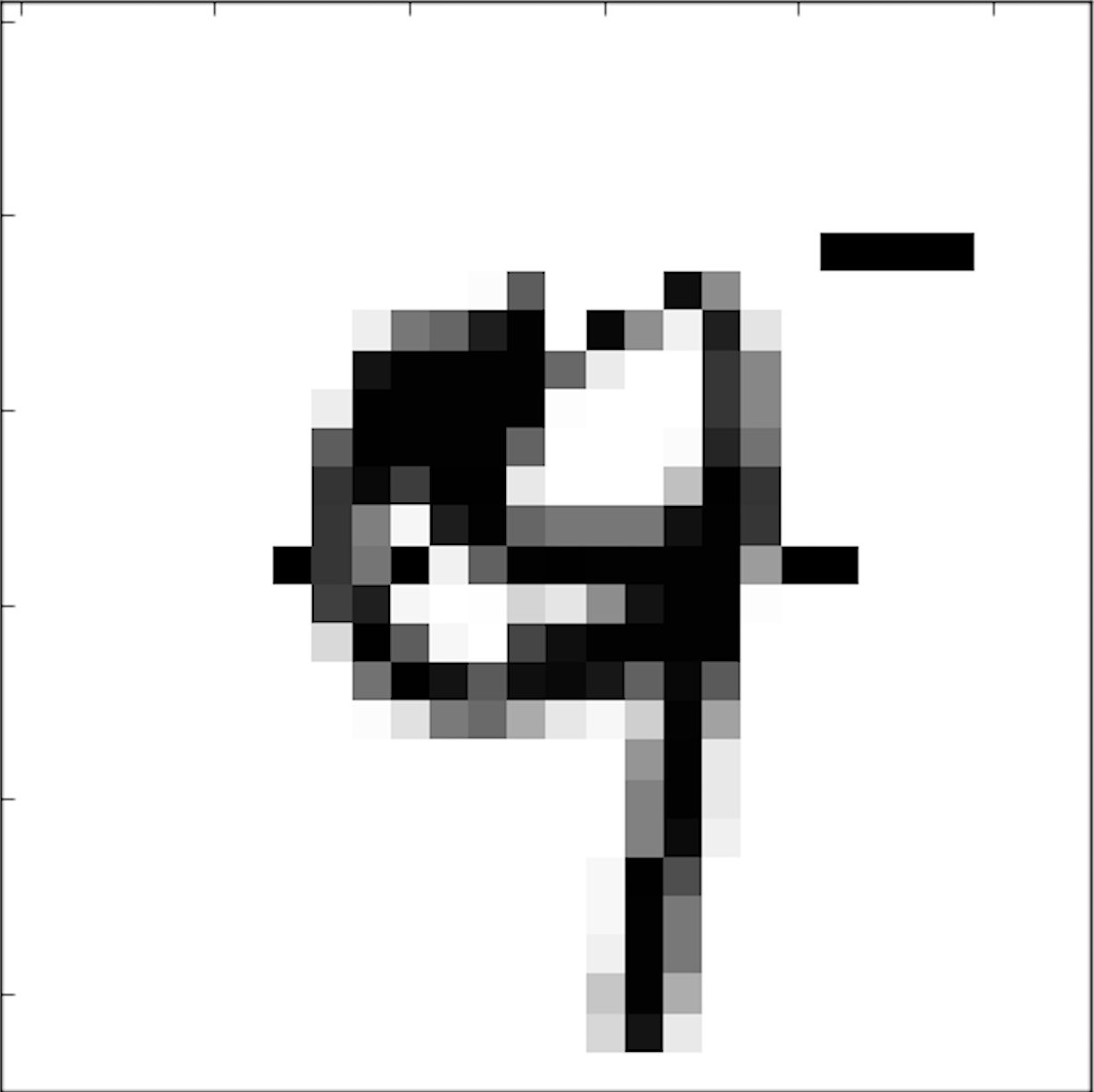}
\text{\hspace{0.8cm}9 to 4}
}
\parbox{2.3cm}{
\includegraphics[width=1.1cm]{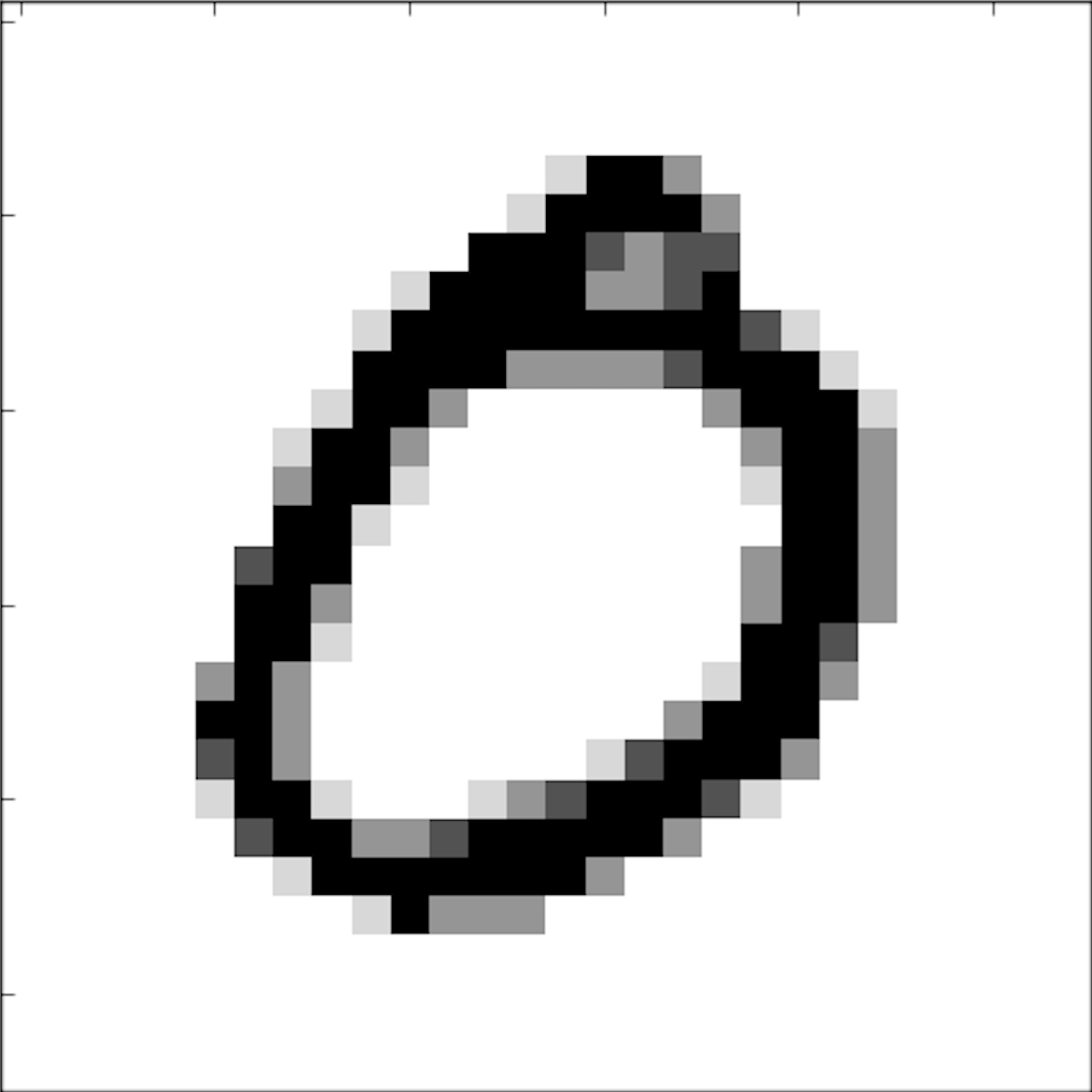}
\includegraphics[width=1.1cm]{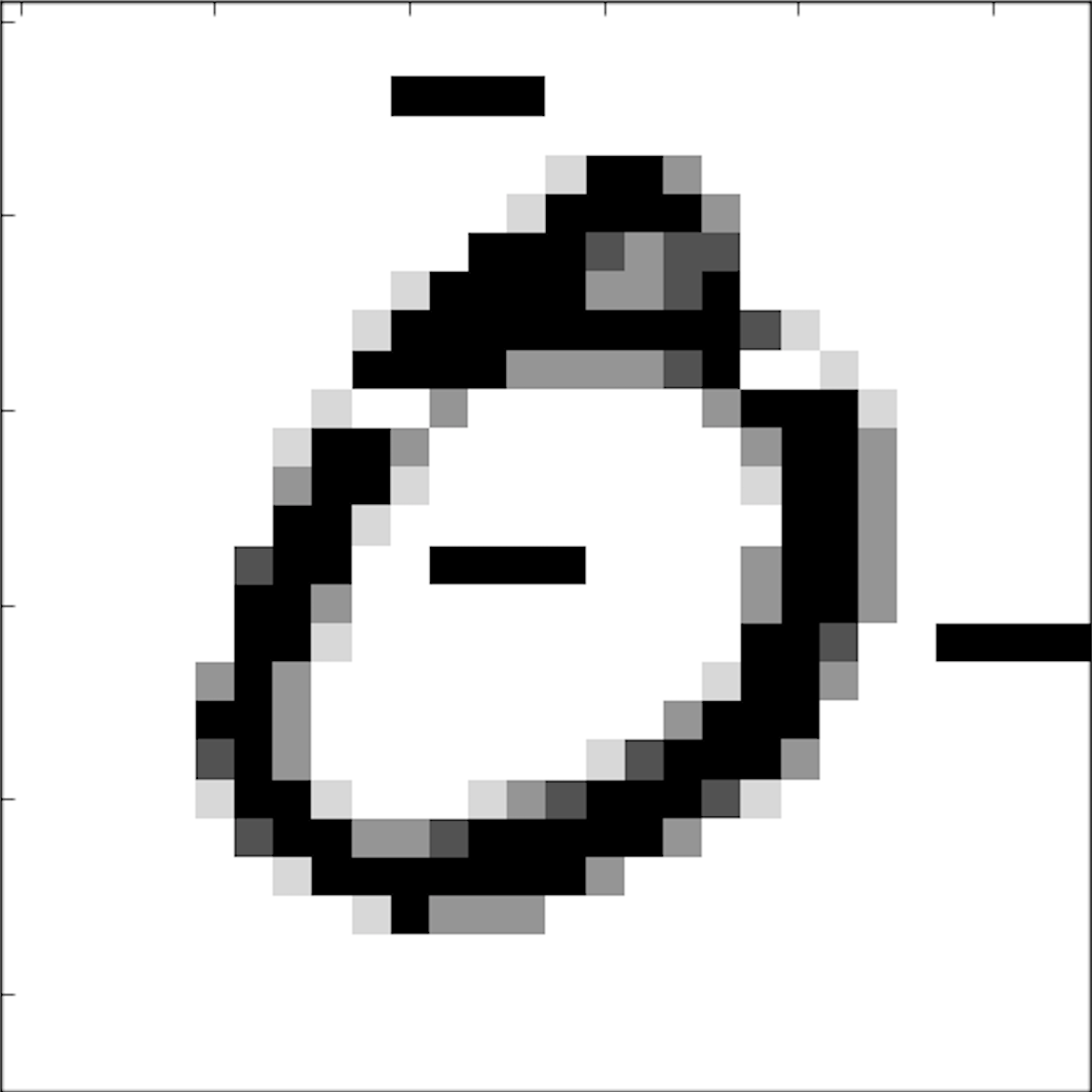}
\text{\hspace{0.8cm}0 to 2}
}
\parbox{2.3cm}{
\includegraphics[width=1.1cm]{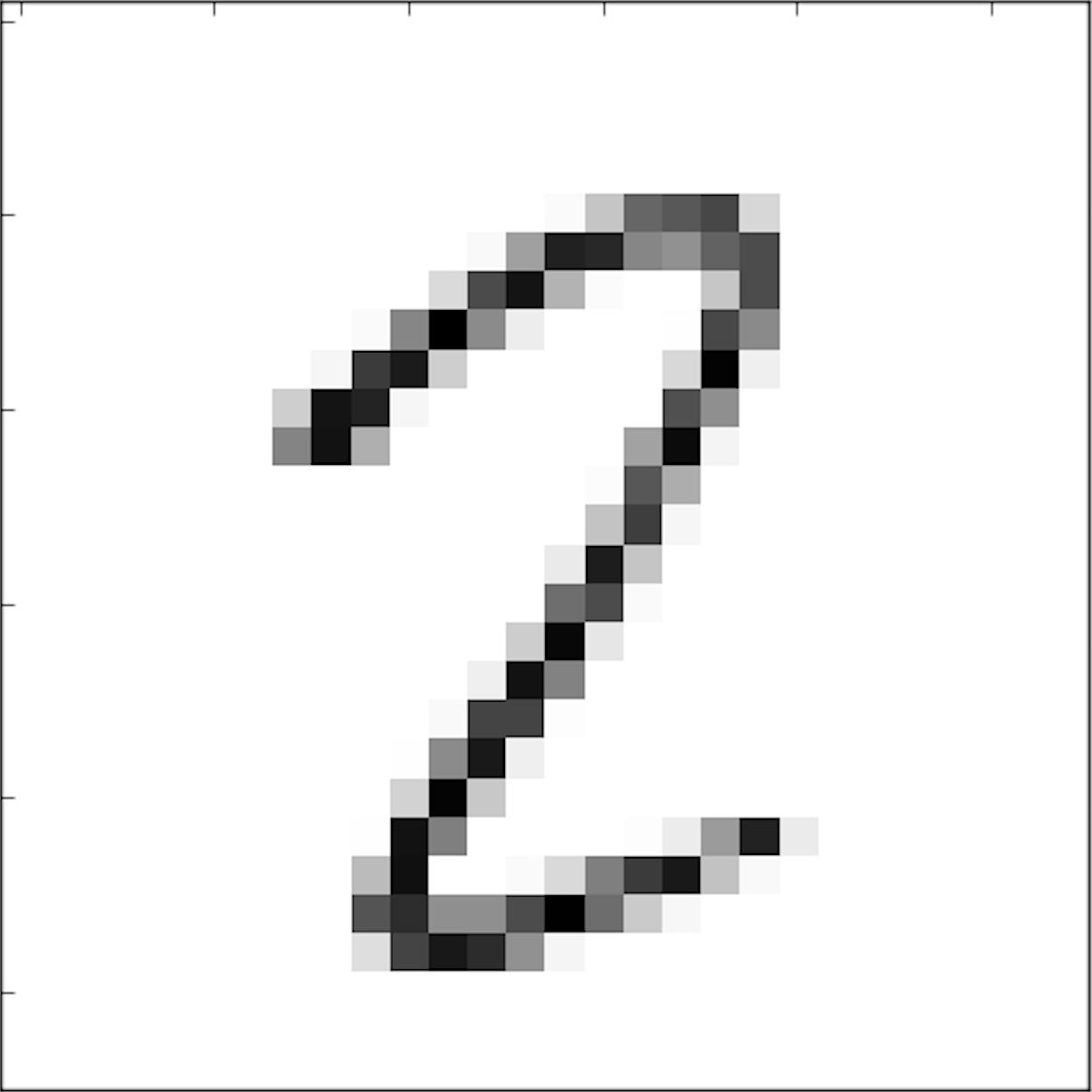}
\includegraphics[width=1.1cm]{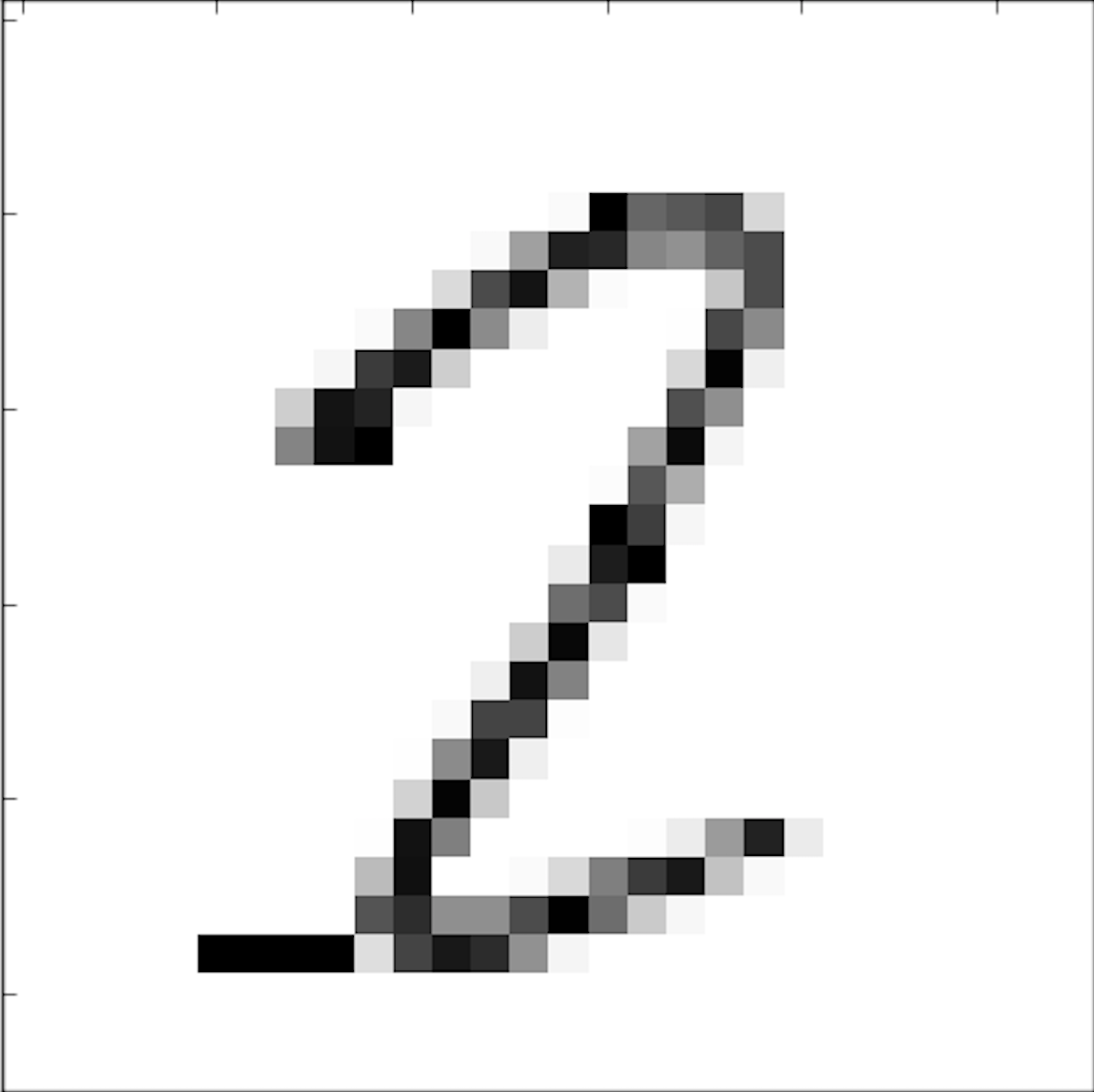}
\text{\hspace{0.8cm}2 to 3}
}
\parbox{2.3cm}{
\includegraphics[width=1.1cm]{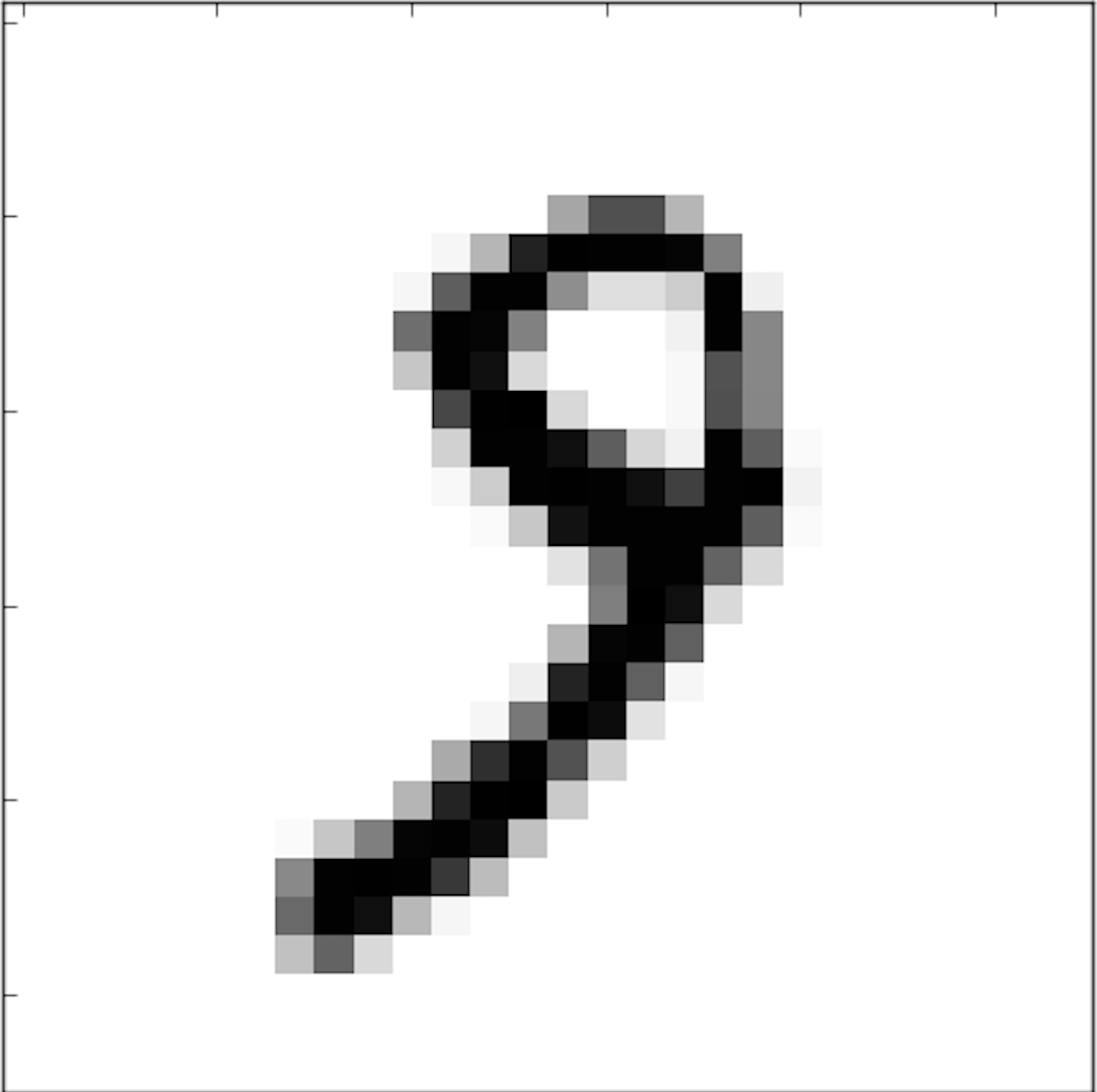}
\includegraphics[width=1.1cm]{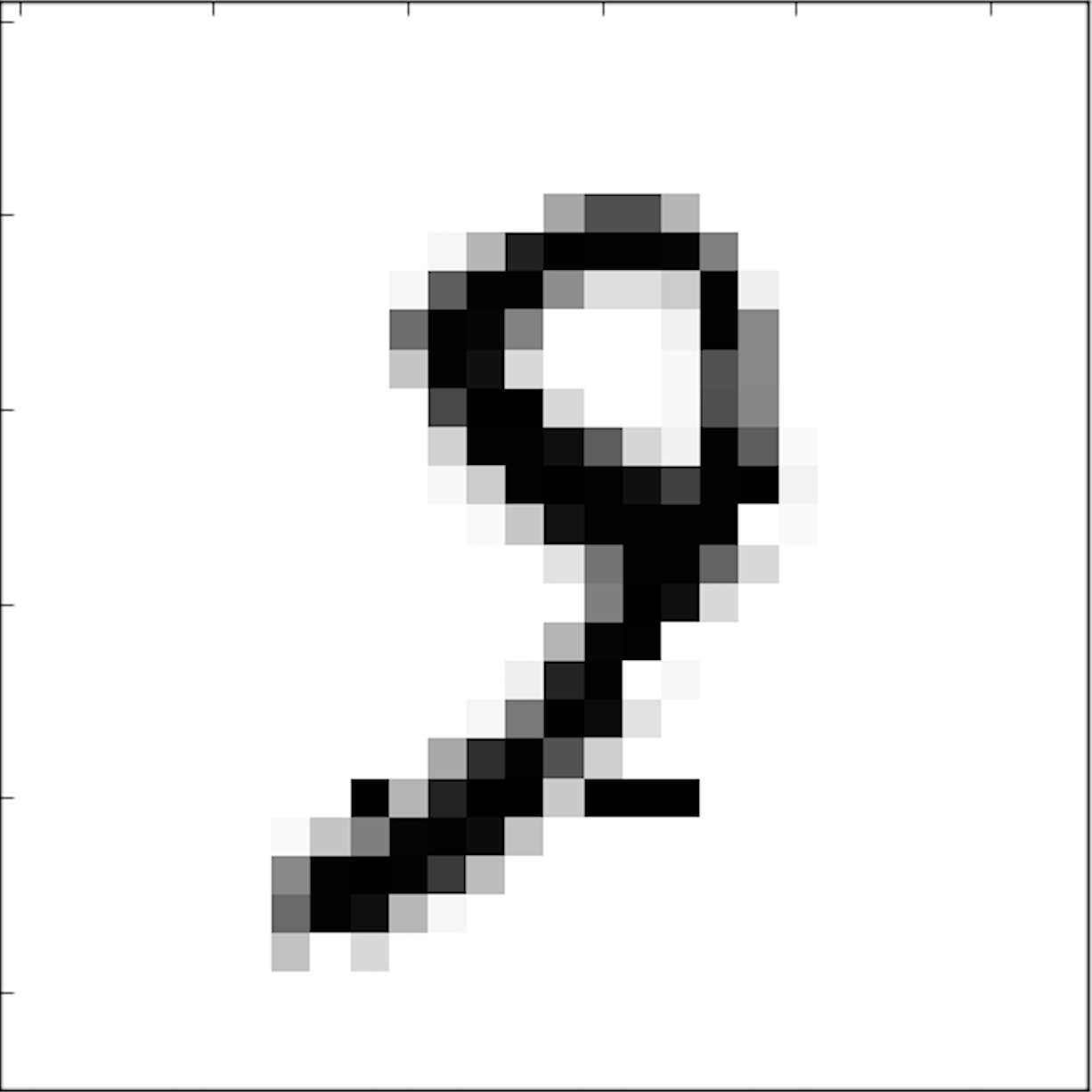}
\text{\hspace{0.8cm}9 to 8}
}
\parbox{2.3cm}{
\includegraphics[width=1.1cm]{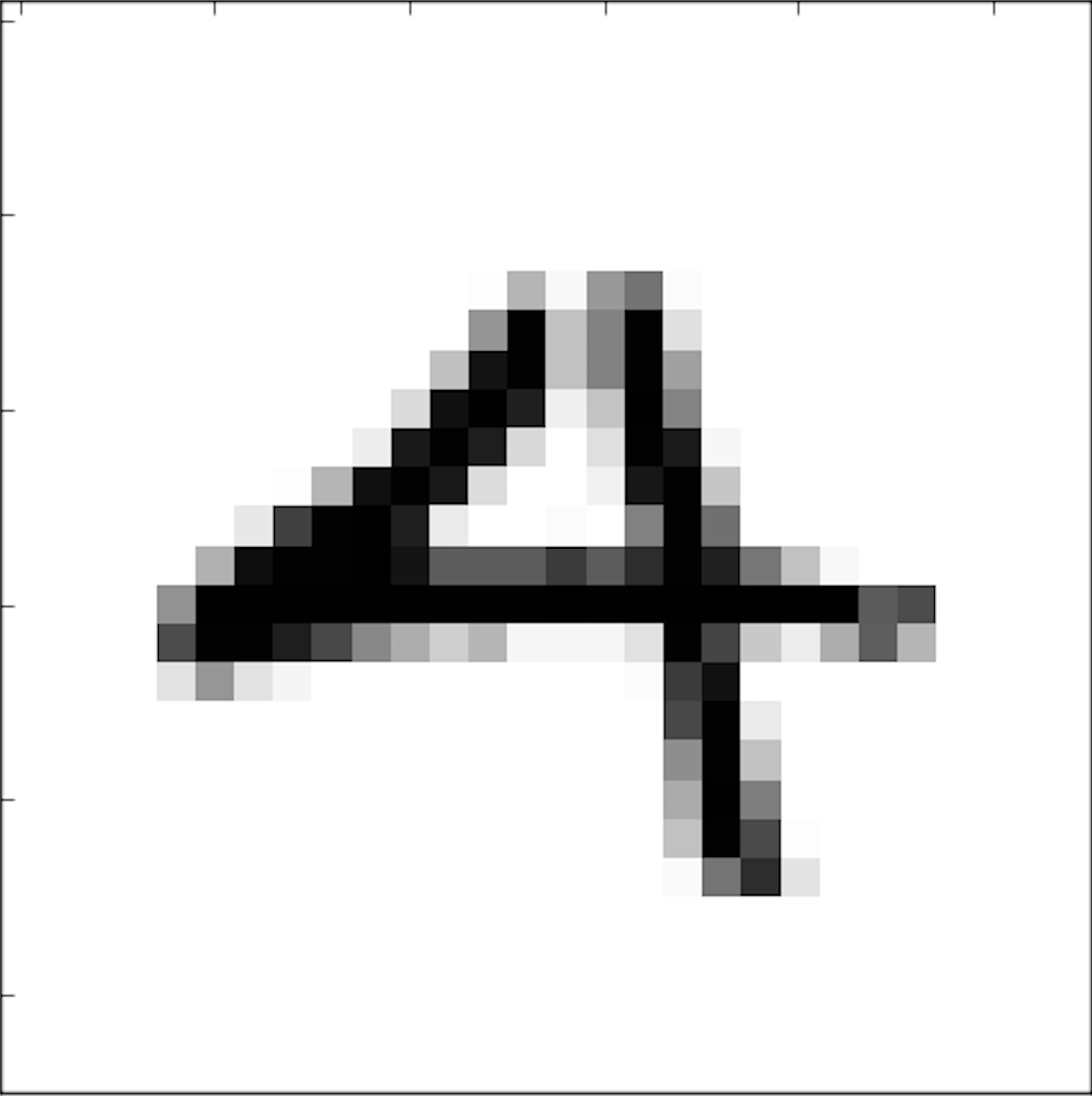}
\includegraphics[width=1.1cm]{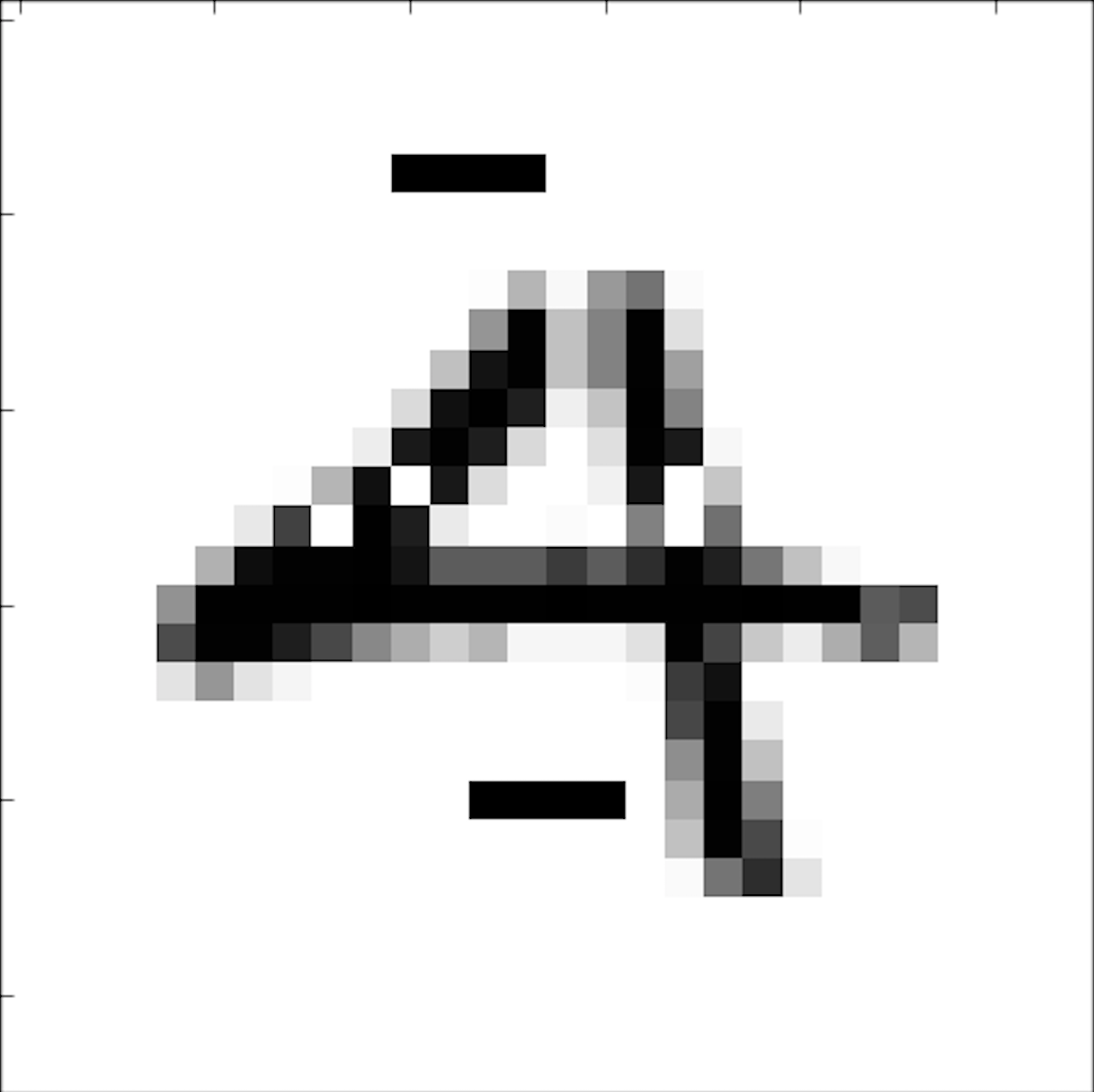}
\text{\hspace{0.8cm}4 to 2}
}
\caption{Adversarial examples for the network trained on the MNIST dataset by multi-path search}
\label{fig:moreMNIST}
\end{figure}

Figure~\ref{fig:cifar10} and Figure~\ref{fig:imageNet} present additional adversarial examples for the CIFAR-10 and ImageNet networks by single-path search. Figure~\ref{fig:moreMNIST} presents adversarial examples for the MNIST network by multi-path search. 

\section{Additional Adversarial Examples for the German Traffic Sign Recognition Benchmark (GTSRB)}

\begin{figure}
\centering
\parbox{3.1cm}{
\includegraphics[width=1.5cm]{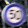}
\includegraphics[width=1.5cm]{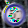}
{speed limit 50 (prohibitory) to speed limit 80 (prohibitory)}
}
\hspace{20pt}
\parbox{3.1cm}{
\includegraphics[width=1.5cm]{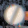}
\includegraphics[width=1.5cm]{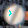}
{restriction ends (other) to restriction ends (80)\\ }
}
\hspace{20pt}
\parbox{3.1cm}{
\includegraphics[width=1.5cm]{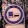}
\includegraphics[width=1.5cm]{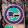}
{no overtaking (trucks) (prohibitory) to speed limit 80 (prohibitory)}
}
\hspace{20pt}
\parbox{3.1cm}{
\includegraphics[width=1.5cm]{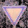}
\includegraphics[width=1.5cm]{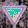}
{give way (other) to priority road (other)\\}
}
\hspace{20pt}
\parbox{3.1cm}{
\includegraphics[width=1.5cm]{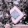}
\includegraphics[width=1.5cm]{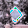}
{priority road (other) to speed limit 30 (prohibitory)}
}
\hspace{20pt}
\parbox{3.1cm}{
\includegraphics[width=1.5cm]{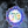}
\includegraphics[width=1.5cm]{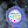}
{speed limit 70 (prohibitory) to speed limit 120 (prohibitory)}
}
\hspace{20pt}
\parbox{3.1cm}{
\includegraphics[width=1.5cm]{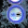}
\includegraphics[width=1.5cm]{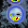}
{no overtaking (prohibitory) to go straight (mandatory)}
}
\hspace{20pt}
\parbox{3.1cm}{
\includegraphics[width=1.5cm]{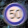}
\includegraphics[width=1.5cm]{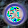}
{speed limit 50 (prohibitory) to stop (other)\\}
}
\hspace{20pt}
\parbox{3.1cm}{
\includegraphics[width=1.5cm]{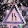}
\includegraphics[width=1.5cm]{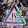}
{road narrows (danger) to construction (danger)\\}
}
\hspace{20pt}
\parbox{3.1cm}{
\includegraphics[width=1.5cm]{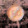}
\includegraphics[width=1.5cm]{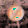}
{restriction ends 80 (other) to speed limit 80 (prohibitory) \\}
}
\hspace{20pt}
\parbox{3.1cm}{
\includegraphics[width=1.5cm]{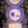}
\includegraphics[width=1.5cm]{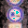}
{no overtaking (trucks) (prohibitory) to speed limit 80 (prohibitory)\\}
}
\hspace{20pt}
\parbox{3.1cm}{
\includegraphics[width=1.5cm]{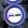}
\includegraphics[width=1.5cm]{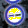}
{no overtaking (prohibitory) to restriction ends (overtaking (trucks)) (other)}
}
\hspace{20pt}
\parbox{3.1cm}{
\includegraphics[width=1.5cm]{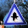}
\includegraphics[width=1.5cm]{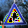}
{priority at next intersection (danger) to speed limit 30 (prohibitory)}
}
\hspace{20pt}
\parbox{3.1cm}{
\includegraphics[width=1.5cm]{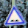}
\includegraphics[width=1.5cm]{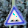}
{uneven road (danger) to traffic signal (danger)\\}
}
\hspace{20pt}
\parbox{3.1cm}{
\includegraphics[width=1.5cm]{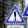}
\includegraphics[width=1.5cm]{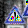}
{danger (danger) to school crossing (danger)\\}
}
\caption{Adversarial examples for the GTSRB dataset by single-path search}
\label{fig:moreGtsrb}
\end{figure}

Figure~\ref{fig:moreGtsrb} presents adversarial examples obtained when selecting single-path search.

\section{Architectures of Neural Networks}

\begin{figure}
\center
\includegraphics[width=0.4\textwidth]{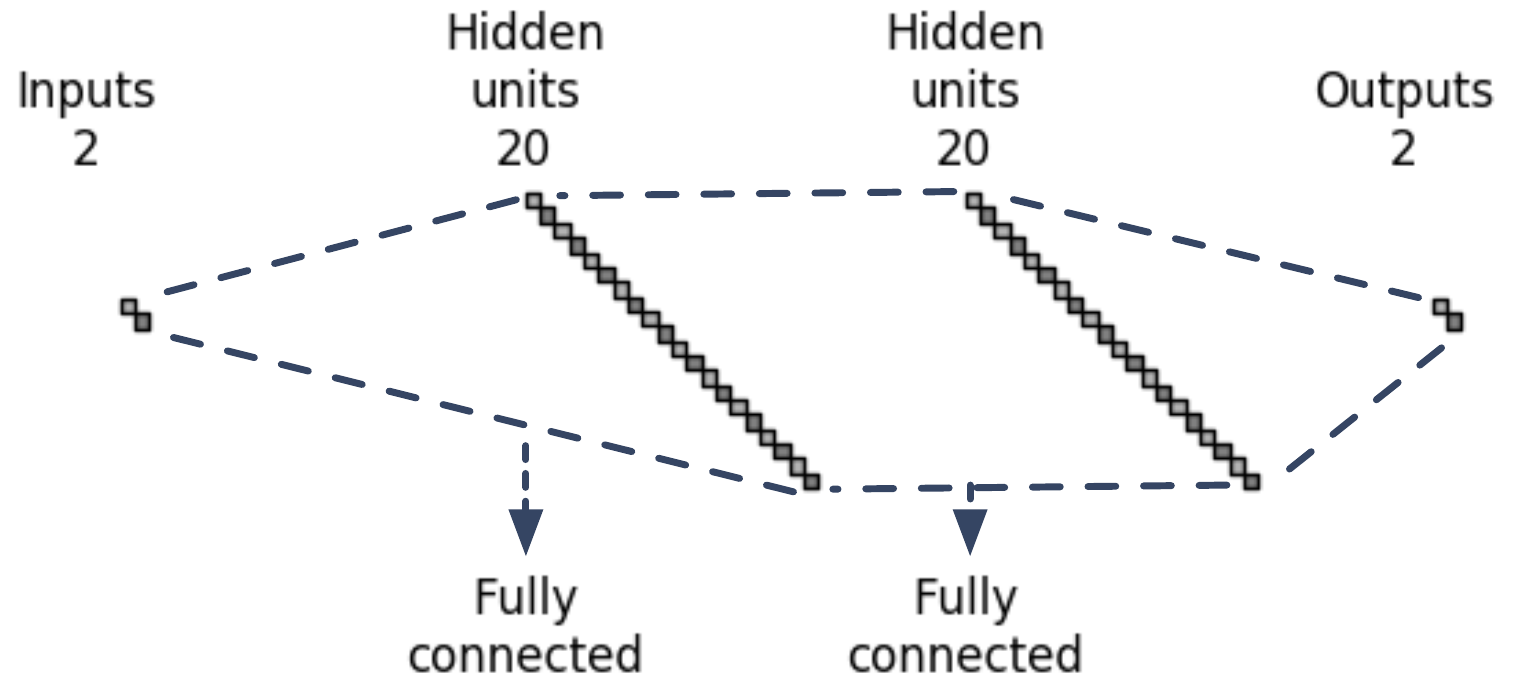}
\caption{Architecture of the neural network for two-dimensional point classification}
\label{fig:archi_curve}
\end{figure}

\begin{figure}
\center
\includegraphics[width=\textwidth]{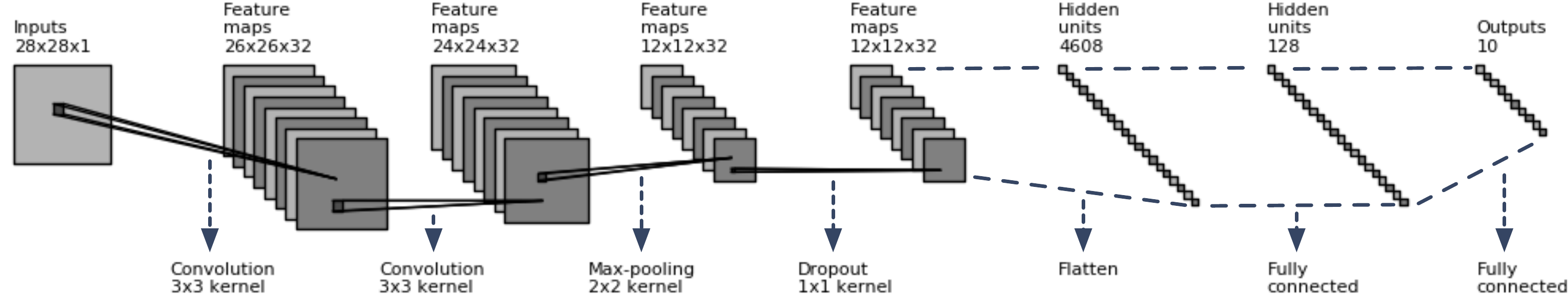}
\caption{Architecture of the neural network for the MNIST dataset}
\label{fig:archi_mnist}
\end{figure}

\begin{figure}
\center
\includegraphics[width=\textwidth]{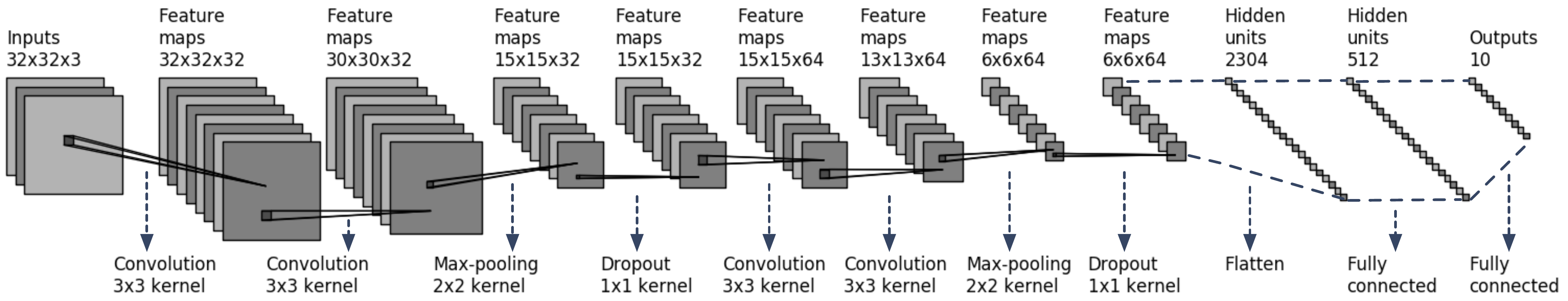}
\caption{Architecture of the neural network for the CIFAR-10 dataset}
\label{fig:archi_cifar}
\end{figure}

\begin{figure}
\center
\includegraphics[width=\textwidth]{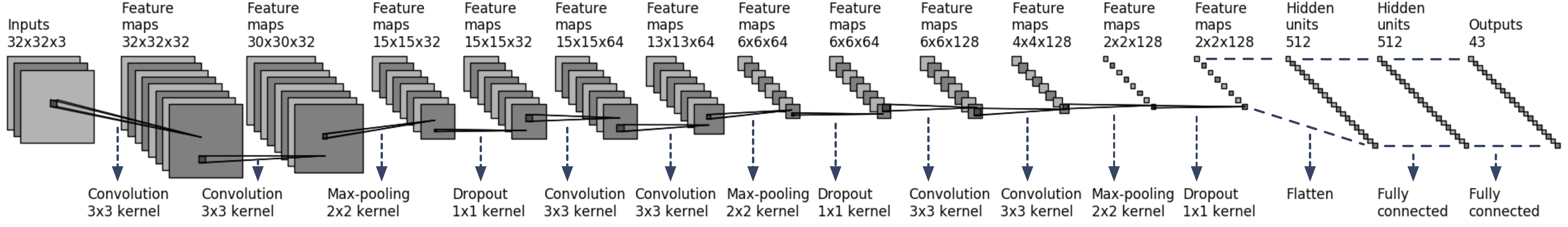}
\caption{Architecture of the neural network for the GTSRB dataset}
\label{fig:archi_gtsrb}
\end{figure}

Figure~\ref{fig:archi_curve}, Figure~\ref{fig:archi_mnist}, Figure~\ref{fig:archi_cifar}, and Figure~\ref{fig:archi_gtsrb} present architectures of the networks we work with in this paper. The network for the ImageNet dataset is from \cite{SZ2014}. 

\end{document}